\documentclass[10pt,twocolumn,letterpaper]{article}
\usepackage[pagenumbers]{cvpr_arxiv} 

\usepackage{graphicx}
\usepackage{amsmath}
\usepackage{amssymb}
\usepackage{booktabs}
\usepackage[pagebackref,breaklinks,colorlinks]{hyperref}

\usepackage[capitalize]{cleveref}
\crefname{section}{Sec.}{Secs.}
\Crefname{section}{Section}{Sections}
\Crefname{table}{Table}{Tables}
\crefname{table}{Tab.}{Tabs.}

\usepackage{appendix}
\usepackage{multirow}
\usepackage{comment}
\usepackage{boldline} 
\usepackage[]{caption}
\usepackage{soul}
\usepackage{xcolor}

\newcommand*{\imagenet}{\textsc{ImageNet}\xspace}
\newcommand*{\resnet}{\textsc{ResNet}\xspace}
\newcommand*{\resnets}{\textsc{ResNet}s\xspace}
\newcommand*{\resnetfifty}{\textsc{ResNet50}\xspace}
\newcommand*{\resnetOneFiftyTwo}{\textsc{ResNet152}\xspace}
\newcommand*{\resnetEighteen}{\textsc{ResNet18}\xspace}
\newcommand*{\inception}{\textsc{Inception}\xspace}
\newcommand*{\deit}{\textsc{DeiT}\xspace}
\newcommand*{\deits}{\textsc{DeiT}s\xspace}
\newcommand*{\deitsmall}{\textsc{DeiT-S}\xspace}
\newcommand*{\deitTiny}{\textsc{DeiT-T}\xspace}
\newcommand*{\deitBase}{\textsc{DeiT-B}\xspace}
\newcommand*{\swin}{\textsc{SWIN}\xspace}
\newcommand*{\swins}{\textsc{SWIN}s\xspace}
\newcommand*{\aptos}{\textsc{APTOS2019}\xspace}
\newcommand*{\ddsm}{\textsc{DDSM}\xspace}
\newcommand*{\isic}{\textsc{ISIC}\xspace}
\newcommand*{\chexpert}{\textsc{CheXpert}\xspace}
\newcommand*{\camelyon}{\textsc{PatchCamelyon}\xspace}
\newcommand*{\wtst}{WT-ST-$n$/$m$\xspace}
\newcommand*{\knn}{$k$-NN\xspace}
\newcommand*{\ltwo}{$\ell_2$\xspace}

\begin{document}

\title{
\vspace{-7mm}\hspace{13mm}
What Makes Transfer Learning Work For Medical Images: \newline Feature Reuse \& Other Factors
\vspace{-4mm}
}

\author{\noindent
{Christos Matsoukas}$^{1,2,3}$\textsuperscript{
\thanks{Corresponding author: Christos Matsoukas \textless{}matsou@kth.se\textgreater{}}}, 
\vspace{1mm}
{Johan Fredin Haslum} $^{1,2,3}$, 
{Moein Sorkhei} $^{1,2}$,
{Magnus Söderberg} $^{3}$, 
{Kevin Smith} $^{1,2}$\\[2mm]
\normalfont{\noindent
$^{1}$ KTH Royal Institute of Technology, Stockholm, Sweden} \\
$^{2}$ Science for Life Laboratory, Stockholm, Sweden \\
$^{3}$ AstraZeneca, Gothenburg, Sweden \\
}

\maketitle

\begin{abstract}

Transfer learning is a standard technique to transfer knowledge from one domain to another. 
For applications in medical imaging, transfer from ImageNet has become the de-facto approach, despite differences in the tasks and image characteristics between the domains.
However, it is unclear what factors determine whether -- and to what extent -- transfer learning to the medical domain is useful.
The long-standing assumption that features from the source domain get reused has recently been called into question. 
Through a series of experiments on several medical image benchmark datasets, we explore the relationship between transfer learning, data size, the capacity and inductive bias of the model, as well as the distance between the source and target domain.
Our findings suggest that transfer learning is beneficial in most cases, and we characterize the important role feature reuse plays in its success.

\end{abstract}
\section{Introduction}
\label{intro}

\let\thefootnote\relax\footnotetext{\vspace{-6pt}\textit{\\Originally published at the Computer Vision and Pattern Recognition Conference (CVPR), 2022.}}

\begin{figure}[ht!]
\begin{center}

\begin{tabular}{@{}c@{}c@{}c@{}}
    \includegraphics[width=0.47\columnwidth]{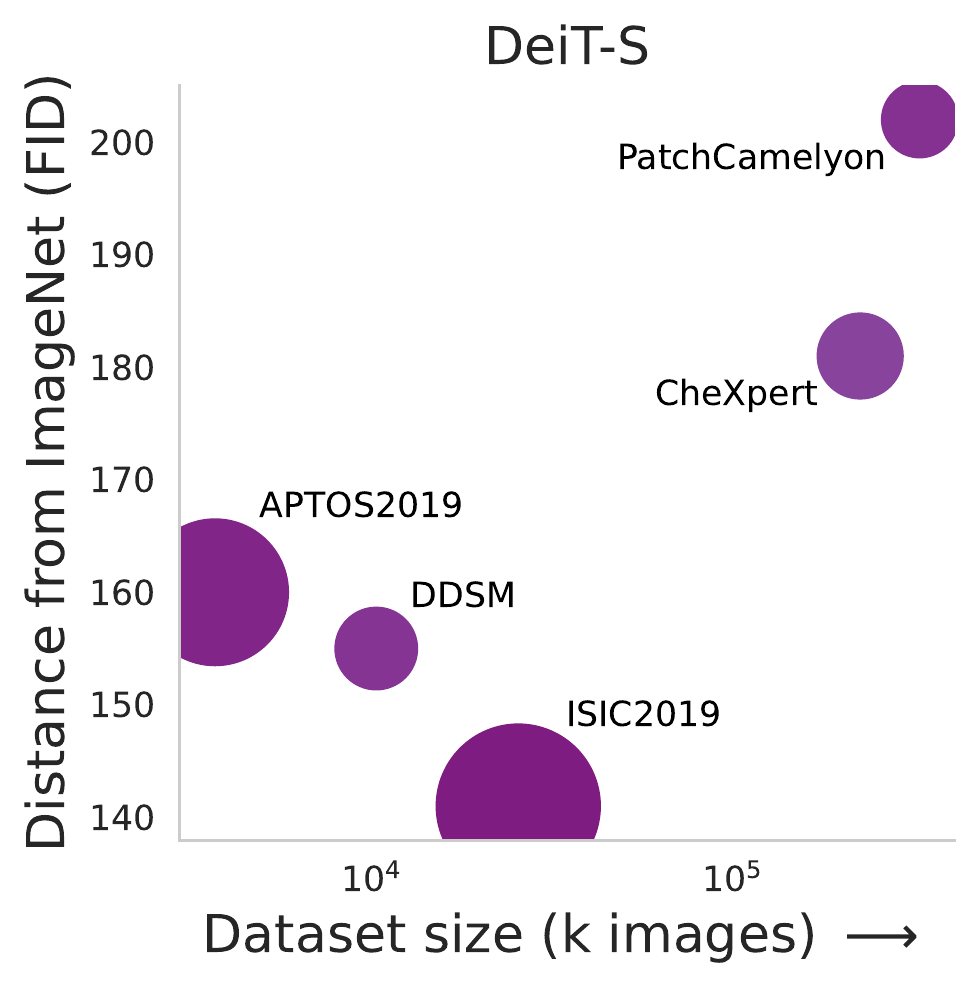} & 
    \includegraphics[width=0.47\columnwidth]{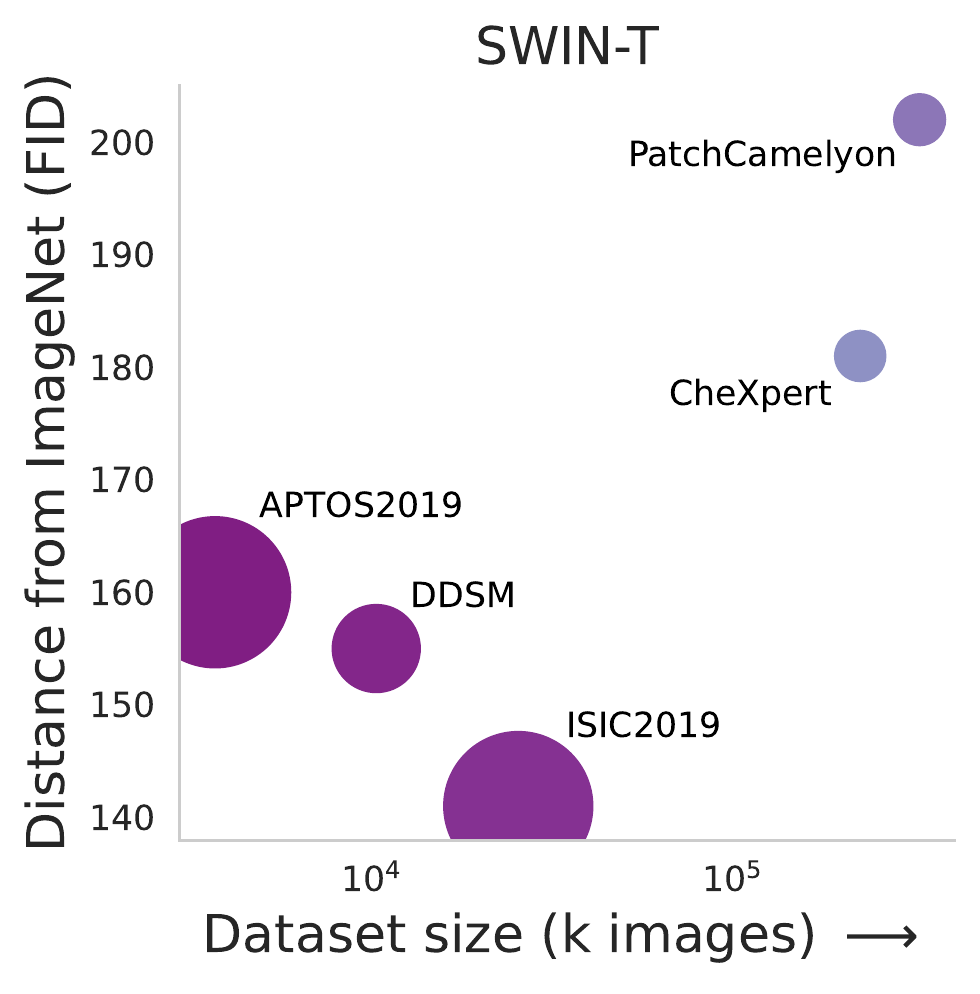}  &
    \multirow{2}{*}[3cm]{\includegraphics[width=0.1 \columnwidth ]{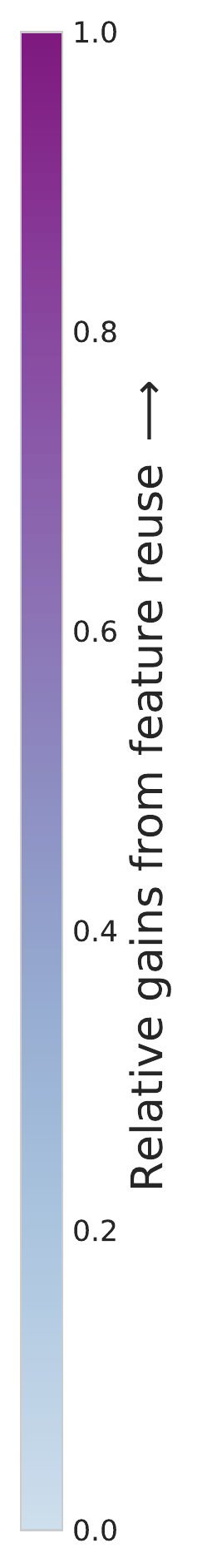}} 
    \\[-1.5mm]
    \includegraphics[width=0.47\columnwidth]{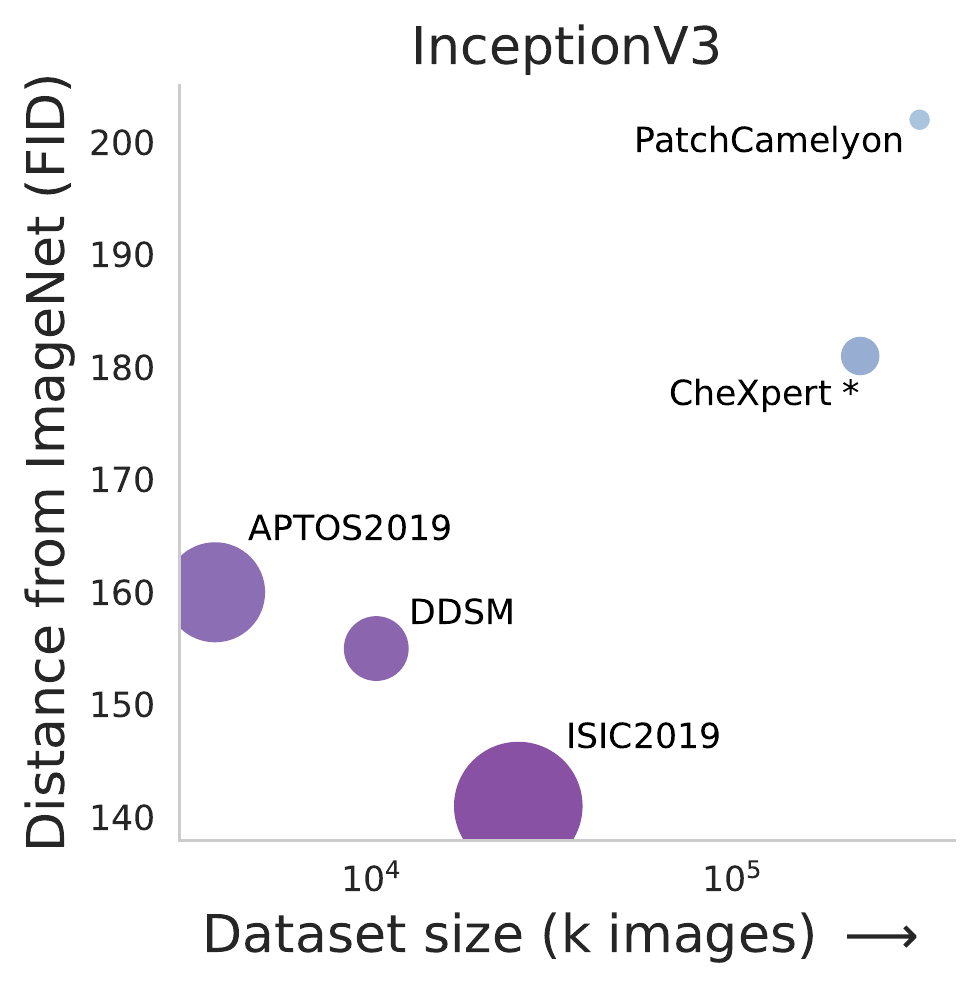} &    
    \includegraphics[width=0.47\columnwidth]{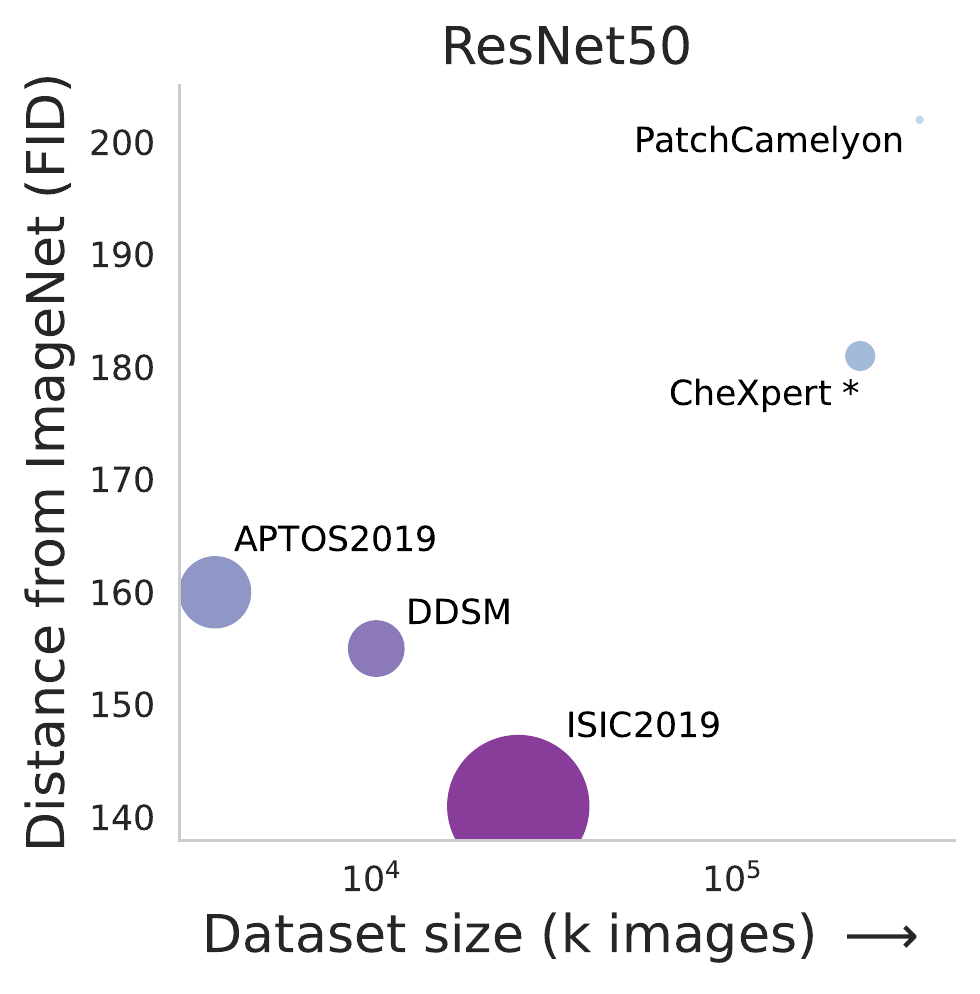} &
    \\[-1.5mm] 
\end{tabular}
\\[2.5mm]
\begin{tabular}{@{}c@{\hspace{0.5mm}}c@{\hspace{0.5mm}}c@{\hspace{0.5mm}}c@{\hspace{0.5mm}}c@{}}
    \includegraphics[width=0.1933\columnwidth]{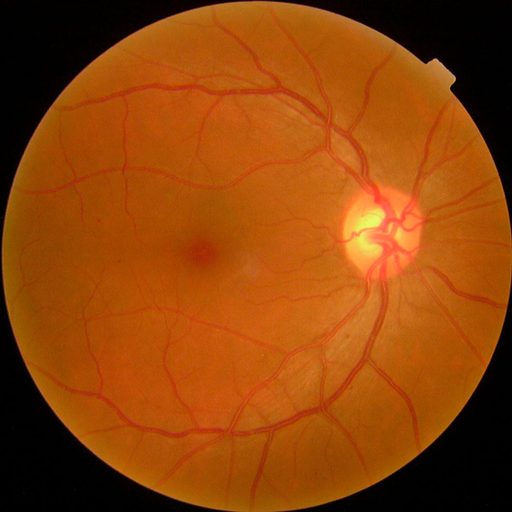} & 
    \includegraphics[width=0.1933\columnwidth]{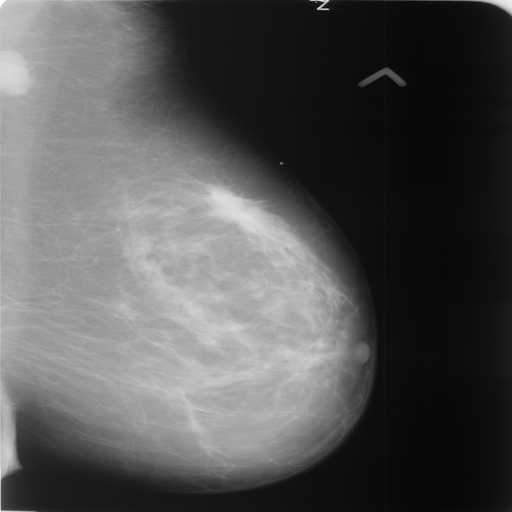} & 
    \includegraphics[width=0.1933\columnwidth]{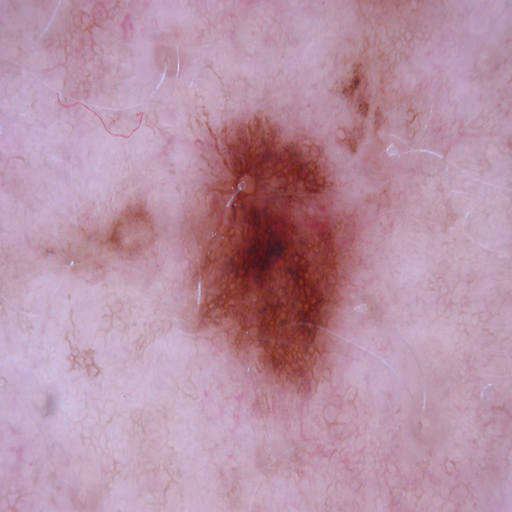} & 
    \includegraphics[width=0.1933\columnwidth]{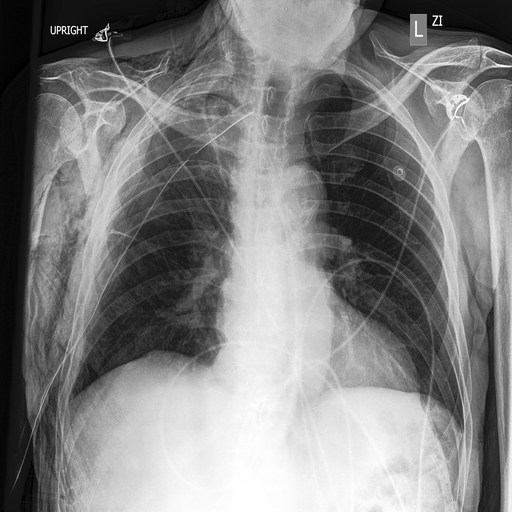} & 
    \includegraphics[width=0.1933\columnwidth]{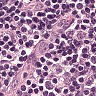}\\[-1.5mm] 
    {\scriptsize APTOS 2019} & {\scriptsize CBIS-DDSM} & {\scriptsize ISIC 2019} & {\scriptsize CheXpert} & {\scriptsize PatchCamelyon} \\    
\end{tabular}
\end{center}

\vspace{-5mm}

\caption{\emph{Factors affecting the utility of transfer learning from ImageNet to medical domains.} 
The size of each dot represents relative increase in performance ($\frac{WT}{RI}$) achieved transferring weights   from \imagenet (WT) compared to random initialization (RI).
The color of the dot indicates how much of the gain can be attributed to feature reuse (relative gains $\frac{WT-ST}{WT}$ from Table \ref{tab:wt_vs_st}, normalized between the minimum and the maximum value for all settings, see Section \ref{methods} for details).
Each panel shows the gains observed by a different model over five runs, in order of increasing inductive biases: {\deitsmall}, {\swin}, {\inception} and {\resnetfifty}.
The benefits from transfer learning increase with (1) reduced data size,  (2)  smaller  distances  between  the  source  and  target domain, and (3) less inductive bias.
Moreover, feature reuse correlates strongly with observed gains from transfer learning, suggesting that feature reuse plays an essential role -- especially for ViTs which lack the inductive biases of CNNs.
(*) indicates cases where feature reuse is less important, uncovered in \cite{transfusion,neyshabur2020being}.}
\label{fig:feature_reuse}
\vspace{-12mm}
\end{figure}

The goal of transfer learning is to reuse knowledge gained in one domain, the \emph{source} domain, to improve performance in another, the \emph{target} domain. 
Transfer learning is often used when data from the target domain is limited.
Such is the case for medical imaging, where the expense of acquisition,
the rareness of the disease, as well as legal and ethical issues limit data size.
The lack of large public datasets 
has led to the widespread adoption of transfer learning from \imagenet \cite{imagenet_cvpr09} to improve performance on medical tasks \cite{tajbakhsh2016convolutional,morid2020scoping,matsoukas2021time}.

Despite its pervasive use, we do not yet fully understand what makes transfer learning from the natural to the medical domain work.
In this paper, we endeavor to paint a more complete picture of which factors enable a successful transfer.
Through a series of comprehensive experiments, we study the effectiveness of transfer learning as a function of dataset size, the distance from the source domain, the model's capacity, and the model's inductive bias.
Our findings, summarized in Figure \ref{fig:feature_reuse}, show that \emph{the benefits from transfer learning increase} with:
\begin{itemize}
    \vspace{-1.0mm}
    \item reduced data size
    \vspace{-1.5mm}
    \item smaller distance between the source and target
    \vspace{-1.5mm}
    \item models with fewer inductive biases
    \vspace{-1.5mm}
    \item models with more capacity, to a lesser extent.
    \vspace{-1.0mm}
\end{itemize}
We also find a strong correlation between the observed benefits from transfer learning and evidence for \emph{feature reuse}.

Much of our understanding about how transfer learning works was, until recently, based on the feature reuse hypothesis.
The feature reuse hypothesis assumes that weights learned in the source domain yield features that can readily be used in the target domain.
In practice, this means that weights learned on ImageNet provide useful features in the target domain, and do not change substantially during fine-tuning
despite differences between the domains \cite{bengio2012deep, bengio2013representation, girshick2014rich, raghu2019rapid}. 
This hypothesis was recently challenged when
Raghu \etal demonstrated that gains observed transferring to a medical task could largely be attributed to weight scaling and low-level statistics \cite{transfusion}, which was later confirmed in \cite{neyshabur2020being}.

We aim to bring some clarity to the role of feature reuse in this work.
Because feature reuse is difficult to measure precisely, we examine it from multiple perspectives through a series of experiments.
We find that \emph{when transfer learning works well}: 
(1) \emph{weight statistics cannot account the majority of the gains}
(2) \emph{evidence for feature reuse is strongest.} 

Our findings do not contradict those of \cite{transfusion, neyshabur2020being}, rather, we show that they uncovered an isolated case (* in Figure \ref{fig:feature_reuse})\footnote{A limitation of \cite{transfusion, neyshabur2020being} was that they only considered CNNs applied to \chexpert, one of the largest publicly available medical imaging datasets (and a similarly large private retinal image dataset in \cite{transfusion}).} where feature reuse is less important: a large dataset, distant from \imagenet.
In this scenario, transfer learning  yields only marginal benefits which can largely be attributed to the weight statistics.
Our work paints a more complete picture, considering datasets with more variety in size and distance to the source domain, and concludes that feature reuse plays an important role in nearly all cases.

We add to this picture with the finding that vision transformers (ViTs), a rising class of models with fewer inductive biases \cite{dosovitskiy2020image, deit}, show a strong dependence on feature reuse in all the datasets we tested.
We select four families of CNNs and ViTs with progressively stronger inductive biases and find that models with less inductive bias rely more heavily on feature reuse.
Moreover, the \emph{pattern of feature reuse} changes in models with less inductive bias.
Specifically, feature reuse in ViTs is concentrated in early layers, whereas CNNs reuse features more consistently throughout the network.

We share the code to reproduce our experiments, available at
\href{https://github.com/ChrisMats/feature-reuse}{github.com/ChrisMats/feature-reuse}.

\section{Problem Formulation and Methodology}
\label{methods}

The aim of this work is to examine transfer learning from the natural to the medical image domain.
Our central question is:~\emph{what factors determine if transferred representations are effective in the medical domain?}
Under what conditions do they yield improved performance?
Is this affected by the size of the target dataset?
The similarity/dissimilarity to the source dataset?
What role does feature reuse play?
Which of the source features are reused?
And finally, what roles do the model's architecture and inductive biases play?

To investigate these questions, we conduct a series of experiments considering a variety of medical image datasets, initialization strategies, and architectures with different levels of inductive bias.
We also perform several ablation studies to characterize feature reuse at different depths throughout each network.
The details of our methodology are described below.

\vspace{-2mm}
\paragraph{Datasets.}
We select datasets that help us characterize how the efficacy of transfer learning varies with properties of the data.
For the source domain, we use \imagenet throughout this work.
For the target domain, we select a representative set of five standard medical image classification datasets. 
They cover a variety of imaging modalities and tasks, ranging from  a few thousand examples to the largest public medical imaging datasets.
\begin{itemize}
\vspace{-1mm}
\item \textbf{\aptos} $(N = 3,662)$ High-resolution diabetic retinopathy images where the task is classification into 5 categories of disease severity \cite{kaggle}.
\vspace{-1mm}
\item \textsc{\textbf{CBIS-DDSM}} $(N=10,239)$ 
A mammography dataset in which the task is to detect the presence of masses \cite{CBIS_DDSM_Citation,DDSM}.
\vspace{-1mm}
\item \textbf{\isic 2019} $(N=25,331)$ Dermoscopic images -- the task is to classify among 9 different diagnostic categories of skin lesions \cite{tschandl2018ham10000,codella2018skin,combalia2019bcn20000}.
\vspace{-1mm}
\item \textbf{\chexpert} $(N=224,316)$ Chest X-rays with labels over 14 categories of diagnostic observations \cite{chexpert}.
\vspace{-1mm}
\item \textbf{\camelyon} $(N=327,680)$ Patches of H\&E stained WSIs of lymph node sections. The task is to classify each patch as cancerous or normal \cite{bejnordi2017diagnostic, veeling2018rotation}.
\vspace{-1mm}
\end{itemize}
We compute the Fréchet Inception Distance (FID) \cite{fid} between \imagenet and the datasets listed above to measure similarity to the source domain (Figure \ref{fig:feature_reuse} and Table \ref{tab:wt_vs_st}).
Although it may not be a perfect measure \cite{lucic2017gans, borji2019pros}, it gives a reasonable indication of relative distances between datasets.

\vspace{-2mm}
\paragraph{Architectures.}
To study the role of network architecture we selected two representative ViT models, \textsc{DeiT} \cite{deit} and \textsc{SWIN} \cite{liu2021swin}, and two representative CNN models, \textsc{ResNet}s \cite{resnet} and \textsc{Inception} \cite{szegedy2016rethinking}.
We selected these model types because they are widely studied and commonly used  as backbones for other networks.
To ensure a fair comparison we select architectural variants that are similar in capacity for our main experiments.

Aside from their popularity, another reason we chose these models is to study the role of \emph{inductive bias} in transfer learning -- as each model has a unique set of inductive biases built in.
The models, in increasing order of inductive bias are: \deit, \swin, \inception, and \resnet.
We start with the model with the least inductive biases,  the \textsc{DeiT} family.
Like the original ViT \cite{dosovitskiy2020image}, \deit is similar in spirit to a pure transformer -- doing away with nearly all image-specific inductive biases, \eg locality, translational equivariance, and hierarchical scale.
According to Dosovitskiy \etal, this causes pure ViTs like \deit\footnote{We use DEIT  without the distillation token \cite{deit}.} to generalize poorly when trained on insufficient amounts of data \cite{dosovitskiy2020image}.
Recently, \swin transformers were shown to outperform \deits and other ViTs on \imagenet by reintroducing many inductive biases of CNNs. 
Combining self-attention with a hierarchical structure that operates locally at different scales, \swin transformers have built locality, translational equivariance, and hierarchical scale into ViTs.
Moving to CNNs, we include \inception, an older CNN which features an inception block that processes the signal in parallel at multiple scales before propagating it to the next layer.
Finally, we selected the \resnet family, as it is the most common and highly cited CNN backbone, and recent works have shown that \textsc{ResNet}s are competitive with recent SOTA CNNs when modern training methods are applied \cite{bello2021revisiting}.

\vspace{-2mm}
\paragraph{Initialization methods.}
To understand the mechanism driving the success of transfer learning from \imagenet to the medical domain, we need to assess \emph{to what extent improvements from transfer learning can be attributed to feature reuse}.
Transfer learning is typically performed by taking an  architecture, along with its \imagenet pretrained weights, and then fine-tuning it on the target task.
Two things are transferred via this process: the model architecture and its learned weights.
Raghu \etal showed that the actual values of the weights are not always necessary for good transfer learning performance \cite{transfusion}.
One can achieve similar performance by initializing the network using its \emph{weight statistics}.
In this setting, transfer amounts to providing a good range of values to randomly initialize the network -- \emph{eliminating feature reuse as a factor}.

To isolate the contribution of feature reuse vs.~weight statistics, we employ three initialization strategies:
\begin{itemize}
\vspace{-1mm}
\item \emph{Weight transfer (WT)} -- transferring \imagenet  pre-trained weights,
\vspace{-1mm}
\item \emph{Stats transfer (ST)} -- sampling weights from a normal distribution whose mean and variance are taken layer-wise from an \imagenet pre-trained model,
\vspace{-1mm}
\item \emph{Random init.~(RI)} -- Kaiming initialization \cite{he2015delving}.
\vspace{-1mm}
\end{itemize}
Interrogating the differences between models initialized with these methods gives an indication as to what extent the transferred model reuses \imagenet features.
Furthermore, we can investigate \emph{where feature reuse is beneficial within the network} by transferring weights (WT) up to block $n$ and initializing the remaining $m$ blocks using ST.
We denote this setup WT-ST. 
For example, a \resnetfifty with weight transfer up to \texttt{conv1} is written ResNet50-WT-ST-1/5 \footnote{The number of blocks differs for each model; for CNNs $n=1$ corresponds to the first convolutional layer, for ViTs it refers to the patchifier.}. 

\vspace{-2mm}
\paragraph{Representational similarity.}
Looking more closely at feature reuse within the network, we ask the questions: 
\emph{how are features organized before and after fine-tuning -- are they similar? 
Can feature similarity reveal feature reuse, or lack thereof?}
To answer these questions, we use Centered Kernel Alignment (CKA) to compute similarity between features within and across networks \cite{kornblith2019similarity}.
CKA's properties of invariance to orthogonal transformations and isotropic scaling allow meaningful quantitative comparisons between representations of different size.
We compute CKA pairwise between every layer (in a single network or pair of networks) to provide a visual overview of network similarity.

\vspace{-2mm}
\paragraph{Resilience of the transfer.}
It is difficult to directly measure whether transferred features are reused after fine-tuning.
But, by investigating how ``sticky'' the transfer was -- how much the weights drifted from their initial transferred values during fine-tuning -- we can gain some insights.
We use two different strategies to quantify the ``stickiness'' of the transfer:
\emph{(1)} we compute the \ltwo distance between the initial weights and the weights after fine-tuning;
\emph{(2)} we measure the impact of resetting a layer's weights to their initial values, a property called \emph{re-initialization robustness} by Zhang \etal \cite{zhang2019all}.
Layers that undergo critical changes during fine-tuning (and thus exhibit low robustness) have either not re-used the transferred weights well or adapted strongly to the new domain.

\vspace{-2mm}
\paragraph{Analyzing transferred representations layer-wise.}
The next questions we wish to address are: \emph{Which parts of the network produce/reuse low-level vs.~high-level features?}
And 
\textit{how do differences in representation between CNNs and ViTs impact transfer learning?} 
The representational power and the effective receptive field of CNNs increase with depth.
ViTs, on the other hand, ``see'' differently \cite{raghu2021vision} -- they maintain more uniform representations throughout, and can utilize both local and global features at every layer.

To investigate these questions, we assess the representational power of the transferred features throughout the network.
After initialization with WT, ST, and WT-ST, we fine-tune on the target dataset and apply a $k$-NN evaluation protocol at the layers in question \cite{caron2021emerging}.
This compares the embedded representation of test samples to the $k = 200$
closest embeddings from the training set using cosine similarity.
Essentially, this test allows us to see when high-level features emerge within the network.
For CNNs, the embedding is obtained using global average pooling at the layer in question.
For ViTs we follow a similar procedure, but with special modifications to handle the \texttt{cls} token in \deits.
The \texttt{cls} token processes information differently than the spatial tokens, carrying much of the information necessary for classification \cite{dosovitskiy2020image, deit, raghu2021vision}. 
Therefore we construct the embeddings in three different ways: \textit{(1)} using only the \texttt{cls} token's activations, \textit{(2)} using activations from the spatial tokens, \textit{(3)} concatenating \textit{(1)} and \textit{(2)}.

\begin{figure}[t]
\begin{center}
\vspace{-2mm}
\begin{tabular}{@{}c@{}c@{}c@{}}
    \includegraphics[width=0.333\columnwidth]{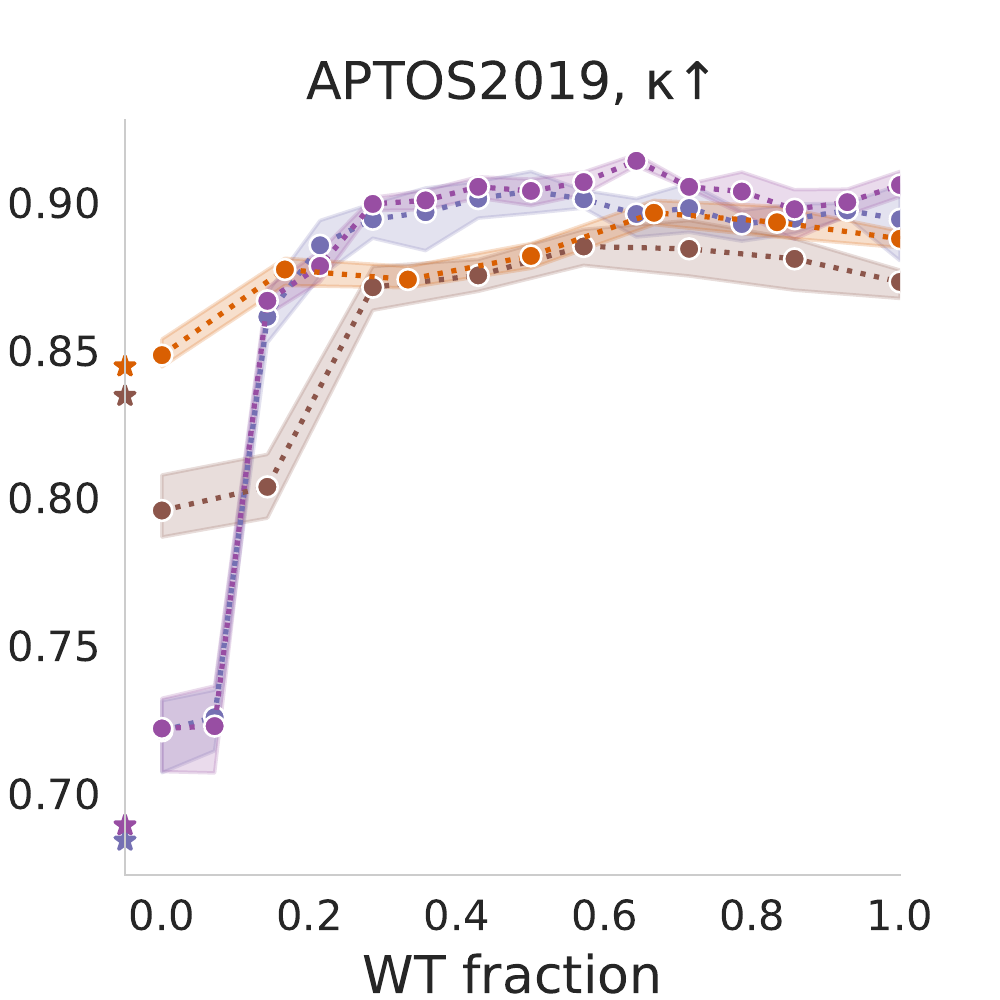} & 
    \includegraphics[width=0.333\columnwidth]{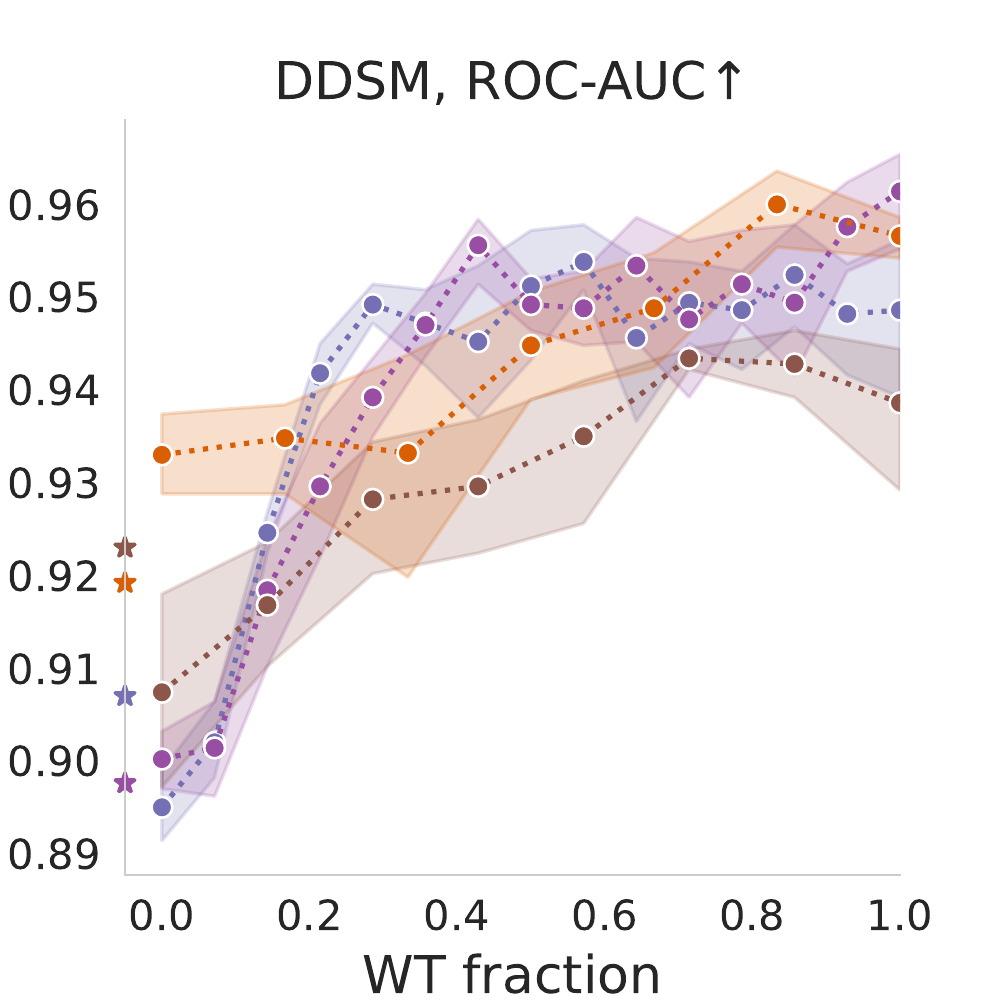} & 
    \includegraphics[width=0.333\columnwidth]{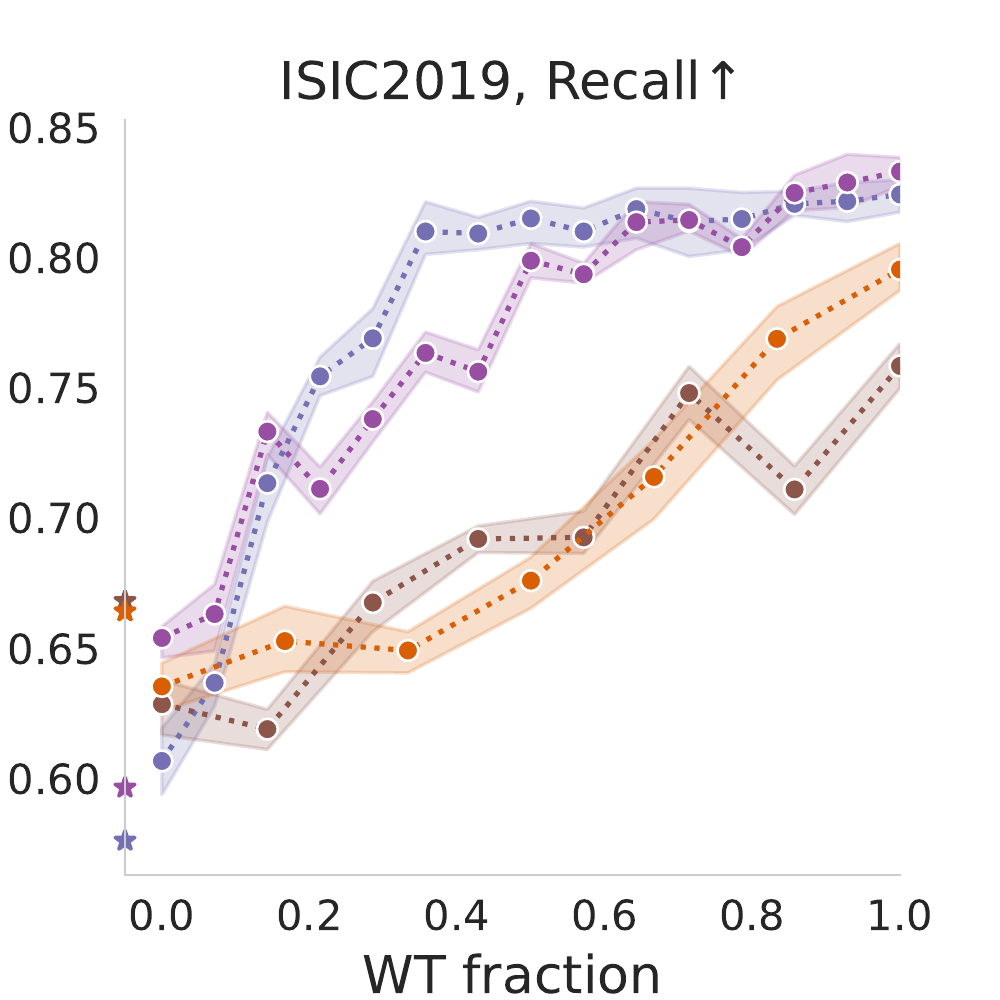}\\[-1.5mm] 
    \includegraphics[width=0.333\columnwidth]{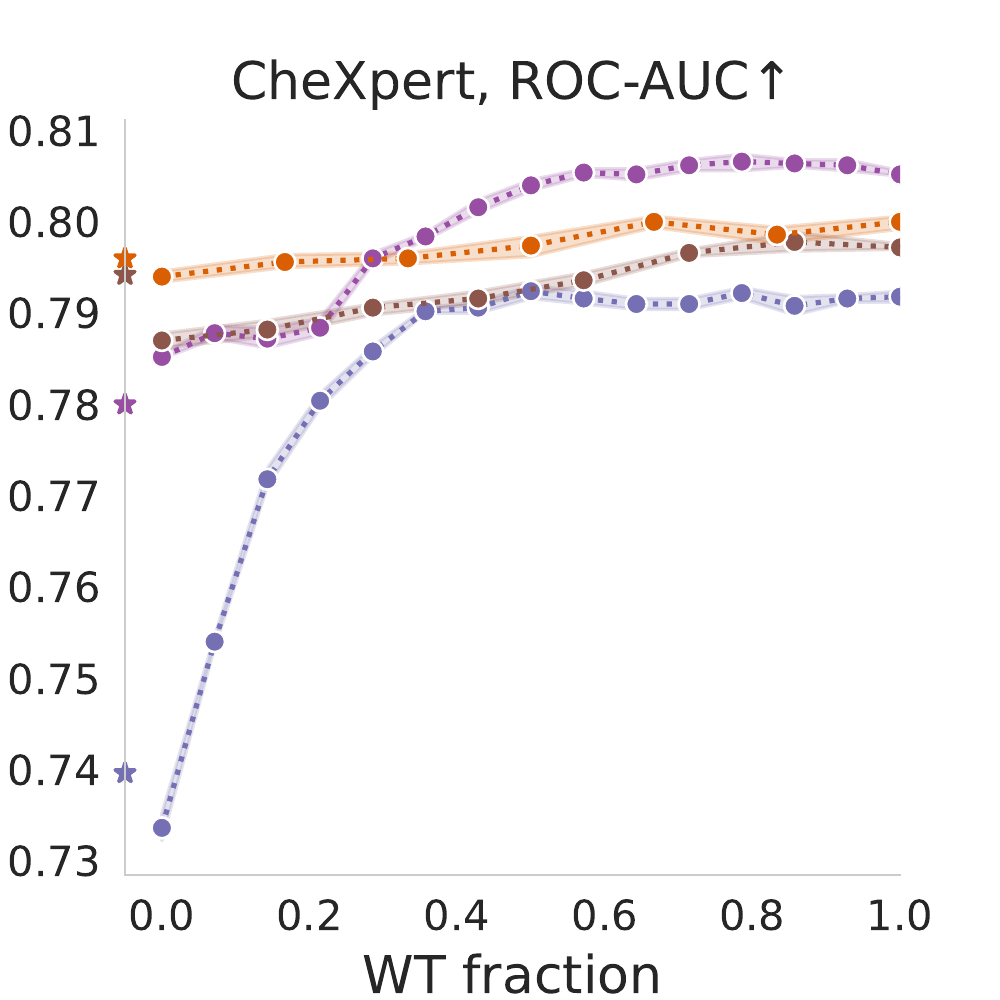} & 
    \includegraphics[width=0.333\columnwidth]{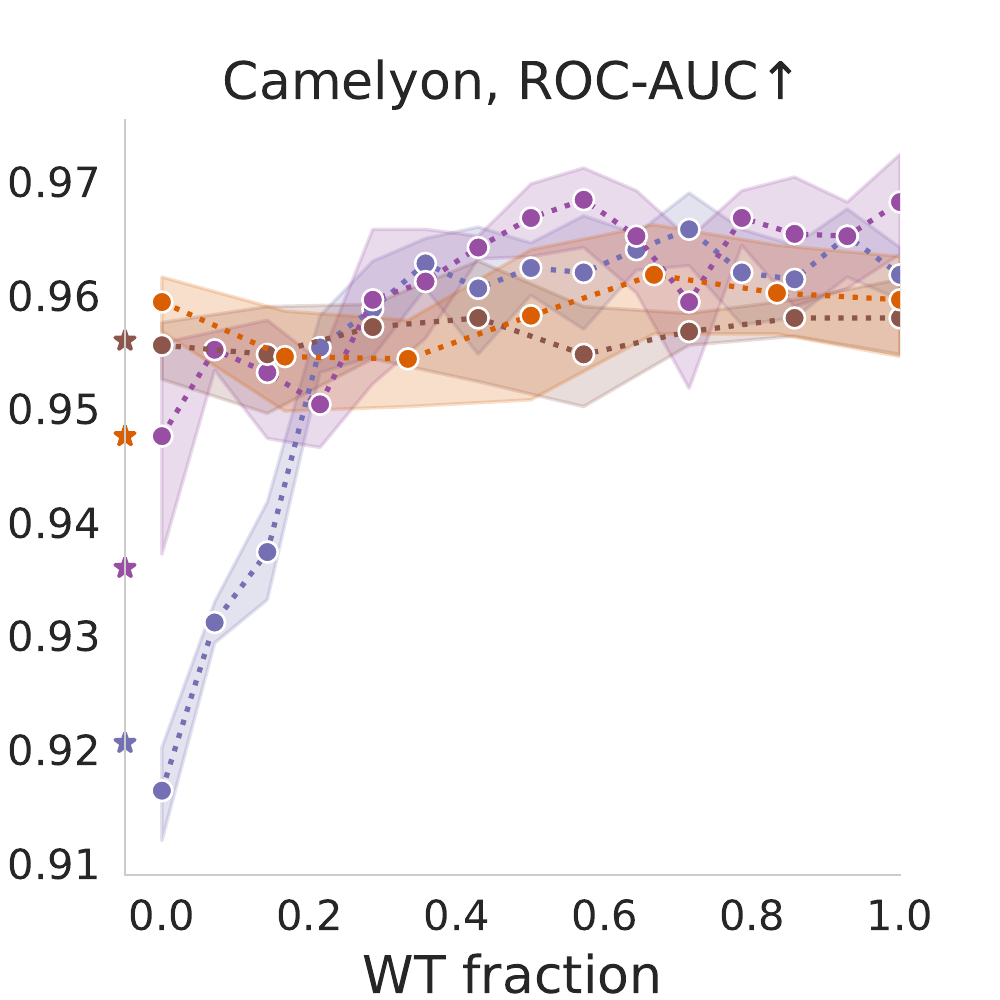} &
    \includegraphics[width=0.333\columnwidth]{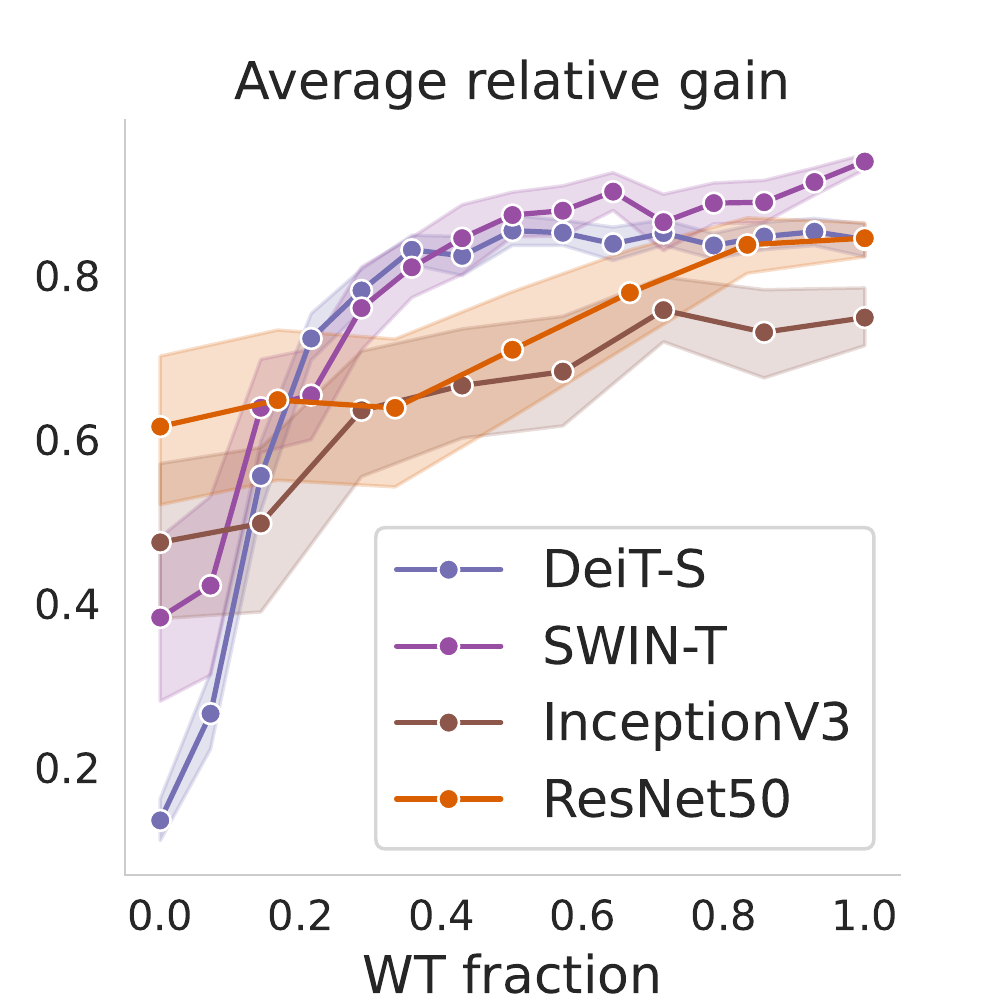}\\[-1.5mm] 
\end{tabular}
\end{center}
\vspace{-3mm}
\caption{\emph{Which layers benefit from feature reuse?} We evaluate the impact of weight transfer when using WT-ST initialization 
(WT fraction from 0 to 1, where 0 = ST and 1 = WT).
Lower performance on the left indicates that the network relies on transferred weights.
$\star$ = RI.
The last panel reports the average relative gains for each model type averaged over all datasets. 
Details of WT-ST initialization can be found in Appendix \ref{appdx:wtst_details}.}
\label{fig:wst_all}
\vspace{-4mm}
\end{figure}

\vspace{-2mm}
\paragraph{Training procedure.} 
Unless otherwise specified, we used the following training procedure for all experiments.
Each dataset was divided into 80/10/10 train/test/validation splits, with the exception of APTOS2019, which was divided 70/15/15 due to its small size.
Images were normalised
and resized to $256\times256$ with the following augmentations applied: color jitter, random vertical and horizontal flips, and random crops $224\times224$ after rescaling.
\imagenet-pretrained weights were either available in PyTorch \cite{NEURIPS2019_9015} or downloaded from the official repositories, in the cases of {\deit} and {\swin}.
CNN and ViT models were trained with the Adam \cite{adam} and AdamW \cite{adamw} optimizers respectively, with a batch size of 64.
We performed independent grid searches to find suitable learning rates, and found that $10^{-4}$ works best for both CNNs and ViTs, except for RI which used $3\times10^{-4}$.
We used these as the base learning rates for the optimizers along with default 1,000 warm-up iterations.
During training, we reduce the learning rate by a factor of 10 when the validation performance saturates, until we reach a final learning rate of $10^{-6}$.
For transformer models, we used the default patch size of $16\times16$ for the {\deit} models and $4\times4$ for  {\swin}.
For each run, we save the initial checkpoint and the checkpoint with highest validation performance.

\section{Experiments}
\label{experiments}

In this section, we report our findings related to transfer learning and feature reuse.
Unless otherwise stated, each experiment is repeated 5 times. 
We report the mean and standard deviation of the appropriate evaluation metric for each dataset: Quadratic Cohen Kappa for \aptos, Recall for \isic, and ROC-AUC for \ddsm, \chexpert, and \camelyon. 

\addtolength{\tabcolsep}{-5pt}  
\begin{table}[t]
\tiny
\begin{tabular}{llccccc}
\toprule
\textbf{Model} & 
\textbf{Init} & 
\textbf{APTOS2019}, $\kappa \uparrow$  &
\textbf{DDSM}, AUC $\uparrow$ &
\textbf{ISIC2019}, Rec. $\uparrow$ & 
\textbf{CheXpert}, AUC $\uparrow$ &
\textbf{Camelyon}, AUC $\uparrow$ 
\\
(\# parameters)
& 
& 
$n =$ 3,662  & 
$n =$ 10,239 &
$n =$ 25,333 & 
$n =$ 224,316 &
$n =$ 327,680 \\
& 
& 
FID = 160  & 
FID = 155 &
FID = 141 & 
FID = 181 &
FID = 202 \\
\midrule

\multirow{3}{*}{\parbox{1cm}{DeiT-S \\ (22M)}}

& RI &
0.684   $\pm$ 0.017 &
0.907   $\pm$ 0.005 &
0.576   $\pm$ 0.013 &
0.740   $\pm$ 0.006 &
0.921   $\pm$ 0.002 
\\
& ST &
0.721   $\pm$ 0.016 &
0.895   $\pm$ 0.005 &
0.607   $\pm$ 0.017 &
0.734   $\pm$ 0.002 &
0.916   $\pm$ 0.005 
\\
& WT & 
0.894   $\pm$ 0.017 &
0.949   $\pm$ 0.011 &
0.824   $\pm$ 0.008 &
0.792   $\pm$ 0.001 &
0.962   $\pm$ 0.003 
\\[0.5em]

\multirow{3}{*}{\parbox{1cm}{SWIN-T \\ (29M)}}
& RI &
0.689   $\pm$ 0.022 &
0.898   $\pm$ 0.005 &
0.597   $\pm$ 0.080 &
0.780   $\pm$ 0.001 &
0.936   $\pm$ 0.002 
\\
& ST &
0.722   $\pm$ 0.017 &
0.900   $\pm$ 0.004 &
0.654   $\pm$ 0.008 &
0.785   $\pm$ 0.000 &
0.948   $\pm$ 0.013 
\\
& WT & 
0.906   $\pm$ 0.005 &
0.961   $\pm$ 0.007 &
0.833   $\pm$ 0.008 &
0.805   $\pm$ 0.000 &
0.968   $\pm$ 0.006 
\\[0.5em]

\multirow{3}{*}{\parbox{1cm}{InceptionV3 \\ (24M)}}
& RI &
0.835   $\pm$ 0.012 &
0.923   $\pm$ 0.003 &
0.668   $\pm$ 0.008 &
0.794   $\pm$ 0.001 &
0.956   $\pm$ 0.006 
\\
& ST &
0.796   $\pm$ 0.014 &
0.907   $\pm$ 0.014 &
0.629   $\pm$ 0.013 &
0.787   $\pm$ 0.001 &
0.956   $\pm$ 0.003 
\\
& WT & 
0.873   $\pm$ 0.007 &
0.939   $\pm$ 0.010 &
0.758   $\pm$ 0.011 &
0.797   $\pm$ 0.000 &
0.958   $\pm$ 0.004 
\\[0.5em]

\multirow{3}{*}{\parbox{1cm}{ResNet50 \\ (25M)}}
& RI &
0.845   $\pm$ 0.022 &
0.919   $\pm$ 0.005 &
0.664   $\pm$ 0.016 &
0.796   $\pm$ 0.000 &
0.948   $\pm$ 0.008 
\\
& ST &
0.848   $\pm$ 0.006 &
0.933   $\pm$ 0.006 &
0.635   $\pm$ 0.012 &
0.794   $\pm$ 0.001 &
0.959   $\pm$ 0.003 
\\
& WT & 
0.888   $\pm$ 0.004 &
0.957   $\pm$ 0.003 &
0.795   $\pm$ 0.011 &
0.800   $\pm$ 0.001 &
0.960   $\pm$ 0.006 
\\[0.5em]

\bottomrule
\end{tabular}
\caption{
\emph{Performance of the models w.r.t different initializations.} 
}
\label{tab:wt_vs_st}
\vspace{-3mm}
\end{table}
\addtolength{\tabcolsep}{3pt}

\vspace{-2mm}
\paragraph{When is transfer learning to medical domains beneficial, and how important is feature reuse?}
To quantify the overall benefit of transfer learning and isolate the contribution of feature reuse, we compare weight transfer (WT), stats transfer (ST), and random initialization (RI).
We also measure the distance between the source domain (\imagenet) and target domains using Fréchet Inception Distance (FID) \cite{fid}.
The results are reported in Table \ref{tab:wt_vs_st} and Figure \ref{fig:feature_reuse}.

The overall trend we observe is the following: the benefits from transfer learning increase with (1) reduced data size, (2) smaller distances between the source and target domain, and (3) models with fewer inductive biases.
We first consider the case where transfer learning is least beneficial: models with strong inductive biases applied to large datasets that poorly resemble \imagenet.
Here, gains from transfer learning are insignificant, \eg for \resnetfifty and \inception applied to \chexpert and \camelyon.
The small benefits we do observe can be largely attributed to the weight statistics (ST), not feature reuse (WT), confirming previous observations \cite{transfusion, neyshabur2020being}.

However, these findings do not carry over to ViTs.
\emph{ViTs appear to benefit far more from feature reuse than CNNs.}
\deit sees a strong boost from transfer learning on \chexpert and \camelyon, wholly attributed to weight transfer, implying strong feature reuse.
\swin, which re-introduces the inductive biases of CNNs, falls somewhere in the middle.
A possible explanation for this behavior is that, owing to \deit's lack of inductive bias, even the largest public medical datasets lack sufficient examples to learn better features than those transferred from \imagenet.

The picture changes when we turn to small datasets.
Here, transfer learning shows noteworthy gains for all models.
However, the strength of the gains and the importance of feature reuse depends on the inductive biases of the model
and  the distance between the domains.
\deit and \swin observe significant gains across the board, strongly attributed to feature reuse.
\resnetfifty and \inception show reasonable gains from transfer learning on \aptos and \ddsm which can be partially attributed to feature reuse.
Finally \isic, the dataset which most closely resembles \imagenet, shows strong benefits for transfer learning and evidence for feature reuse for \emph{all models}.

\begin{figure}[t]
\begin{center}
\begin{tabular}{@{}c@{}c@{}c@{}c@{}c@{}}
    \includegraphics[width=0.20\columnwidth]{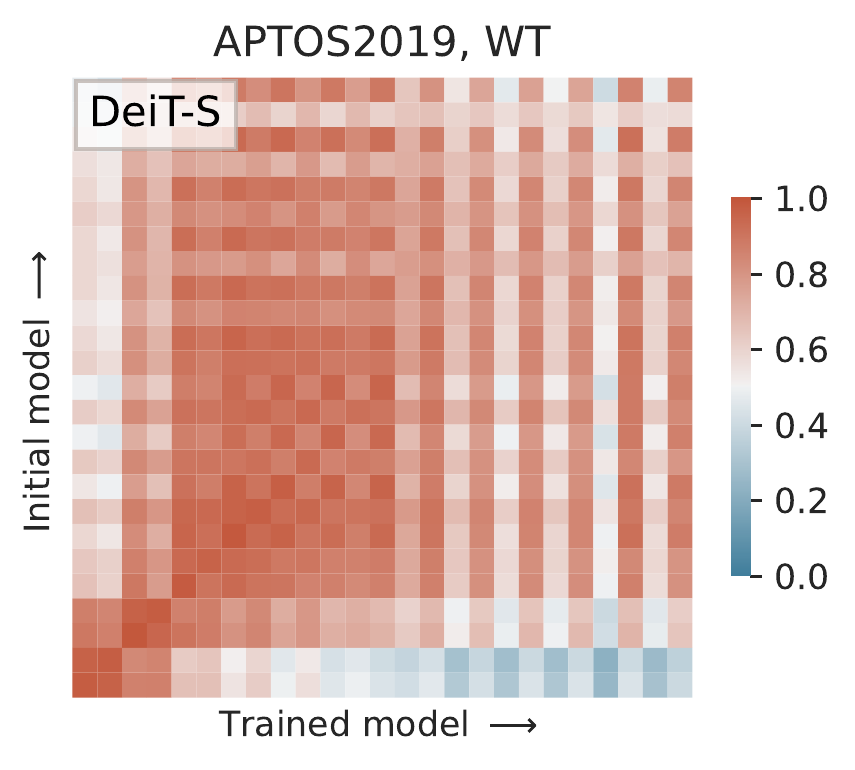} &
    \includegraphics[width=0.20\columnwidth]{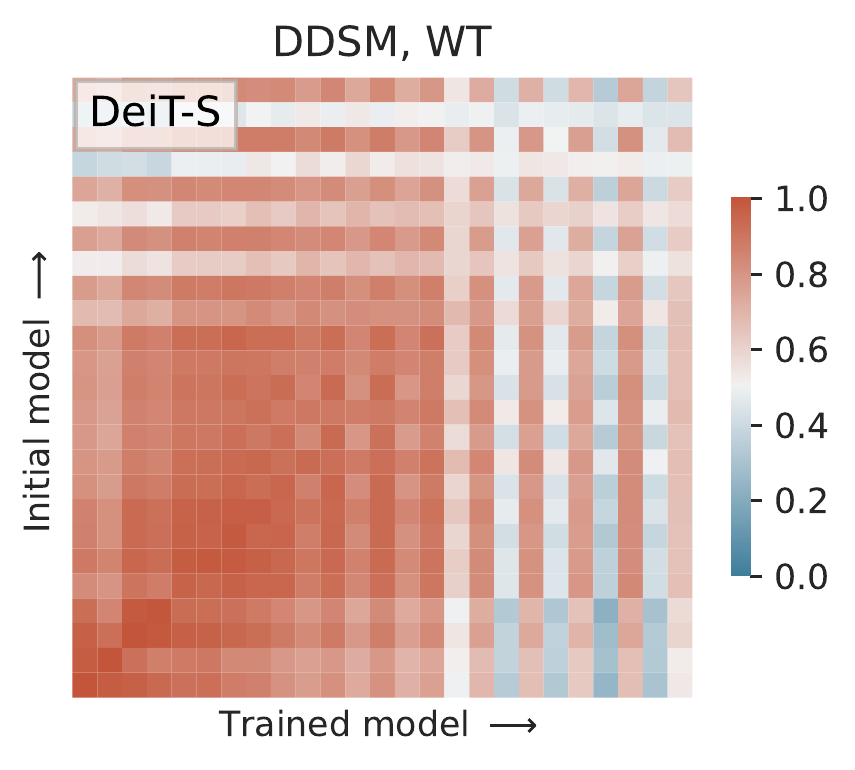} &
    \includegraphics[width=0.20\columnwidth]{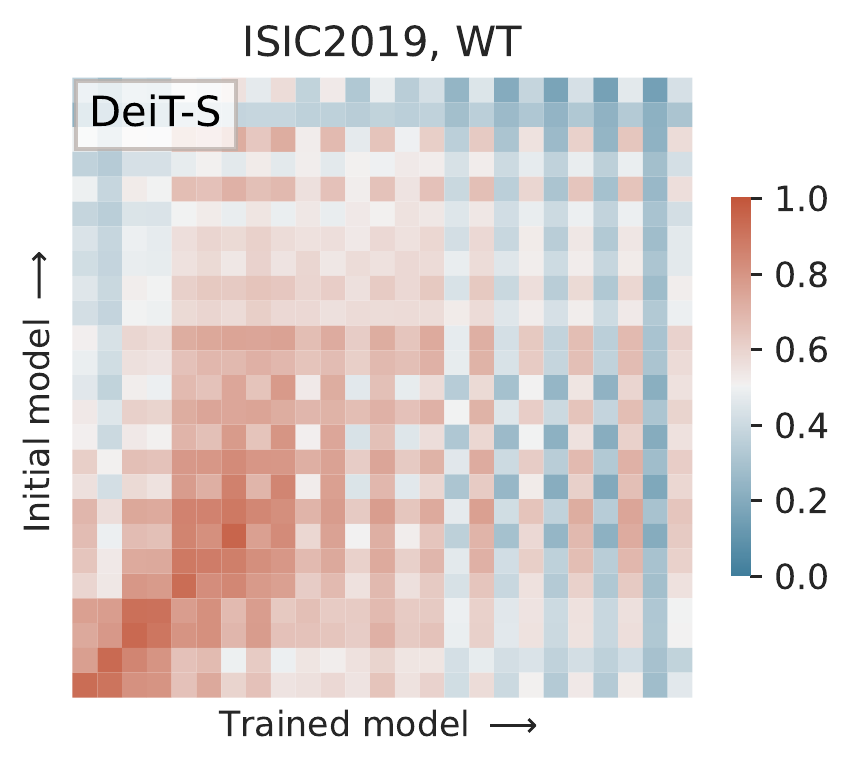} &
    \includegraphics[width=0.20\columnwidth]{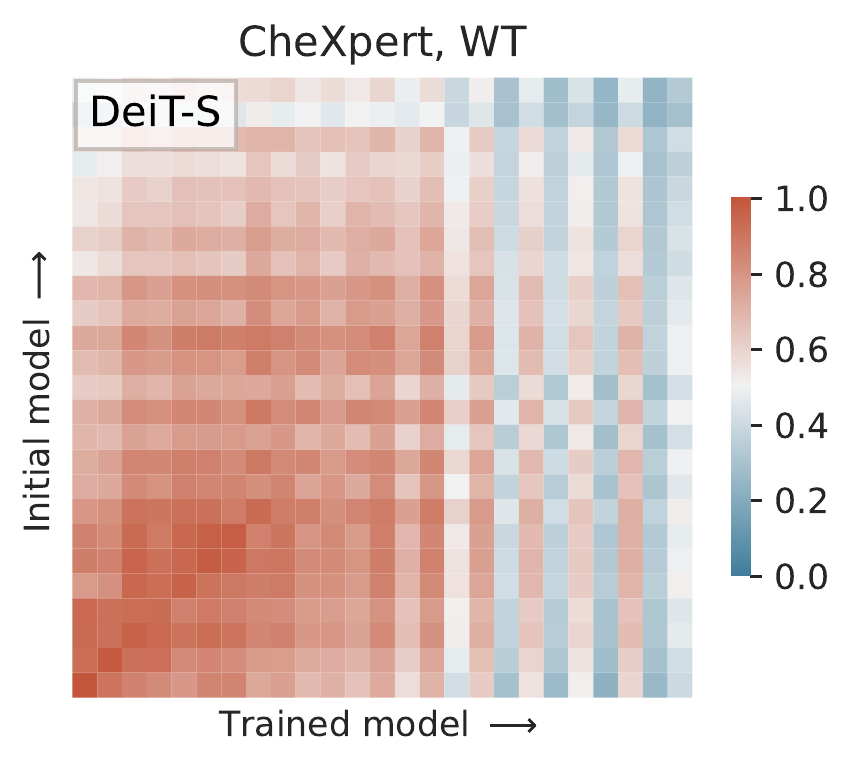} &
    \includegraphics[width=0.20\columnwidth]{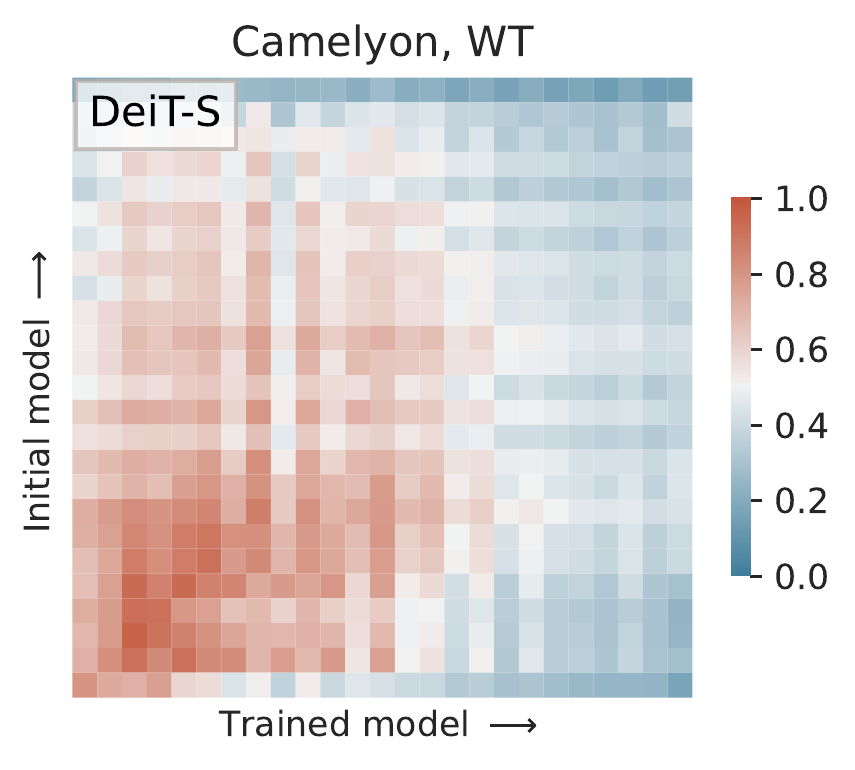}
    \\[-1.5mm] 
    \includegraphics[width=0.20\columnwidth]{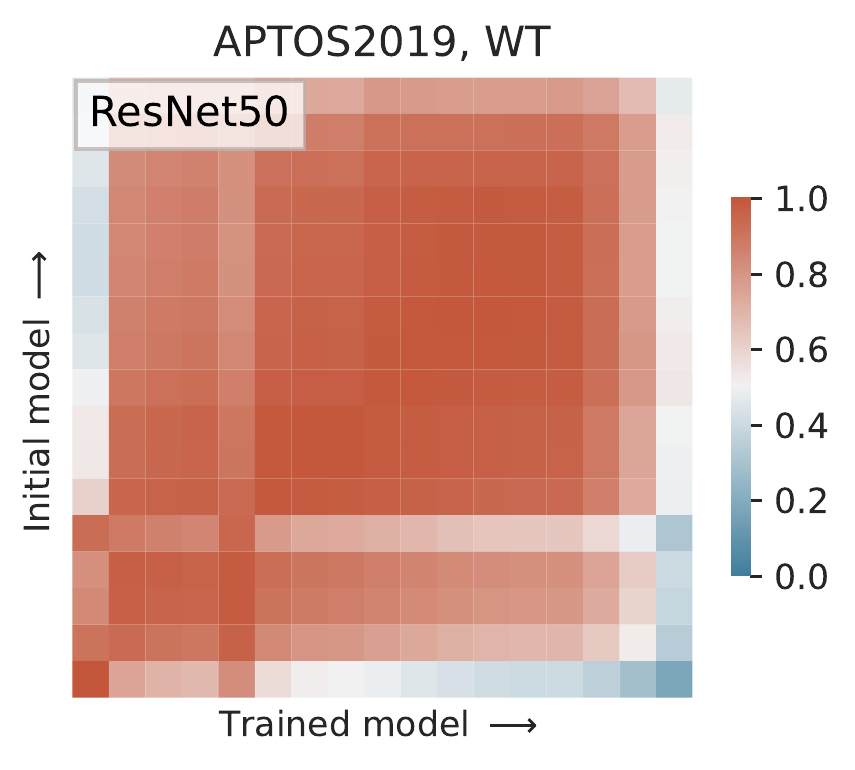} &
    \includegraphics[width=0.20\columnwidth]{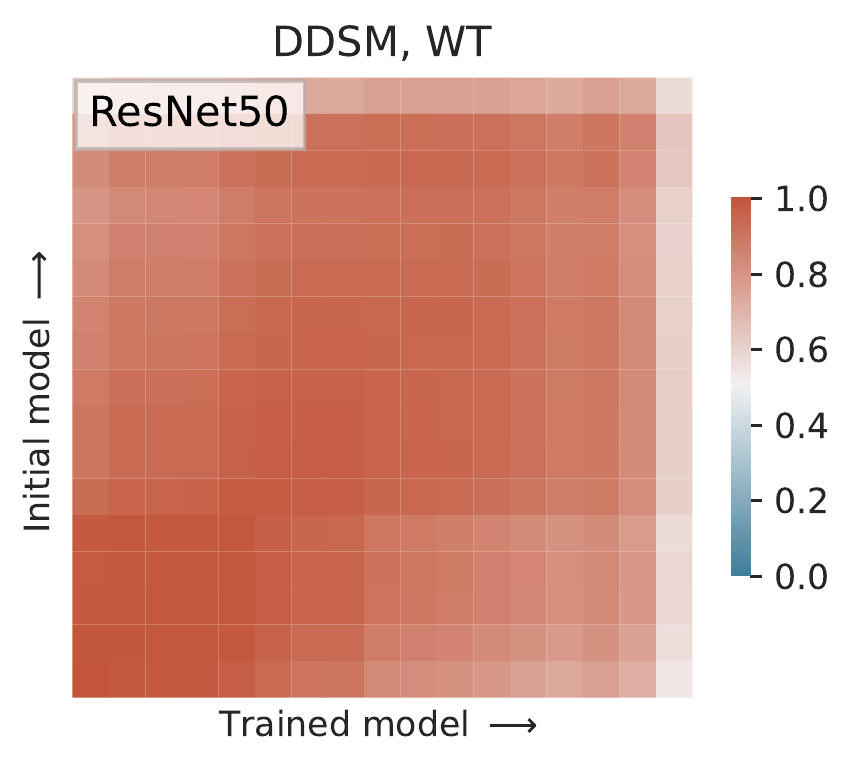} &
    \includegraphics[width=0.20\columnwidth]{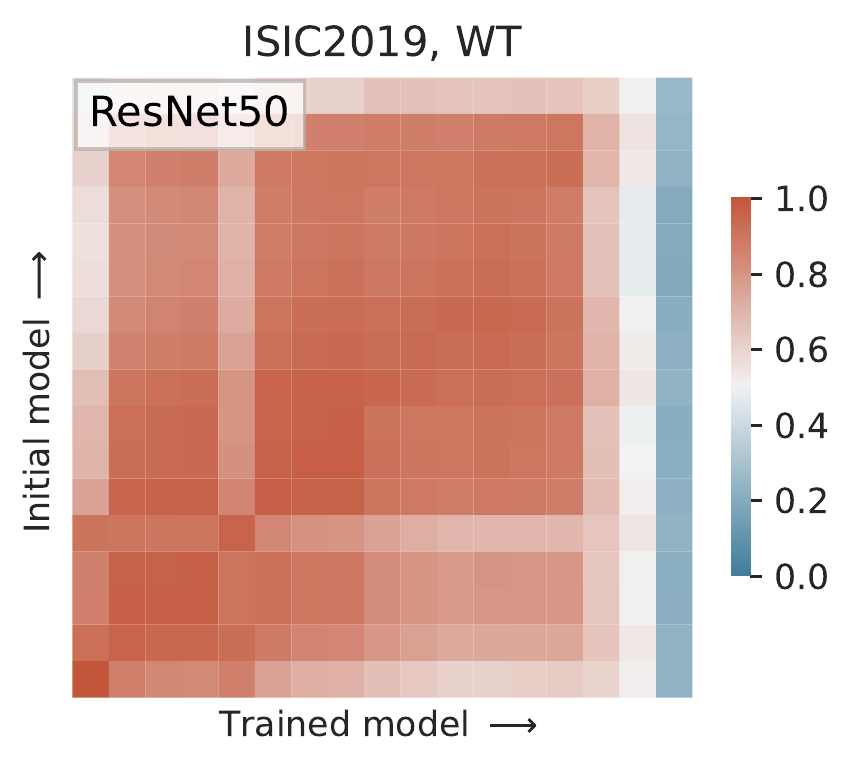} &
    \includegraphics[width=0.20\columnwidth]{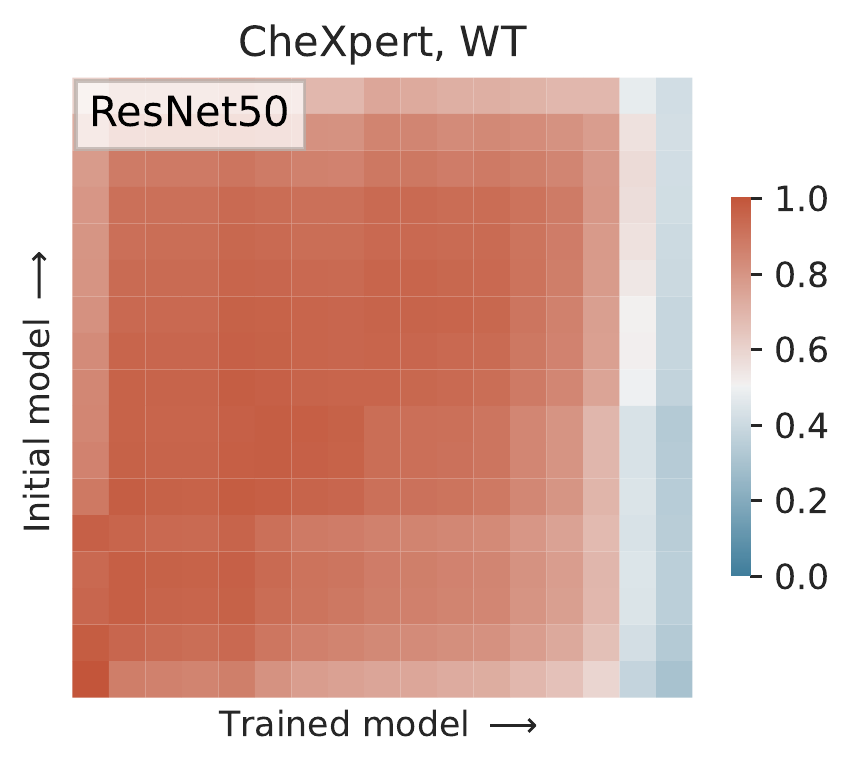} &
    \includegraphics[width=0.20\columnwidth]{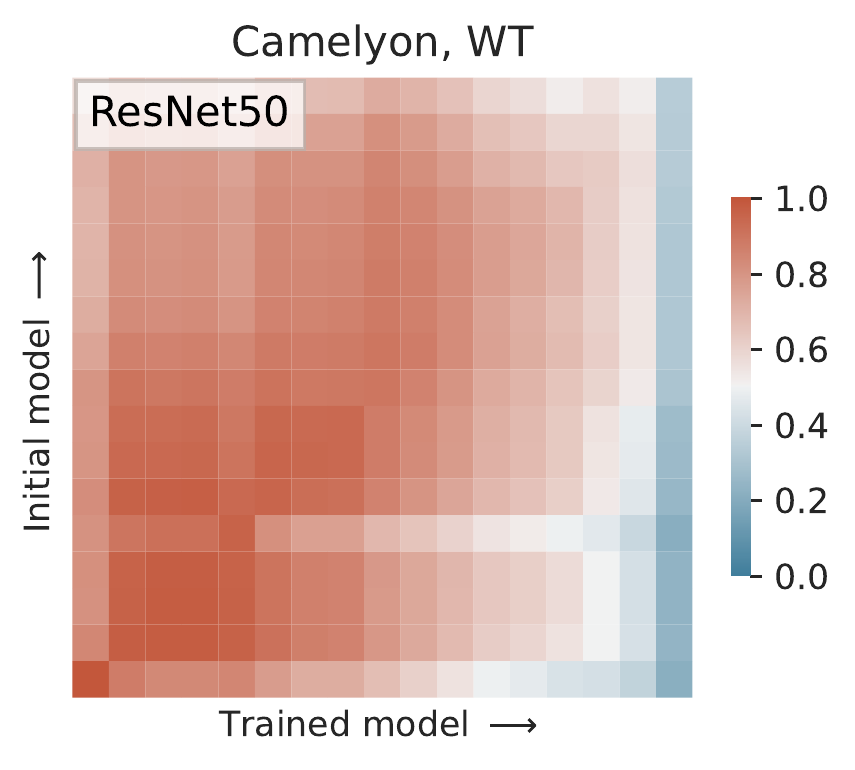}
    \\[-1.5mm] 
    \midrule
    \includegraphics[width=0.20\columnwidth]{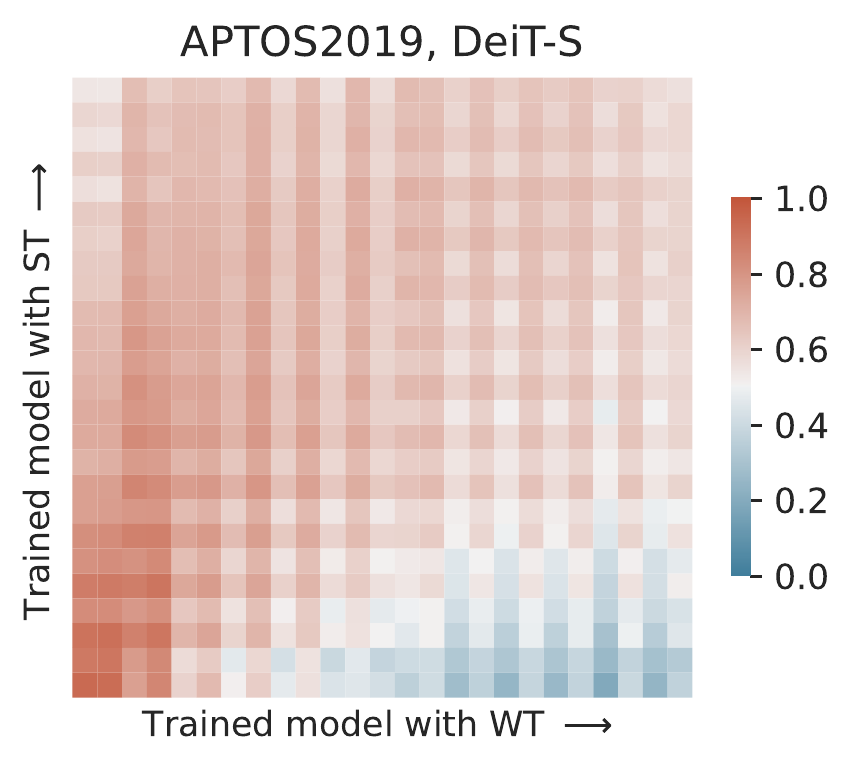} &
    \includegraphics[width=0.20\columnwidth]{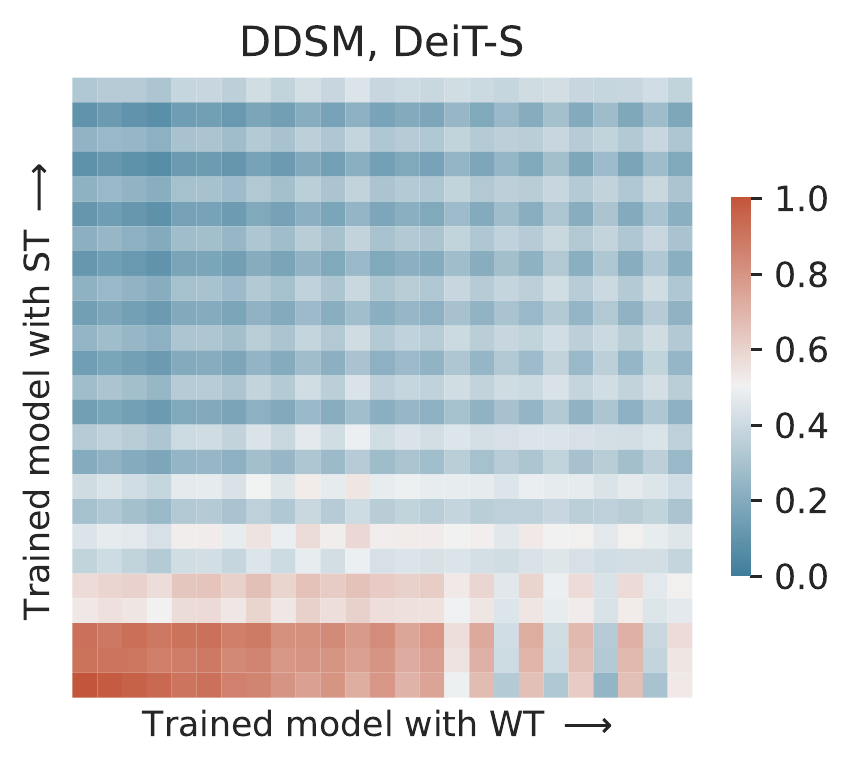} &
    \includegraphics[width=0.20\columnwidth]{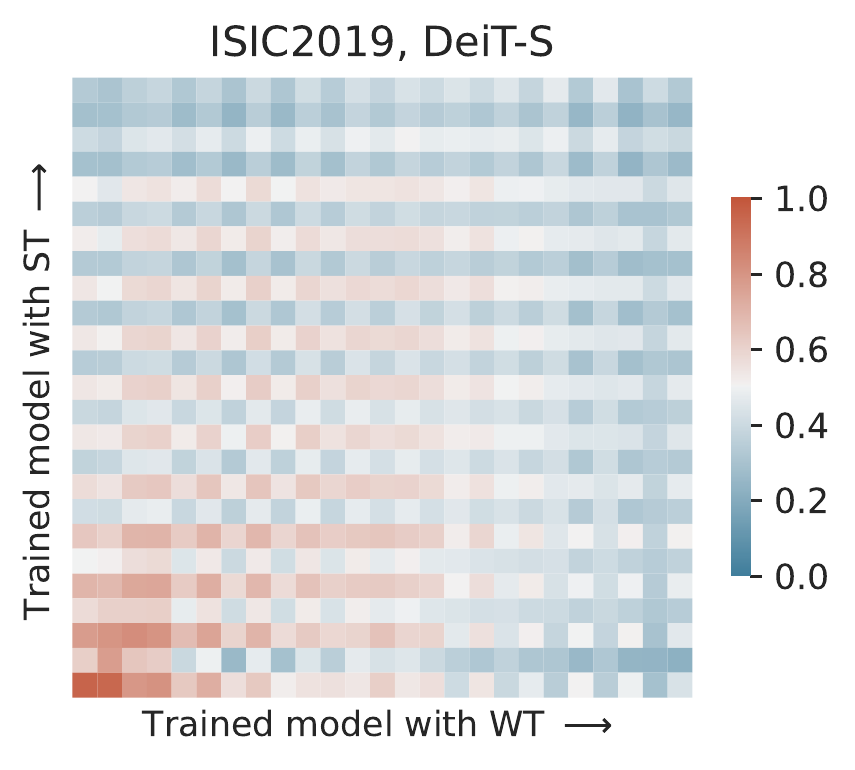} &
    \includegraphics[width=0.20\columnwidth]{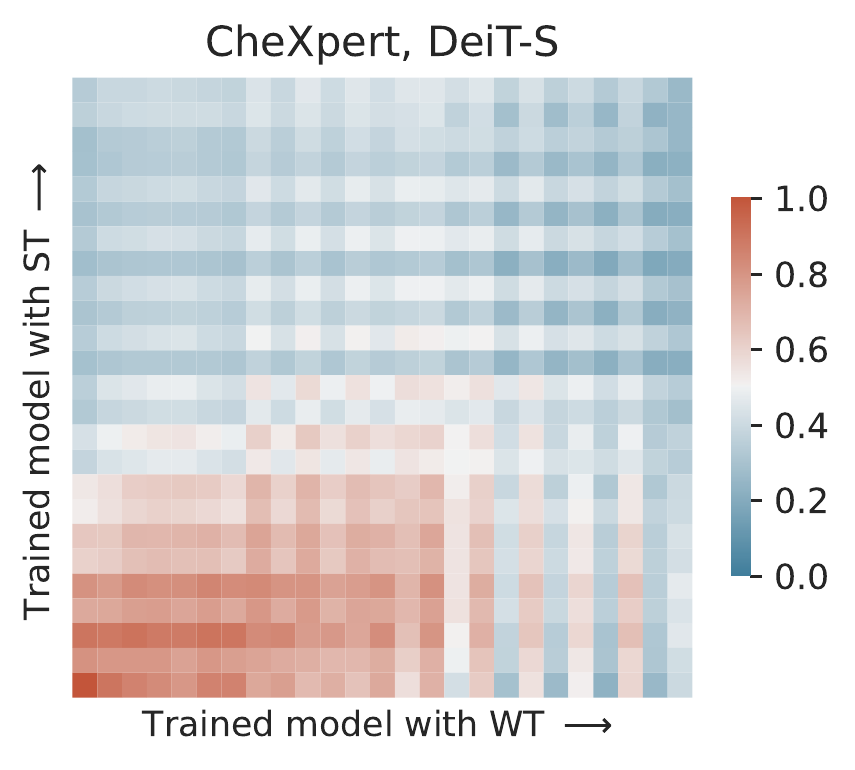} &
    \includegraphics[width=0.20\columnwidth]{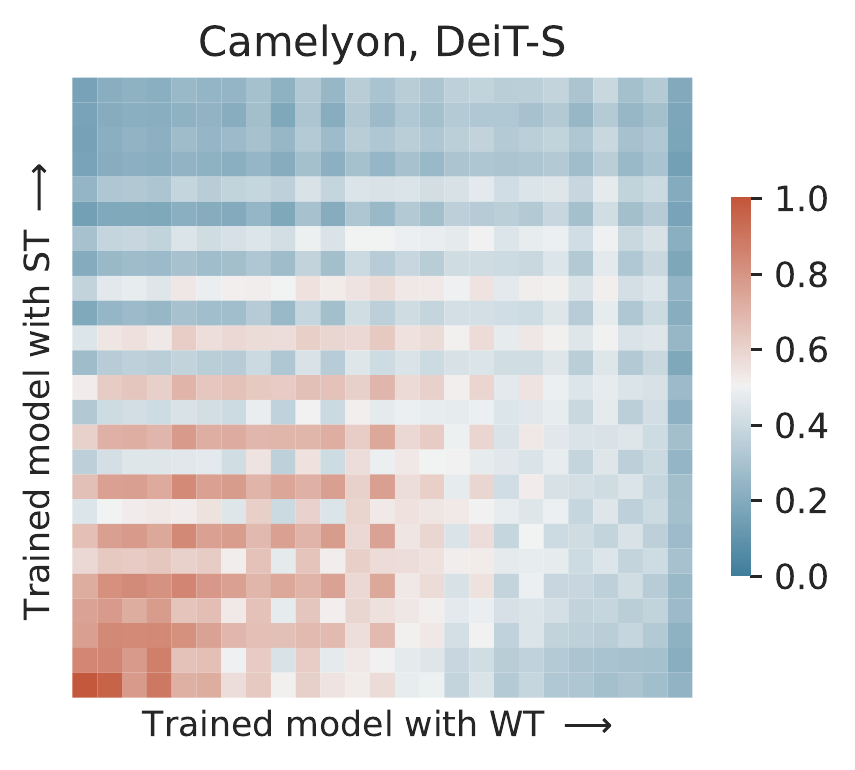}
    \\[-1.5mm]
    \includegraphics[width=0.20\columnwidth]{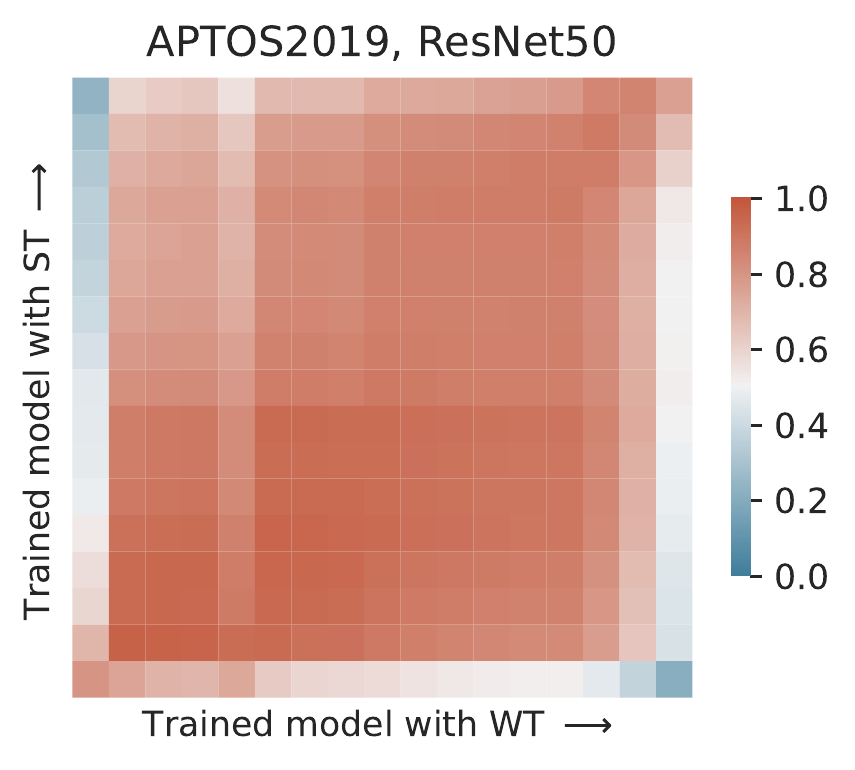} &
    \includegraphics[width=0.20\columnwidth]{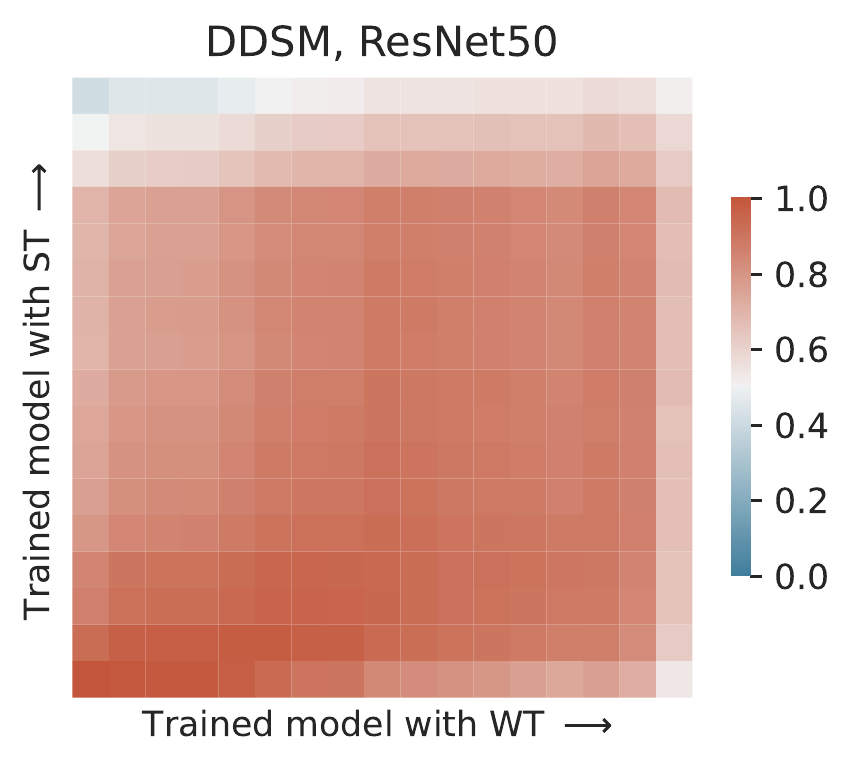} &
    \includegraphics[width=0.20\columnwidth]{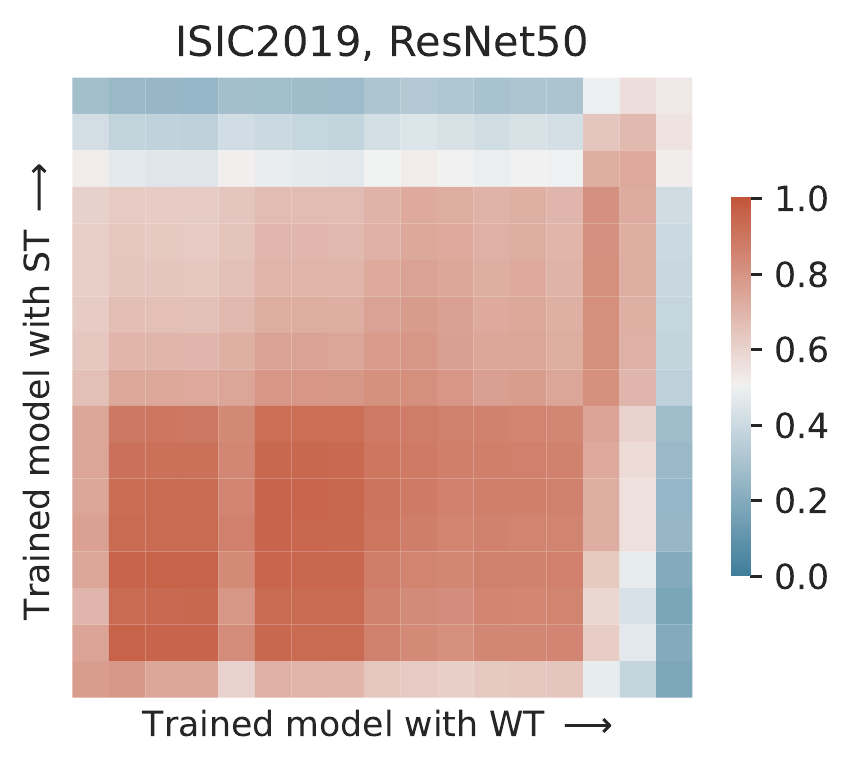} &
    \includegraphics[width=0.20\columnwidth]{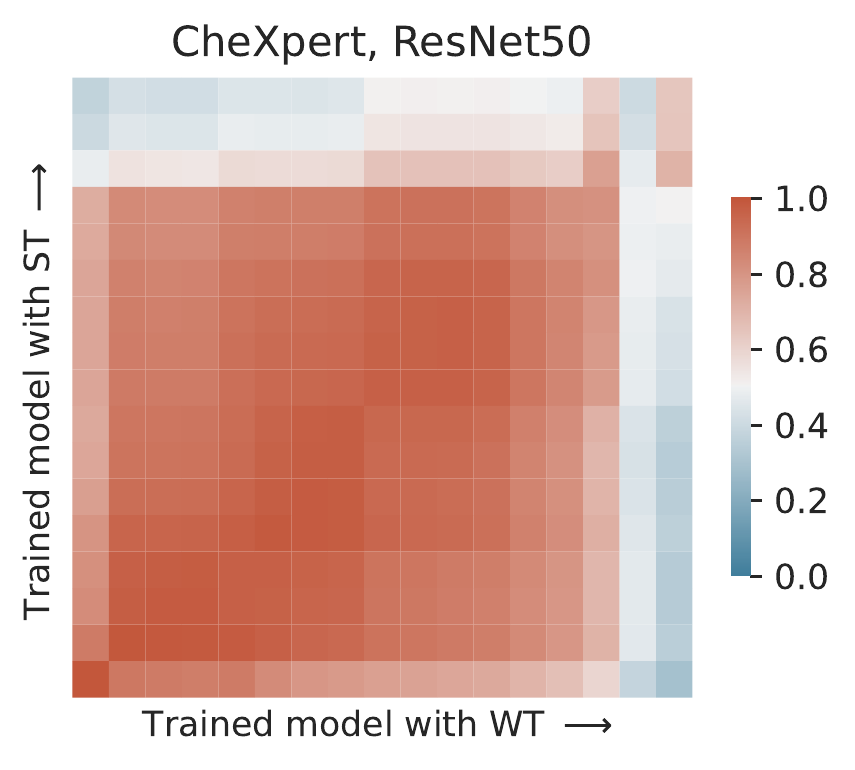} &
    \includegraphics[width=0.20\columnwidth]{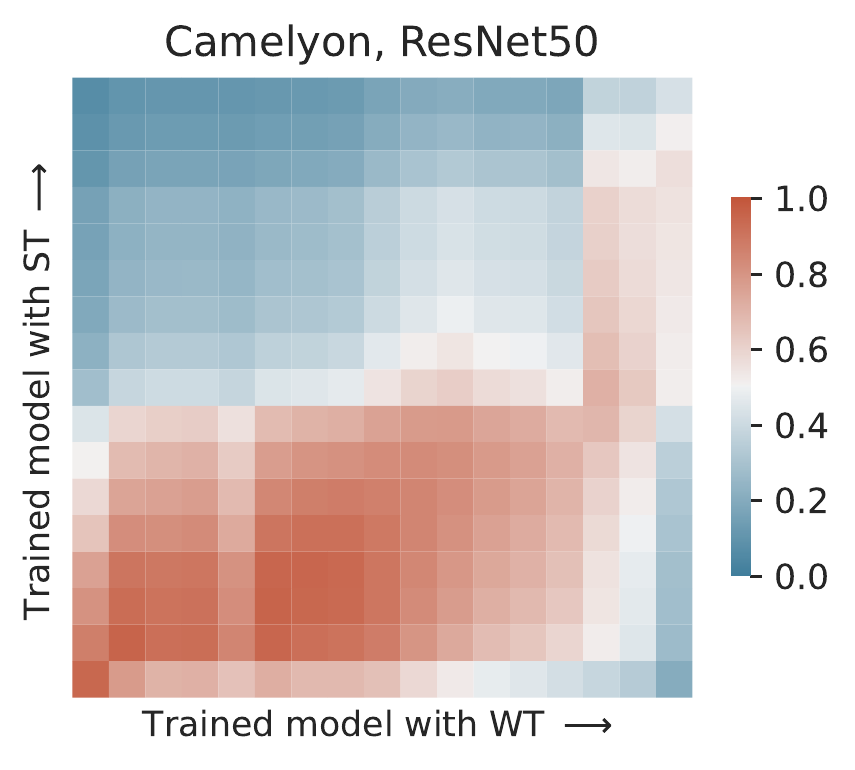} 
    \\[-1.5mm]        
\end{tabular}
\end{center}
\vspace{-3mm}
\caption{\emph{Layer-wise feature similarity using CKA.} \textbf{top:} The CKA representational similarity as a function of model depth for WT initialized {\deitsmall} and {\resnetfifty}, before and after fine-tuning. \textbf{bottom:} Feature similarity between ST and WT initialized models after fine-tuning. See text for details. Full results appear in Appendix \ref{sec:sup-figures-feature-similarity}.}
\label{fig:similarity}
\vspace{-4mm}
\end{figure}

\vspace{-2mm}
\paragraph{Which layers benefit from feature reuse?}
We investigate where feature reuse occurs within the network by transferring weights (WT) up to block $n$ and initializing the remaining $m$ blocks using ST.
The results appear in Figure \ref{fig:wst_all}.
Here, we see distinctive trends revealing the differences between CNNs and ViTs.
On large datasets, CNNs exhibit a relatively flat line indicating that, throughout the network, weight transfer (WT) offers no benefit over the statistics (ST).
Here, most of the benefits of transfer learning come from the statistics, not feature reuse.
For smaller datasets, CNNs show a linear trend implying that every layer sees some modest benefit from feature reuse.
\deit shows a markedly different trend across all datasets -- a sharp jump in performance in early layers -- indicating a strong dependence on feature reuse in these layers.
This fits with previous works that have shown that local attention, which is crucial for good performance, is learned in early layers
\cite{dosovitskiy2020image, caron2021emerging}.
The importance of early layers we observe might be attributed to reuse of these local features which require huge amounts of data to learn \cite{raghu2021vision}.
\swin exhibits properties of both \deit and the CNNs, reflecting its mixture of inductive biases.
On small datasets and those similar to \imagenet \swin closely mirrors \deit, but shows trends resembling a CNN with enough data.
General inductive bias trends can be seen comparing models in the last panel of Figure \ref{fig:wst_all} which shows the average relative gains.
For ViTs, fewer inductive biases necessitates extensive feature reuse but concentrated in the early layers.
CNNs benefit from reused features to a lesser extent, but more consistently throughout the network, reflecting the hierarchical nature of the architecture. 

To summarize the findings thus far: the benefits of transfer learning are tied to feature reuse, and depend on the size of the dataset, proximity to \imagenet, and the model's inductive biases.
Next, we look for further evidence of feature reuse through different perspectives.

\begin{figure}[t]
\begin{center}
\begin{tabular}{@{}c@{}c@{}c@{}c@{}}
    \includegraphics[width=0.25\columnwidth]{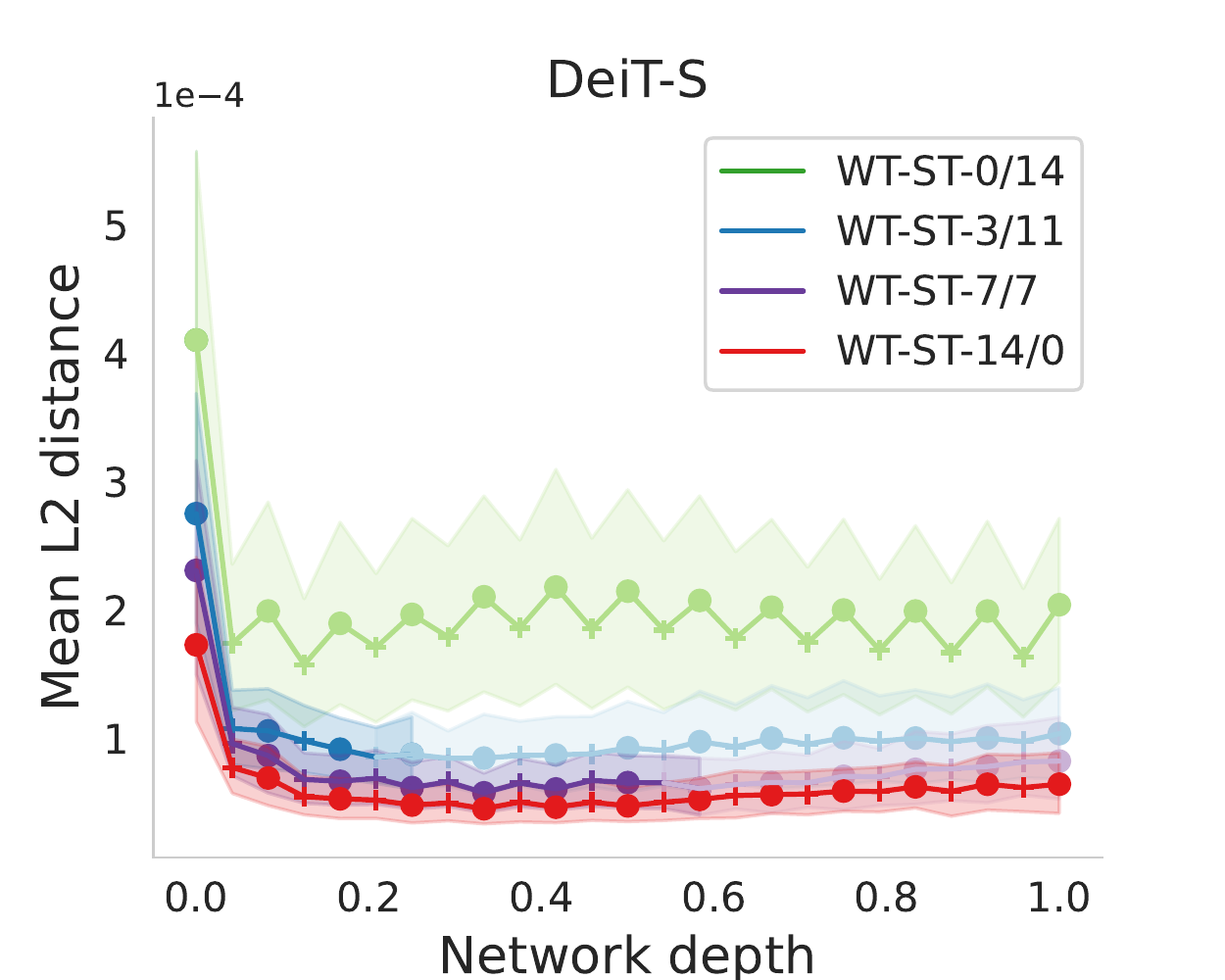} & 
    \includegraphics[width=0.25\columnwidth]{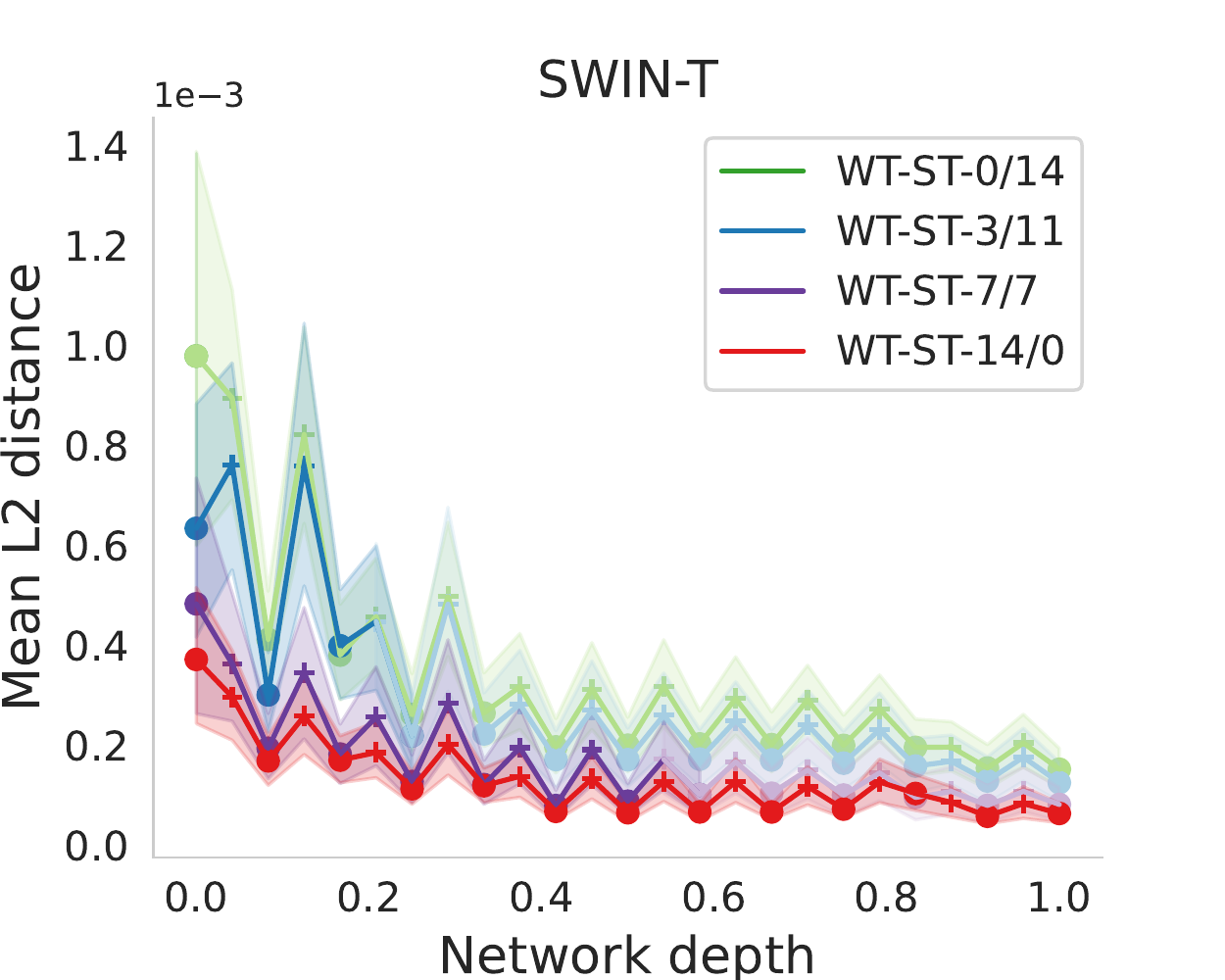} &
    \includegraphics[width=0.25\columnwidth]{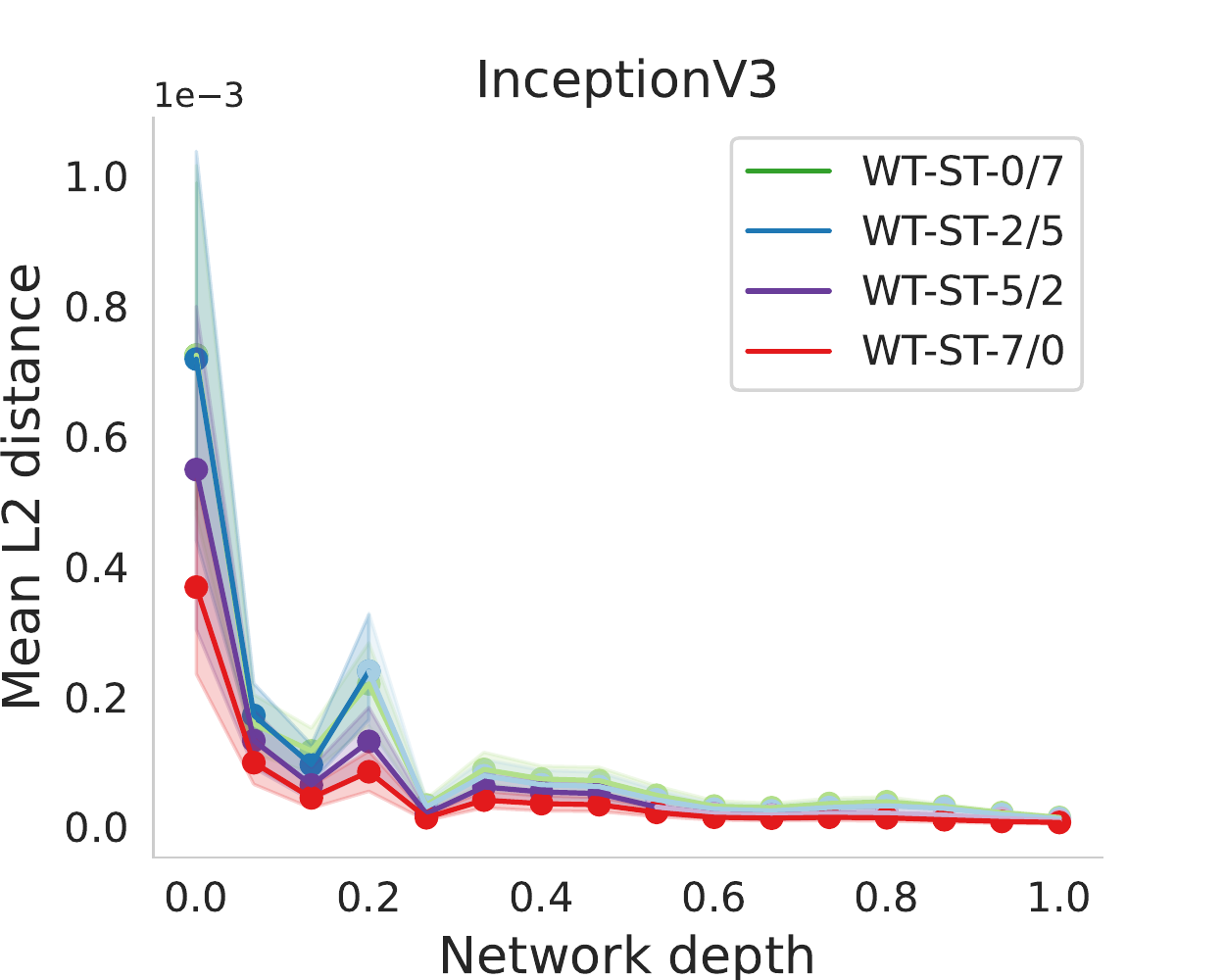} &
    \includegraphics[width=0.25\columnwidth]{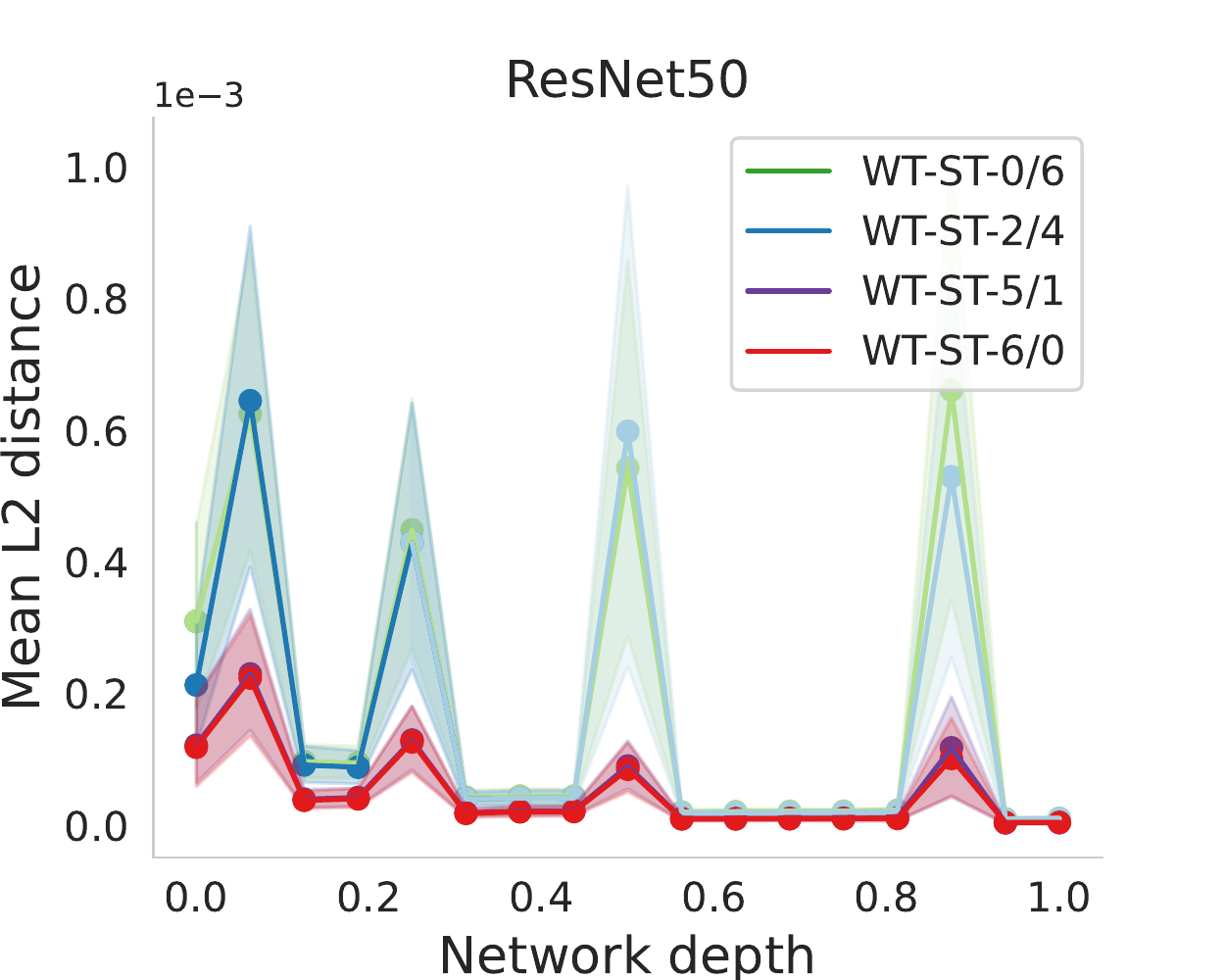}\\[-1.5mm] 
\end{tabular}
\end{center}
\vspace{-3mm}
\caption{\emph{\ltwo distance of weights before and after fine-tuning.}
We report the the mean \ltwo distances between the initial and trained weights for different WT-ST initialization schemes, averaged over all datasets. Increased distances indicate that during training, the network larger changes the layer weights. More results can be found in Figure \ref{fig:L2_appdx} in Appendix \ref{appdx:L2experiment}.}

\label{fig:L2_dist}
\vspace{-3mm}
\end{figure}

\begin{figure}[t]
\begin{center}
\begin{tabular}{@{}c@{}c@{}c@{}c@{}c@{}}
    \includegraphics[width=0.20\columnwidth]{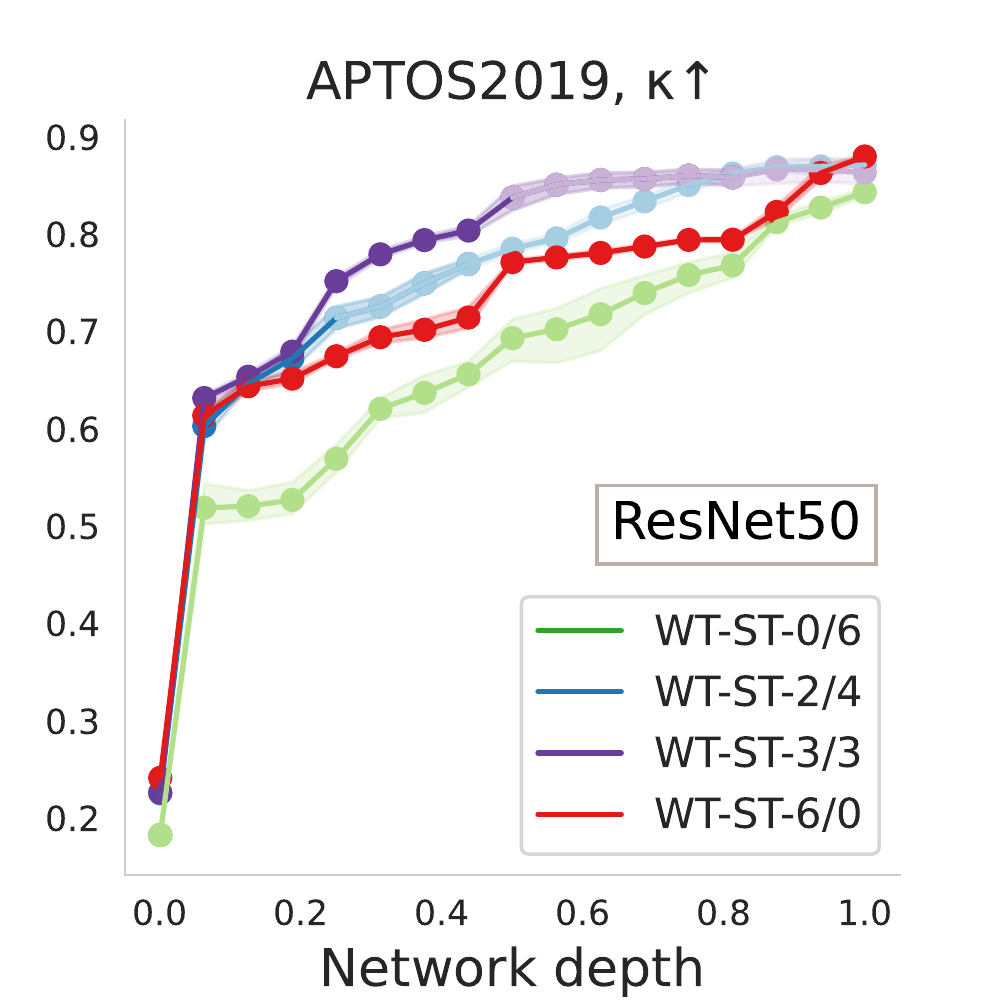} & 
    \includegraphics[width=0.20\columnwidth]{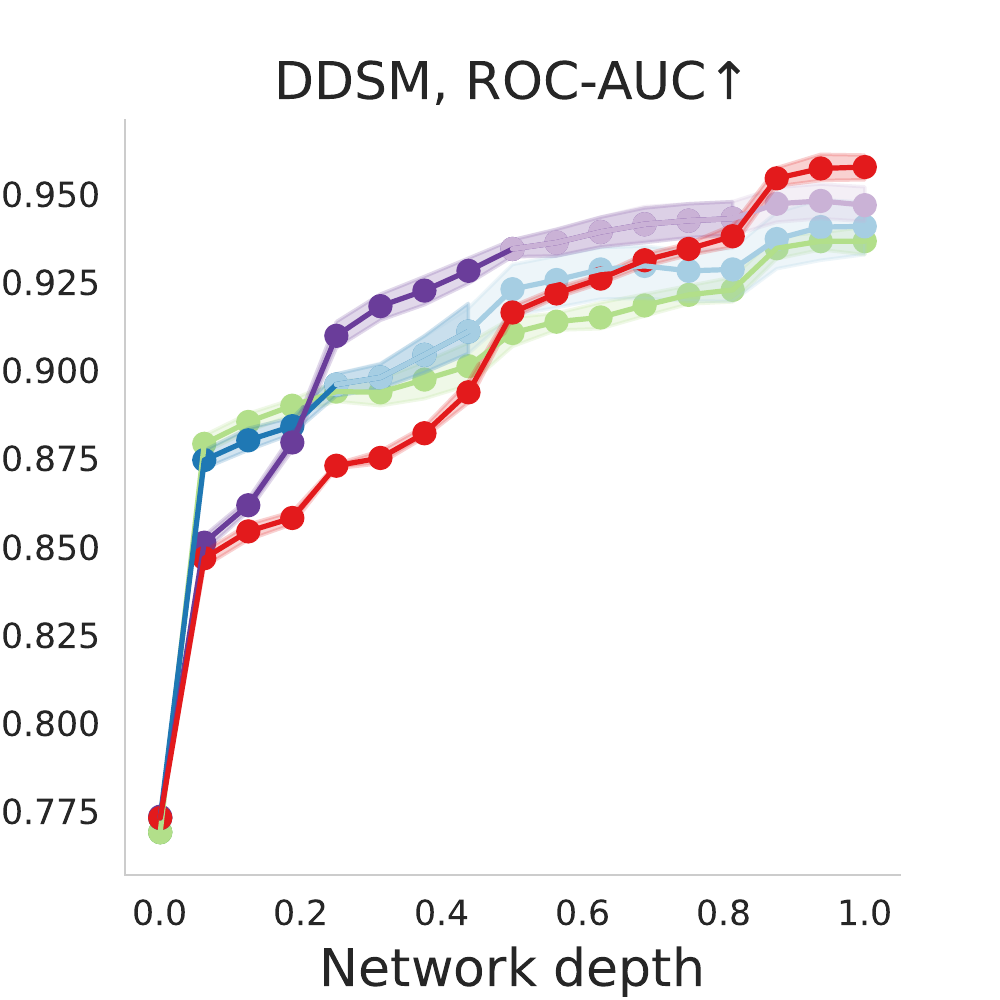} & 
    \includegraphics[width=0.20\columnwidth]{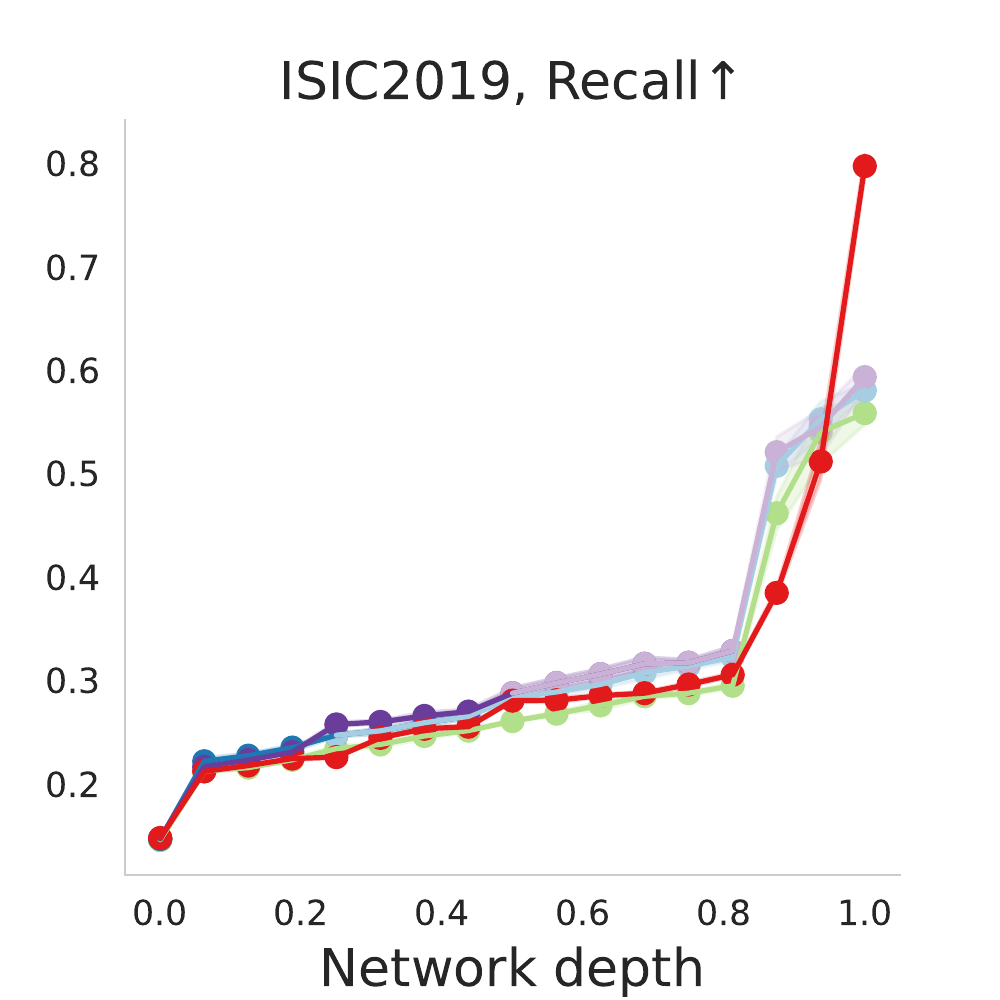} & 
    \includegraphics[width=0.20\columnwidth]{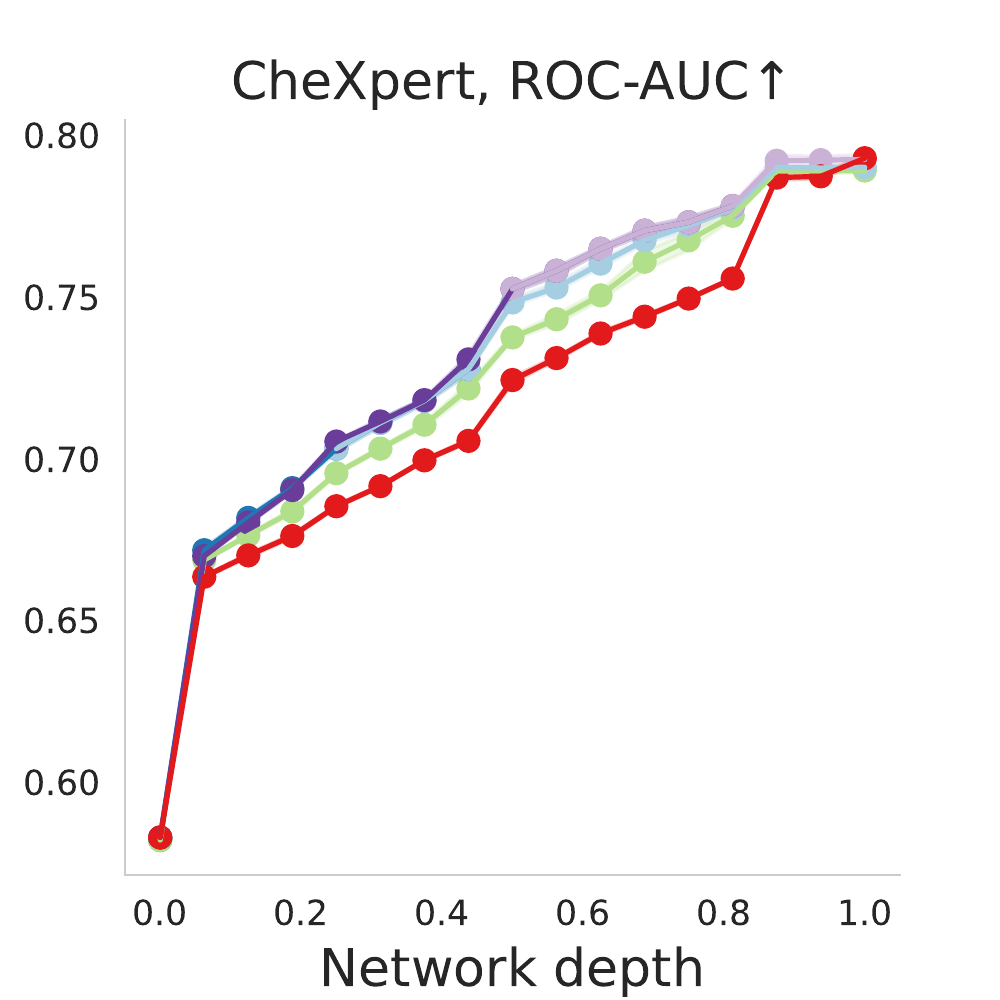} & 
    \includegraphics[width=0.20\columnwidth]{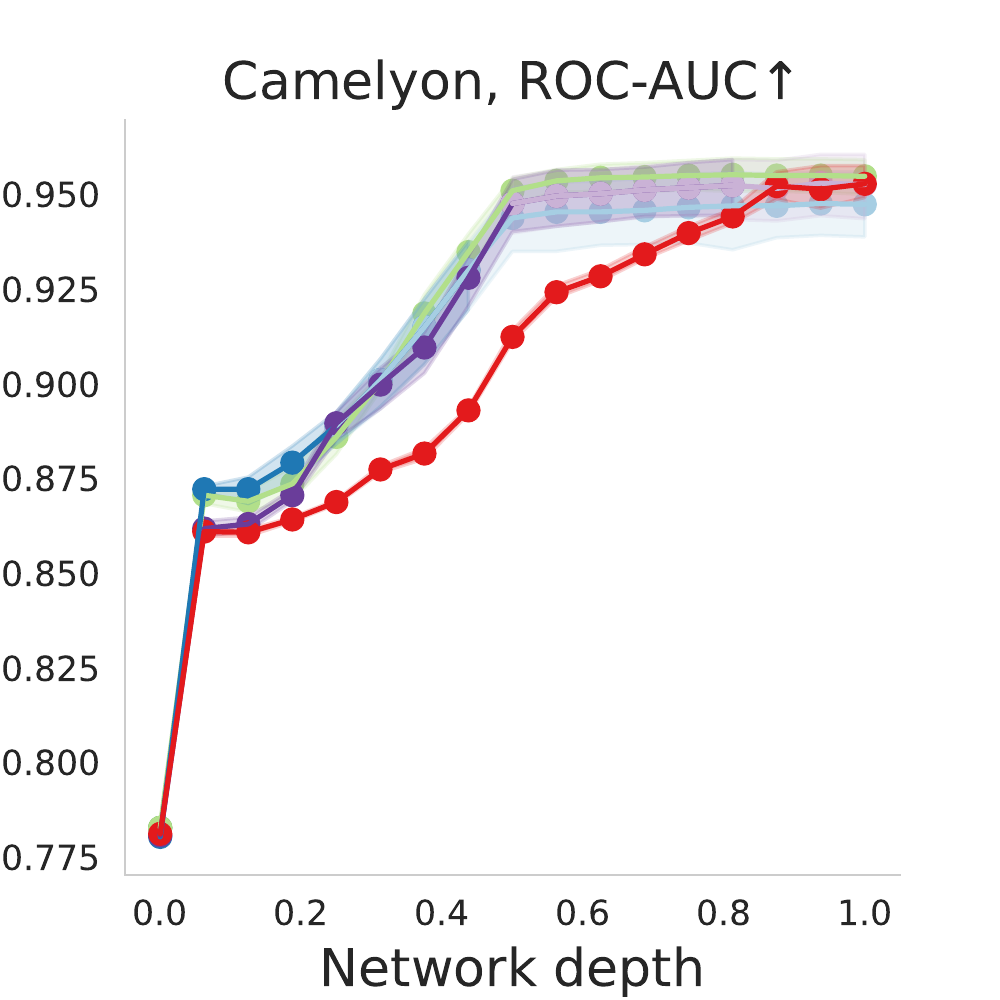}  
    \\[-1.5mm] 
    \includegraphics[width=0.20\columnwidth]{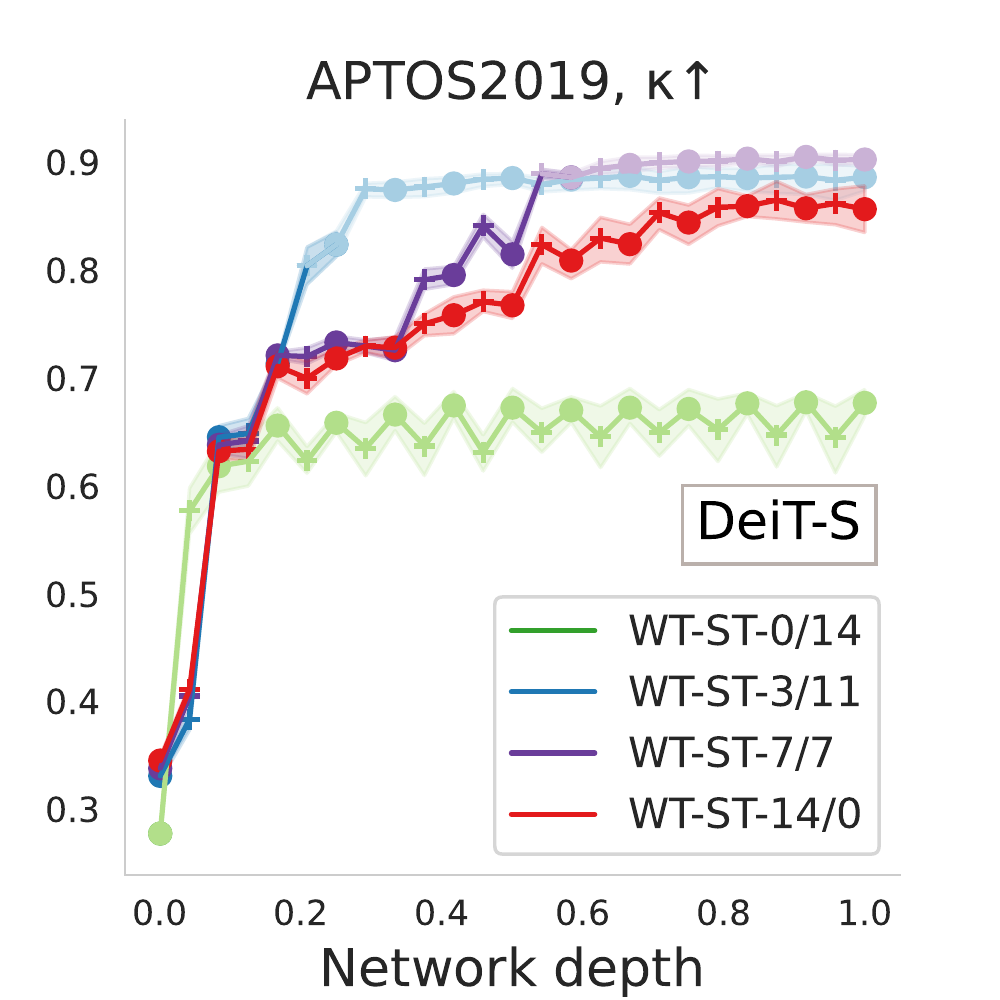} & 
    \includegraphics[width=0.20\columnwidth]{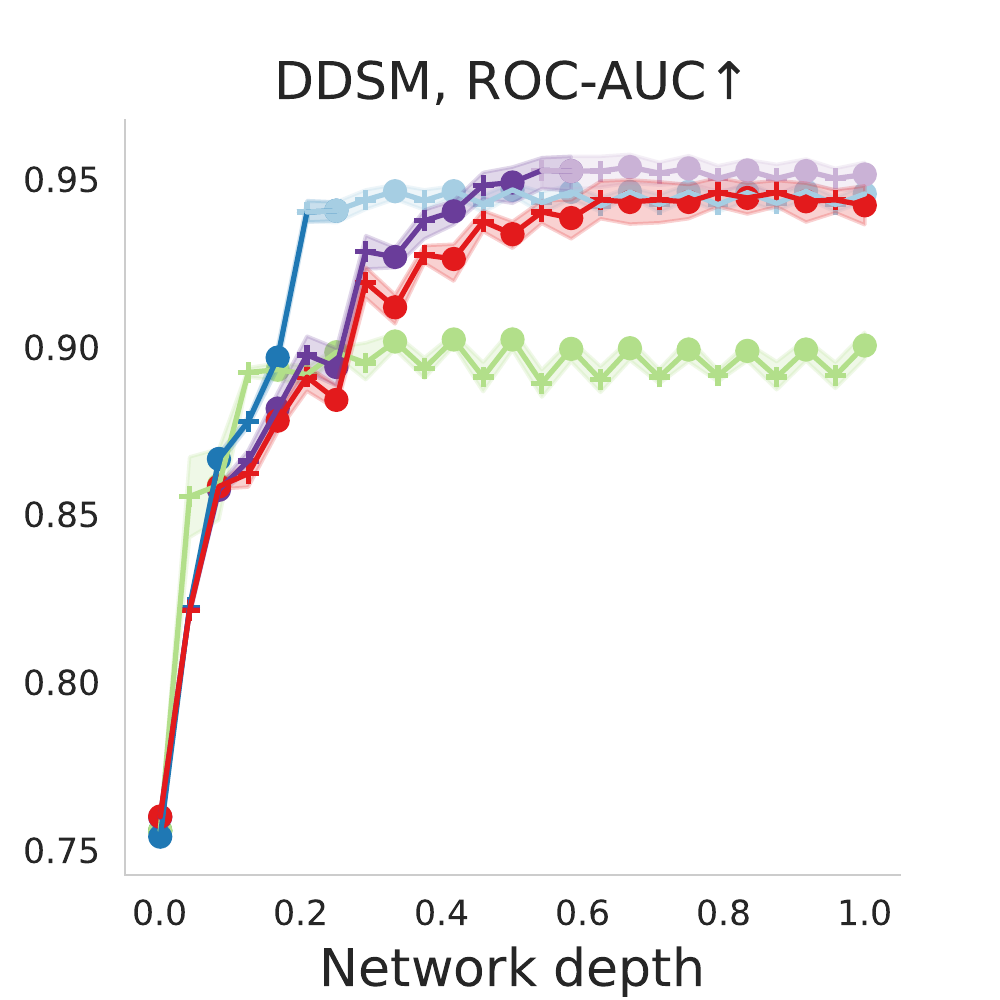} & 
    \includegraphics[width=0.20\columnwidth]{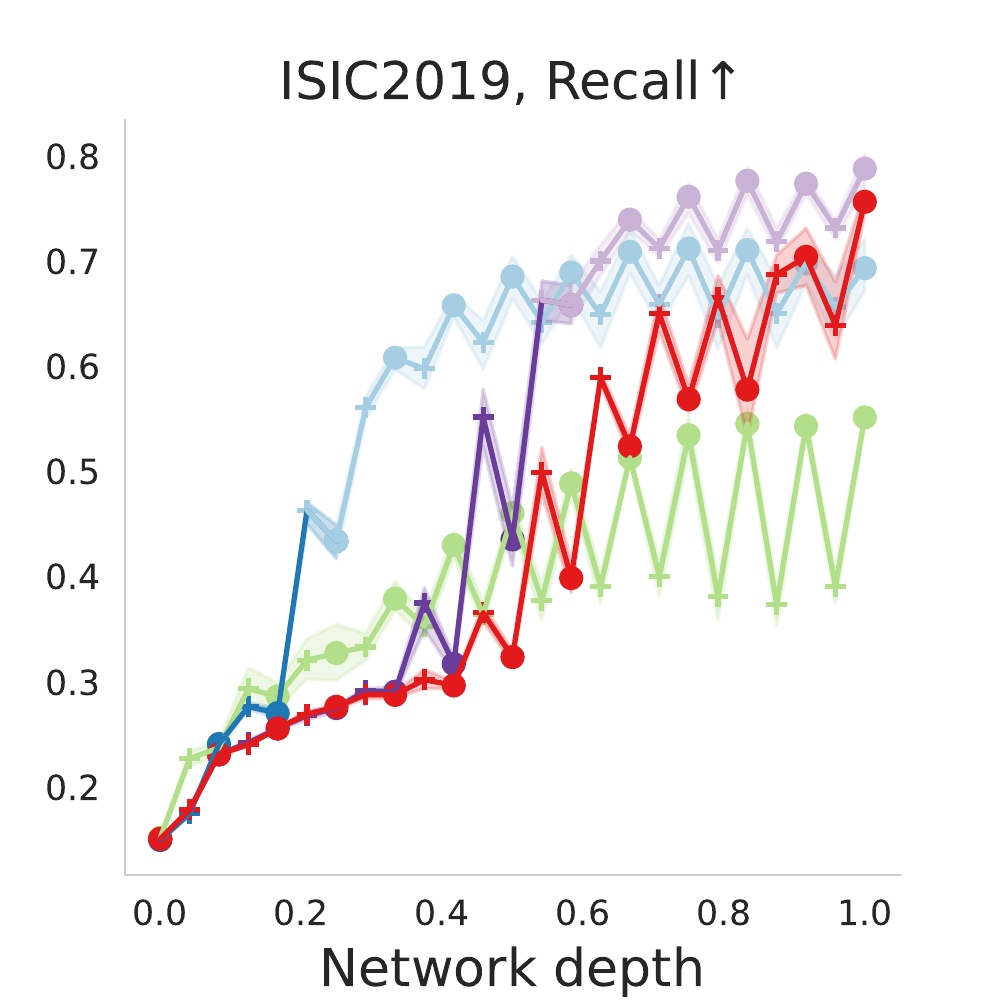} &     
    \includegraphics[width=0.20\columnwidth]{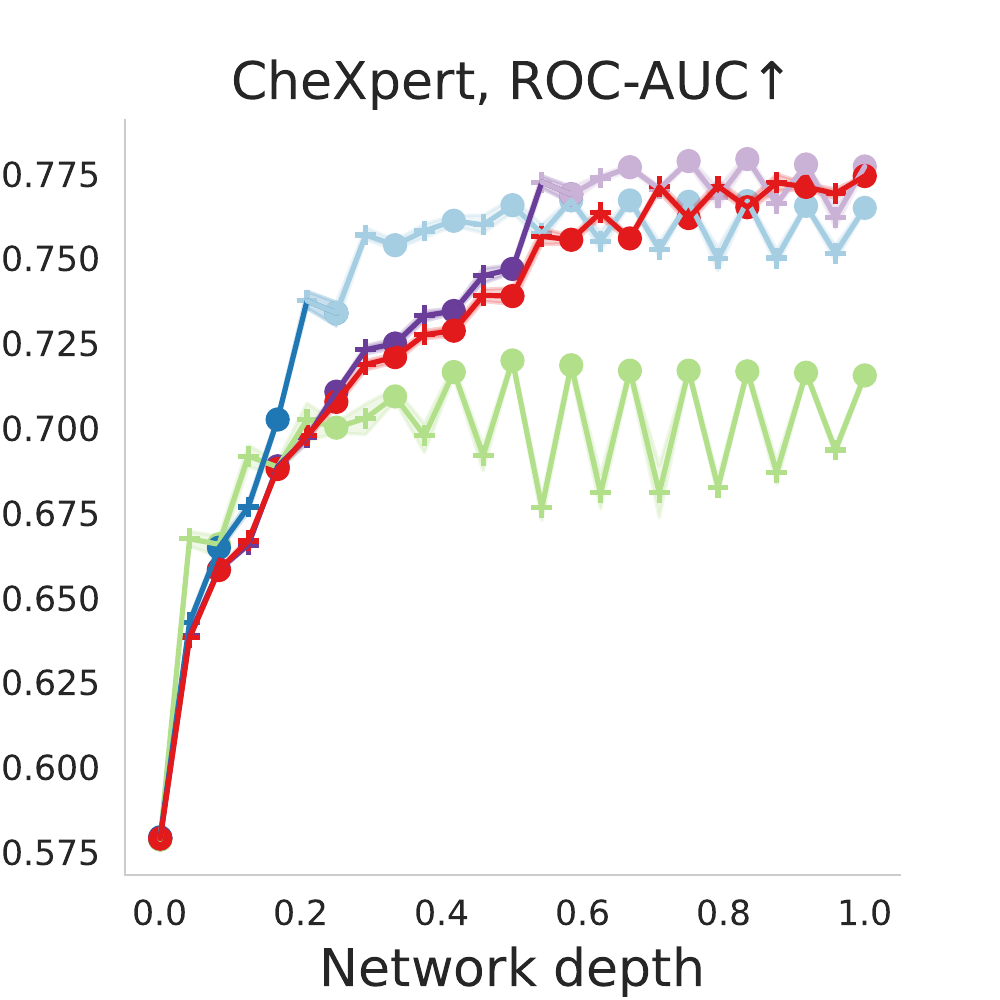} &   
    \includegraphics[width=0.20\columnwidth]{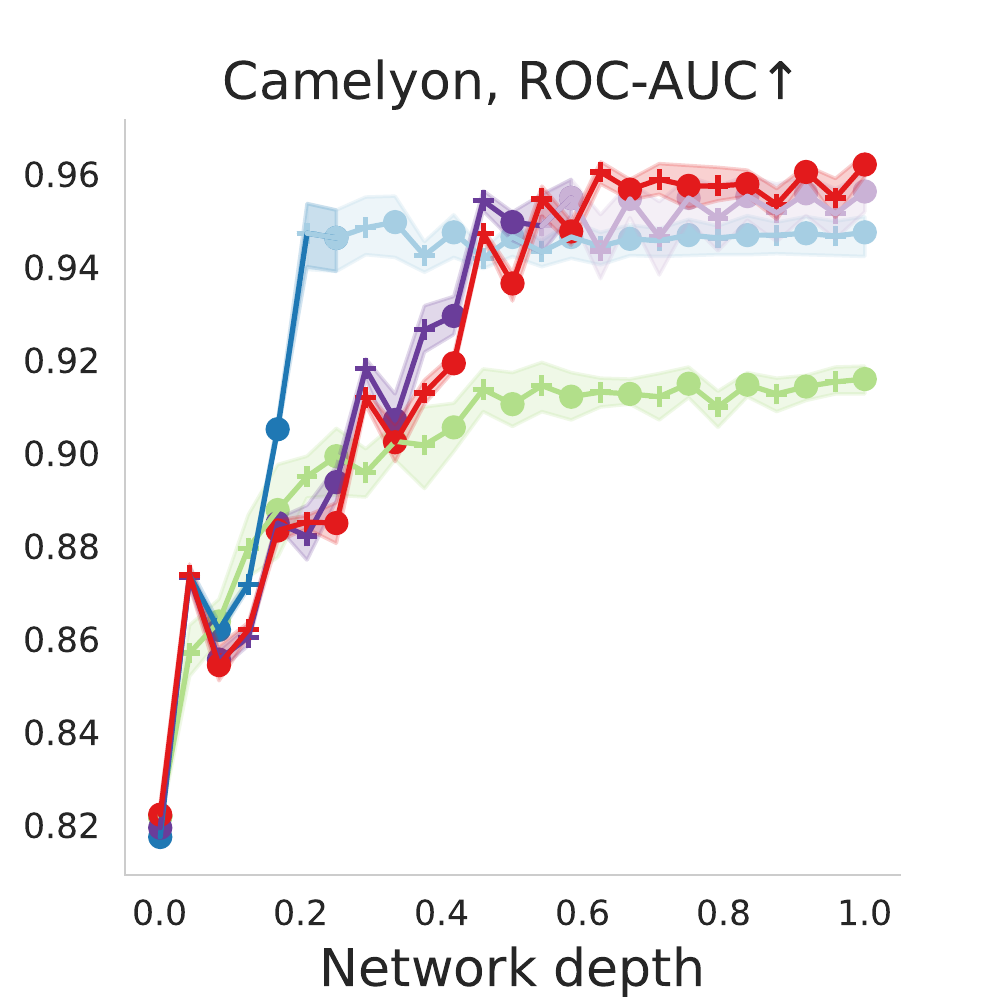}      
    \\[-1.5mm] 
    \midrule
    \includegraphics[width=0.20\columnwidth]{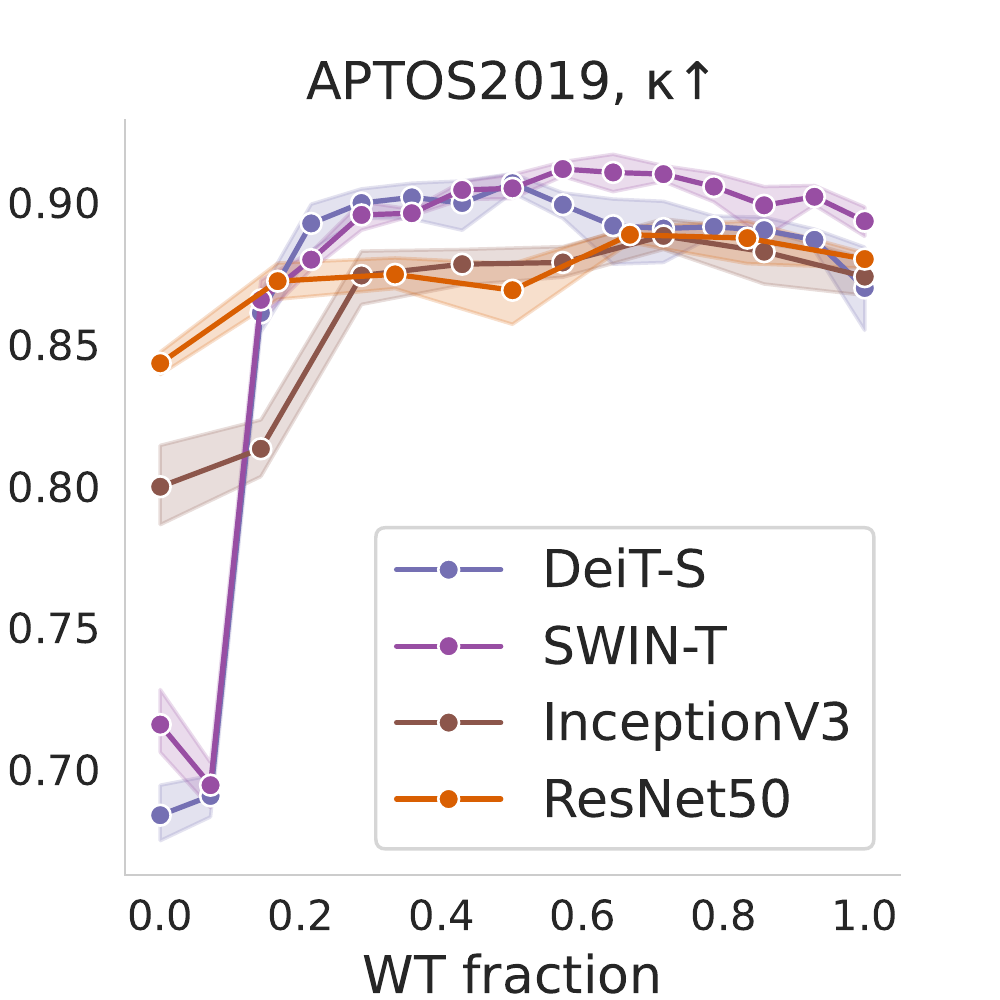} & 
    \includegraphics[width=0.20\columnwidth]{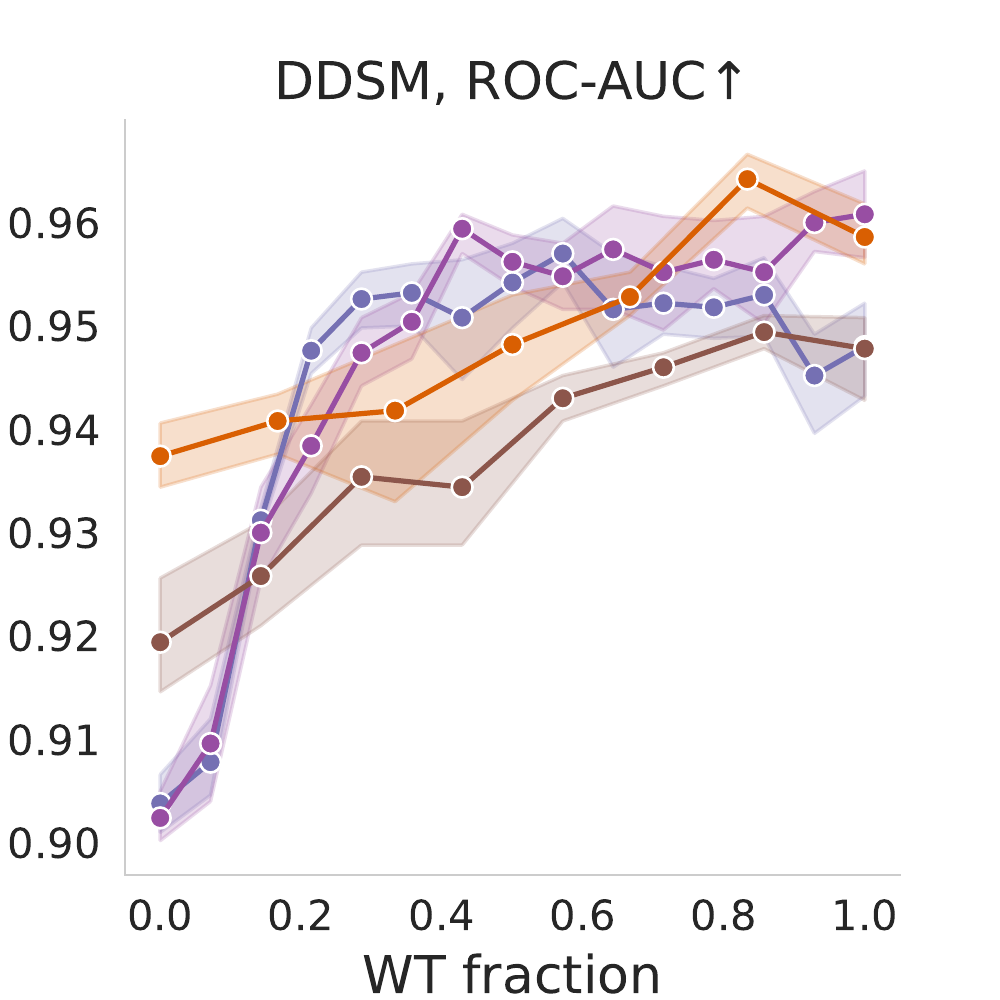} &    
    \includegraphics[width=0.20\columnwidth]{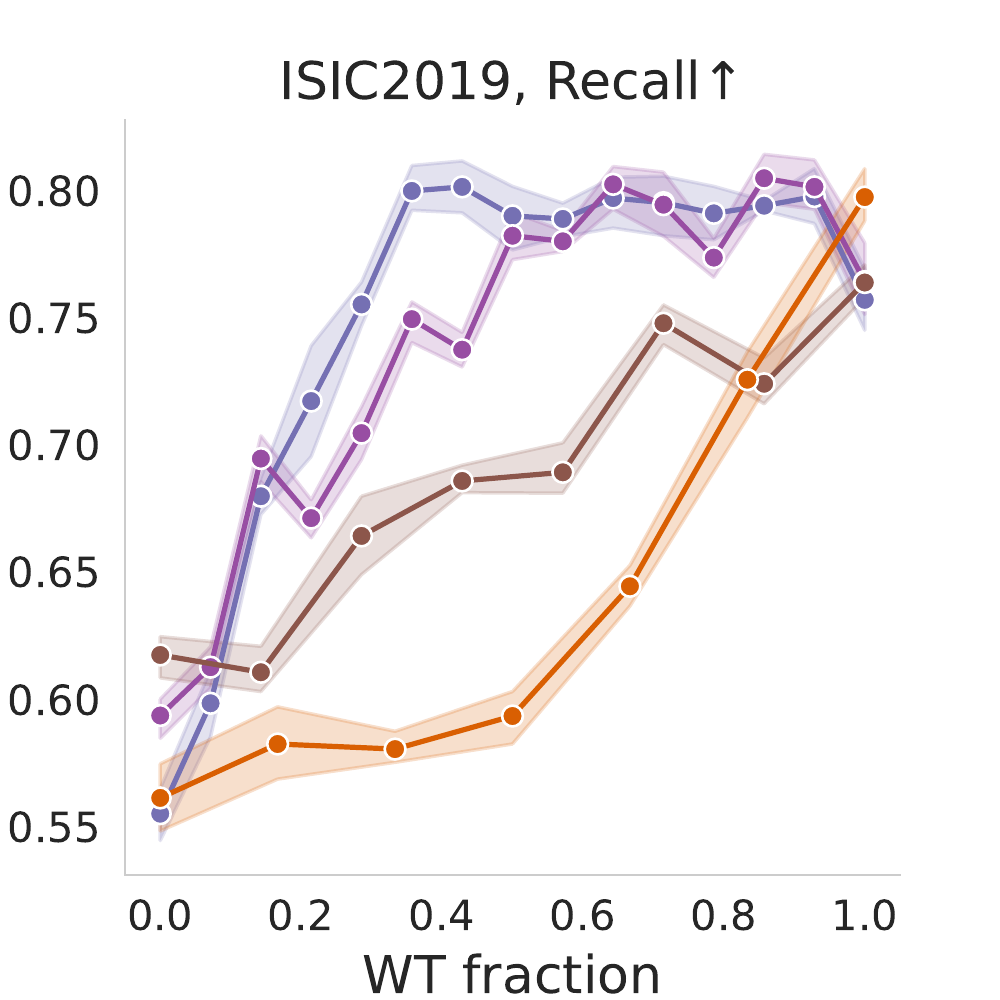} &
    \includegraphics[width=0.20\columnwidth]{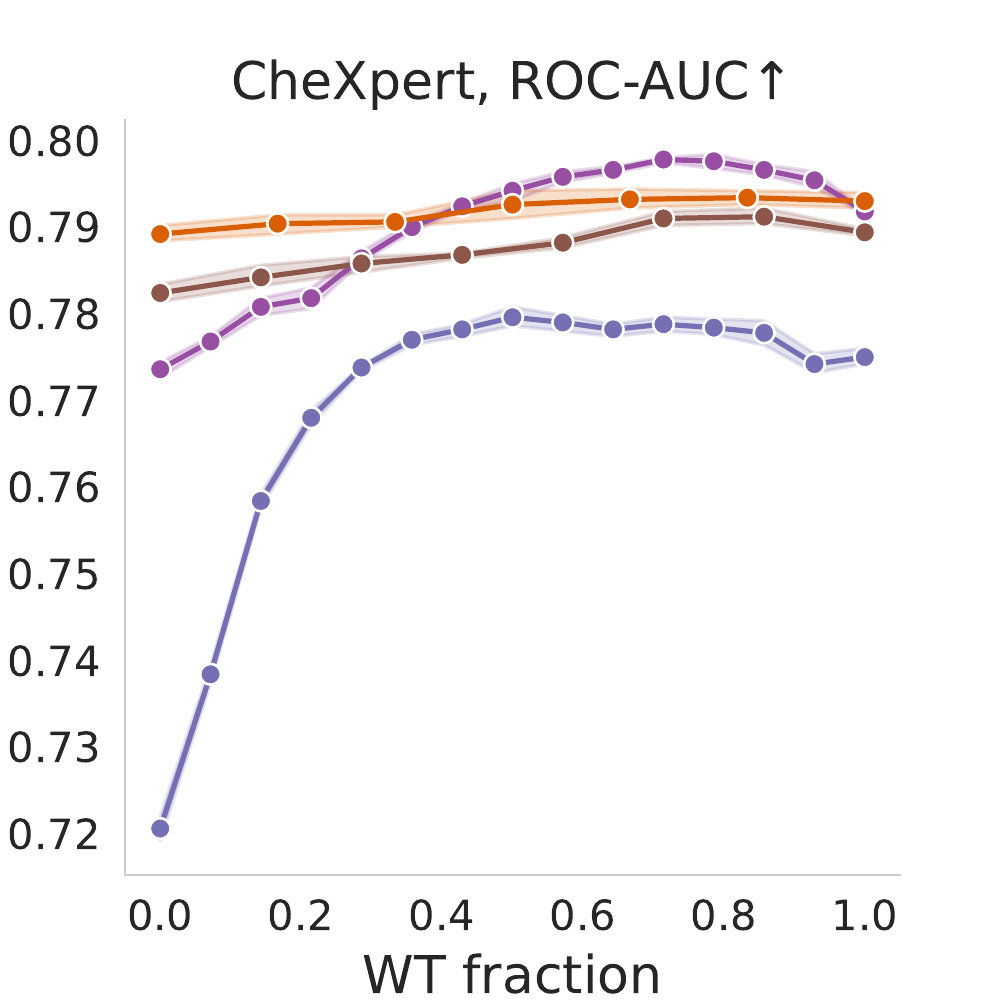} &
    \includegraphics[width=0.20\columnwidth]{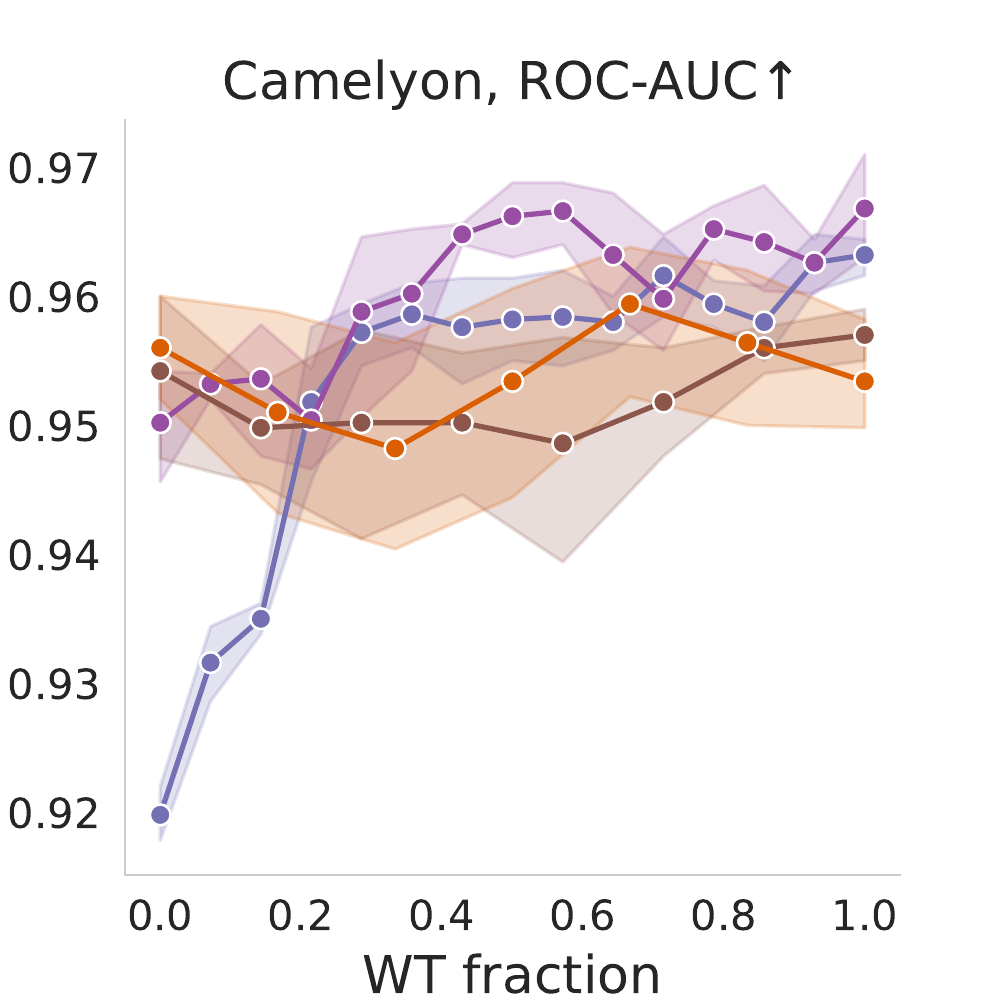}    
    \\[-1.5mm]
    
\end{tabular}
\end{center}
\vspace{-3mm}
\caption{\emph{Predictive performance of features at different depths using $k$-nn evaluation}.
\textbf{top:} $k$-NN evaluation performance at different depths for \resnetfifty  (\textit{row one}) and \deitsmall (\textit{row two}), with varying WT-ST fractions. \textbf{bottom:} Maximum $k$-NN evaluation score achieved at any depth for corresponding WT-ST initialization fraction, for each model type.
See discussion in the text. 
Full results appear in Appendix \ref{sec:sup-figures-knn}.
}
\label{fig:wst_knn}
\vspace{-4mm}
\end{figure}

\vspace{-2mm}
\paragraph{What properties of transfer learning are revealed via feature similarity?}
We investigate where similar features occur within the network using CKA, a similarity measure described in Section \ref{methods}.
In Figure \ref{fig:similarity} (top) and Figure \ref{fig:similarity_apx_wt} in the Appendix, we visualize feature similarity resulting from transfer learning (WT), before and after fine-tuning.
Red indicates high feature similarity.
High feature similarity along the diagonal is evidence for feature reuse in the corresponding layers.
For \deit, we see feature similarity is strongest in the early- to mid-layers.
In later layers, the trained model adapts to the new task and drifts away from the \imagenet features.
\resnetfifty after transfer learning shows more broad feature similarity -- with the exception of the final layers which must adapt to the new task.
This fits with the compositional nature of CNN features, also reflected in layer-by-layer improvements in Figures \ref{fig:wst_all} and \ref{fig:wst_knn}. 
A common trend shared by both ViTs and CNNs is that when more data is available, the transition point from feature reuse to feature adaptation shifts towards earlier layers because the network has sufficient data to adapt more of the transferred \imagenet features to the new task.

\begin{figure}[t]
\begin{center}
\begin{tabular}{@{}c@{}c@{}c@{}c@{}c@{}}
    \includegraphics[width=0.20\columnwidth]{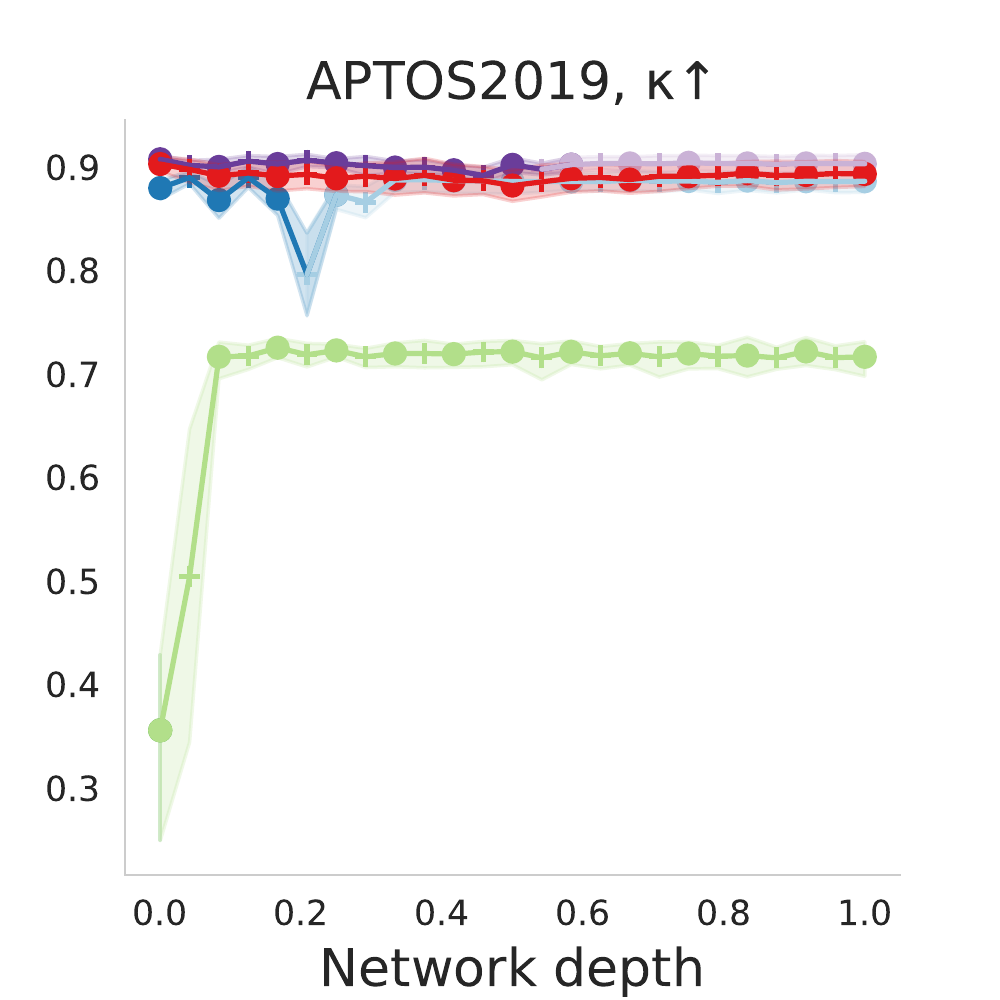} & 
    \includegraphics[width=0.20\columnwidth]{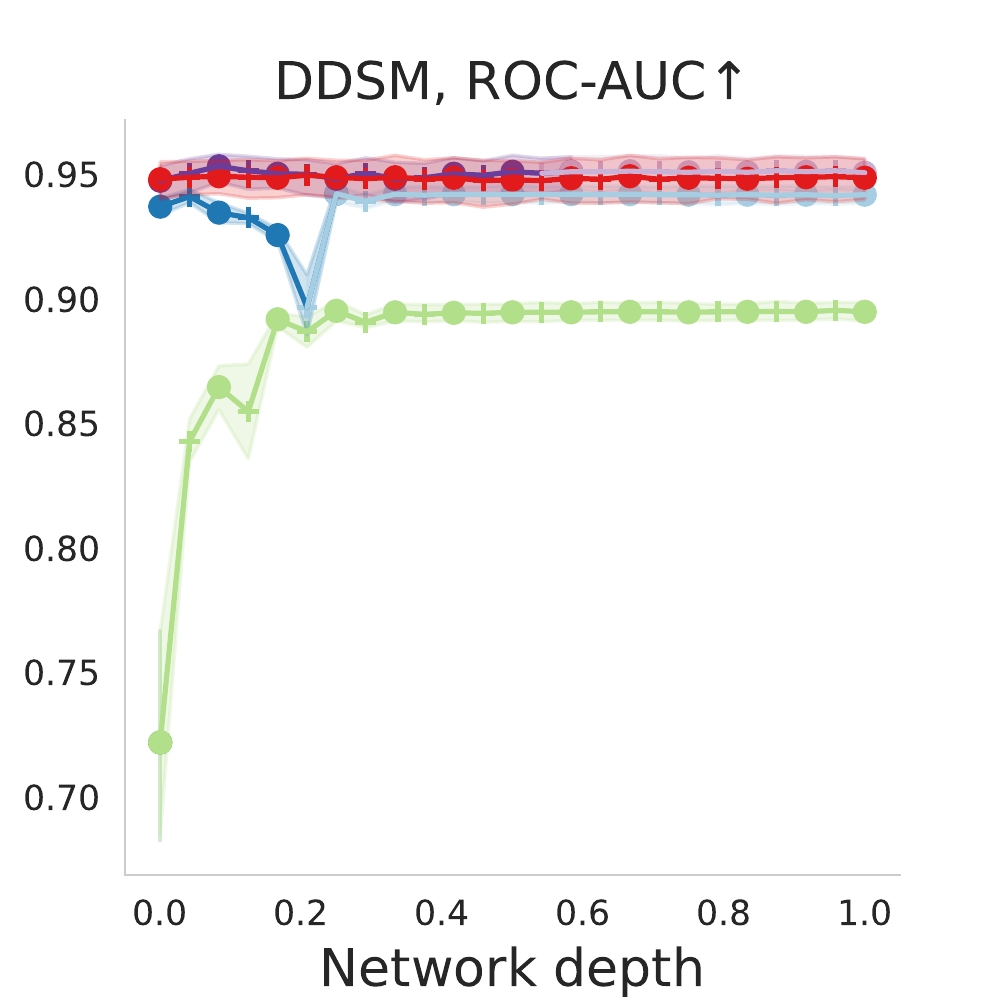} & 
    \includegraphics[width=0.20\columnwidth]{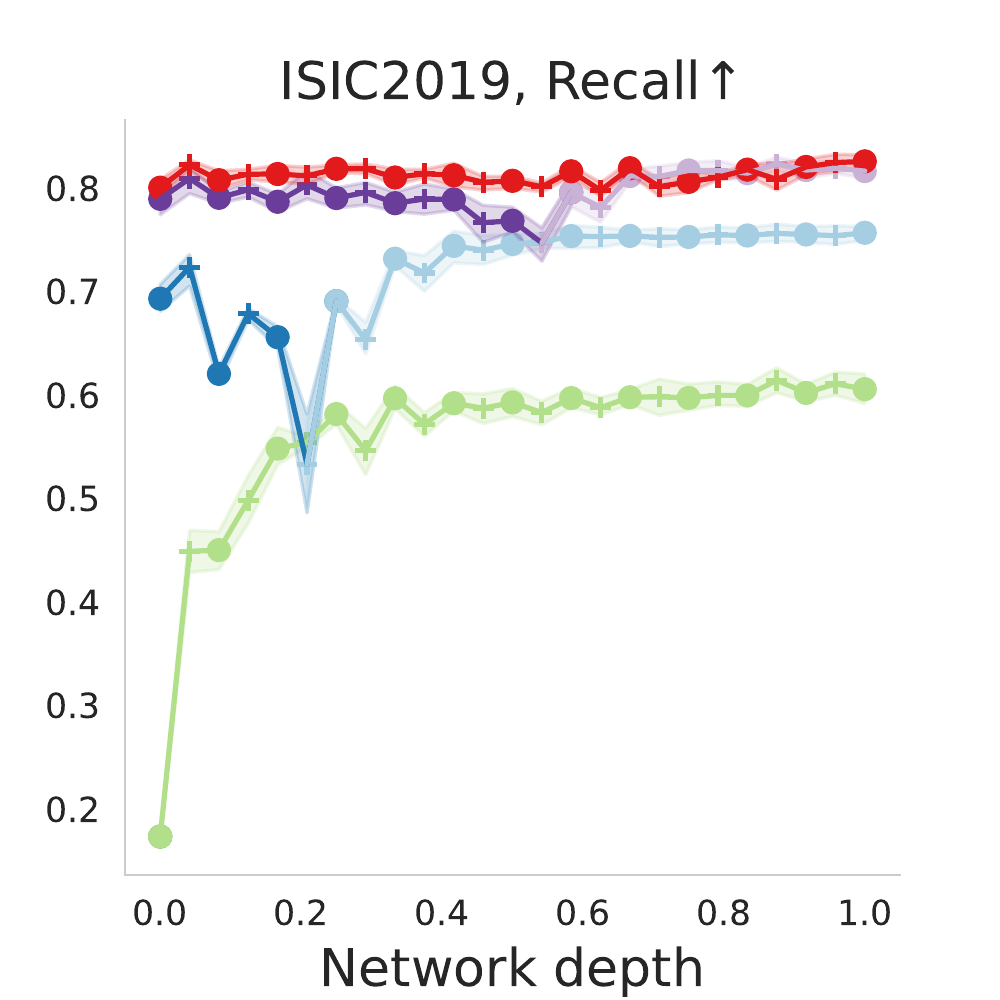} & 
    \includegraphics[width=0.20\columnwidth]{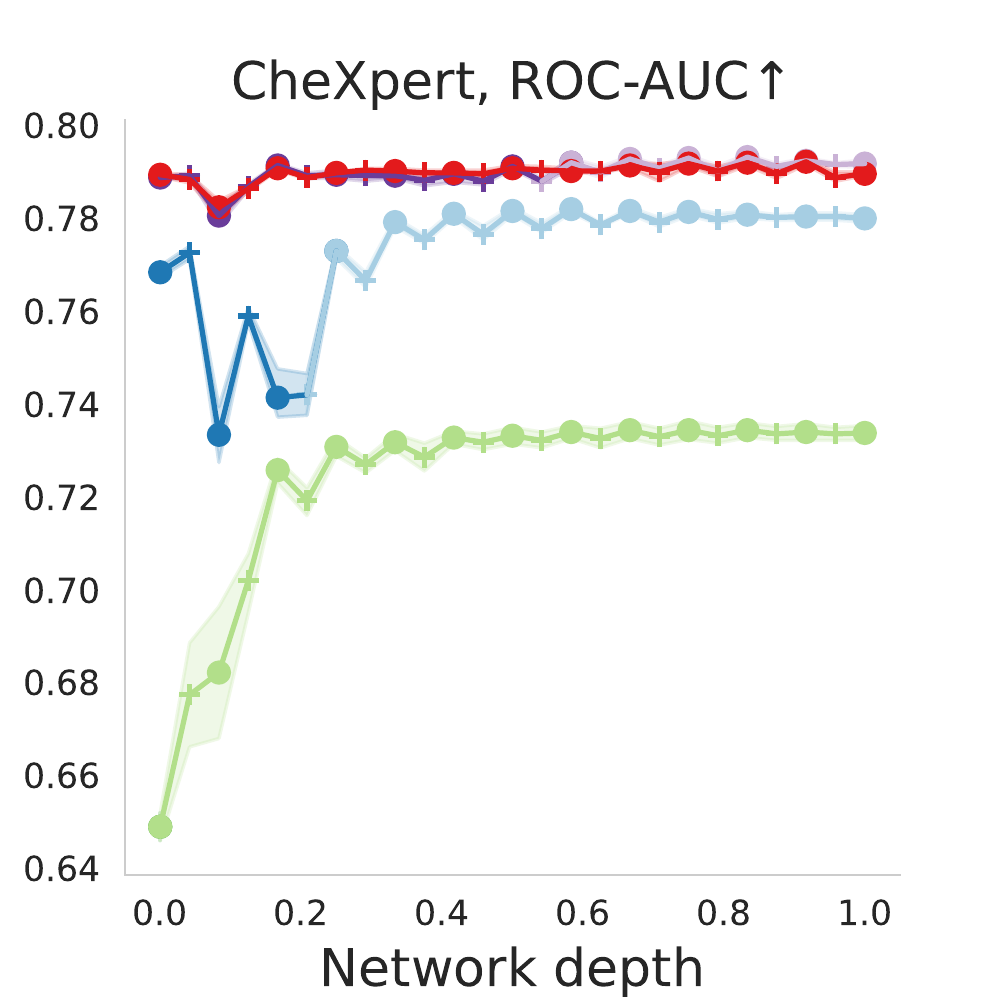} & 
    \includegraphics[width=0.20\columnwidth]{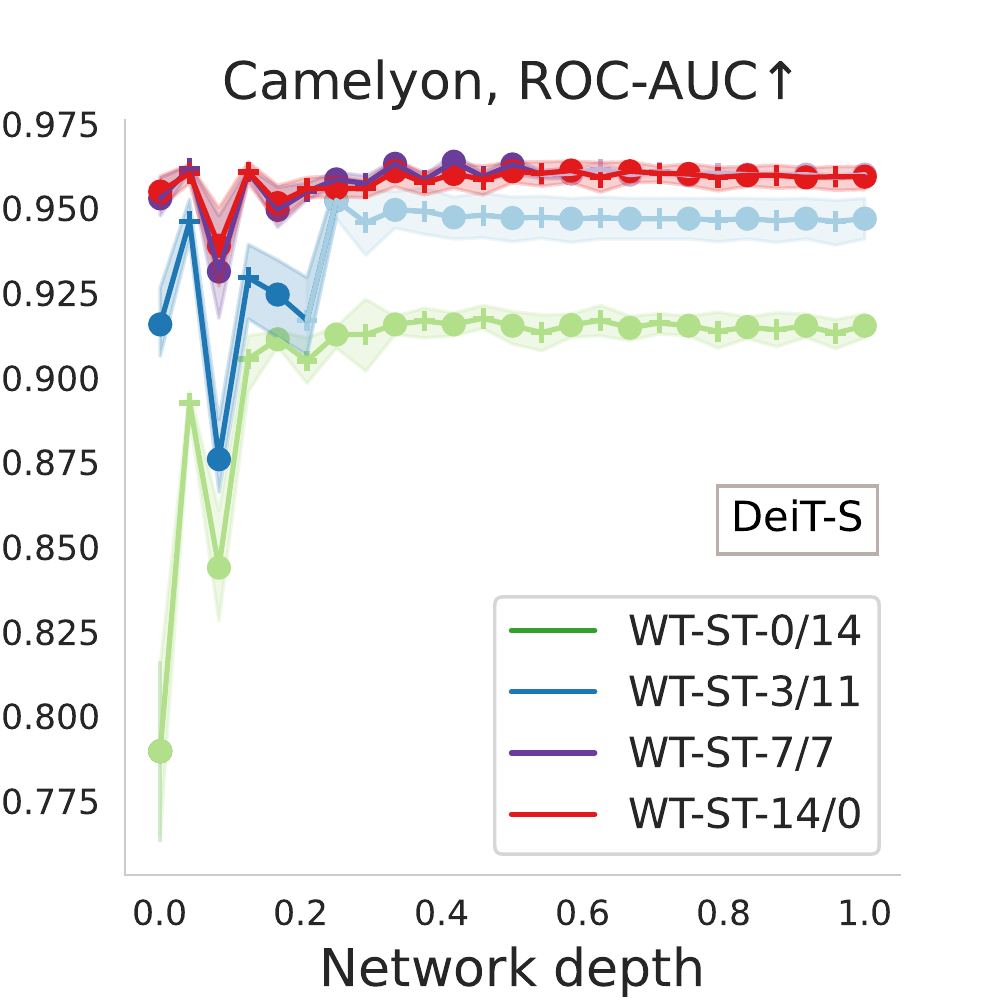}
    \\ [-1.5mm]
    \includegraphics[width=0.20\columnwidth]{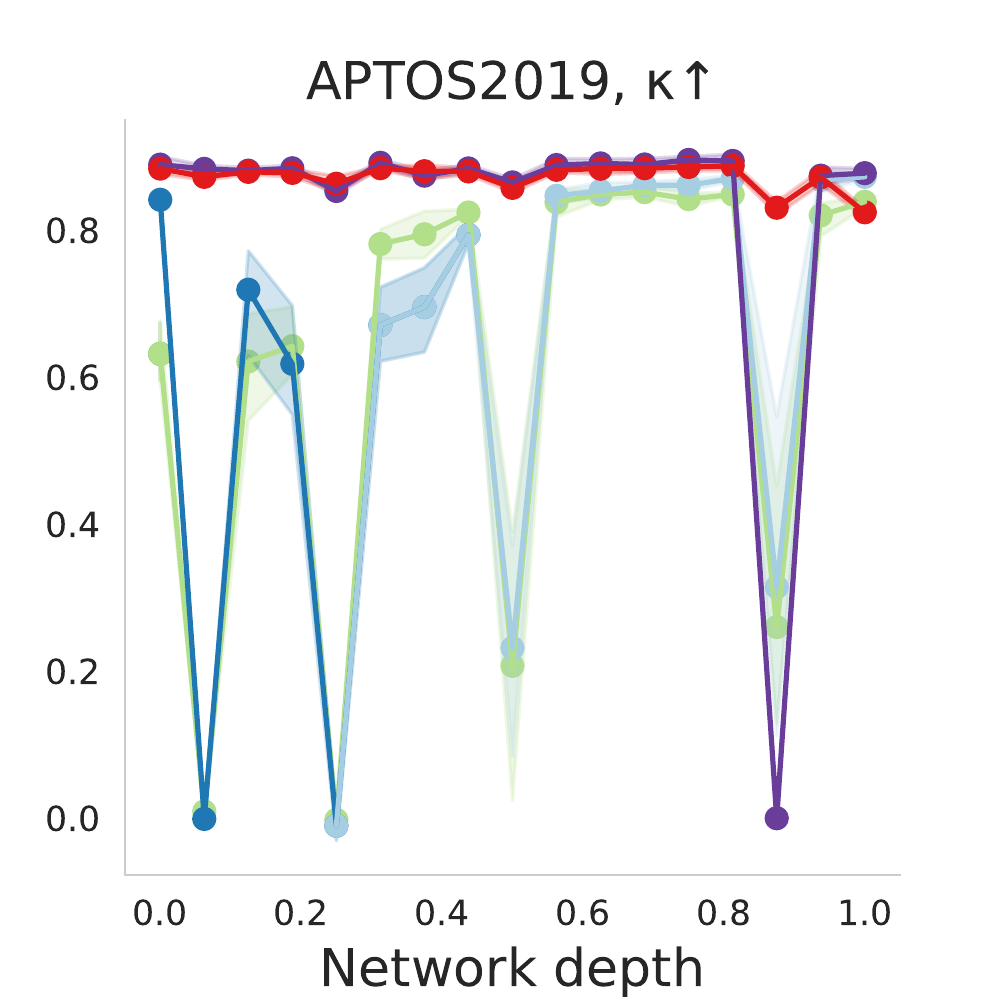} & 
    \includegraphics[width=0.20\columnwidth]{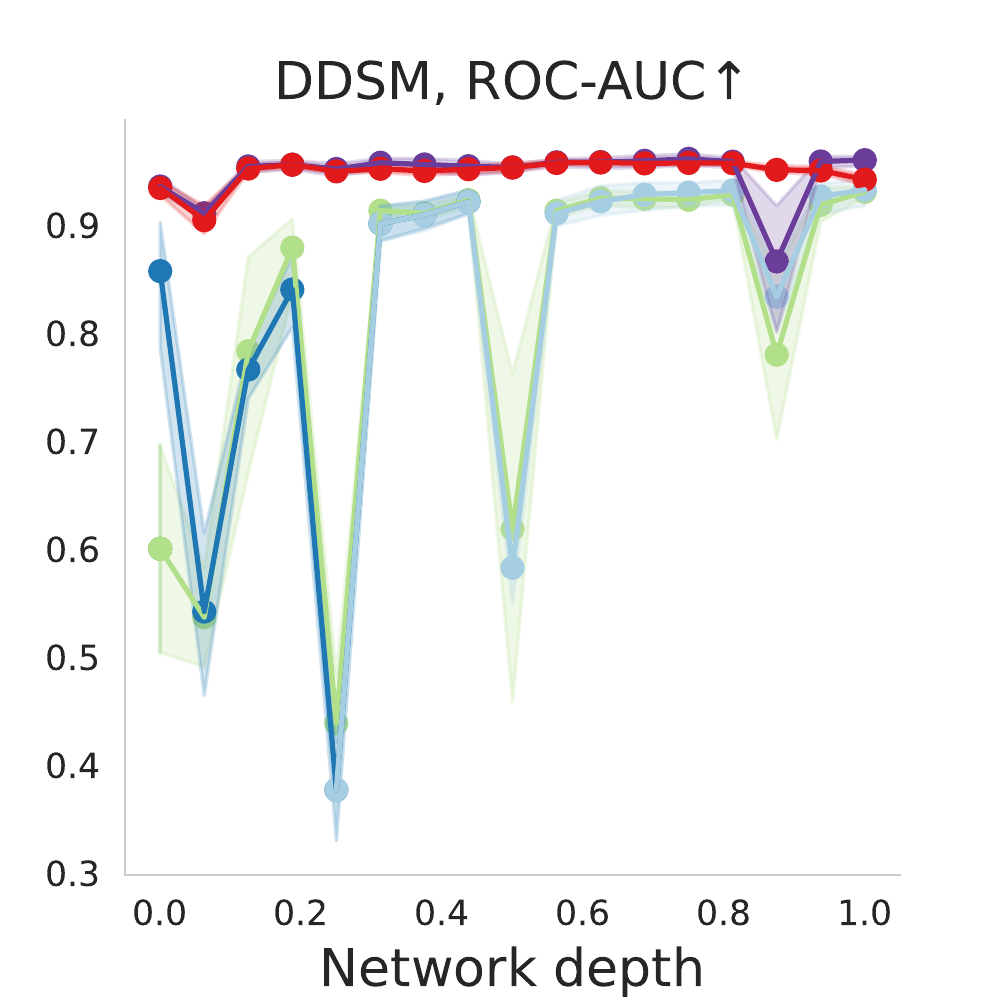} & 
    \includegraphics[width=0.20\columnwidth]{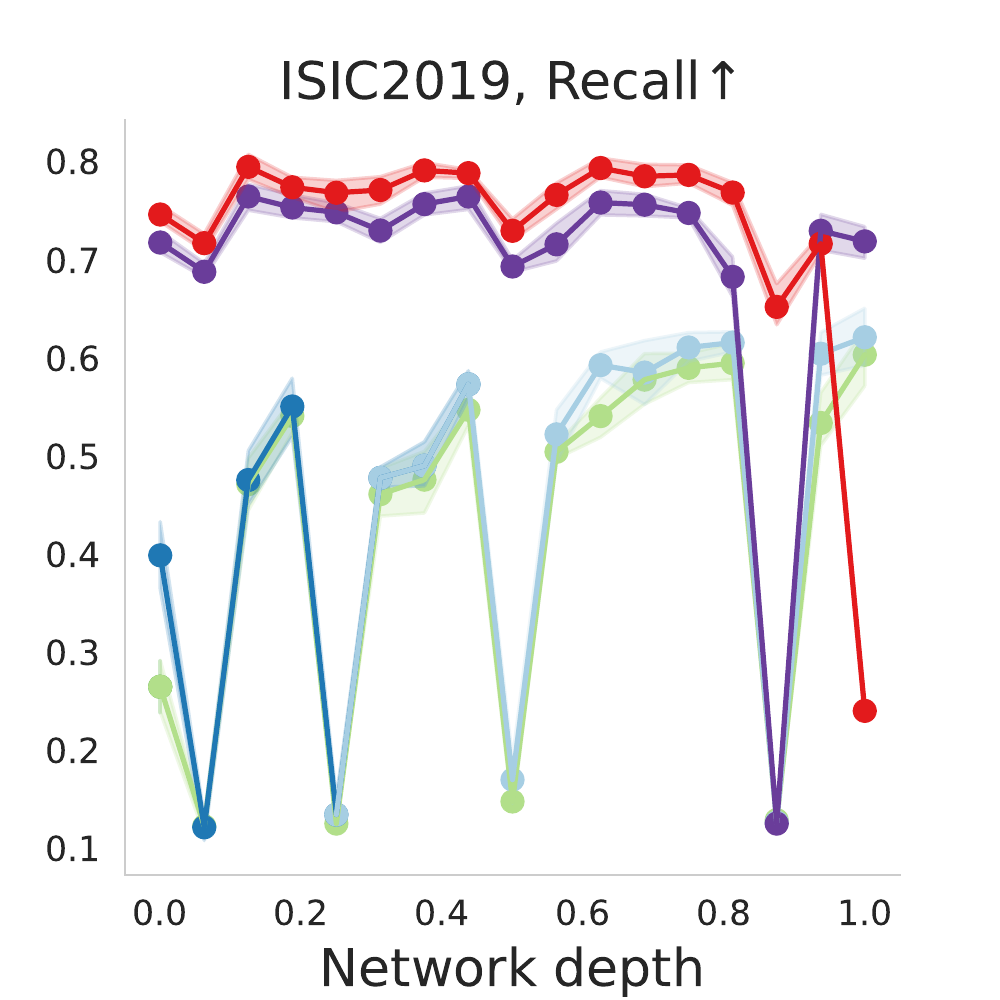} & 
    \includegraphics[width=0.20\columnwidth]{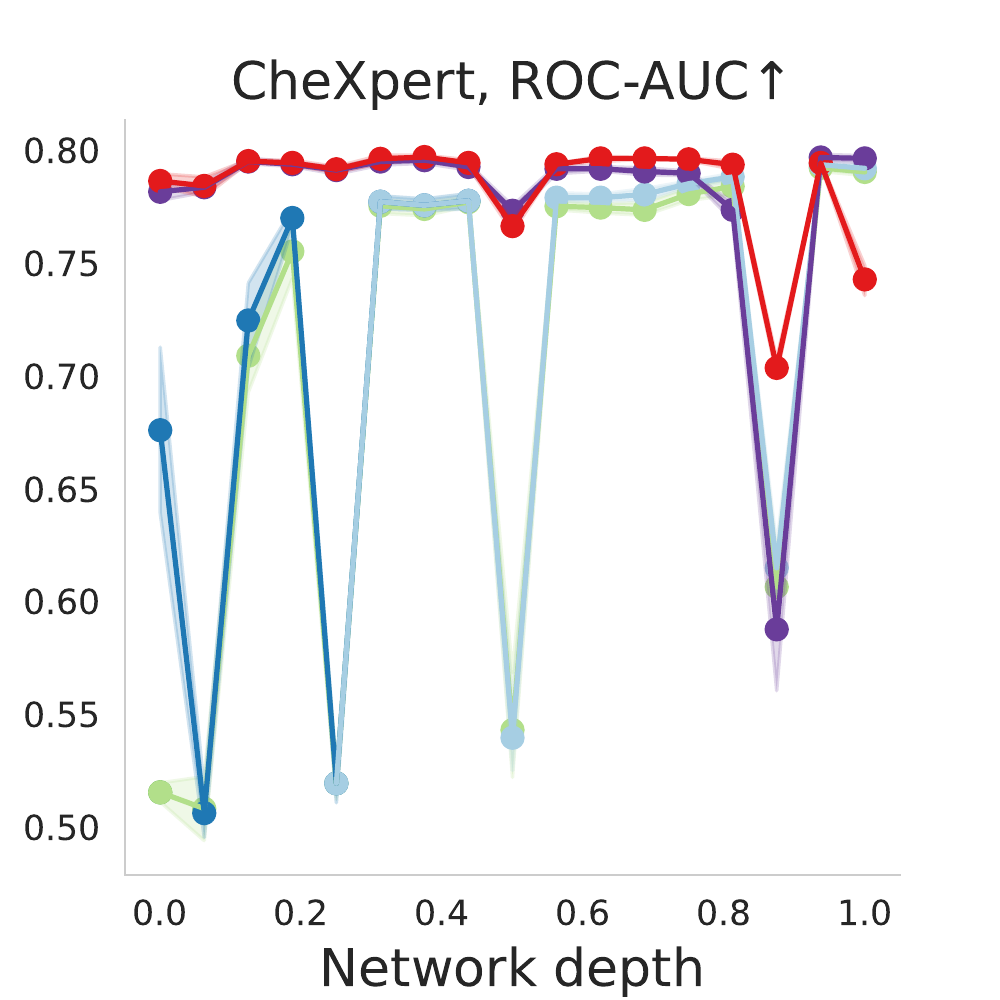} &
    \includegraphics[width=0.20\columnwidth]{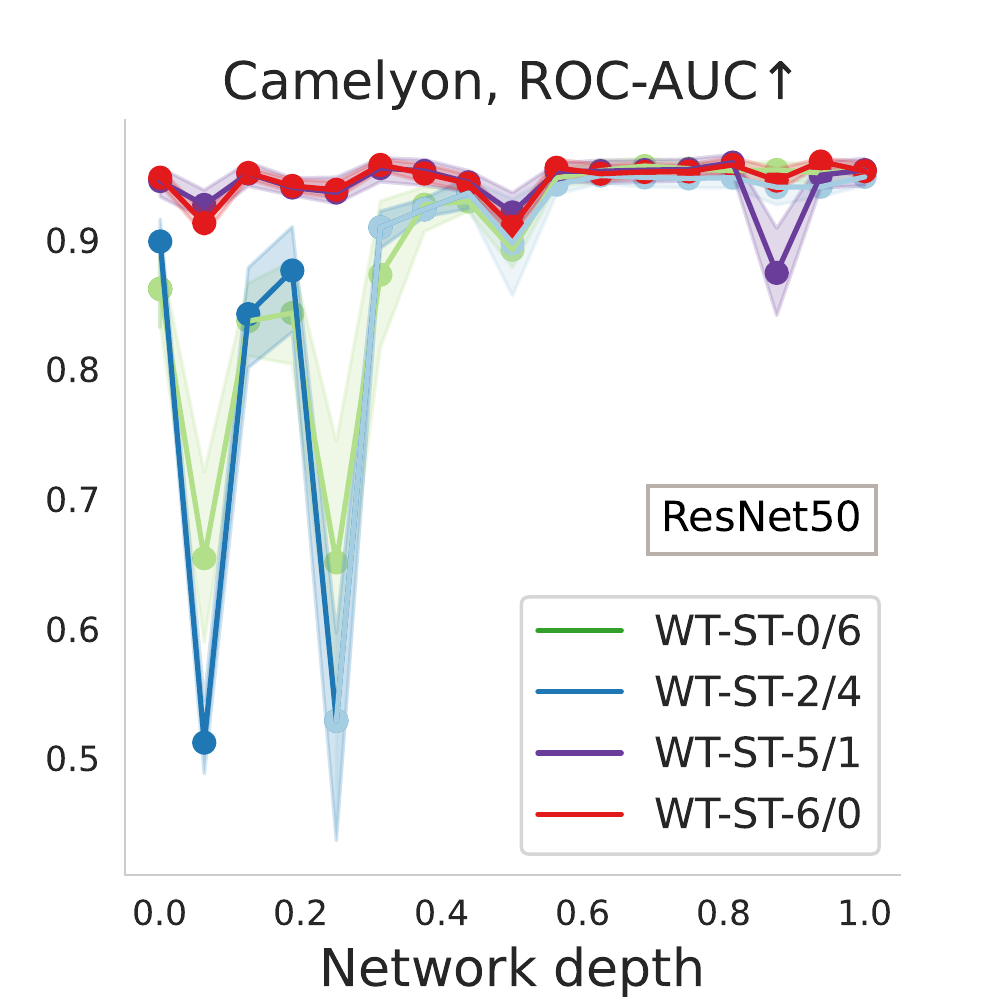}
    \\ [-1.5mm]
\end{tabular}
\end{center}
\vspace{-3mm}

\caption{\emph{Re-initialization robustness}. 
We measure the impact of resetting the model's weights to their initial value, one layer at a time.
Drops in performance indicate that during learning, the network made critical changes to the layer weights which indicate it has 
not reused the transferred weights well.
See text for details.
Full results appear in Appendix \ref{sec:sup-figures-layerwise-importance}.}
\label{fig:layer_importance}
\vspace{-4mm}
\end{figure}

\vspace{-2mm}
\paragraph{Which transferred weights change?}
Another way to investigate feature reuse is to measure how much the weights drifted from their initial values during fine-tuning.
In Figure \ref{fig:L2_dist} and Appendix \ref{appdx:L2experiment} we report the \ltwo distance between the initial weights of each network and the weights after fine-tuning.
The general trend is that transferred weights (WT) remain in the same vicinity after fine-tuning, more so when transfer learning gains are strongest  (Figure \ref{fig:L2_appdx}).
As the network is progressively initialized more with ST, the transferred weights tend to ``stick'' less well.
Certain layers, however, undergo substantial changes regardless -- early layers in ViTs (the patchifier) and \inception, and the first block at each scale in \resnetfifty.
These are the first layers to encounter the data, or a scale change.

The final way we look at feature reuse is to measure the impact of resetting a layer's weights to its initial values, or its \emph{re-initialization robustness}, reported in Figure \ref{fig:layer_importance} and Figure \ref{fig:LI_apx} of the Appendix.
Layers with low robustness underwent critical changes during fine-tuning.
Those transferred weights could not be reused directly and had to be adapted.
Our main finding is that networks with weight transfer (WT) undergo few critical changes, indicating feature reuse.
When transfer learning is least effective (\resnet on \chexpert and \camelyon) the gap in robustness between WT and ST is at its smallest.
Interestingly, in ViTs with partial weight transfer (WT-ST), critical layers often appear at the transition between WT and ST.
Rather than change the transferred weights, the network quickly adapts.
But following this adaptation, no critical layers appear.
As the data size increases, ViTs make more substantial early changes to adapt to the raw input (or partial WT).
Transferred weights in CNNs, on the other hand, tend to be less ``sticky'' than ViTs.
We see the same general trend where WT is the most robust, but unlike ViTs where WT was robust throughout the network, \resnetfifty exhibits poor robustness at the final layers responsible for classification, and also periodically within the network at critical layers where the scale changes, as observed by \cite{zhang2019all}.

\paragraph{Are reused features low-level or high-level?}
Above, we employed multiple techniques to investigate when and where feature reuse occurs within the network.
With those experiments in mind, our aim now is to determine what role the reused features play. 
Are they low-level or high-level features?
A good indicator for a high-level feature is that it can partition the data for the final task -- a property we can measure layer-wise using the $k$-NN evaluation.
Results of the $k$-NN test are given in Figure \ref{fig:wst_knn}.

First, we consider ViTs.
Previously, we observed that early layers are most crucial for ViT performance (Figure \ref{fig:wst_all}).
In the re-initialization experiment (Figure \ref{fig:layer_importance}) we also noticed that critical changes in ViTs occur either directly after the input, or at the transition between WT and ST. 
From the $k$-NN tests in Figure \ref{fig:wst_knn} and \ref{fig:LI_apx} in the Appendix, we see that the relevance of the features increases dramatically within these critical layers.
Later layers do not seem to contribute further to solve the task\footnote{The zig-zag pattern in row 2 of Fig.~\ref{fig:wst_knn} is due to alternating self-attention (+) \& MLP layers ($\cdot$) common in ViT architectures.}.
In the bottom of Figure \ref{fig:wst_knn} we notice that the discriminative power of ViT features increases rapidly as we add more WT layers in the beginning, but it saturates approximately halfway through the network.
Interestingly, in an ablation we present in Appendix \ref{sec:smaller-deit}, we found that the first 5 blocks of \deits performs comparably with the full 12 blocks for transfer learning.
Evidently, early feature reuse in ViTs combined with the small medical data size results in unutilized capacity in the later layers of the ViTs, which can effectively be thrown away.
Thus, we find that \textit{features reused in these critical early layers of ViTs are responsible for the creation of high-level features}.
According to \cite{dosovitskiy2020image, raghu2021vision}, these same critical early layers are responsible for learning a mix of local and global features -- an essential component for good performance which requires very large datasets to learn -- explaining ViT's strong dependence on feature reuse in transfer learning.
In Appendix \ref{appdx:mean_att_distance} we confirm that WT transfer produces a mixture of local and global attention in early ViT layers, whereas ST initialization cannot learn to attend locally.
Next we turn to the CKA experiments at the bottom of Figure \ref{fig:similarity}.
Here, we find that early layers of ST-initialized models are similar to features from the first half of the WT-initialized models.
We see that if the network is denied these essential pre-trained weights, it attempts to learn them rapidly using only a few layers (due to lack of data), resulting in poor performance.

\begin{figure}[t]
\begin{center}
\begin{tabular}{@{}c@{}c@{}c@{}c@{}c@{}}
    \includegraphics[width=0.20\columnwidth]{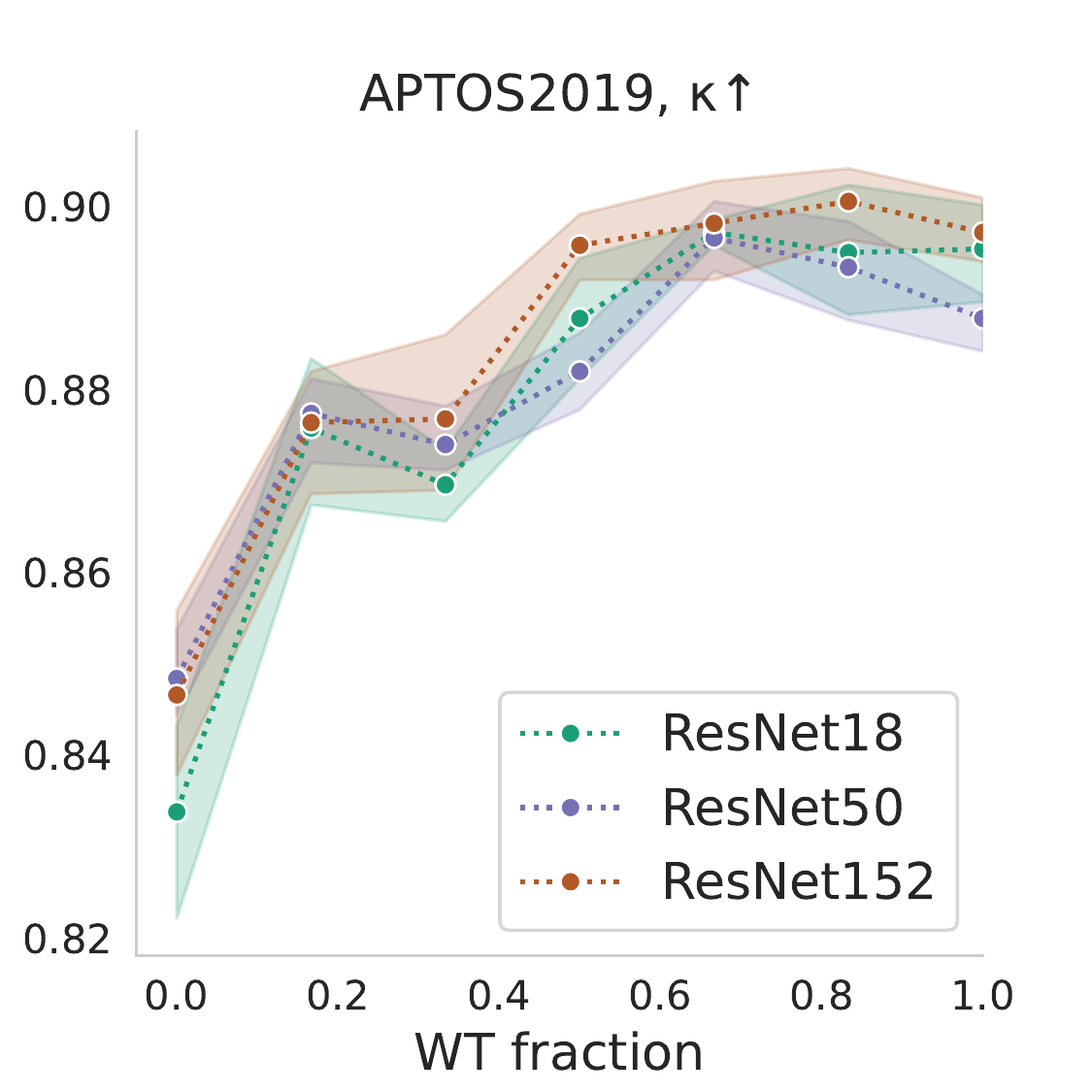} & 
    \includegraphics[width=0.20\columnwidth]{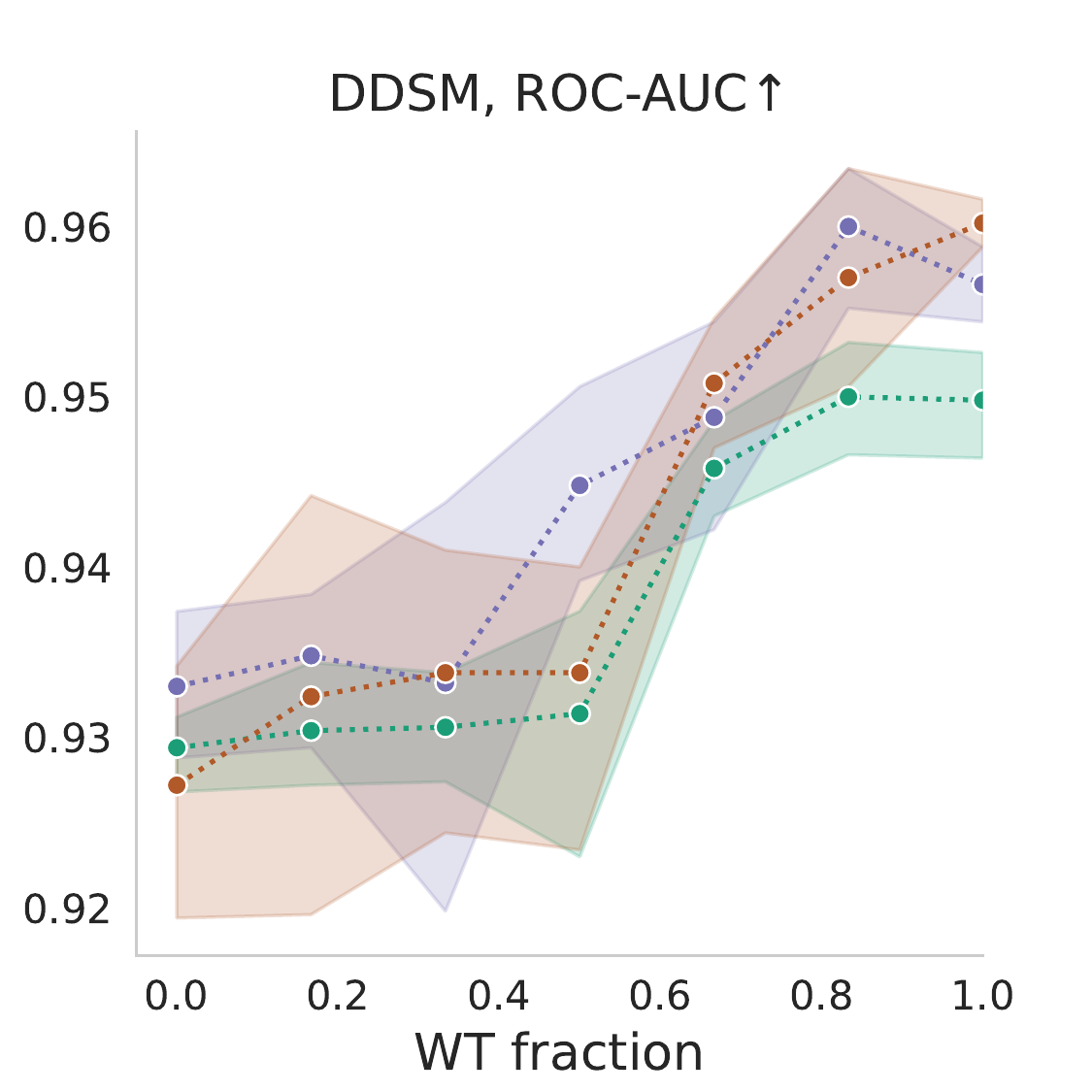} & 
    \includegraphics[width=0.20\columnwidth]{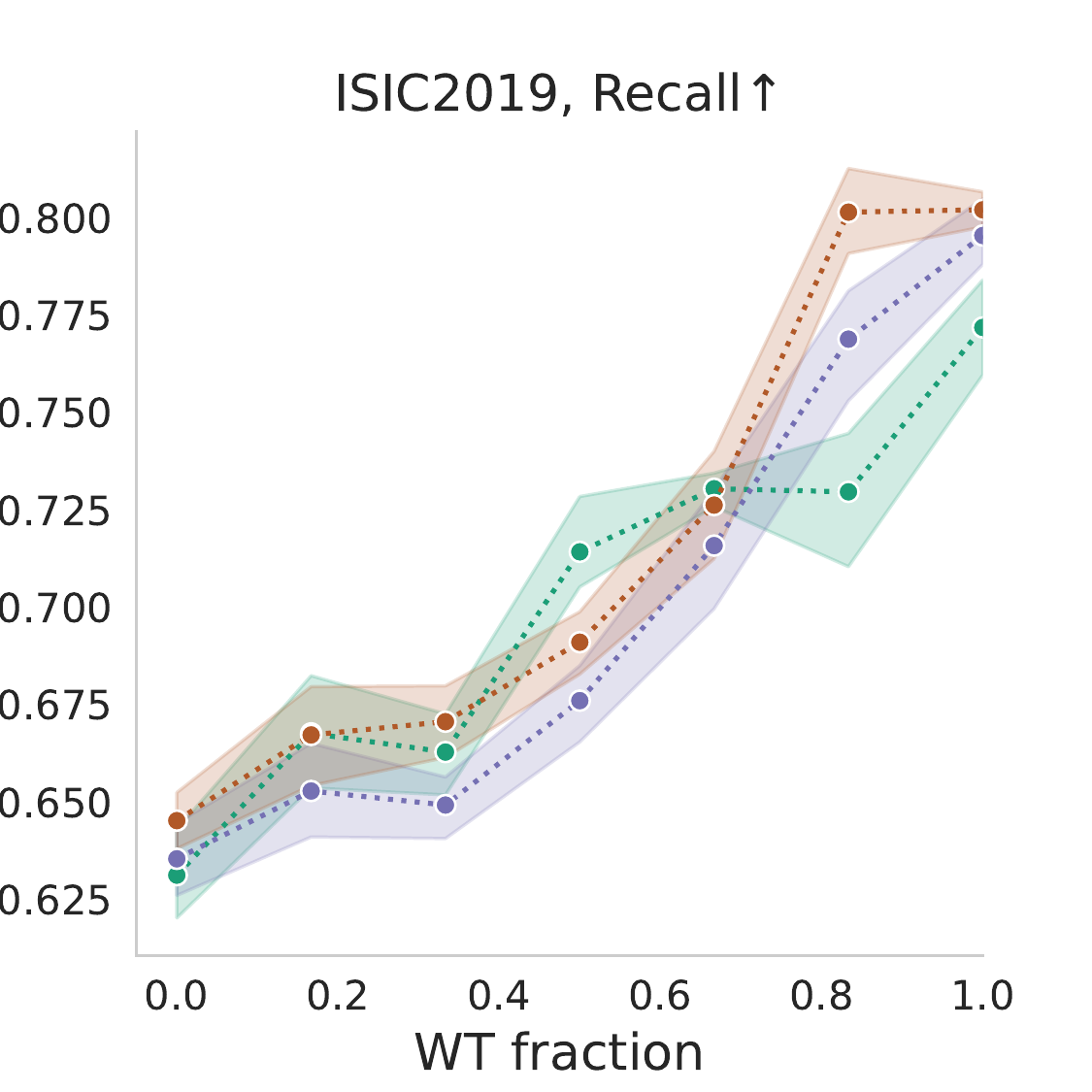} & 
    \includegraphics[width=0.20\columnwidth]{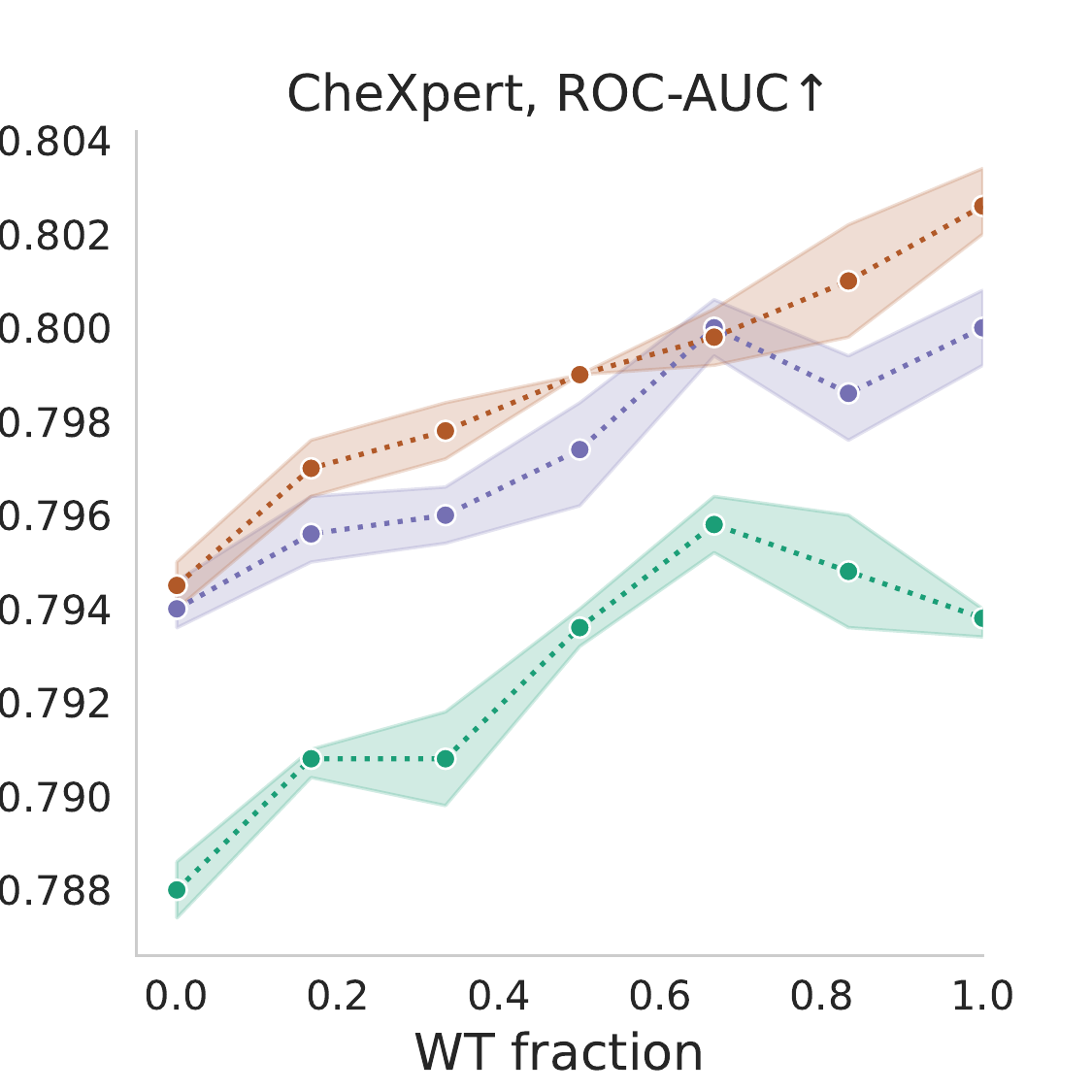} & 
    \includegraphics[width=0.20\columnwidth]{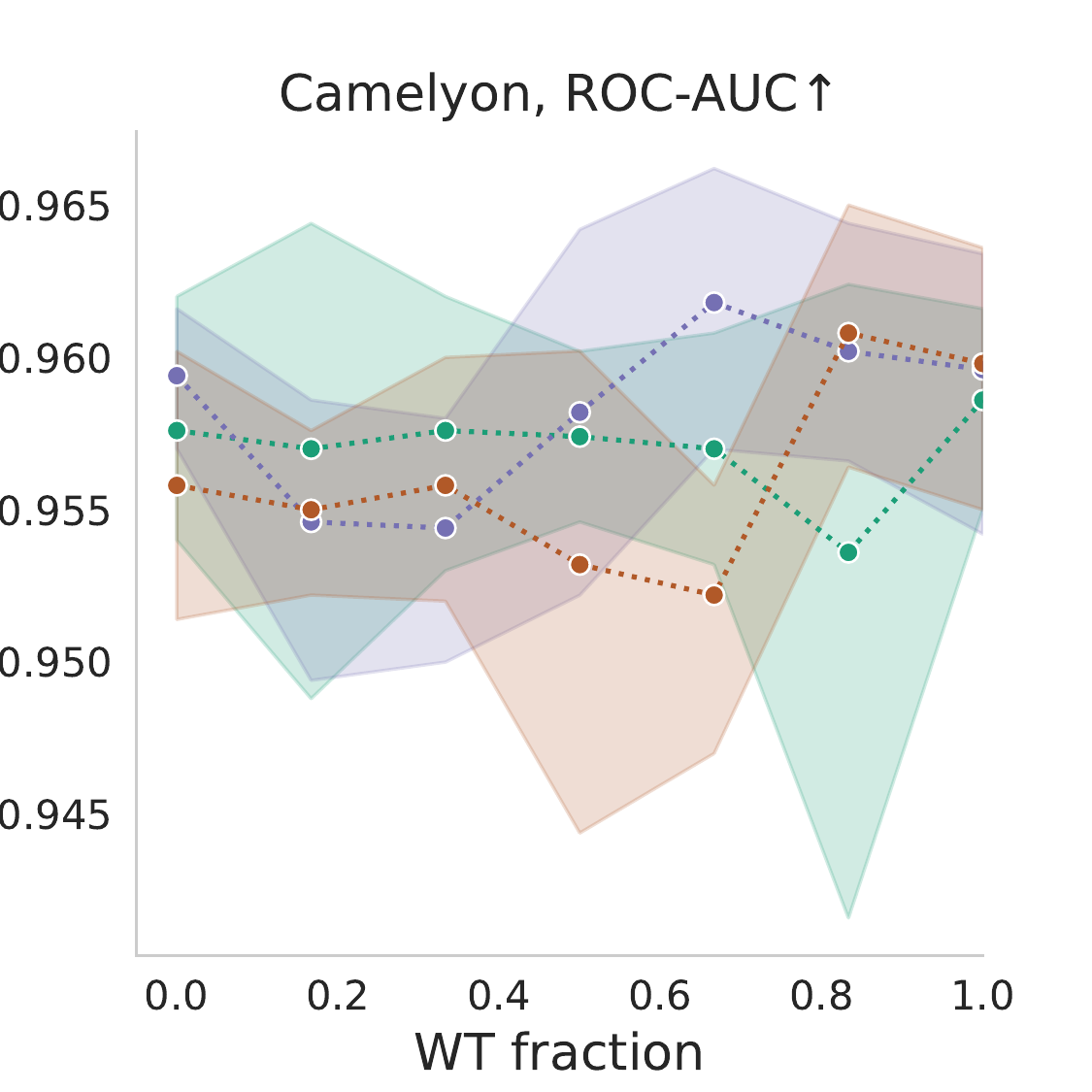}  
    \\[-1.5mm] 
    \includegraphics[width=0.20\columnwidth]{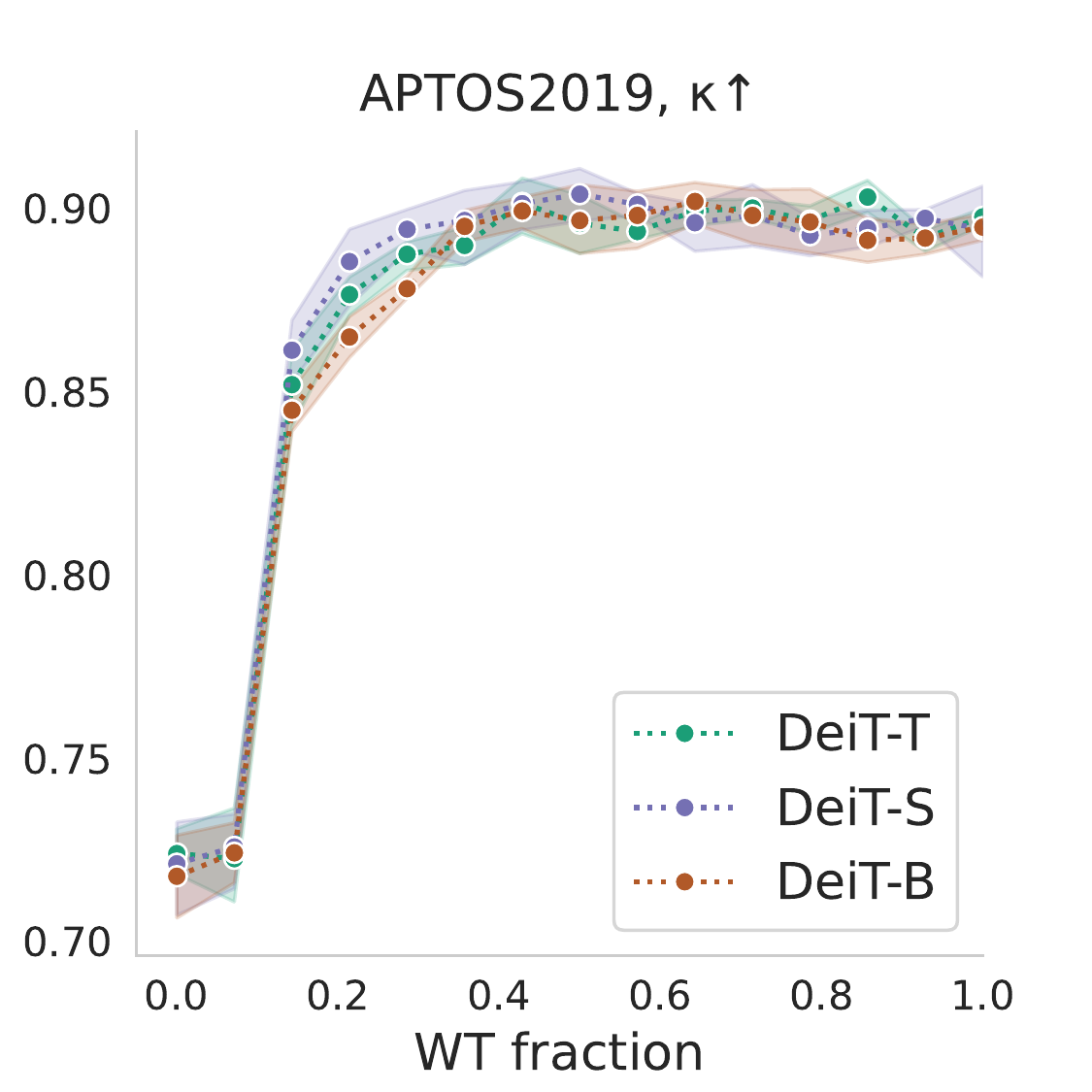} & 
    \includegraphics[width=0.20\columnwidth]{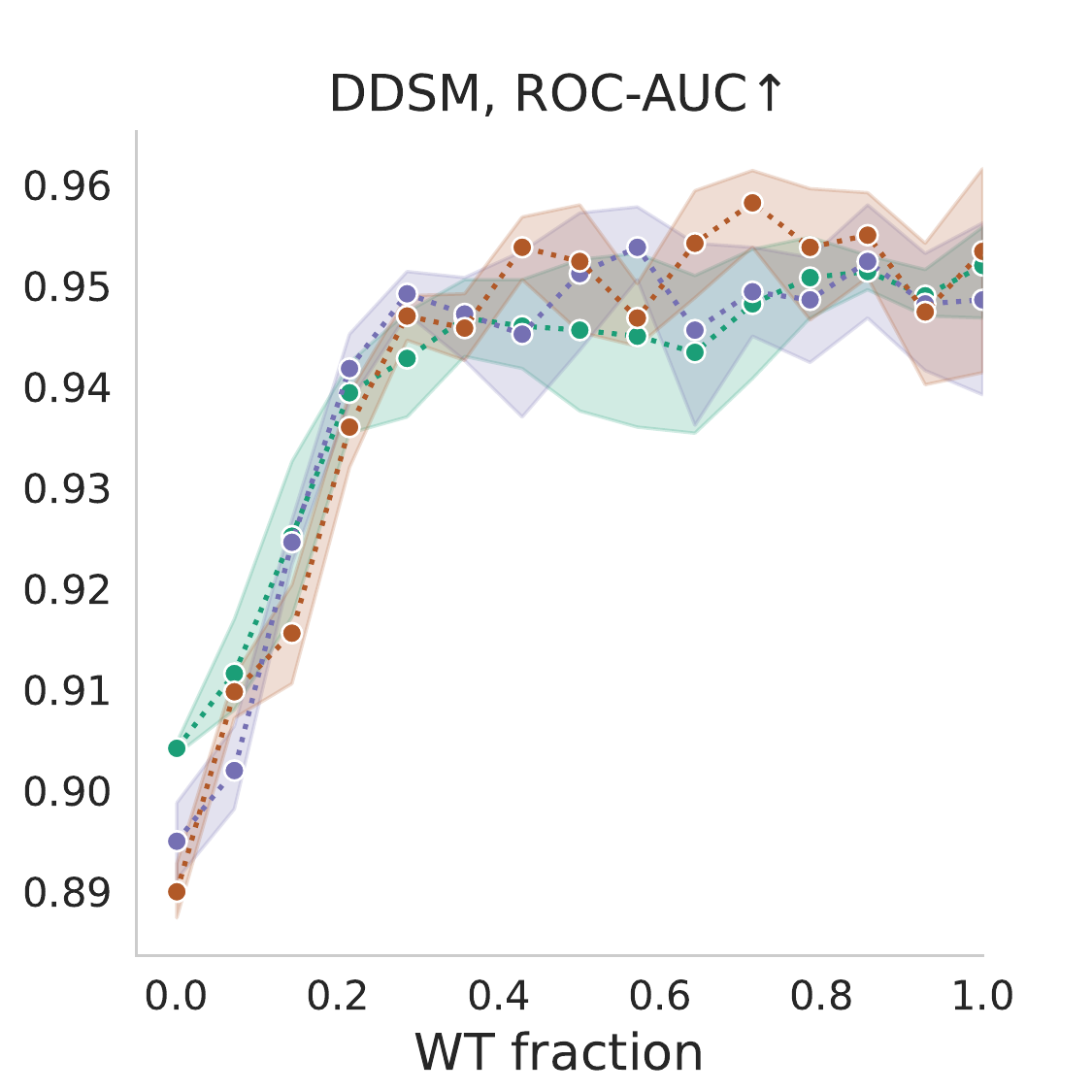} & 
    \includegraphics[width=0.20\columnwidth]{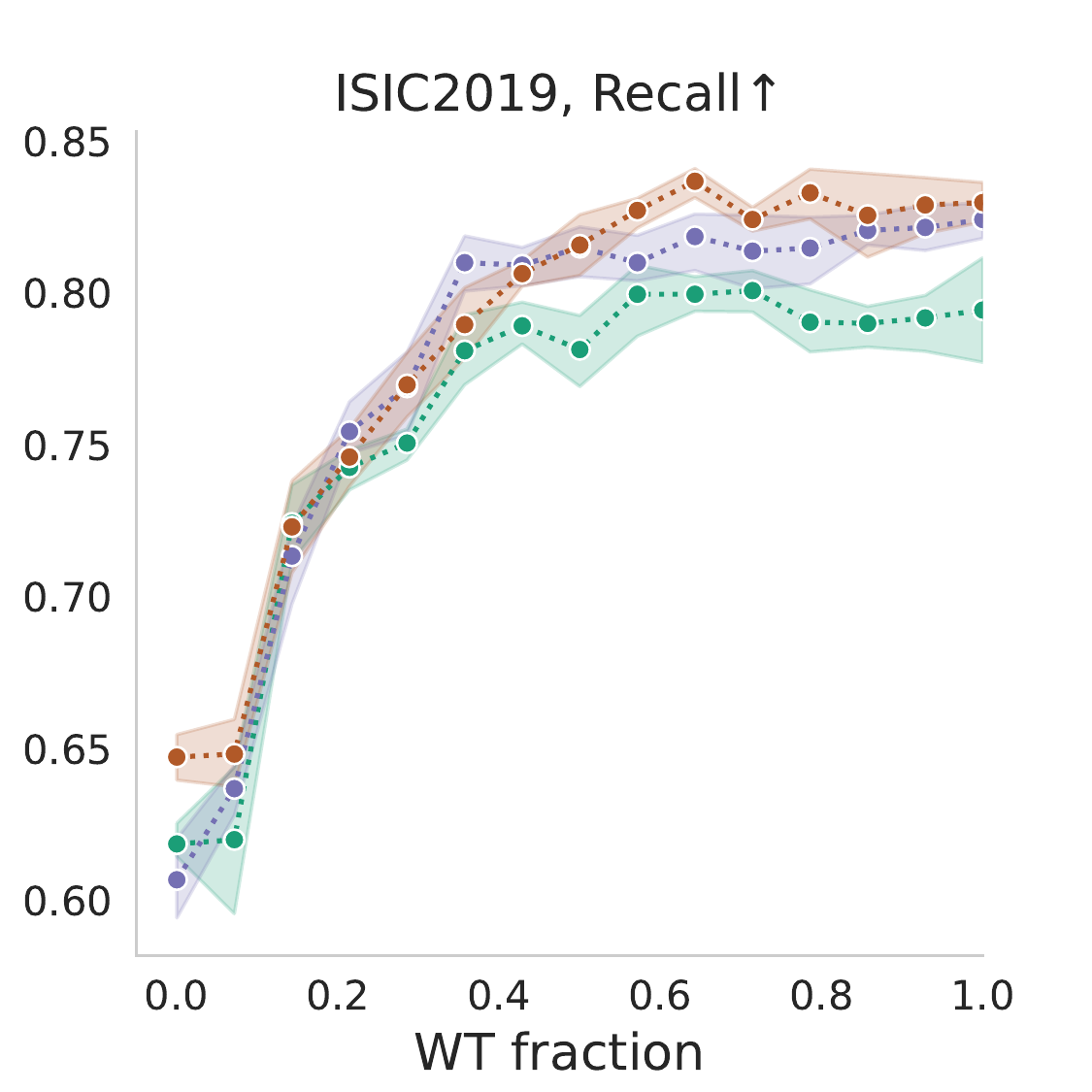} &     
    \includegraphics[width=0.20\columnwidth]{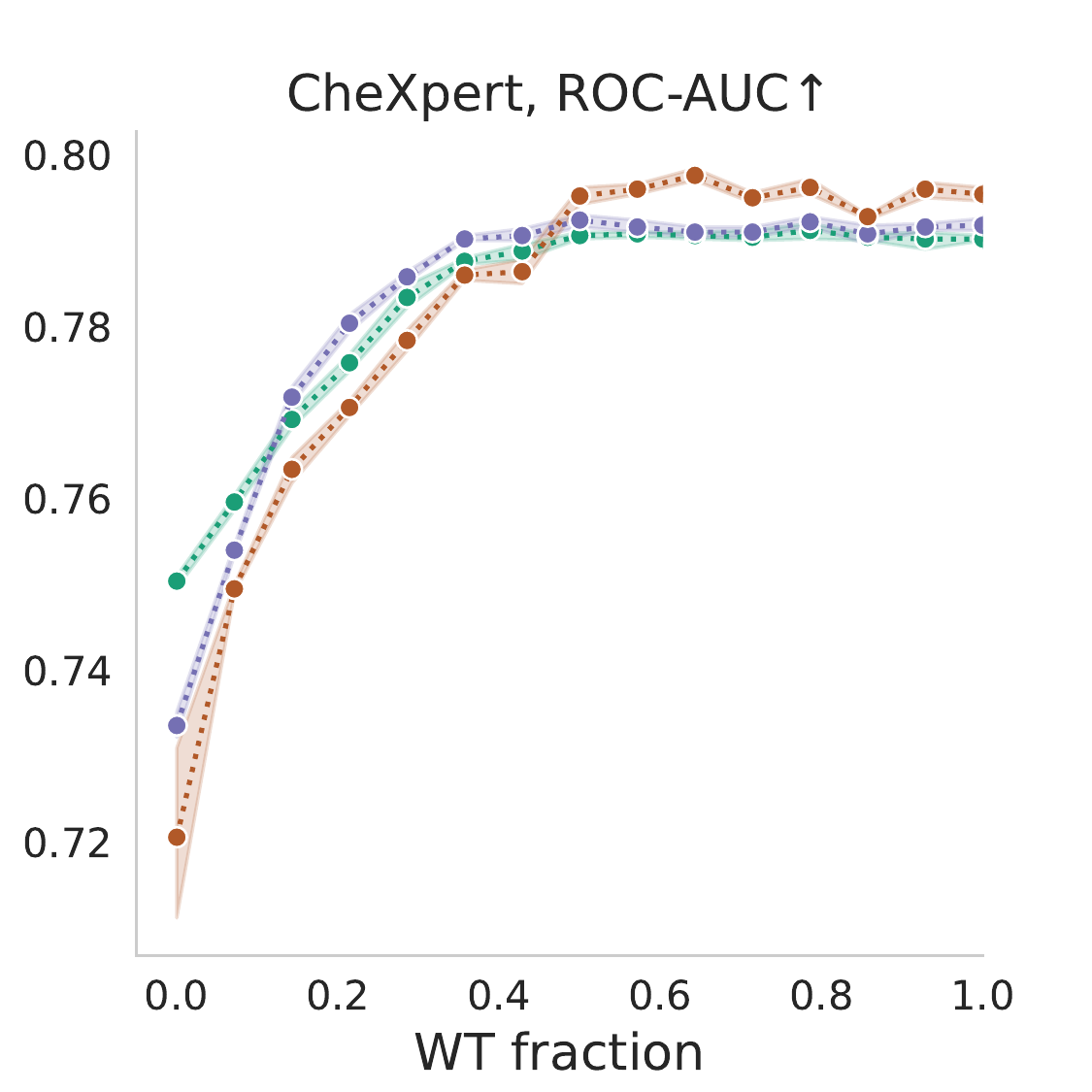} &   
    \includegraphics[width=0.20\columnwidth]{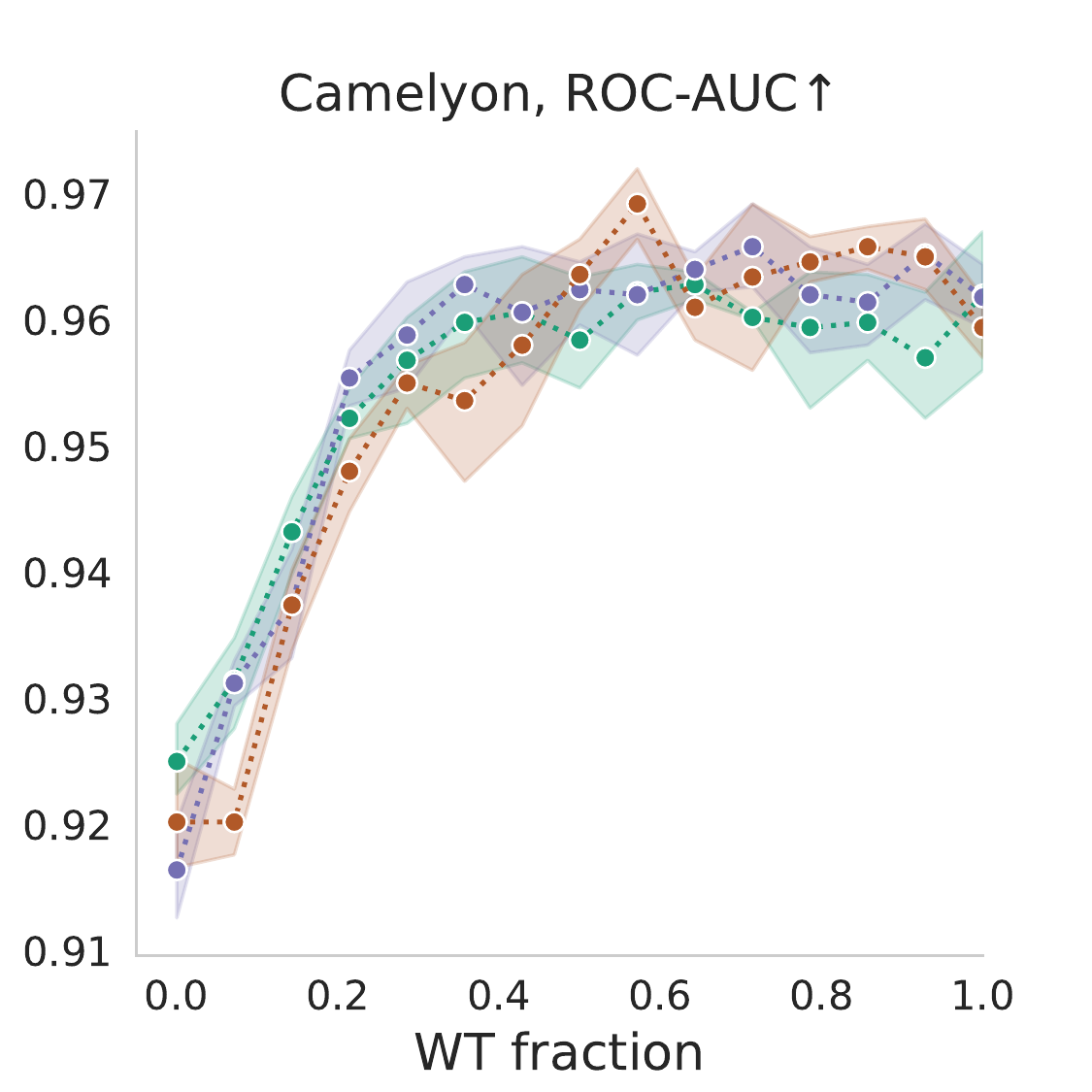}      
    
\end{tabular}
\end{center}
\vspace{-5mm}
\caption{\emph{The impact of weight transfer for different model capacities.} 
We evaluate the impact of weight transfer when using WT-ST initialization as a function of model capacity. 
Larger models benefit more from transfer learning but the same architectures follow similar patterns.
}
\label{fig:wst_capacity}
\vspace{-4mm}
\end{figure}

The role of transferred features in CNNs is different, as one might expect.
We saw in Figure \ref{fig:wst_all} that performance benefits from feature reuse are more evenly distributed throughout CNNs, while the re-initialization experiment in Figure \ref{fig:layer_importance} revealed that the critical layers are also spread out throughout the network.
The \knn test in Figure \ref{fig:wst_knn} further supports these findings -- a jump in early layers corresponding to low-level feature extraction is followed by progressive improvements in the features as each layer adds complexity over the previous, until the final layer.
Large periodic \knn increases correspond to critical layers in Figure \ref{fig:layer_importance}. 
These trends nicely follow our understanding of compositional learning in CNNs.
A notable outlier is \isic, where \knn improvement is delayed.
This is likely due to \isic's similarity to \imagenet, which allows mid-level transferred features to be reused more readily.
From the bottom row of Figure \ref{fig:similarity} we further observe that CNNs seem to learn similar features from different initializations, suggesting that their inductive biases may somehow naturally lead to these features (although the final layers used for classification diverge).
We also observe a trend where, given more data, the ST-initialization is able to learn some novel mid- to high-level features not found in \imagenet.

\vspace{-2mm}
\paragraph{Capacity and convergence.} 
In addition to the other transfer learning factors investigated thus far we consider model capacity.
We repeat our main experiments using \deits and \resnets with different capacities 
and report the results in Figure \ref{fig:wst_capacity}.
We observe slight increases in transfer learning performance as model size increases, but the patterns exhibited by the individual architectures do not change.

\begin{figure}[t]
\begin{center}
\begin{tabular}{@{}c@{}c@{\hspace{0.5mm}}c@{\hspace{2mm}}|c@{}}
    \includegraphics[width=0.3\columnwidth]{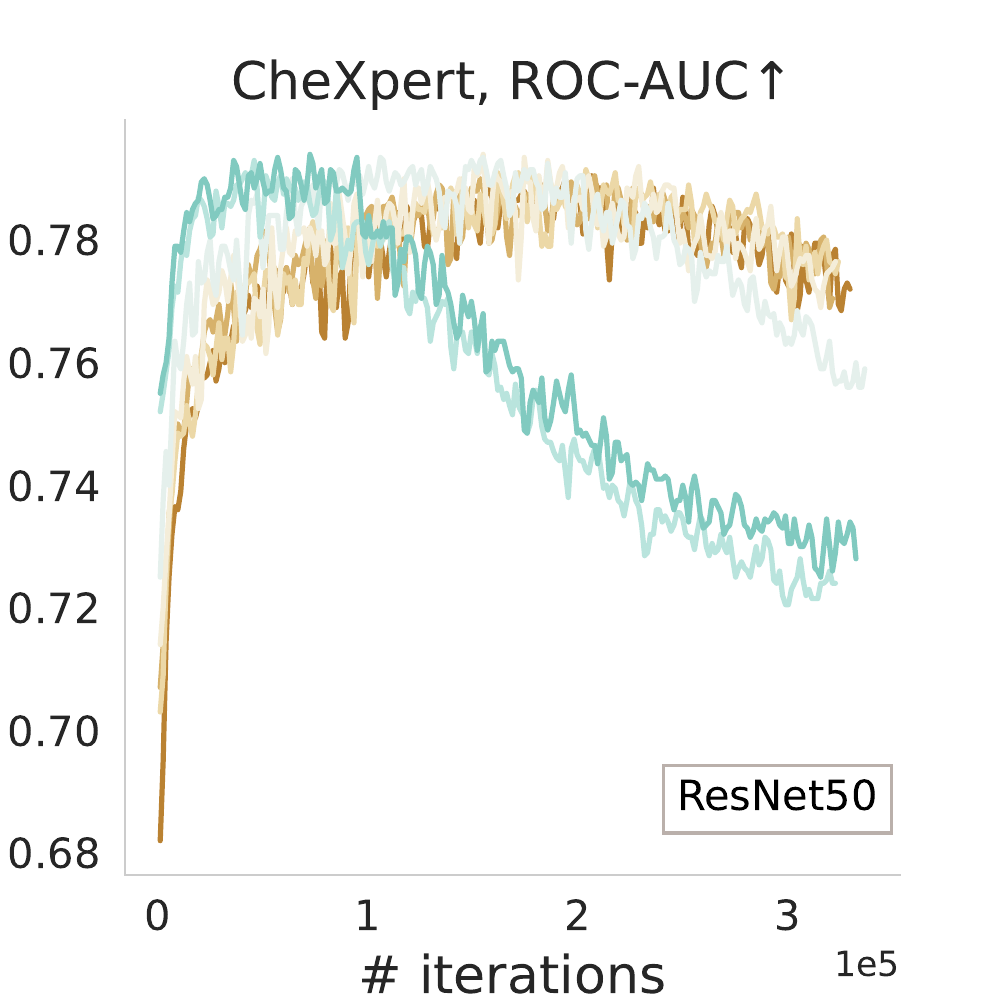} & 
    \includegraphics[width=0.3\columnwidth]{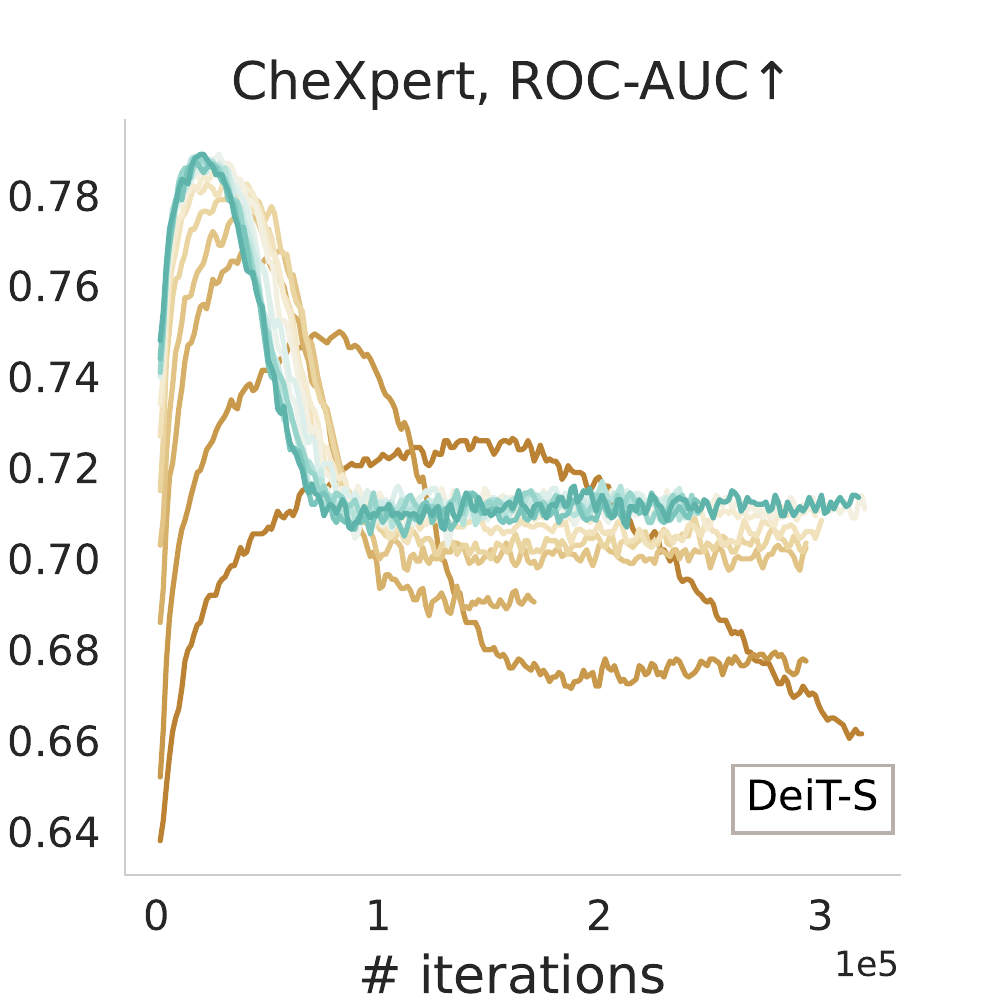} &
    \multirow{1}{*}[2.0cm]{\includegraphics[width=0.045\columnwidth]{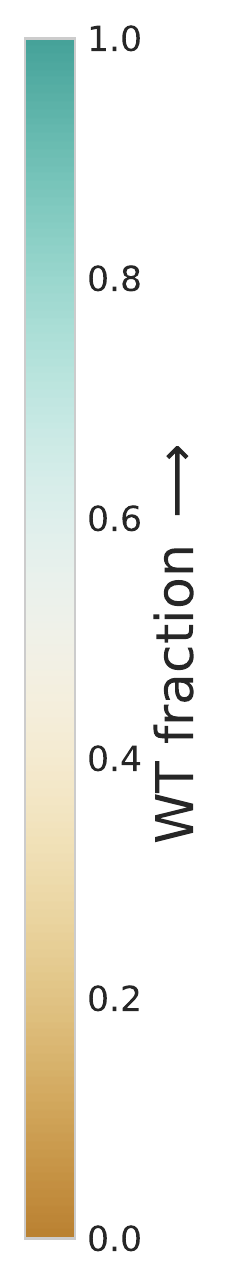}} &
    \includegraphics[width=0.3\columnwidth]{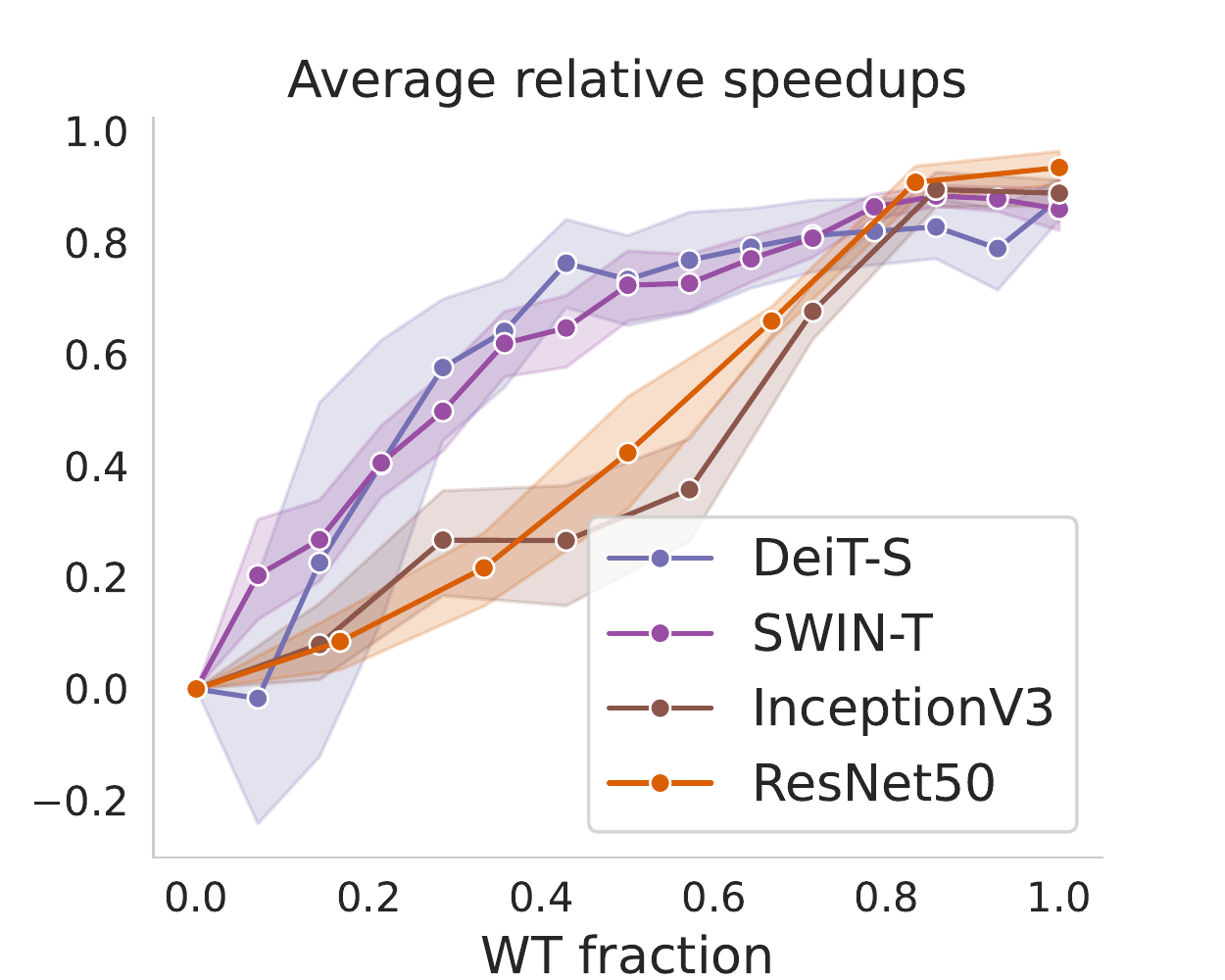}\\[-1.5mm] 
\end{tabular}
\end{center}
\vspace{-3mm}
\caption{\emph{Transfer learning and convergence speed.}
\textbf{left}: Validation curves of \resnetfifty and \deitsmall on \chexpert using a constant learning rate. \textbf{right}: Relative convergence speedups as a function of WT transferred layers. As we transfer more layers, the convergence speed of CNNs see increases linearly with the depth, while for ViTs rapid increases are observed for the first half of the network followed by a plateau.
}
\label{fig:convergence}
\vspace{-4mm}
\end{figure}

Finally, we investigate the impact of transfer learning on convergence speed.
Validation curves in Figure \ref{fig:convergence} demonstrate the speed-up from transfer learning, which we measure in the last panel.
We observe that convergence speed monotonically increases with the number of WT layers, in line with the finding of \cite{transfusion}.
Furthermore, we observe that CNNs converge faster at a roughly linear rate as we include more WT layers, while vision transformers see a rapid increase in convergence speed for the first half of the network but diminishing returns are observed after that.

\section{Discussion}
\label{discussion}
\paragraph{Related work.}
Recent investigations have looked into which situations transfer learning works and when it fails \cite{he2019rethinking,kornblith2019better, ngiam2018domain, geirhos2018imagenet}.
Kornblith \etal \cite{kornblith2019better} illustrate that transferred features may be less generally applicable than previously thought. 
In \cite{he2019rethinking}, He \etal show that transfer learning does not necessarily result in performance improvements, even for similar tasks. 
Yosinski \etal \cite{yosinski2014transferable} found early features tend to be more general, and as we move deeper into the network the features become more task-specific.
They also showed, along with Azizpour \etal, that benefits from transfer learning diminish as the distance between the source and target domain increases \cite{azizpour2015factors, yosinski2014transferable}.
Mustafa \etal showed that transfer learning using large architectures and orders of magnitude larger natural image pre-training datasets can yield significant improvements in medical imaging domains \cite{mustafa2021supervised}.

Most relevant to our work is the work of Raghu \etal \cite{transfusion}, which investigates transfer learning from \imagenet to \chexpert and a large proprietary retinopathy dataset similar to \aptos.
While they found that transfer from \imagenet provided a speed-up during training, they observed very little performance gains from transfer learning.
The authors argued that the main benefits of transfer learning to the medical domain is not due to feature reuse, but rather to the weight statistics 
and over-parameterization of the models.
Neyshabur \etal followed up their work, claiming that the observed speed-ups are due to the initialization and the low-level statistics of the data \cite{neyshabur2020being}.

In this paper, we go beyond the previous works, providing a more comprehensive analysis that delves into feature reuse \emph{within the network}.
We add clarity to the previous findings by exploring a wider range of datasets, showing that transfer learning is indeed effective for smaller datasets.
We consider new angles that the previous works ignored -- \eg the role of inductive biases and how transfer learning works for ViTs, as well as the how domain distance factors into transfer learning for medical images.

\vspace{-2mm}
\paragraph{Factors of transferability.}

In this work, we paint a more complete picture 
of transfer learning to the medical domain 
by considering more medical modalities, data sizes, and the model's capacity and inductive bias.
It is our conclusion that, for the majority of situations, transfer learning from \imagenet yields significant performance gains.
Our findings do not contradict those of \cite{transfusion, neyshabur2020being}, rather, we show that they uncovered an isolated case where the yields from transfer learning are minimal and feature reuse is less important.

We identify four factors that influence transfer learning from \imagenet to the medical domain. 
The data size and distance from the source domain are important factors that should not be overlooked.
Smaller datasets always benefit from transfer learning, and so do datasets that are close to the source domain.
The model's capacity has a small effect, but inductive bias is another important factor -- the benefits from transfer learning are negatively correlated with the strength of the model's inductive biases.
Looking at the extremes from our study: \deits, with the weakest inductive bias, heavily depend on transfer learning across the board.
\resnets, the models primarily used in previous works with the strongest inductive bias, show only limited improvement for large datasets and datasets that are distant from \imagenet.
But when the data size is smaller (as is often the case in medical tasks) or more similar to \imagenet, even \resnet's benefits become significant.

\vspace{-2mm}
\paragraph{The role of feature reuse.}
The importance of feature reuse in transfer learning has also been recently questioned \cite{transfusion}.
In order to better understand what drives transfer learning, we examined feature reuse from a number of different angles.
Our main take-away is that \textit{when transfer learning works well, there is strong evidence of feature reuse}.
Beyond this, we characterized feature reuse within the network in a number of ways.
We identified that certain critical features are ``sticky'' and less prone to change through transfer learning -- though which particular features stick depends on the architecture.
We observed that early layers are most crucial for ViT performance, which reuse a mixture of local and global features learned on \imagenet to perform competitively.
ViT's inability to relearn these essential features on small medical data sizes explains their strong dependence on feature reuse.
We also found that this pattern of early feature reuse in ViTs means that later layers can be discarded without strongly affecting performance.
CNNs benefit differently from feature reuse.
In CNNs, feature reuse occurs more uniformly, marked by progressive improvements in the features as each layer adds complexity over the previous.
The slope of the improvement varies with data characteristics -- it can even become flat, as found in \cite{transfusion, neyshabur2020being}.
We confirmed that these differences are primarily associated with model's inductive bias, rather than capacity, through a series of ablations.

\vspace{-2mm}
\paragraph{Limitations and potential negative societal impact.}
An exhaustive study of the factors that impact transfer learning is impossible -- countless models and datasets could have been included.
Nevertheless, we tried to select relevant and representative datasets and model types, covering a more diverse selection than previously studied.
A potential pitfall of this work is the use of FID \cite{fid}, which may not provide a perfect measure of distance between datasets \cite{lucic2017gans, borji2019pros}.

Despite well-meaning intentions, applying deep learning to medical data opens the possibility of unanticipated negative impacts. 
Without proper consideration, models can learn to replicate unwanted biases in the data. 
Failures can erode the public trust, and models that operate on medical data must take care not to reveal patient information.

\section{Conclusions}
\label{conclusions}

In this work we evaluate the benefits from transfer learning when working with medical images and how feature reuse and other factors, like the dataset and model characteristics, affect its usefulness.
We show that when transfer learning works, it is because of increased reuse of learned representations, and that models with less inductive bias, small datasets  and datasets that are closer to \imagenet see greater gains from it. 
We demonstrate that models with low inductive bias rely on reuse of local representations, composed mainly in early layers, to perform competitively with models with high inductive bias, which benefit from feature reuse throughout the network, but often to a lesser extent.
Our work focuses on transfer to the medical domain, but we believe that our findings may apply to other domains, which we leave for future work.
\paragraph{Acknowlegements.}
This work was supported by the Wallenberg Autonomous Systems Program (WASP), Stockholm County (HMT 20200958), and
the Swedish Research Council (VR) 2017-04609.
We thank Emir Konuk for the thoughtful discussions.

{\small
\bibliographystyle{ieee_fullname}
\bibliography{References}
}

\clearpage
\begin{appendices}

\begin{figure}[t]
\begin{center}
\begin{tabular}{@{}c@{}c@{}c@{}c@{}}
    \includegraphics[width=0.25\columnwidth]{images/similarity/WS-APTOS2019-deit_small-tul_12-trained_to_init} &
    \includegraphics[width=0.25\columnwidth]{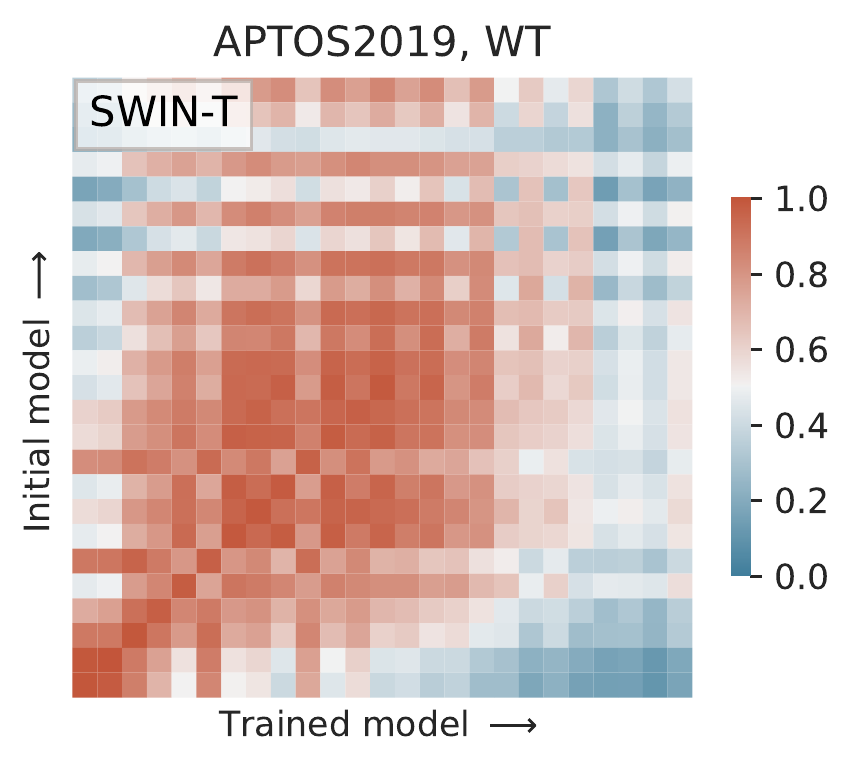} &
    \includegraphics[width=0.25\columnwidth]{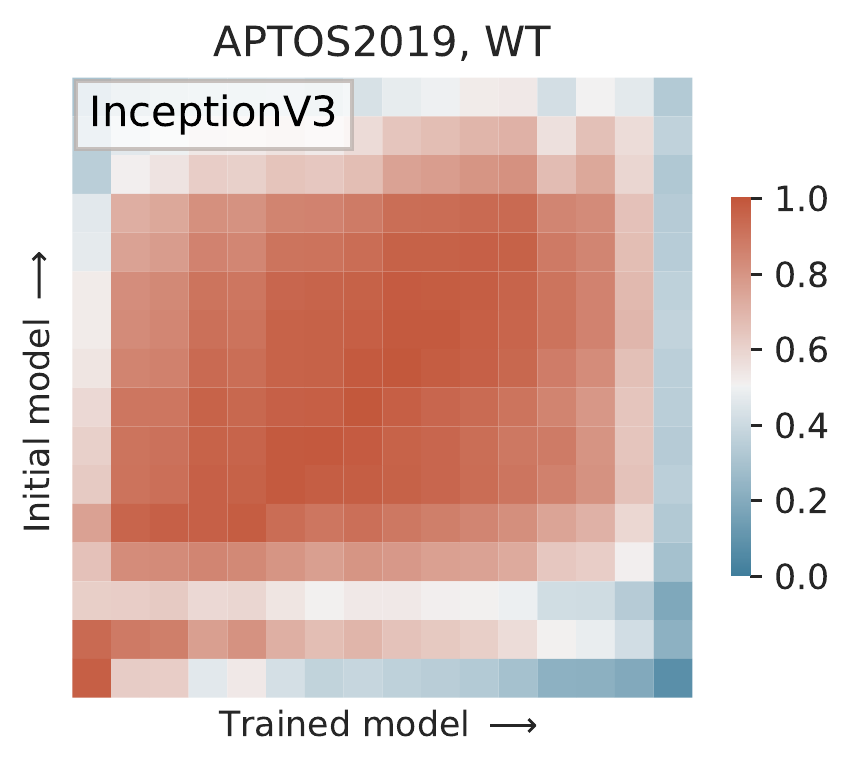} &
    \includegraphics[width=0.25\columnwidth]{images/similarity/WS-APTOS2019-resnet50-tul_4-trained_to_init} \\[-1.5mm] 
    \includegraphics[width=0.25\columnwidth]{images/similarity/WS-DDSM-deit_small-tul_12-trained_to_init} &
    \includegraphics[width=0.25\columnwidth]{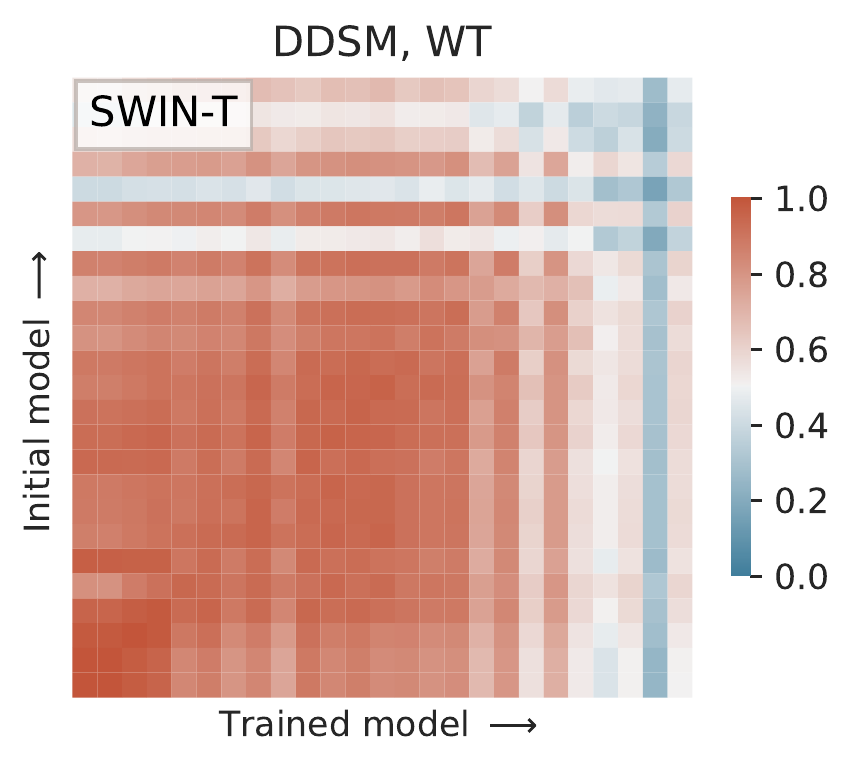} &
    \includegraphics[width=0.25\columnwidth]{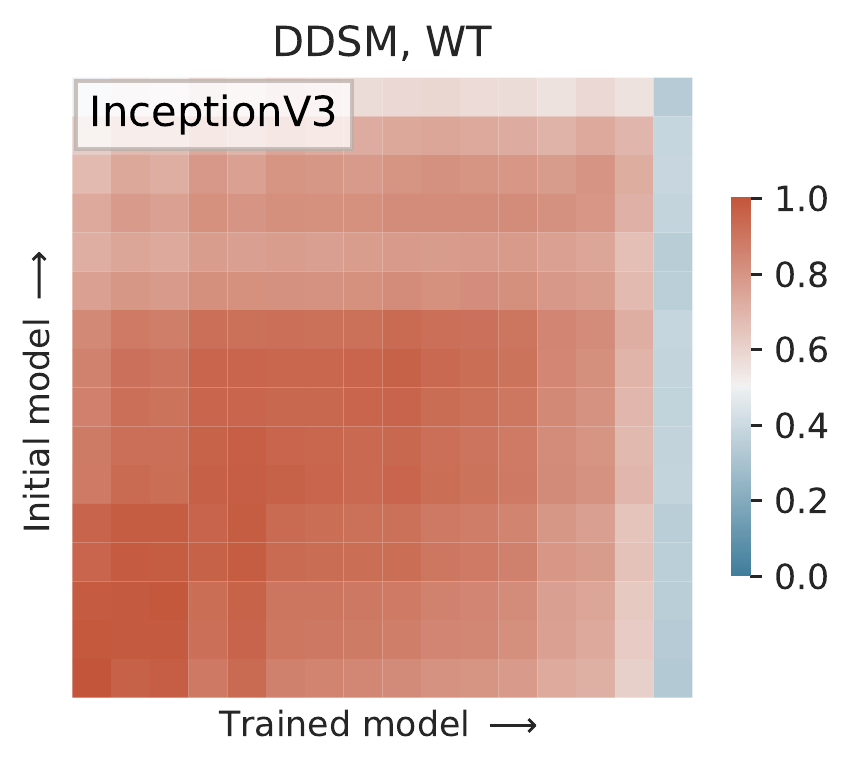} &
    \includegraphics[width=0.25\columnwidth]{images/similarity/WS-DDSM-resnet50-tul_4-trained_to_init} \\[-1.5mm] 
    \includegraphics[width=0.25\columnwidth]{images/similarity/WS-ISIC2019-deit_small-tul_12-trained_to_init} &
    \includegraphics[width=0.25\columnwidth]{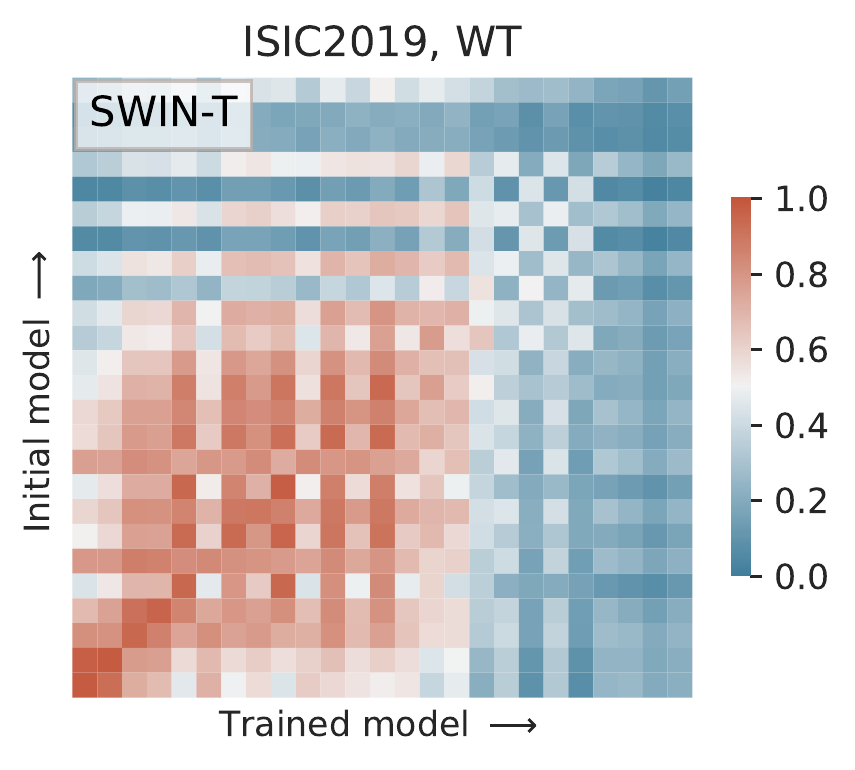} &
    \includegraphics[width=0.25\columnwidth]{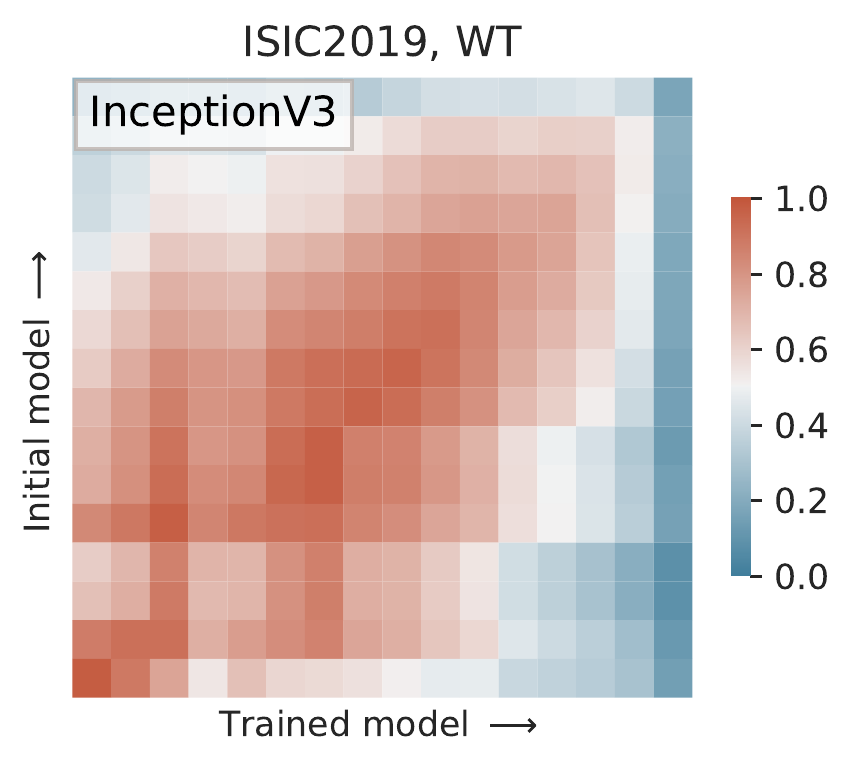} &
    \includegraphics[width=0.25\columnwidth]{images/similarity/WS-ISIC2019-resnet50-tul_4-trained_to_init} \\[-1.5mm] 
    \includegraphics[width=0.25\columnwidth]{images/similarity/WS-CheXpert-deit_small-tul_12-trained_to_init} &
    \includegraphics[width=0.25\columnwidth]{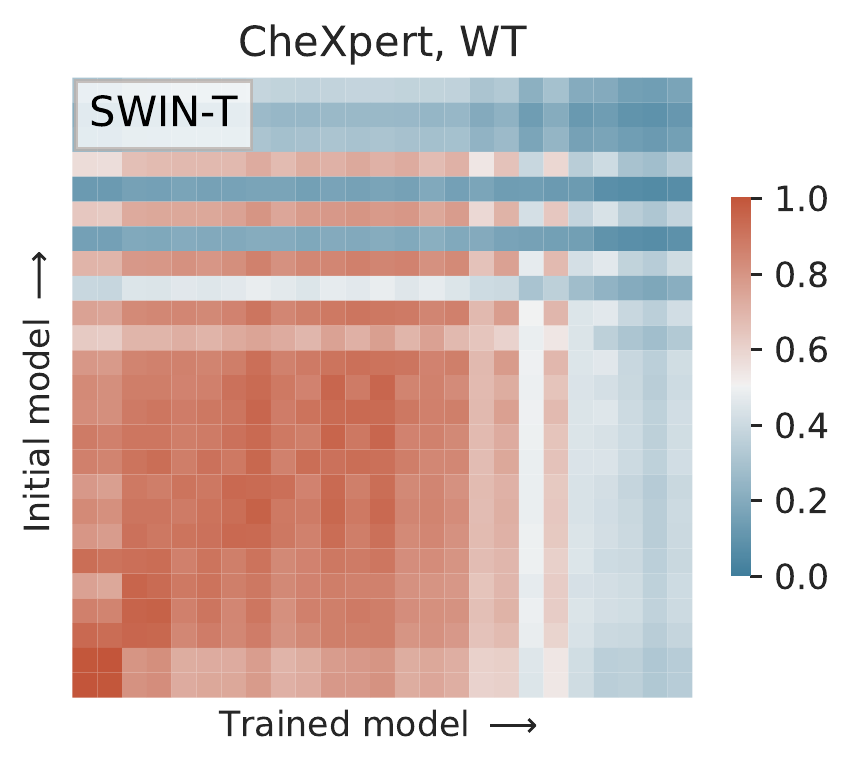} &
    \includegraphics[width=0.25\columnwidth]{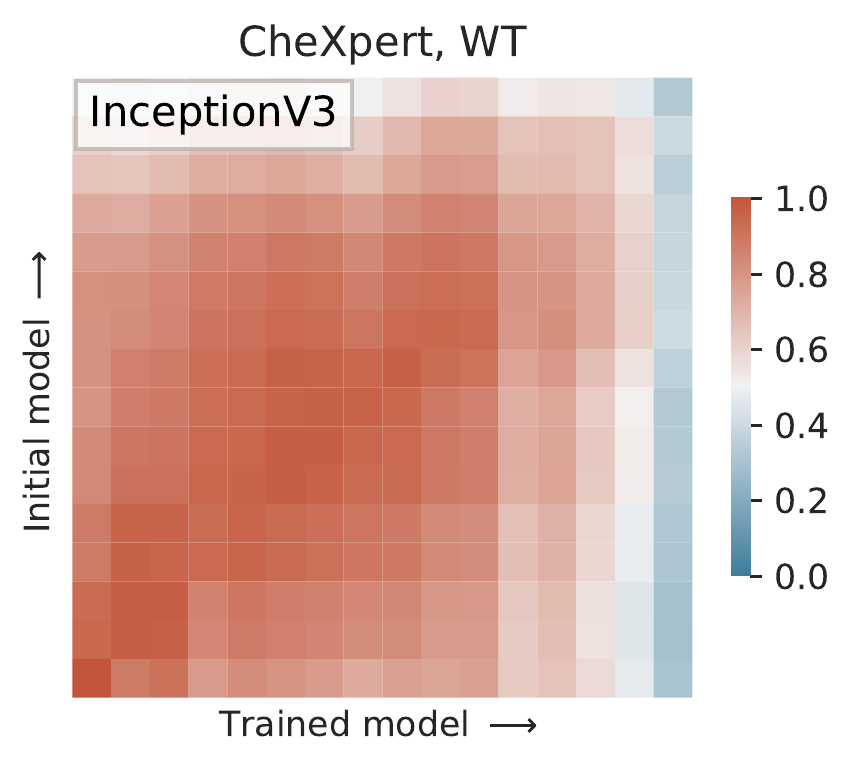} &
    \includegraphics[width=0.25\columnwidth]{images/similarity/WS-CheXpert-resnet50-tul_4-trained_to_init} \\[-1.5mm] 
    \includegraphics[width=0.25\columnwidth]{images/similarity/WS-Camelyon-deit_small-tul_12-trained_to_init} &
    \includegraphics[width=0.25\columnwidth]{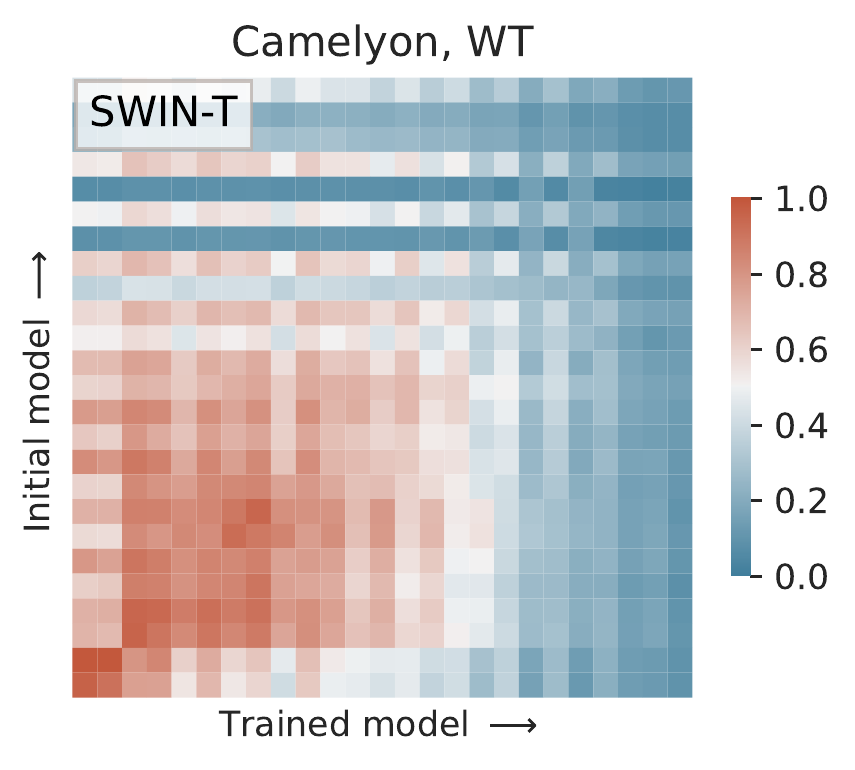} &
    \includegraphics[width=0.25\columnwidth]{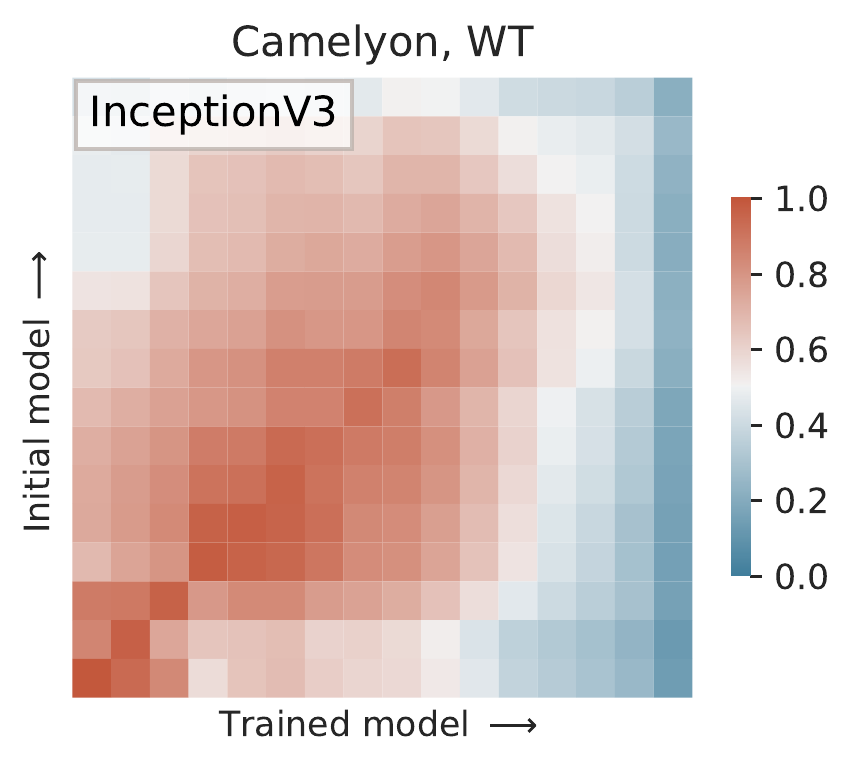} &
    \includegraphics[width=0.25\columnwidth]{images/similarity/WS-Camelyon-resnet50-tul_4-trained_to_init} \\[-1.5mm] 
\end{tabular}
\end{center}
\vspace{-3mm}
\caption{\emph{Feature similarity between the initial and the fine-tuned model for WT initialization.} CKA feature similarity comparison between WT initialized models before and after fine-tuning. Reported for each dataset (rows) and model type (columns). 
Evidently, models with low inductive bias (fist two columns) indicate strong feature reuse on the early layers. On the other hand, models with strong inductive biases (last two columns) appear a more uniform pattern throughout the network. Note that for all models, more data seem to affect more high-level features.}
\label{fig:similarity_apx_wt}
\vspace{-4mm}
\end{figure}

\begin{figure}[t]
\begin{center}
\begin{tabular}{@{}c@{}c@{}c@{}c@{}}
    \includegraphics[width=0.25\columnwidth]{images/similarity/WS_cross-APTOS2019-deit_small} &
    \includegraphics[width=0.25\columnwidth]{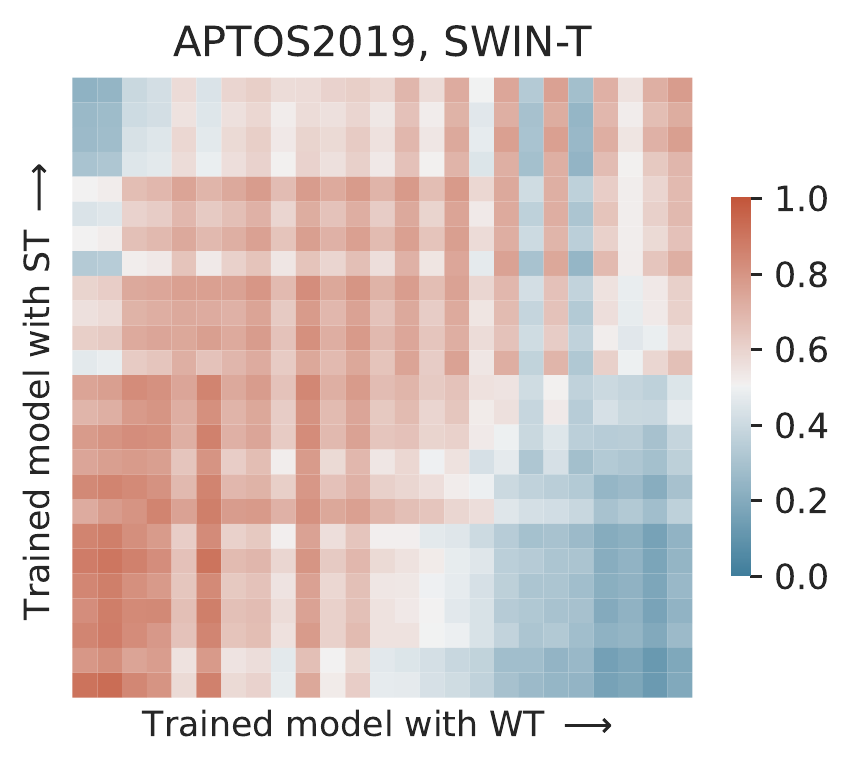} &
    \includegraphics[width=0.25\columnwidth]{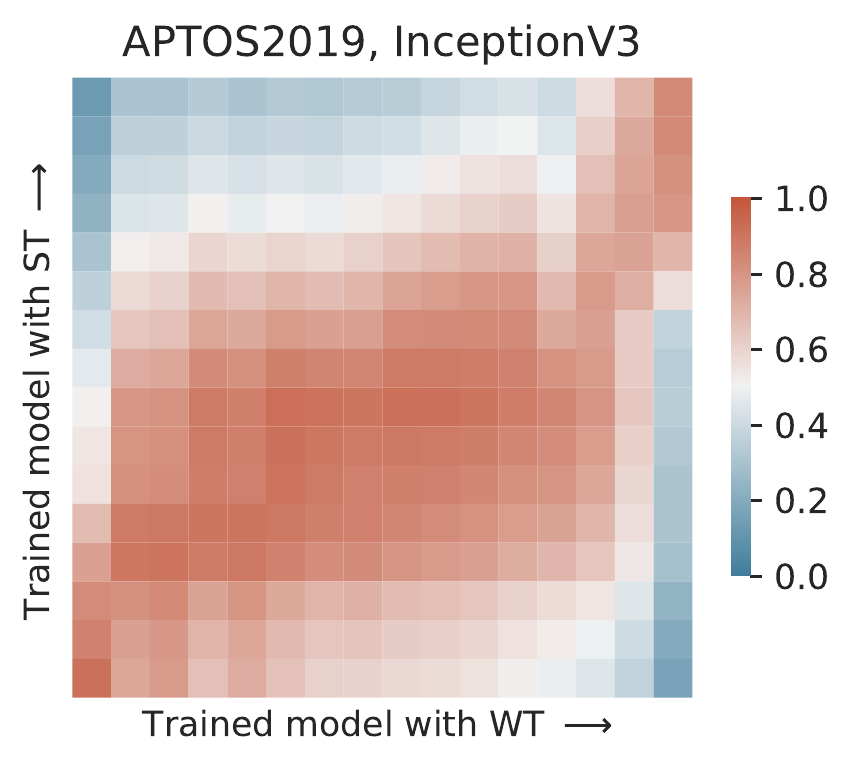} &
    \includegraphics[width=0.25\columnwidth]{images/similarity/WS_cross-APTOS2019-resnet50.pdf} 
    \\[-1.5mm]
    \includegraphics[width=0.25\columnwidth]{images/similarity/WS_cross-DDSM-deit_small} &
    \includegraphics[width=0.25\columnwidth]{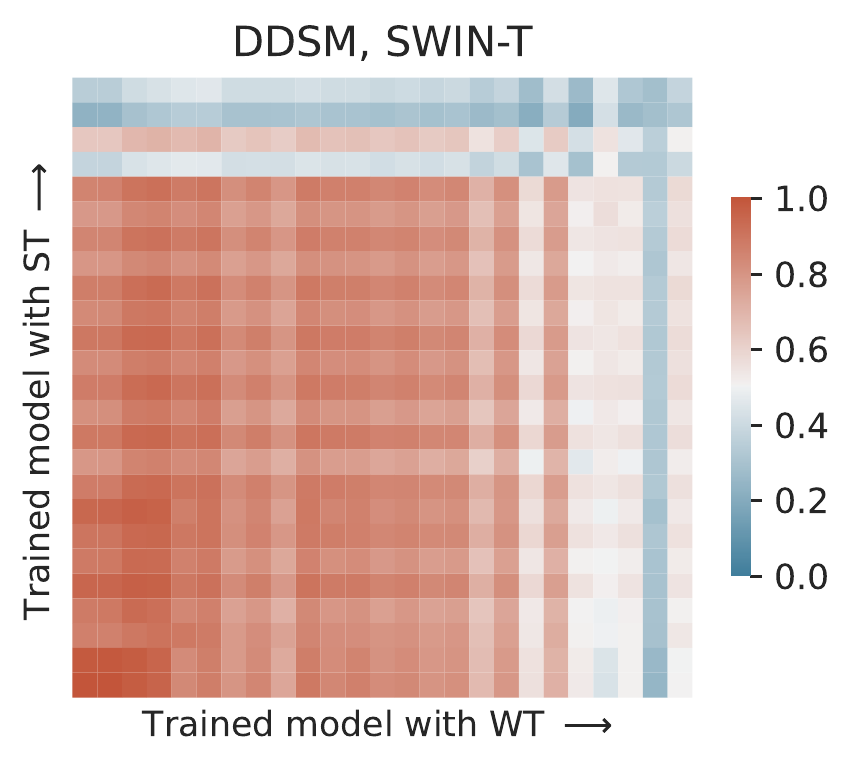} &
    \includegraphics[width=0.25\columnwidth]{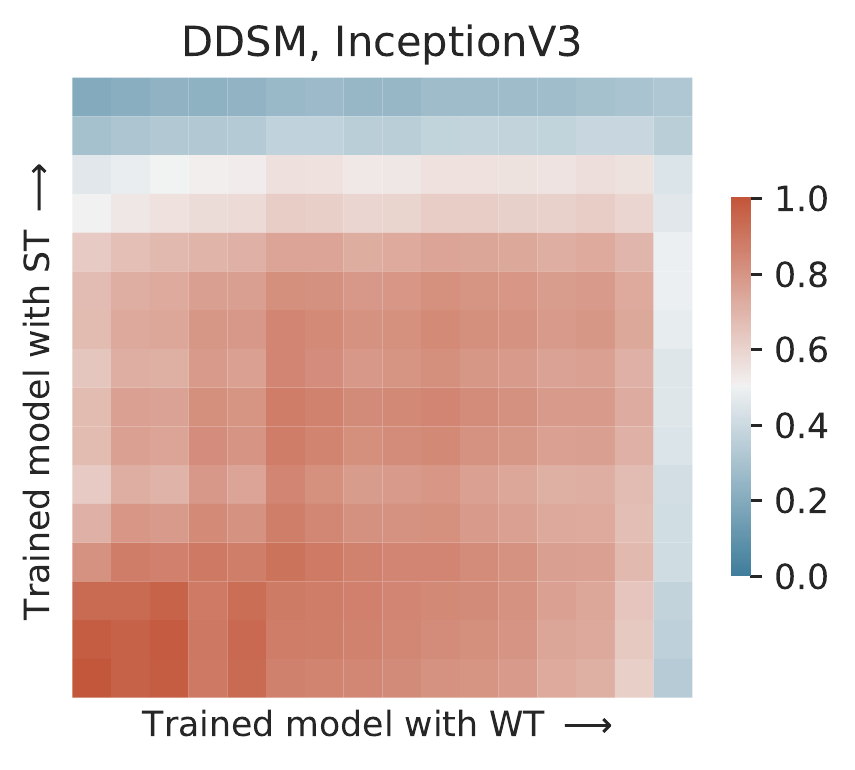} &
    \includegraphics[width=0.25\columnwidth]{images/similarity/WS_cross-DDSM-resnet50.pdf} 
    \\[-1.5mm]
    \includegraphics[width=0.25\columnwidth]{images/similarity/WS_cross-ISIC2019-deit_small} &
    \includegraphics[width=0.25\columnwidth]{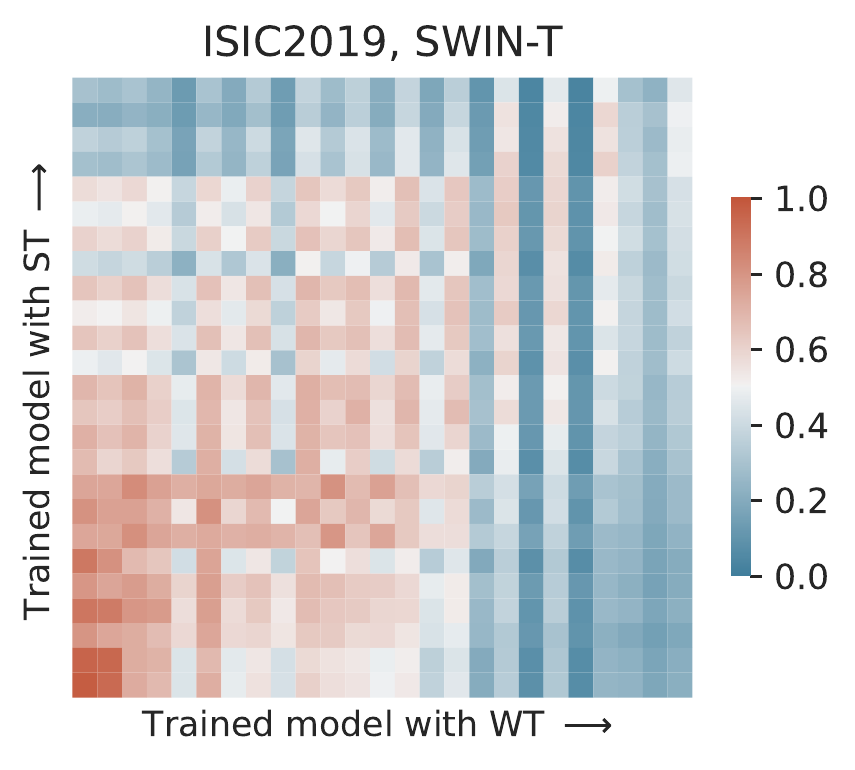} &
    \includegraphics[width=0.25\columnwidth]{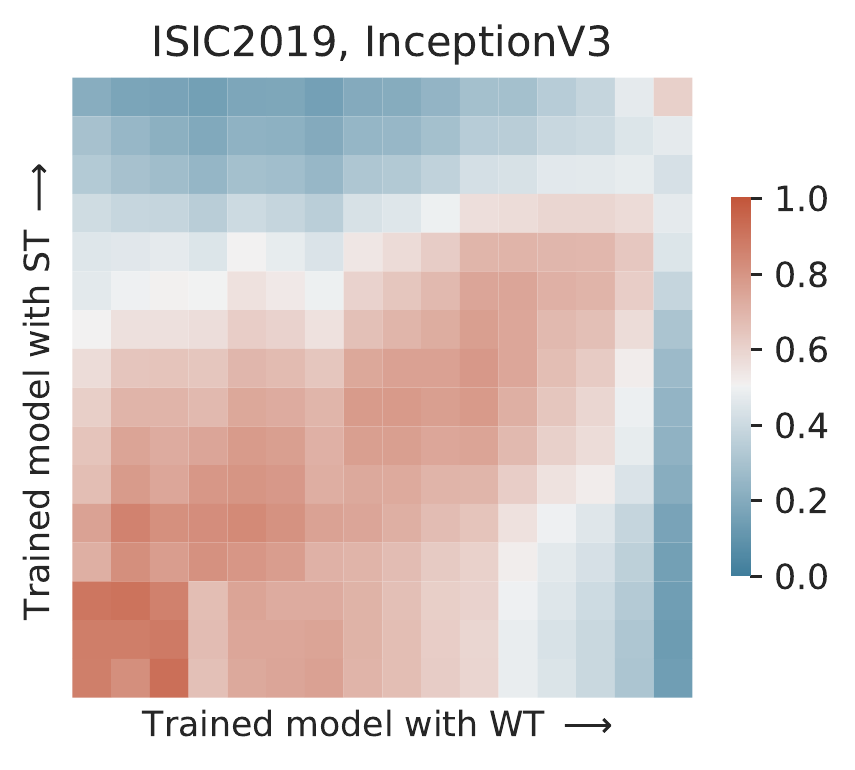} &
    \includegraphics[width=0.25\columnwidth]{images/similarity/WS_cross-ISIC2019-resnet50.pdf} 
    \\[-1.5mm]
    \includegraphics[width=0.25\columnwidth]{images/similarity/WS_cross-CheXpert-deit_small} &
    \includegraphics[width=0.25\columnwidth]{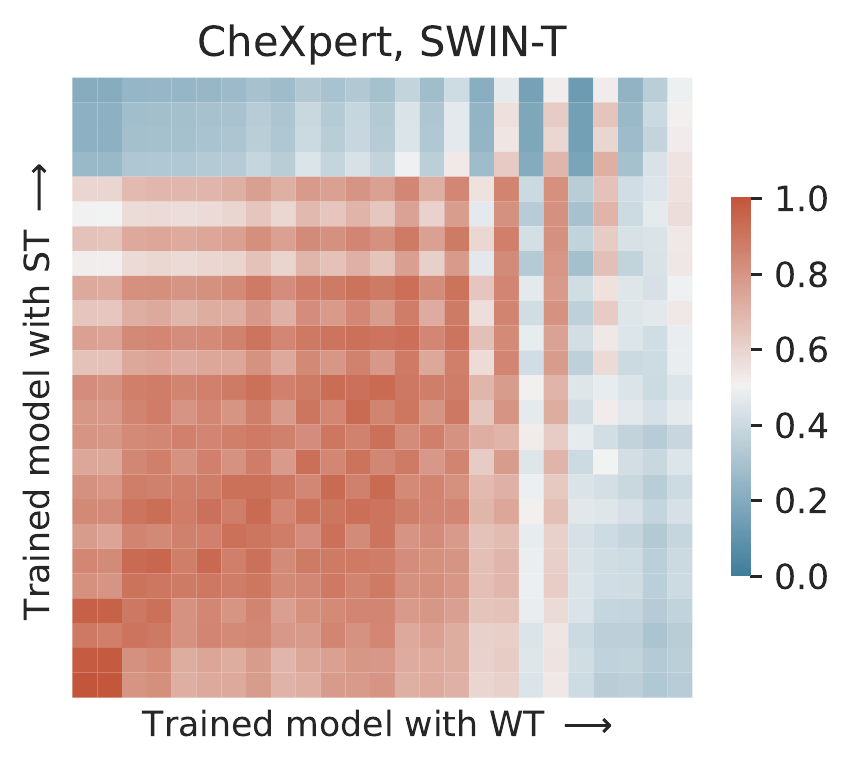} &
    \includegraphics[width=0.25\columnwidth]{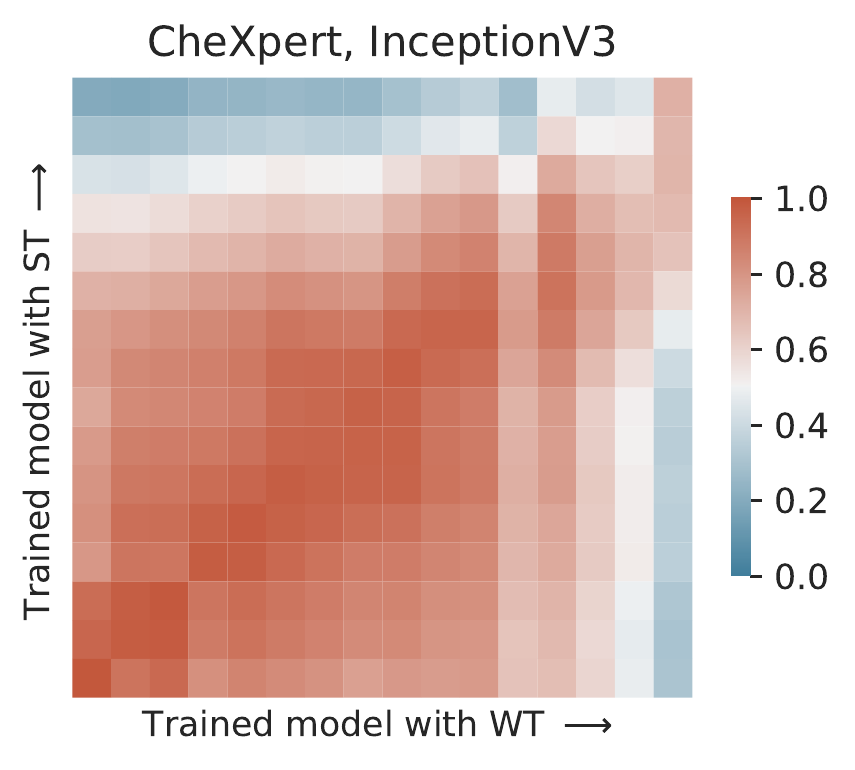} &
    \includegraphics[width=0.25\columnwidth]{images/similarity/WS_cross-CheXpert-resnet50.pdf} 
    \\[-1.5mm]
    \includegraphics[width=0.25\columnwidth]{images/similarity/WS_cross-Camelyon-deit_small} &
    \includegraphics[width=0.25\columnwidth]{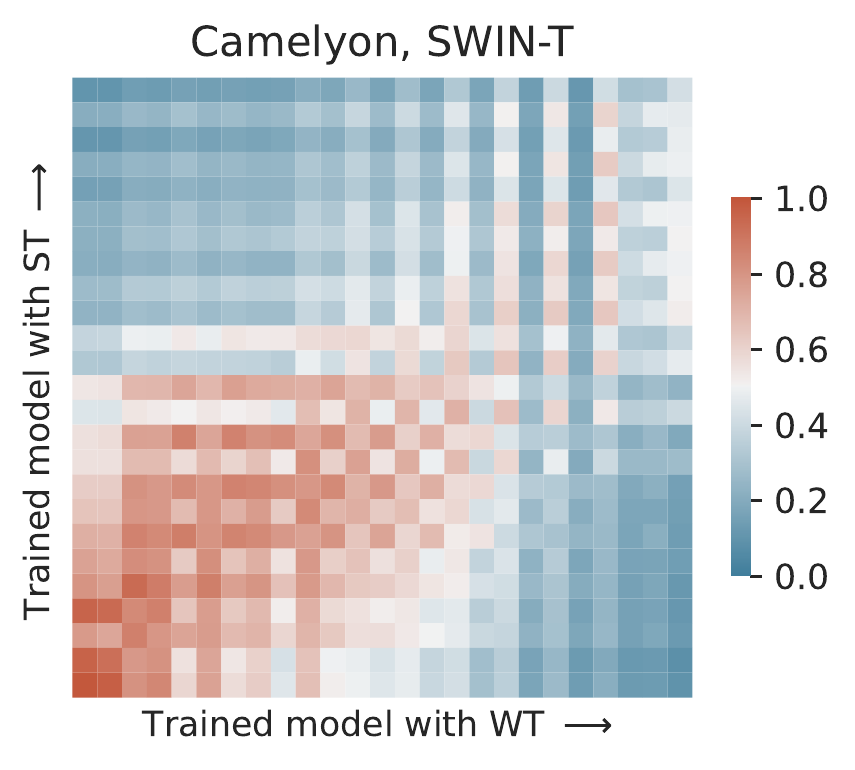} &
    \includegraphics[width=0.25\columnwidth]{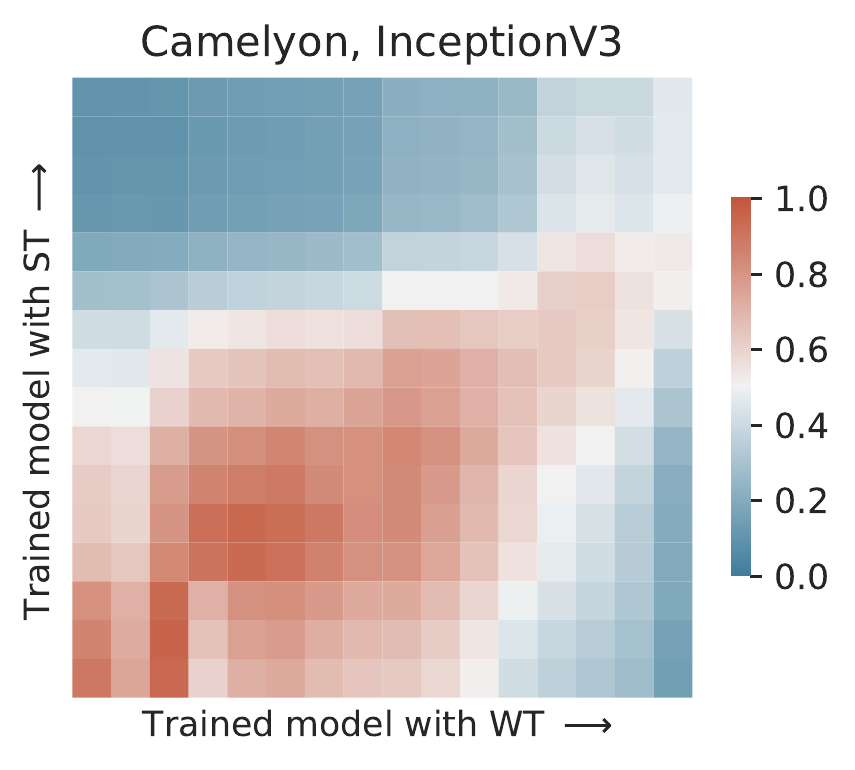} &
    \includegraphics[width=0.25\columnwidth]{images/similarity/WS_cross-Camelyon-resnet50.pdf} 
    \\[-1.5mm]    
\end{tabular}
\end{center}
\vspace{-3mm}
\caption{\emph{Feature similarity between WT and ST fine-tuned models.} CKA feature similarity comparison between WT and ST initialized models, calculated at each layer. Reported for each dataset (rows) and model type (columns).
The results indicate that for models with low inductive bias, like \deits, the early layers of ST-initialized networks have increased similarity with early-to-mid layer features from the WT-initialized model, suggesting that the ST-initialized model learns more global features in the early layers. On the other hand, models with strong inductive biases, like \resnetfifty, seem to learn similar features regardless their initialization, suggesting that inductive biases may somehow naturally lead to similar features.
}
\label{fig:cross_similarity_apx}
\vspace{-4mm}
\end{figure}

\begin{figure}[t]
\begin{center}
\begin{tabular}{@{}c@{}c@{}c@{}c@{}c@{}}
    \includegraphics[width=0.20\columnwidth]{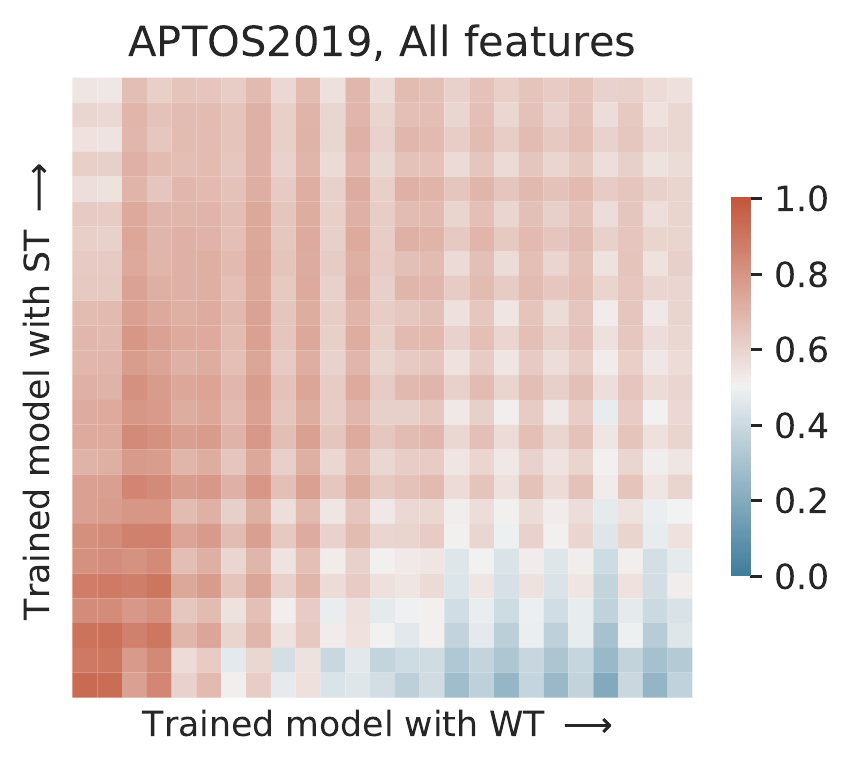} &
    \includegraphics[width=0.20\columnwidth]{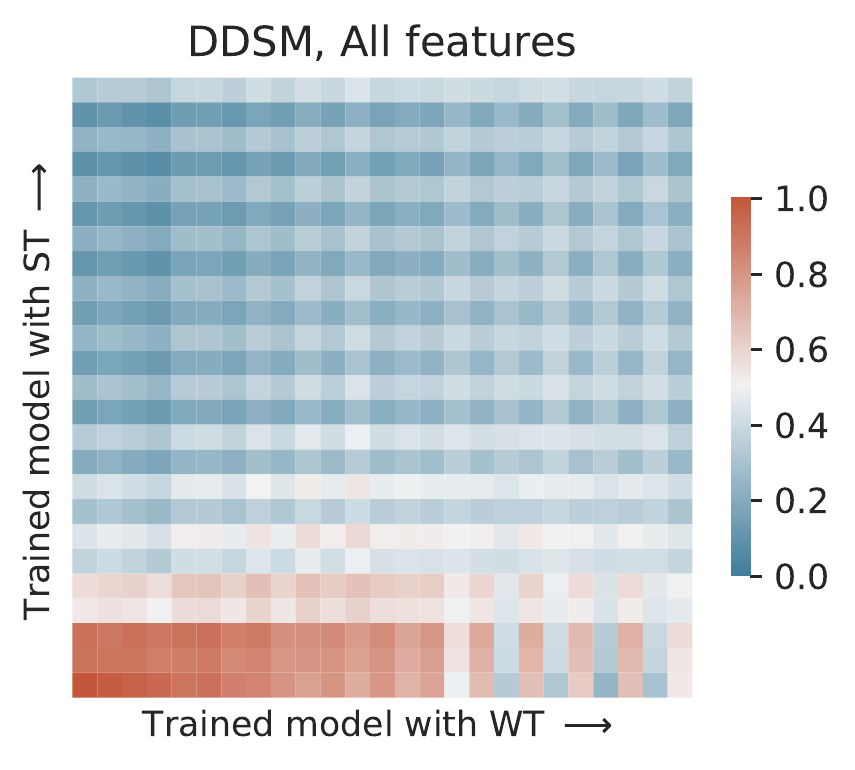} &
    \includegraphics[width=0.20\columnwidth]{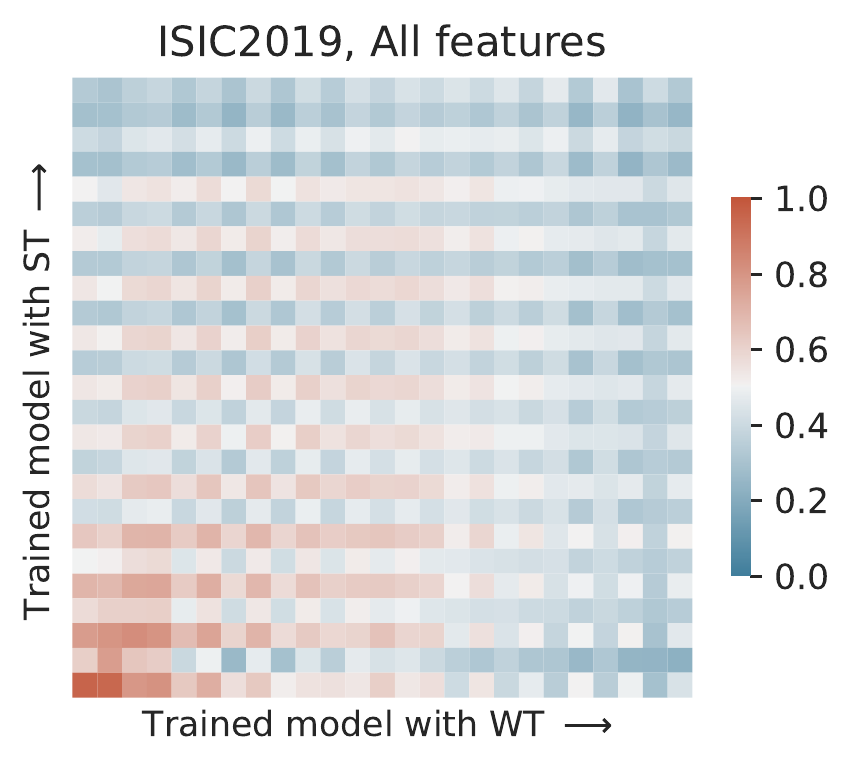} &
    \includegraphics[width=0.20\columnwidth]{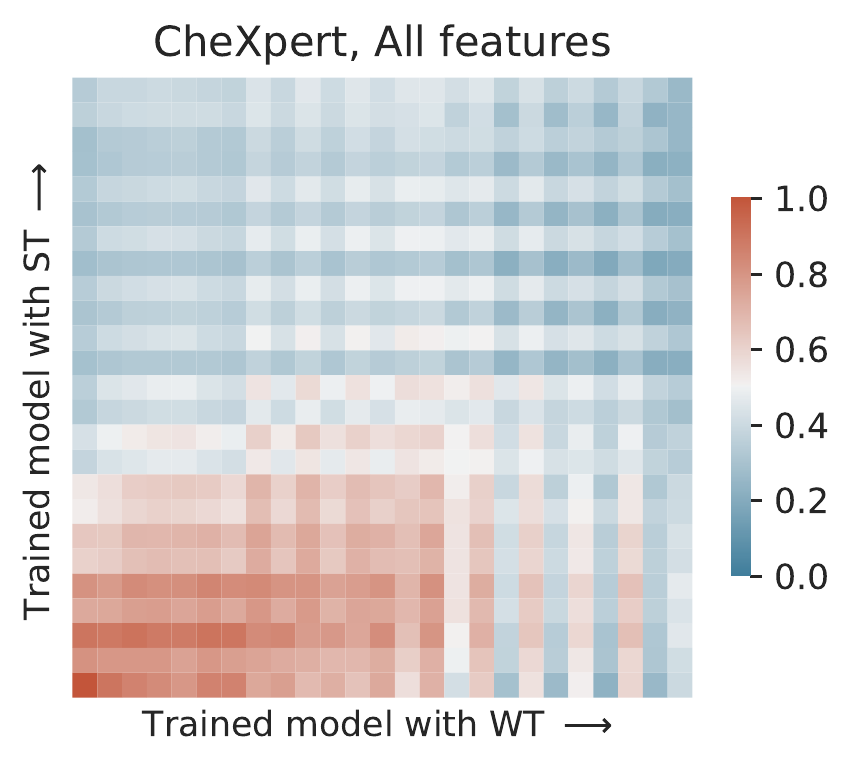} &
        \includegraphics[width=0.20\columnwidth]{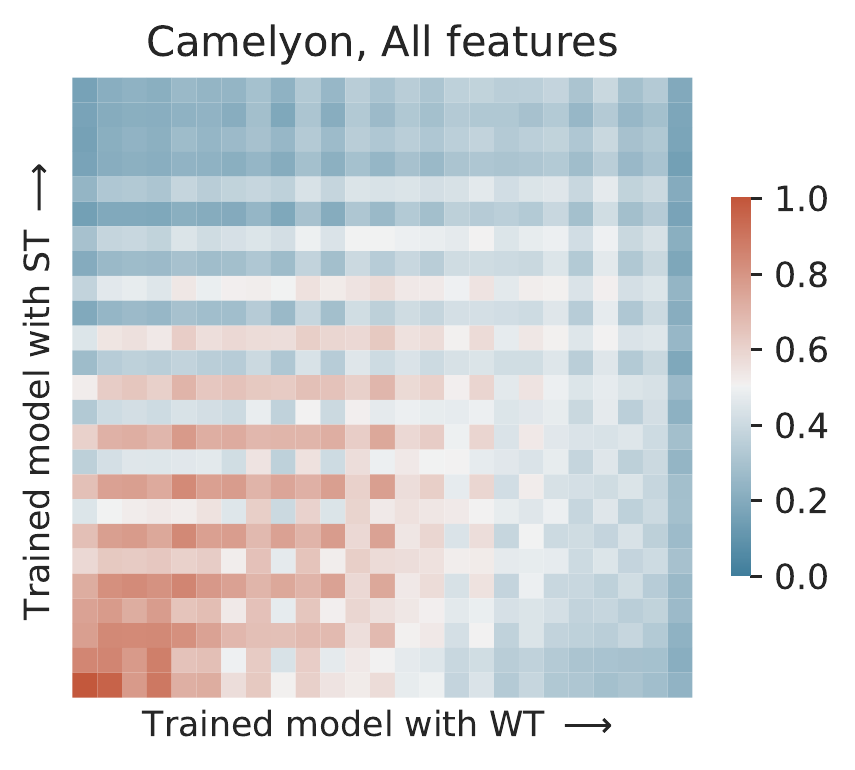} 
    \\[-1.5mm] 
    \includegraphics[width=0.20\columnwidth]{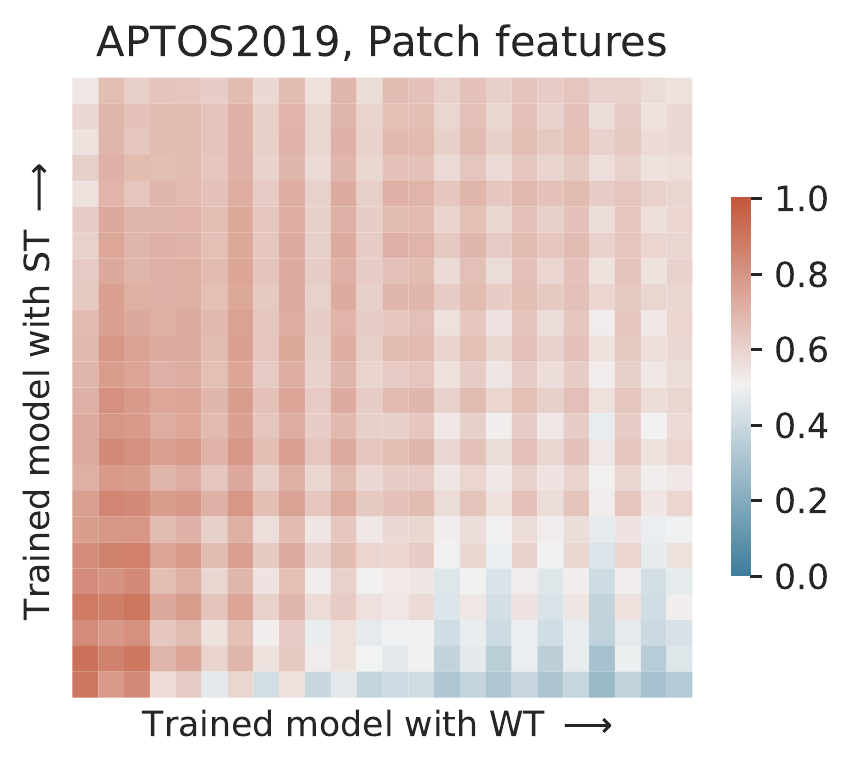} &
    \includegraphics[width=0.20\columnwidth]{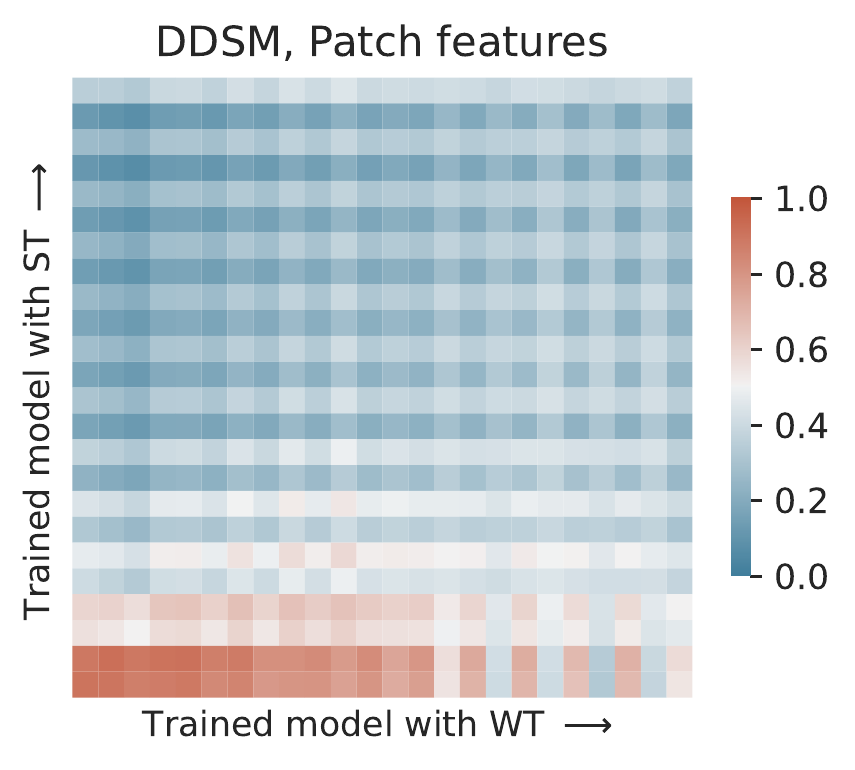} &
    \includegraphics[width=0.20\columnwidth]{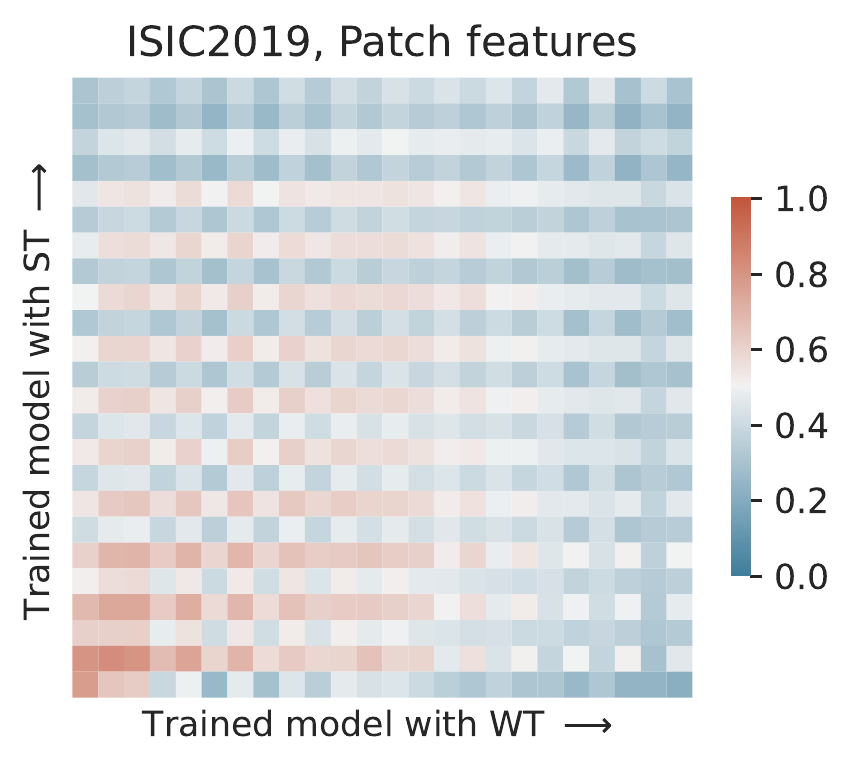} &
    \includegraphics[width=0.20\columnwidth]{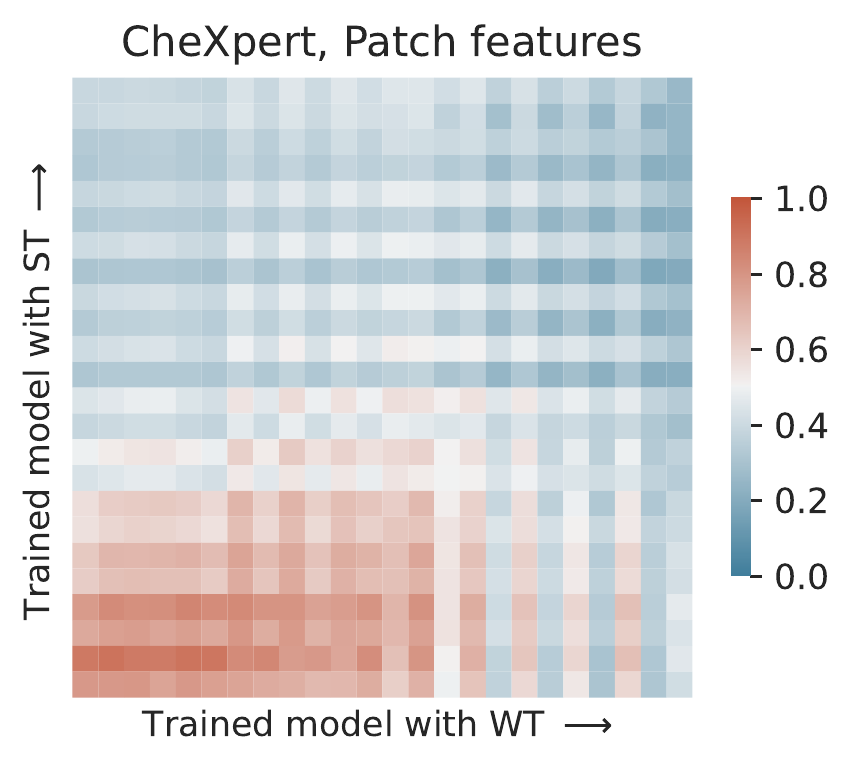} &
        \includegraphics[width=0.20\columnwidth]{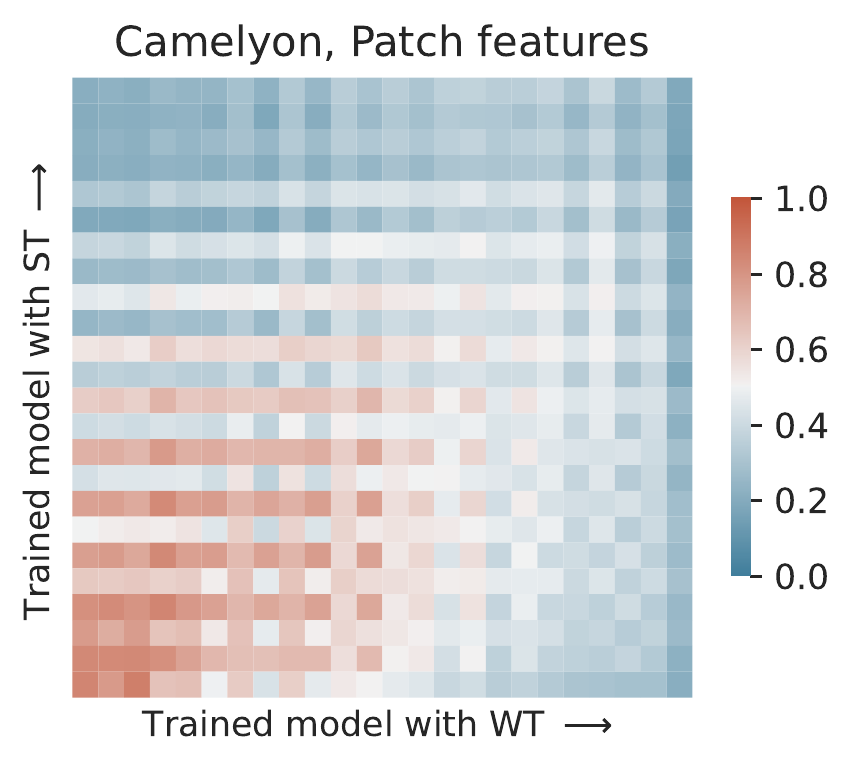} 
    \\[-1.5mm] 
    \includegraphics[width=0.20\columnwidth]{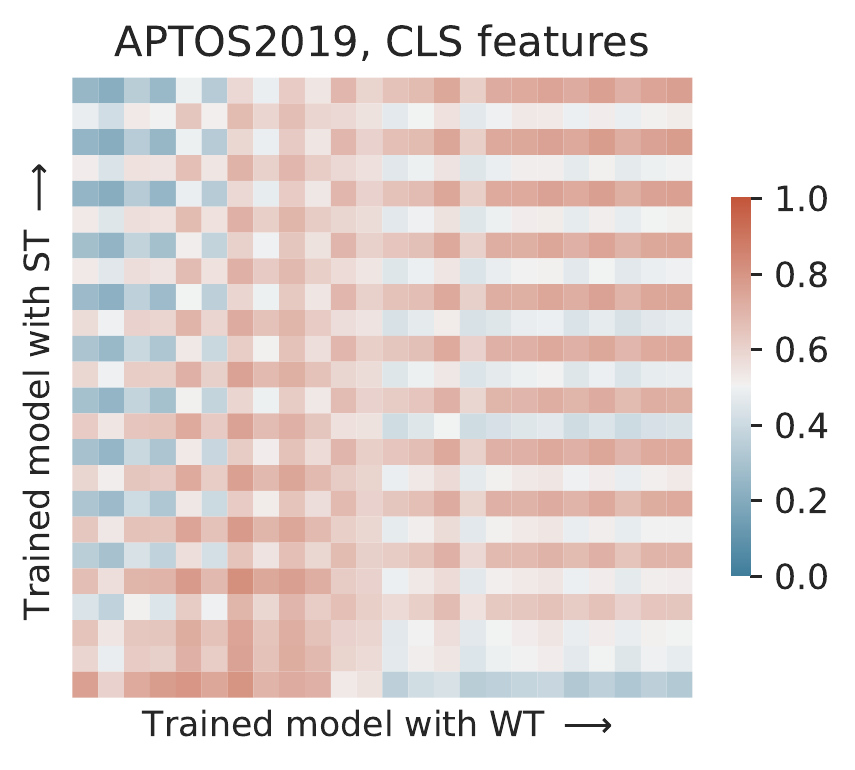} &
    \includegraphics[width=0.20\columnwidth]{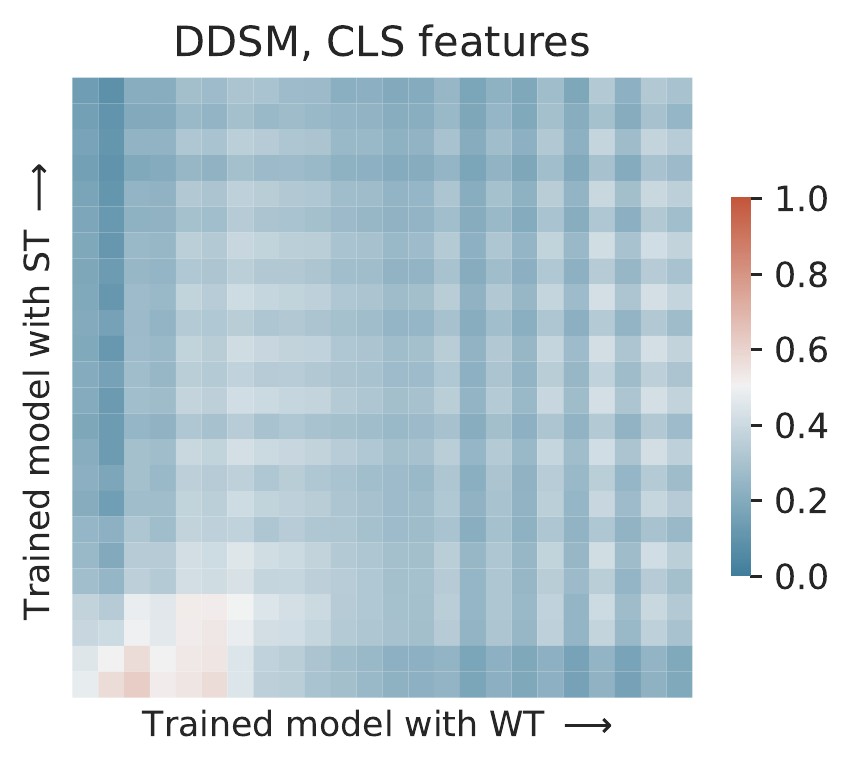} &
    \includegraphics[width=0.20\columnwidth]{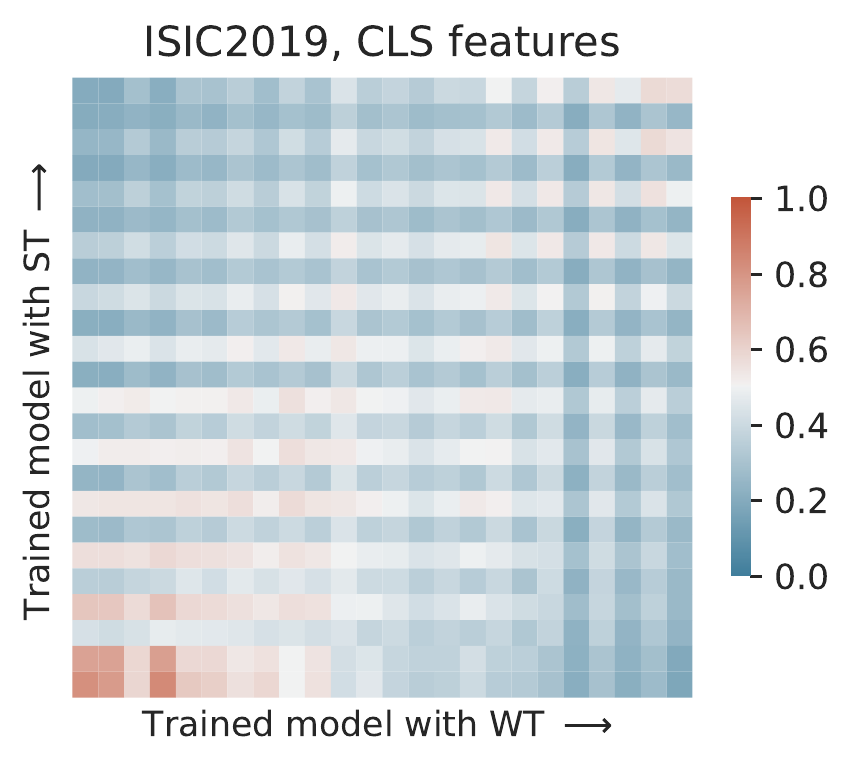} &
    \includegraphics[width=0.20\columnwidth]{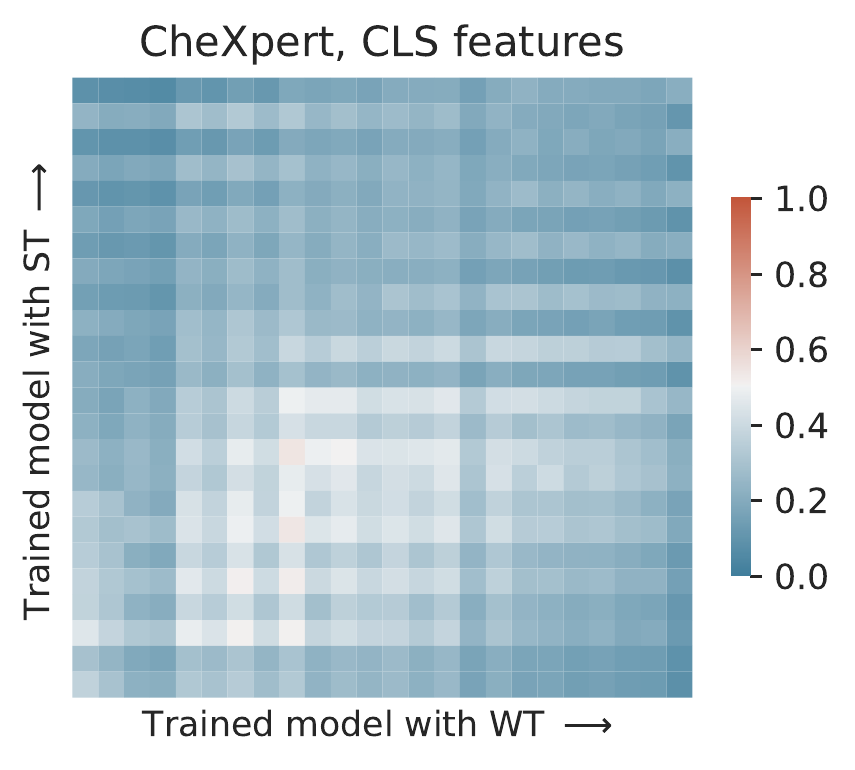} &
        \includegraphics[width=0.20\columnwidth]{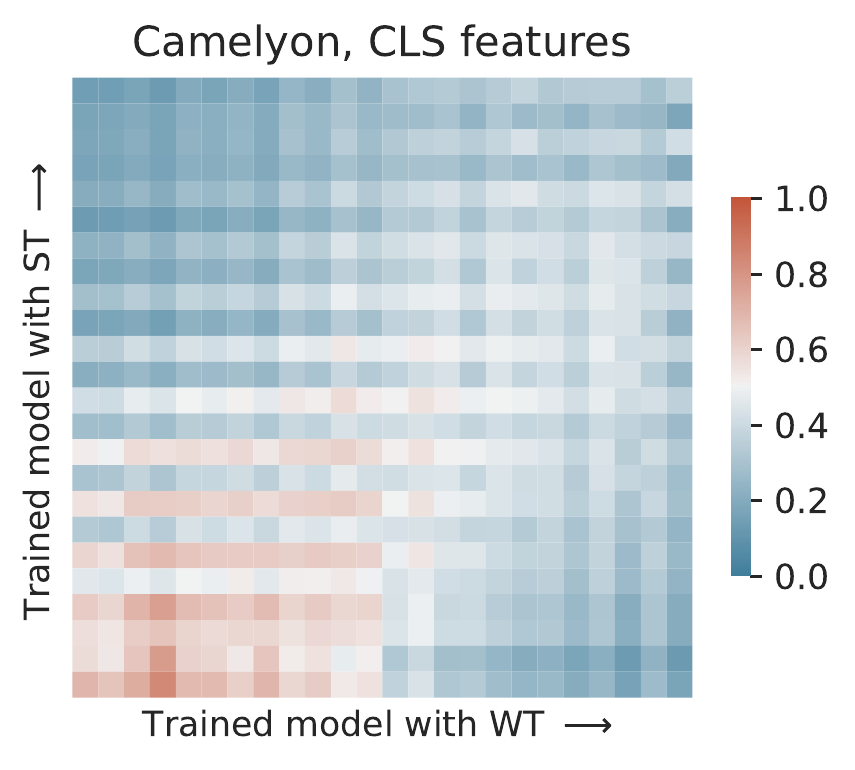} 
    \\[-1.5mm]   
\end{tabular}
\end{center}
\vspace{-3mm}
\caption{\emph{Feature similarity of WT and ST fine-tuned models for different feature-types.} CKA feature similarity comparison between WT and ST initialized models in \deitsmall, using three different token types (CLS, Patches, CLS + Patches), for each dataset (columns) and embedding type (rows).
Evidently, the emergence of global features in the early layers of ST initialized models is associated mainly with the patch tokens (which represent the image patches) whilst the CLS token learns different features depending on its initialization.
}
\label{fig:cross_similarity_deit_apx}
\vspace{-4mm}
\end{figure}

\paragraph{Appendix overview}
Here, we provide further details and results from the experiments carried out in this work. 
Section \ref{sec:sup-figures-feature-similarity} provides further experiment details and supplementary figures for the feature similarity experiments. Section \ref{sec:sup-figures-knn} for the layer-wise {\knn} evaluation experiments. Section \ref{sec:sup-figures-layerwise-importance} for the layer-wise re-initialization and Section \ref{appdx:L2experiment} for the \ltwo weight distance experiments. 
In Section \ref{appdx:mean_att_distance} we show the mean attended distance for \deitsmall using different initialization schemes.
In Sections \ref{sec:convergence-study} and \ref{appdx:capacity-study}, we provide additional details for the convergence speed and the architectures we used to evaluate the behaviour of different model capacities.
In Section \ref{appdx:wtst_details}, we describe in detail the modules used for the \wtst initialization schemes and the intermediate layers that we used for the re-initialization, \ltwo, \knn and representational similarity experiments. 
Finally, in Section \ref{sec:smaller-deit} we provide additional details for the 5-layer \deitsmall model and we show that it performs comparably with the full 12-layer model.

\section{Feature similarity}
\label{sec:sup-figures-feature-similarity}

\begin{figure}[t]
\begin{center}
\begin{tabular}{@{}c@{}c@{}c@{}c@{}}
    \includegraphics[width=0.25\columnwidth]{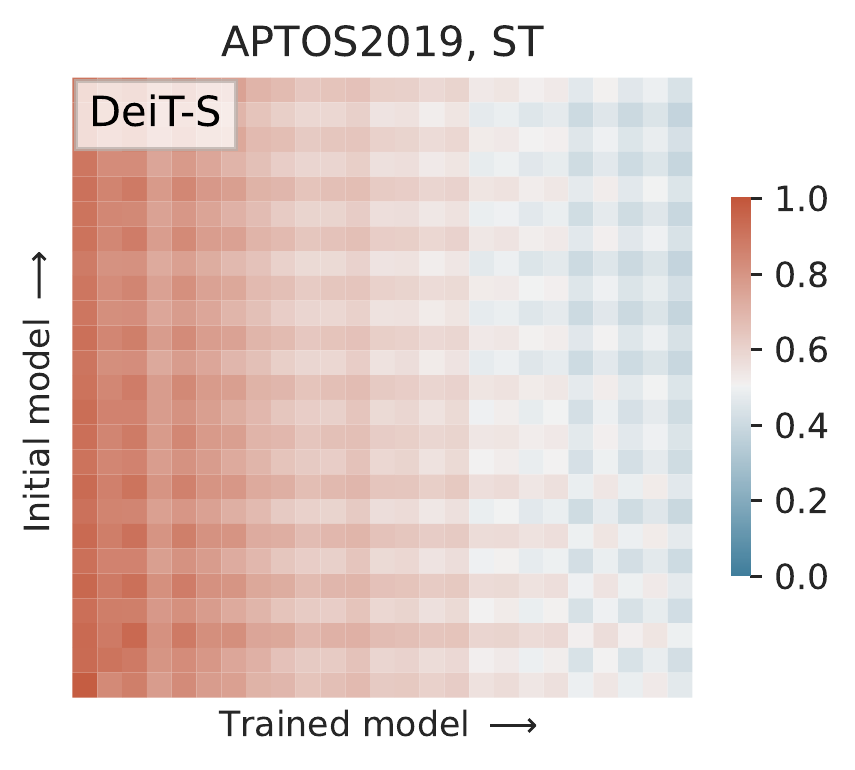} &
    \includegraphics[width=0.25\columnwidth]{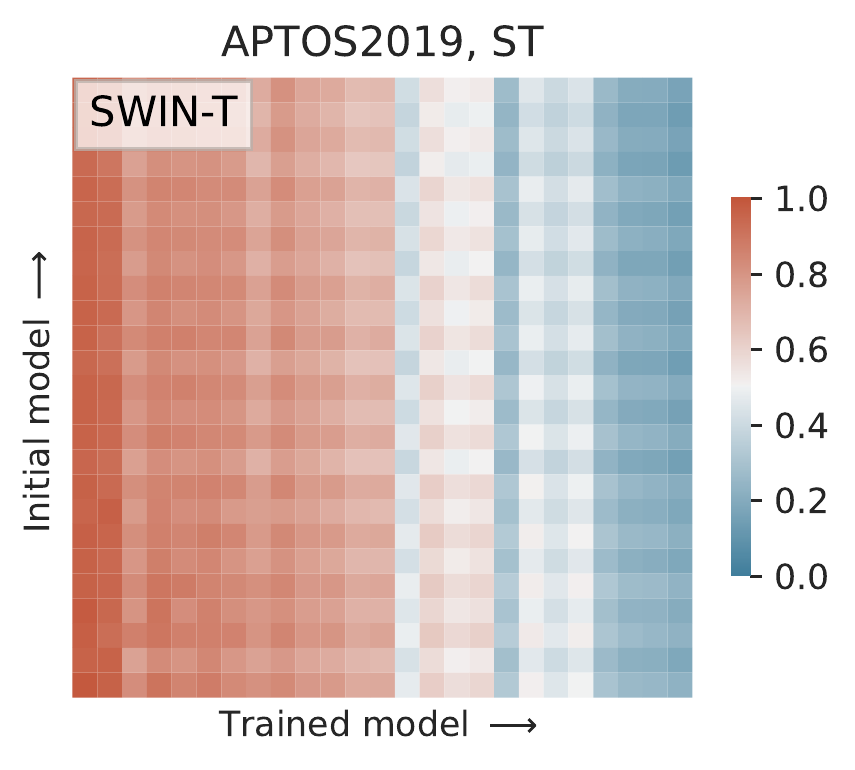} &
    \includegraphics[width=0.25\columnwidth]{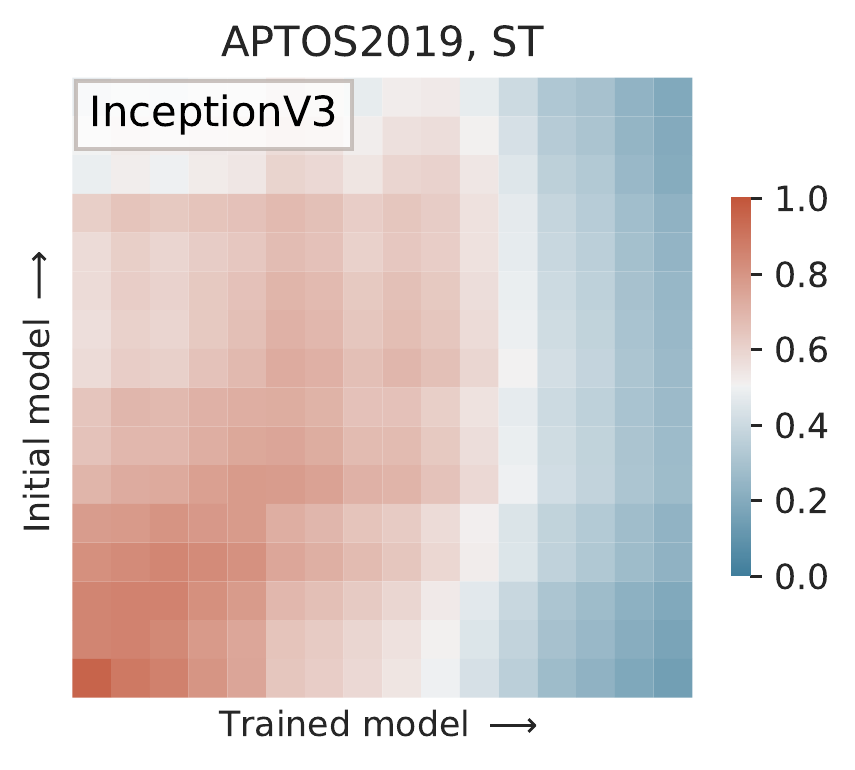} &    
    \includegraphics[width=0.25\columnwidth]{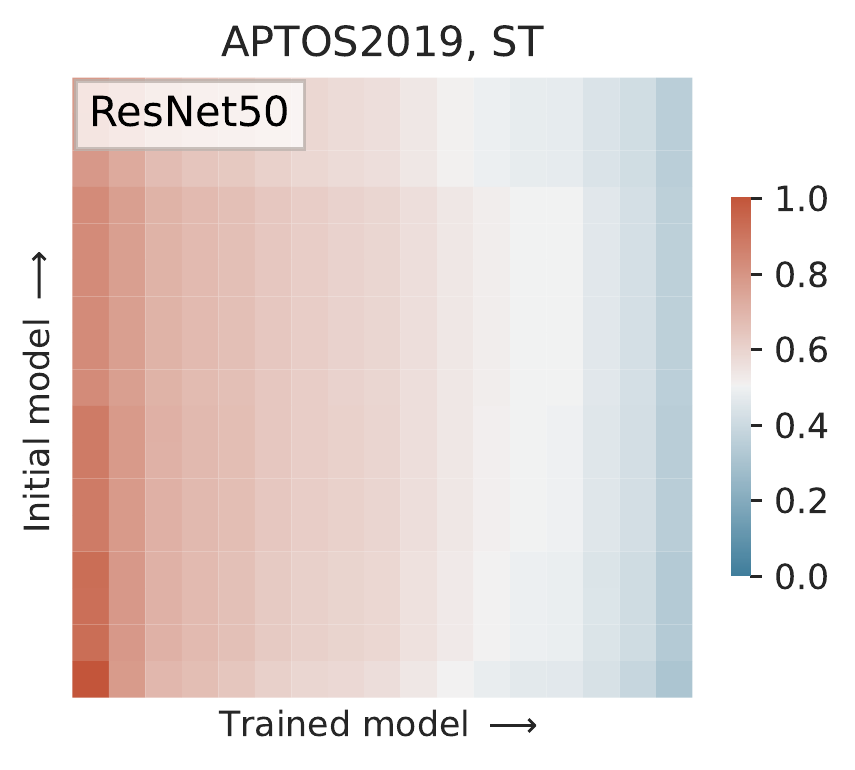} \\[-1.5mm] 
    \includegraphics[width=0.25\columnwidth]{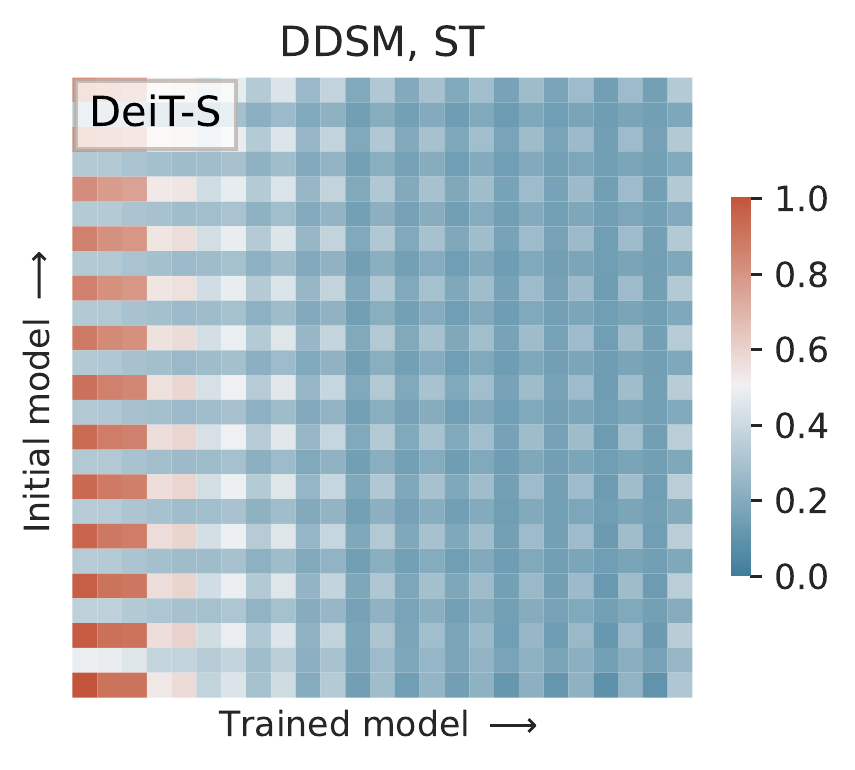} &
    \includegraphics[width=0.25\columnwidth]{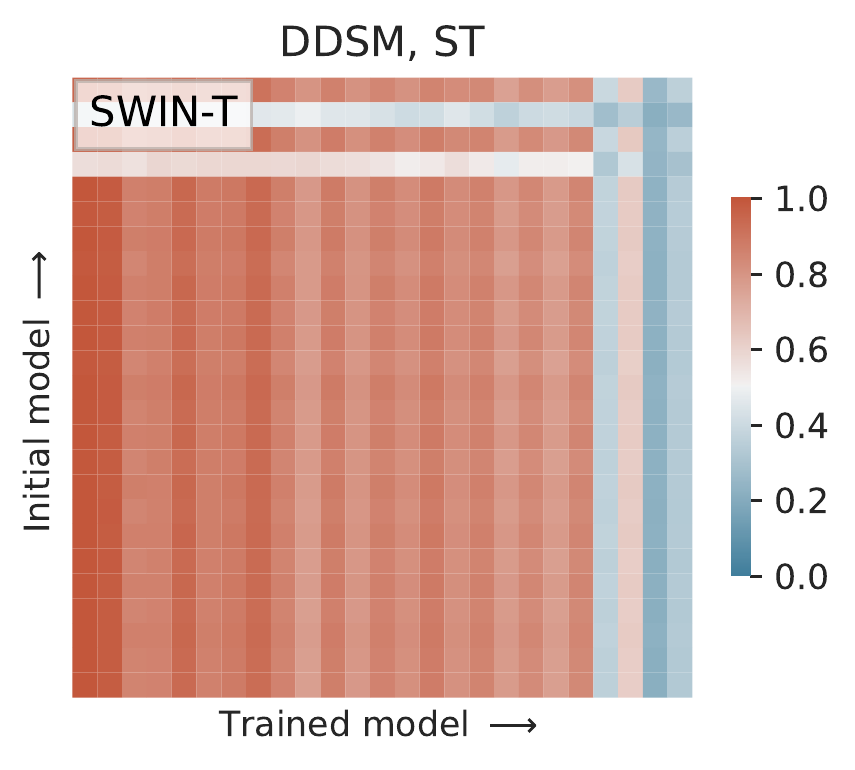} &
    \includegraphics[width=0.25\columnwidth]{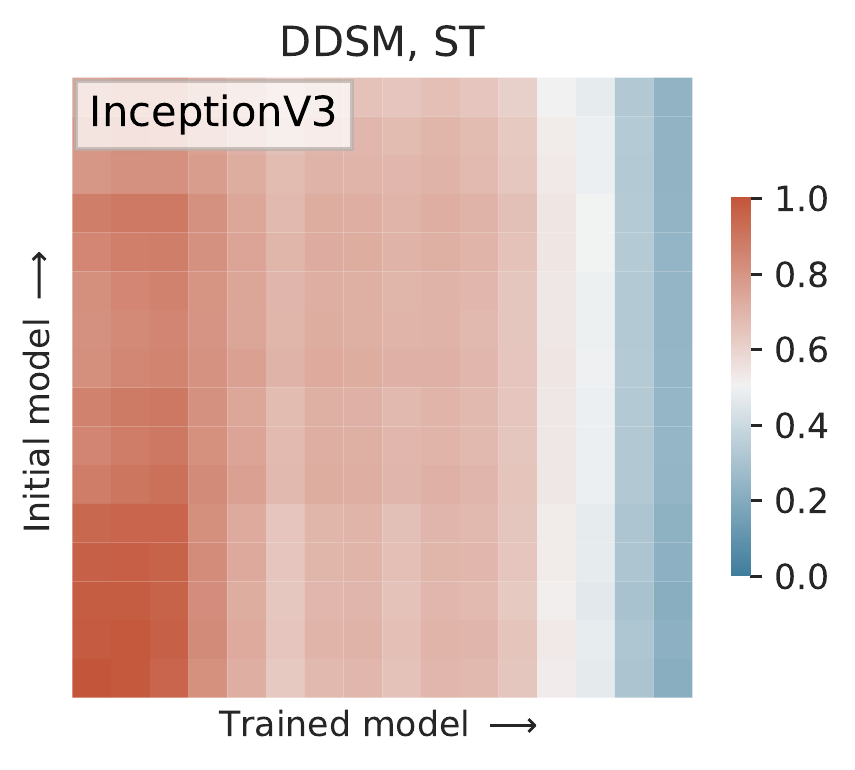} &    
    \includegraphics[width=0.25\columnwidth]{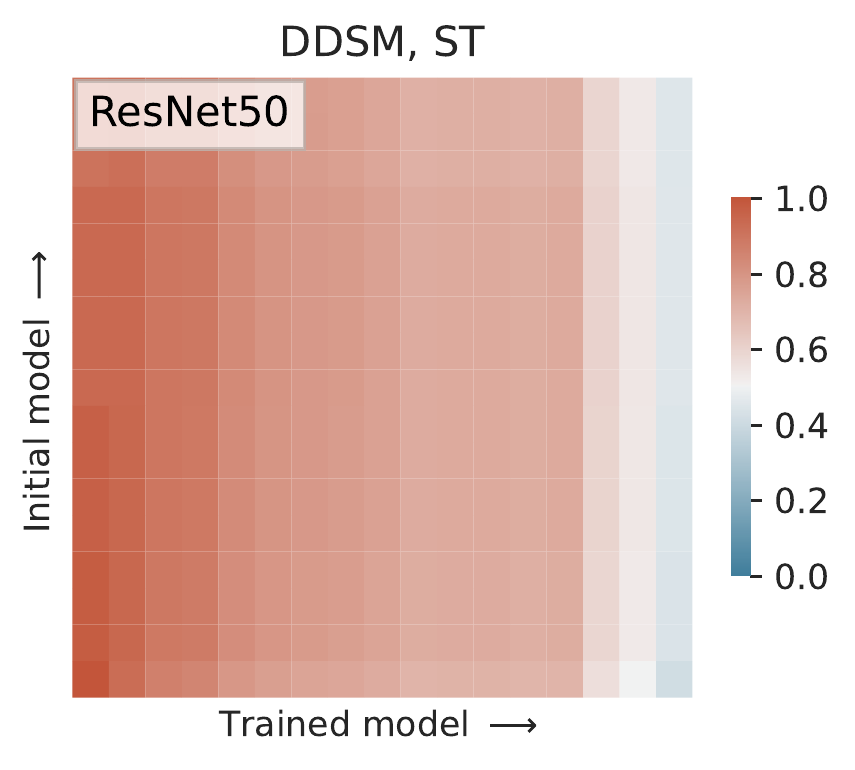} \\[-1.5mm] 
    \includegraphics[width=0.25\columnwidth]{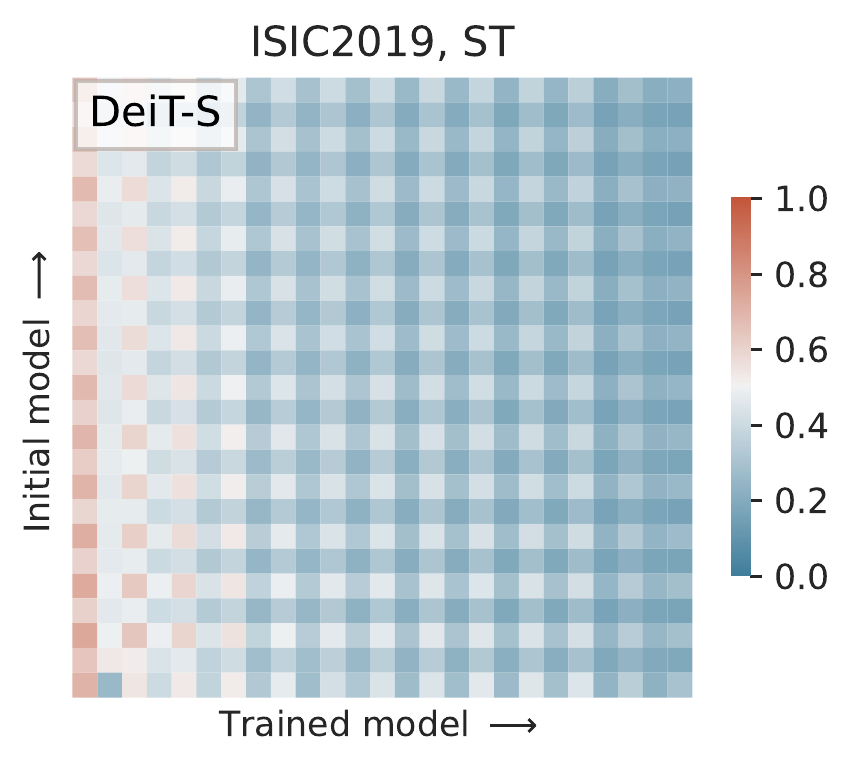} &
    \includegraphics[width=0.25\columnwidth]{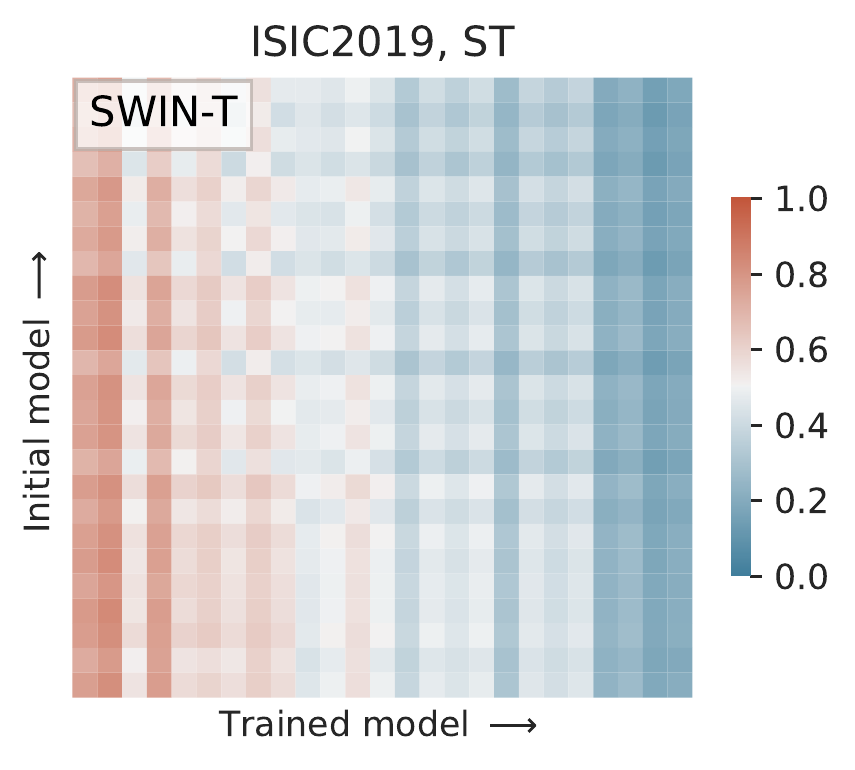} &
    \includegraphics[width=0.25\columnwidth]{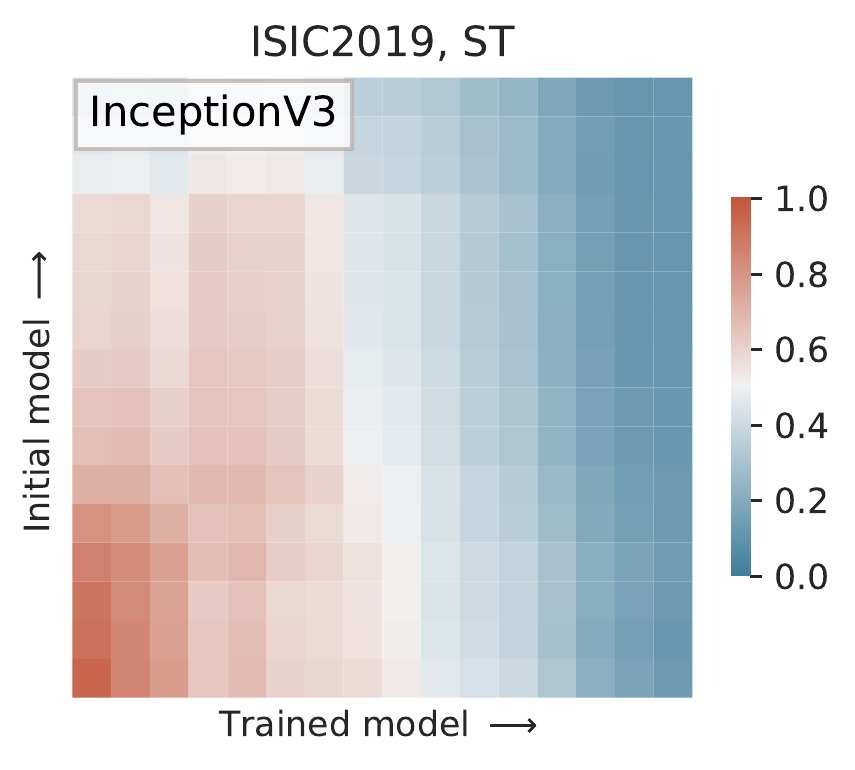} &    
    \includegraphics[width=0.25\columnwidth]{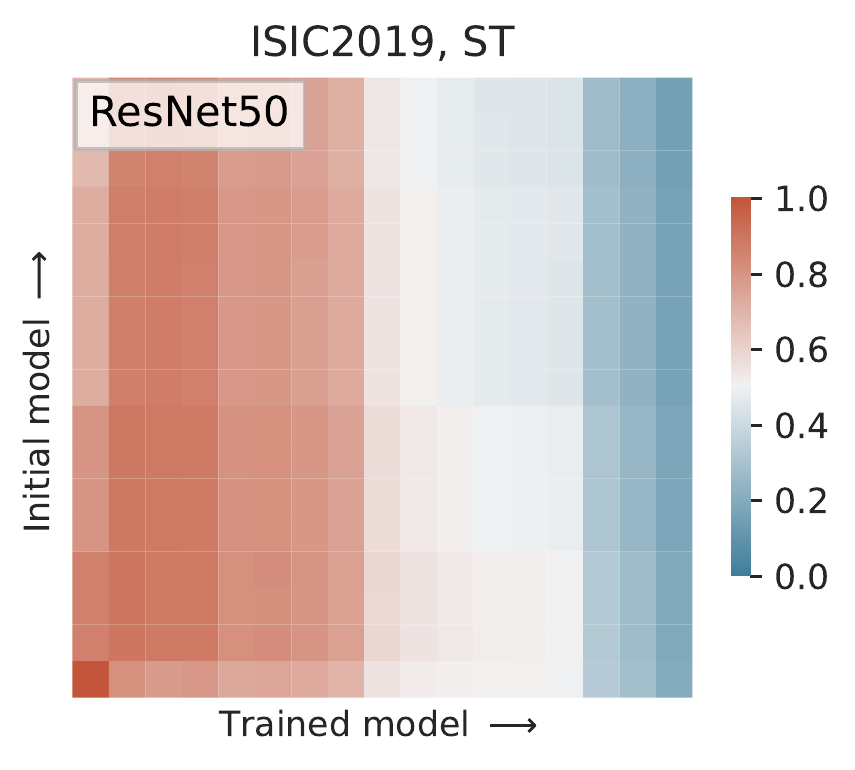} \\[-1.5mm] 
    \includegraphics[width=0.25\columnwidth]{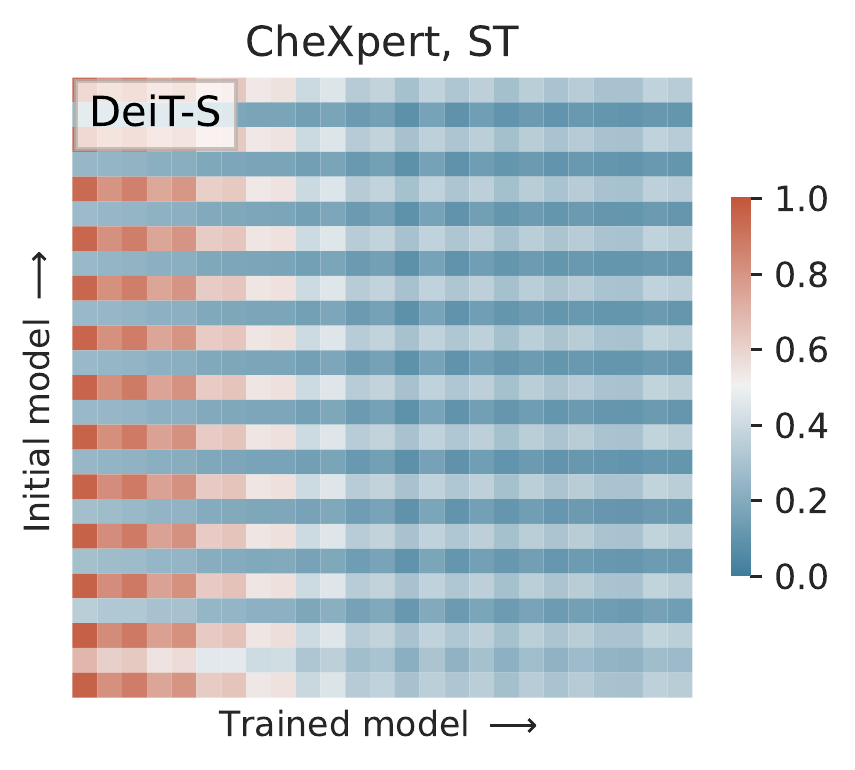} &
    \includegraphics[width=0.25\columnwidth]{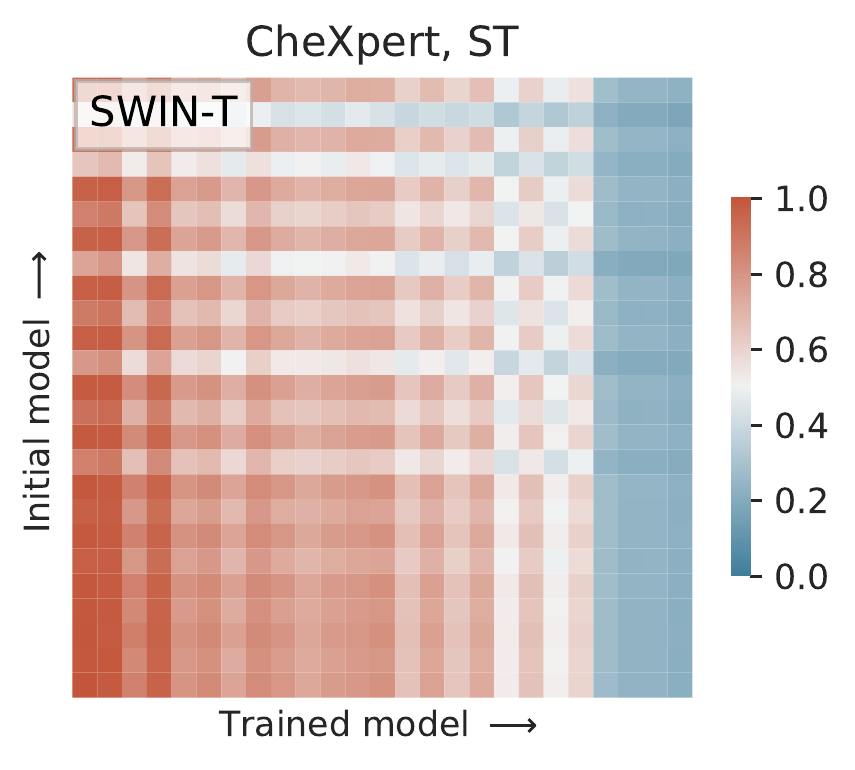} &
    \includegraphics[width=0.25\columnwidth]{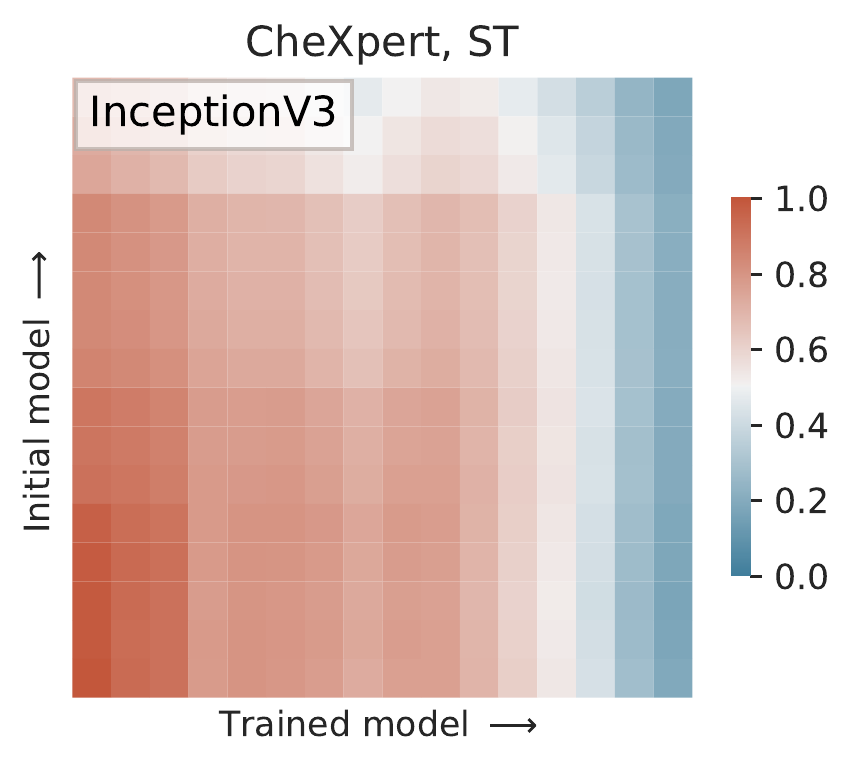} &    
    \includegraphics[width=0.25\columnwidth]{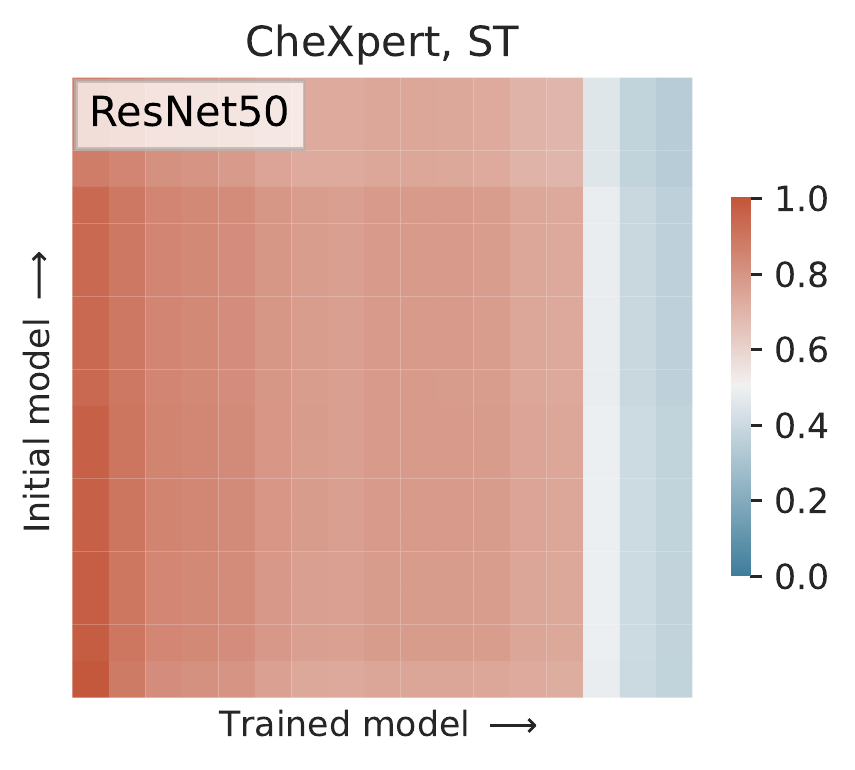} \\[-1.5mm] 
    \includegraphics[width=0.25\columnwidth]{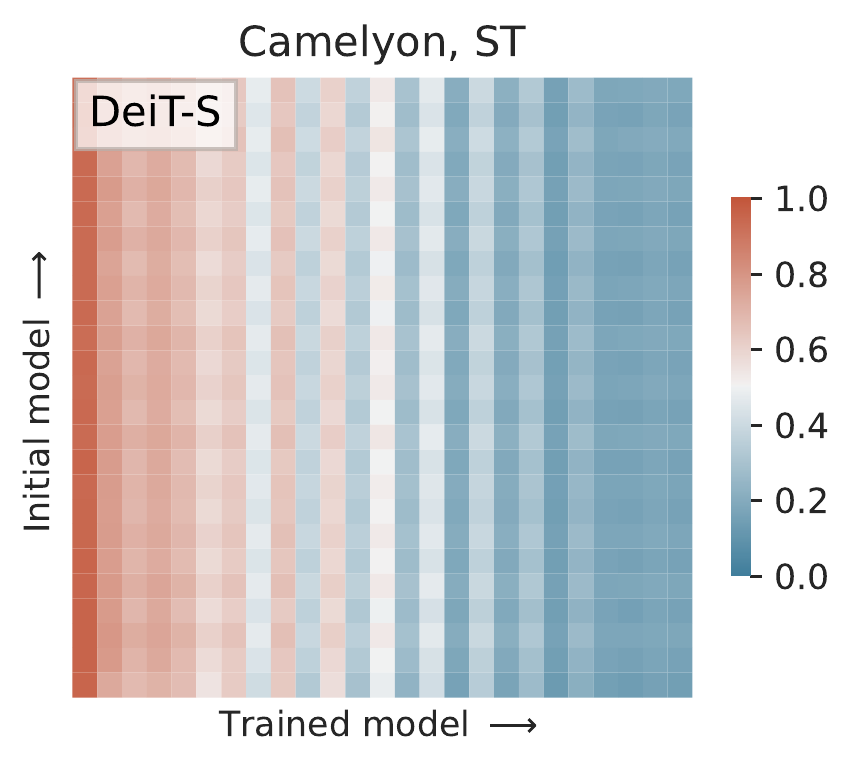} &
    \includegraphics[width=0.25\columnwidth]{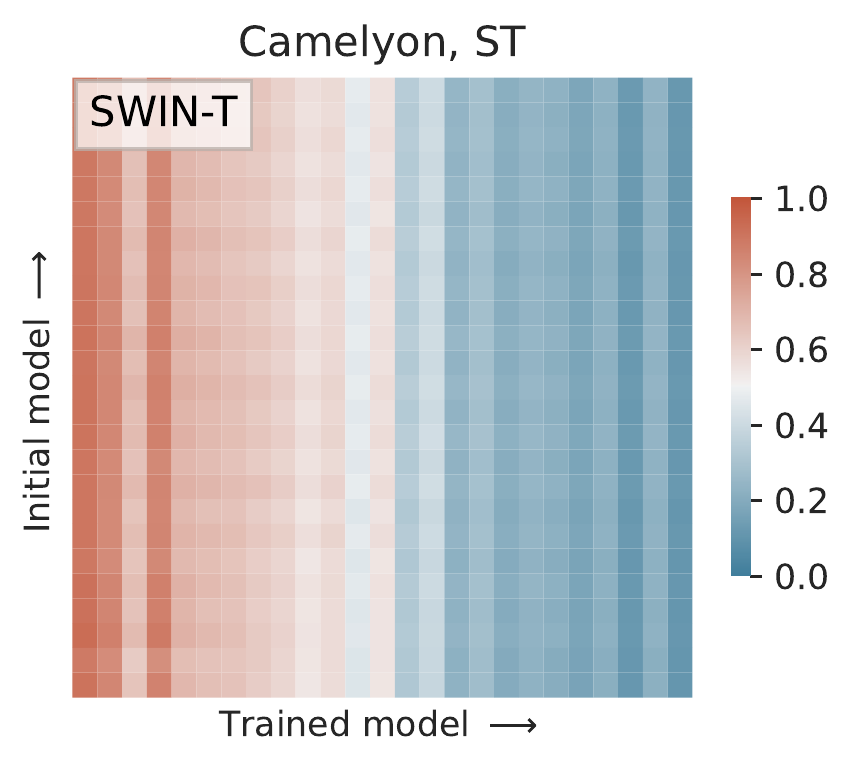} &
    \includegraphics[width=0.25\columnwidth]{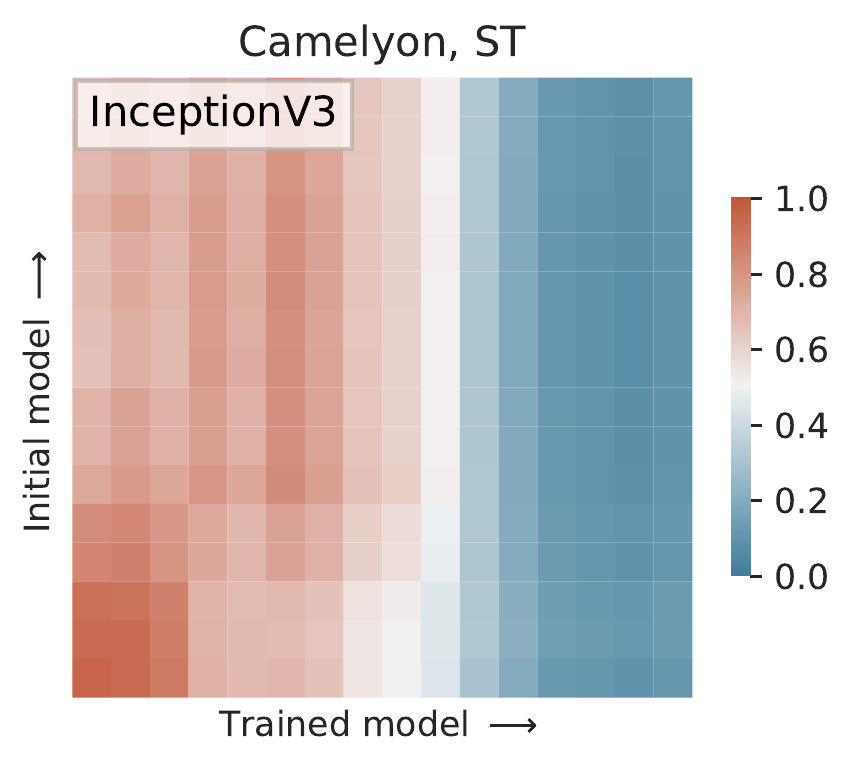} &    
    \includegraphics[width=0.25\columnwidth]{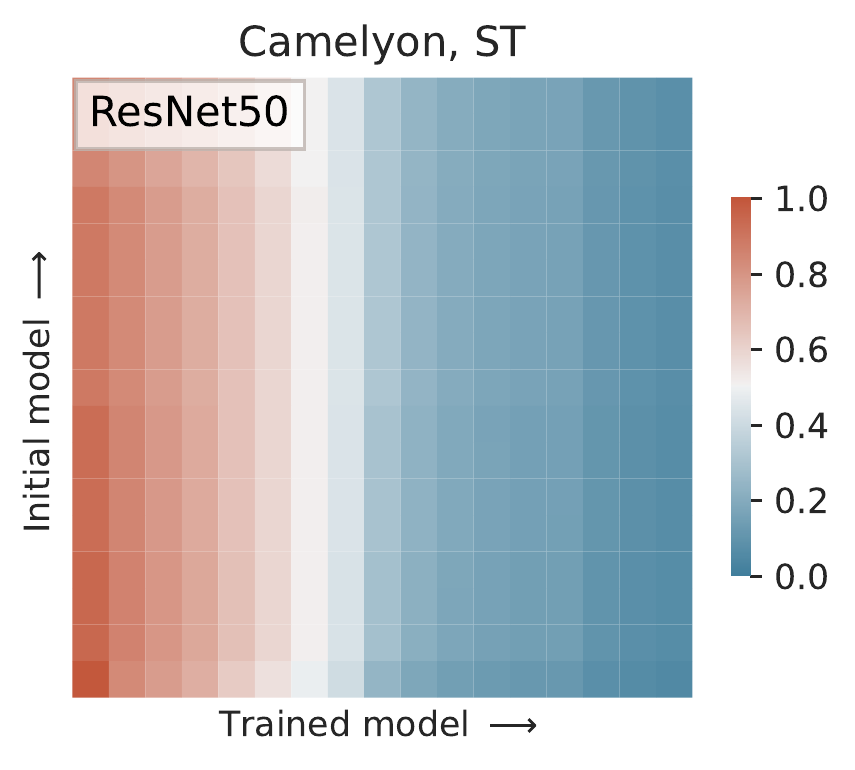} \\[-1.5mm] 
\end{tabular}
\end{center}
\vspace{-3mm}
\caption{\emph{Feature similarity between initial and fine-tuned model for ST initialization.} CKA feature similarity comparison between ST initialized models before and after fine-tuning. Reported for each dataset (rows) and model type (columns).
As we can see, models with no inductive biases exhibit changes throughout the network, while models with increased inductive bias focus more on the mid-to-high layers during training. }
\label{fig:similarity_apx_st}
\vspace{-4mm}
\end{figure}

\paragraph{Details of the feature similarity calculations.}

The feature similarity throughout this work is measured using Centered Kernel Alignment (CKA). 
CKA computes the feature similarity between two representations, allowing us to compare those of different layers and/or different models. 
For a more detailed description of CKA see \cite{raghu2021vision} and \cite{nguyen2020wide}. 
The similarity scores reported in these experiments follow the procedure described in \cite{nguyen2020wide}. 
For each setting the values are calculated by measuring the similarity over the full test set, in batches of $128$. 
This is done for all five runs of each setting, and we report the mean similarity score averaged over all runs. 
The intermediate layers of the models that were used for calculating similarities could be seen in Table \ref{tab:cnn_layer_details} and Table \ref{tab:vit_layer_details} in the Appendix and the results can be found in Figure \ref{fig:similarity} in the main text and Figures \ref{fig:similarity_apx_wt}, \ref{fig:cross_similarity_apx}, \ref{fig:cross_similarity_deit_apx} and \ref{fig:similarity_apx_st}  in Appendix.

\section{{\knn} evaluation}
\label{sec:sup-figures-knn}

\begin{figure}[t!]
\begin{center}
\begin{tabular}{@{}c@{}c@{}c@{}c@{}}
    \includegraphics[width=0.25\columnwidth]{images/knn_wst/knn_wst-APTOS2019-deit_small-all.pdf} & 
    \includegraphics[width=0.25\columnwidth]{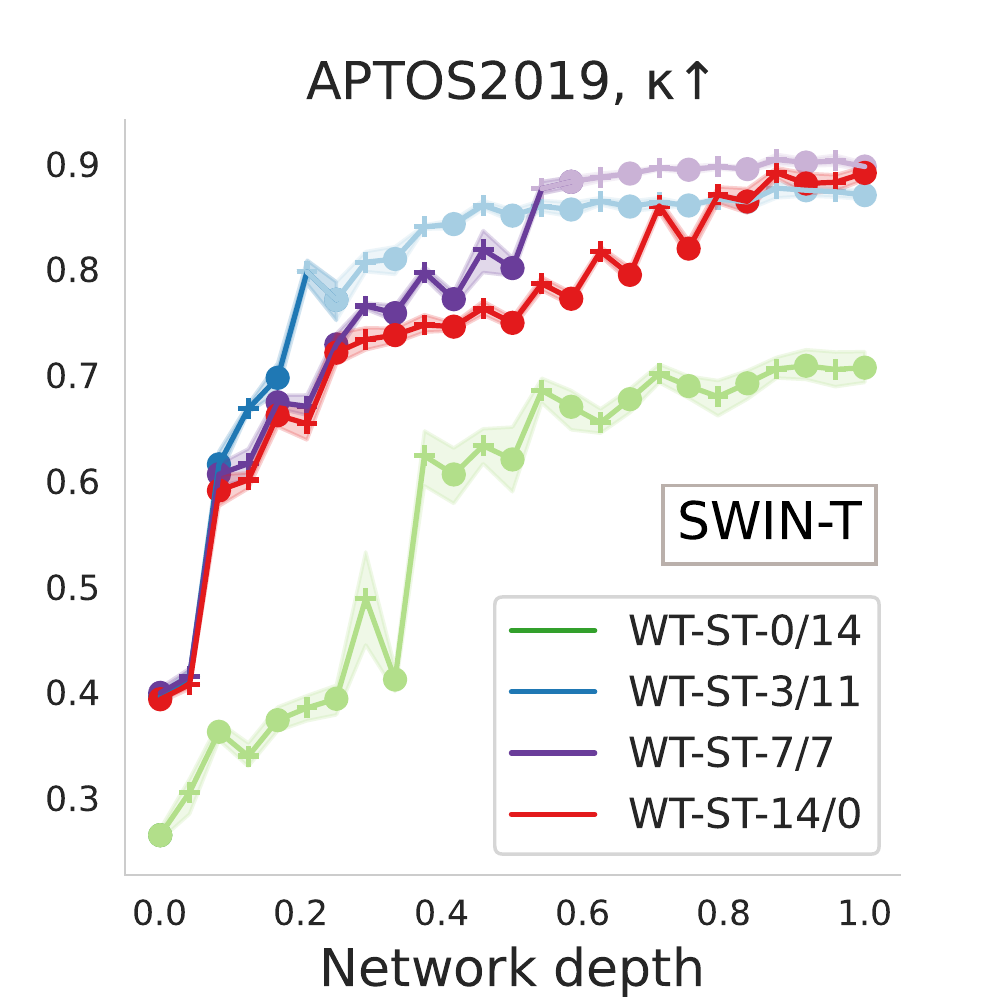} & 
    \includegraphics[width=0.25\columnwidth]{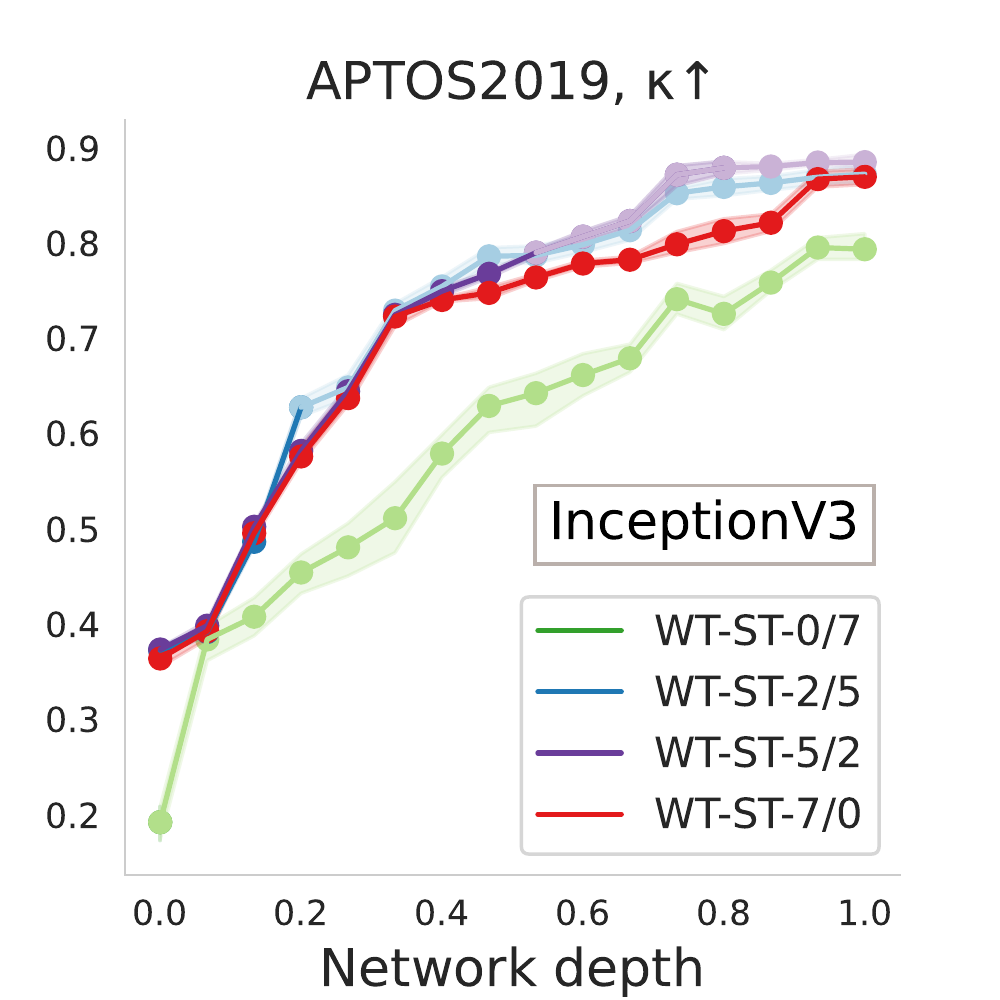} & 
    \includegraphics[width=0.25\columnwidth]{images/knn_wst/knn_wst-APTOS2019-resnet50-all.pdf}\\[-1.5mm]
    \includegraphics[width=0.25\columnwidth]{images/knn_wst/knn_wst-DDSM-deit_small-all.pdf} & 
    \includegraphics[width=0.25\columnwidth]{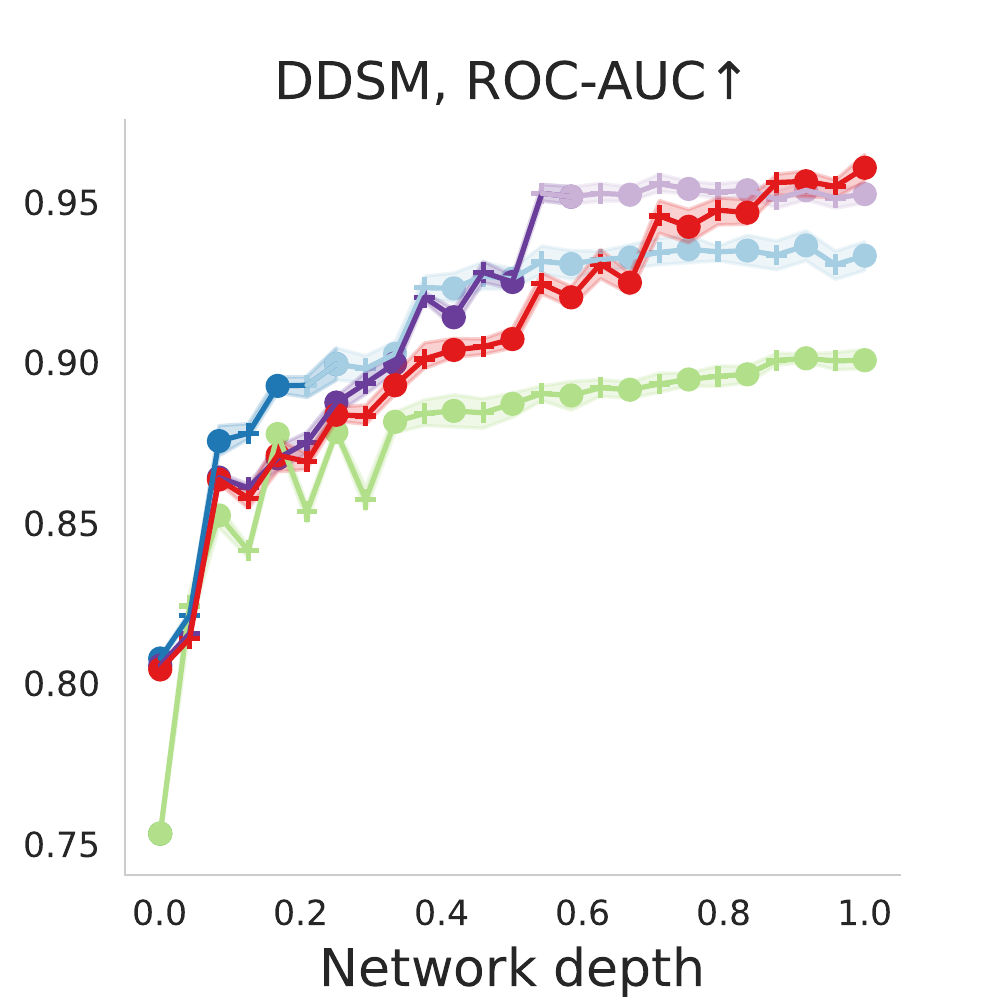} & 
    \includegraphics[width=0.25\columnwidth]{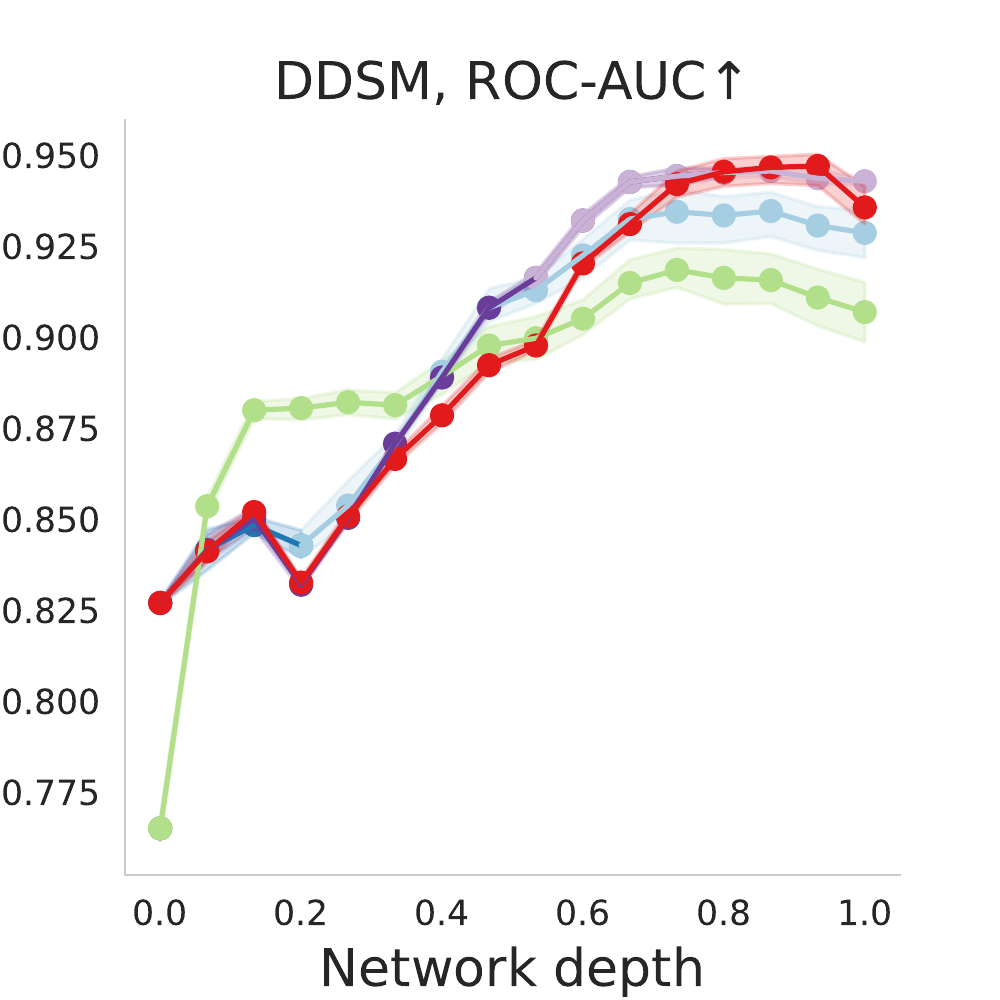} & 
    \includegraphics[width=0.25\columnwidth]{images/knn_wst/knn_wst-DDSM-resnet50-all.pdf}\\[-1.5mm]
    \includegraphics[width=0.25\columnwidth]{images/knn_wst/knn_wst-ISIC2019-deit_small-all.pdf} & 
    \includegraphics[width=0.25\columnwidth]{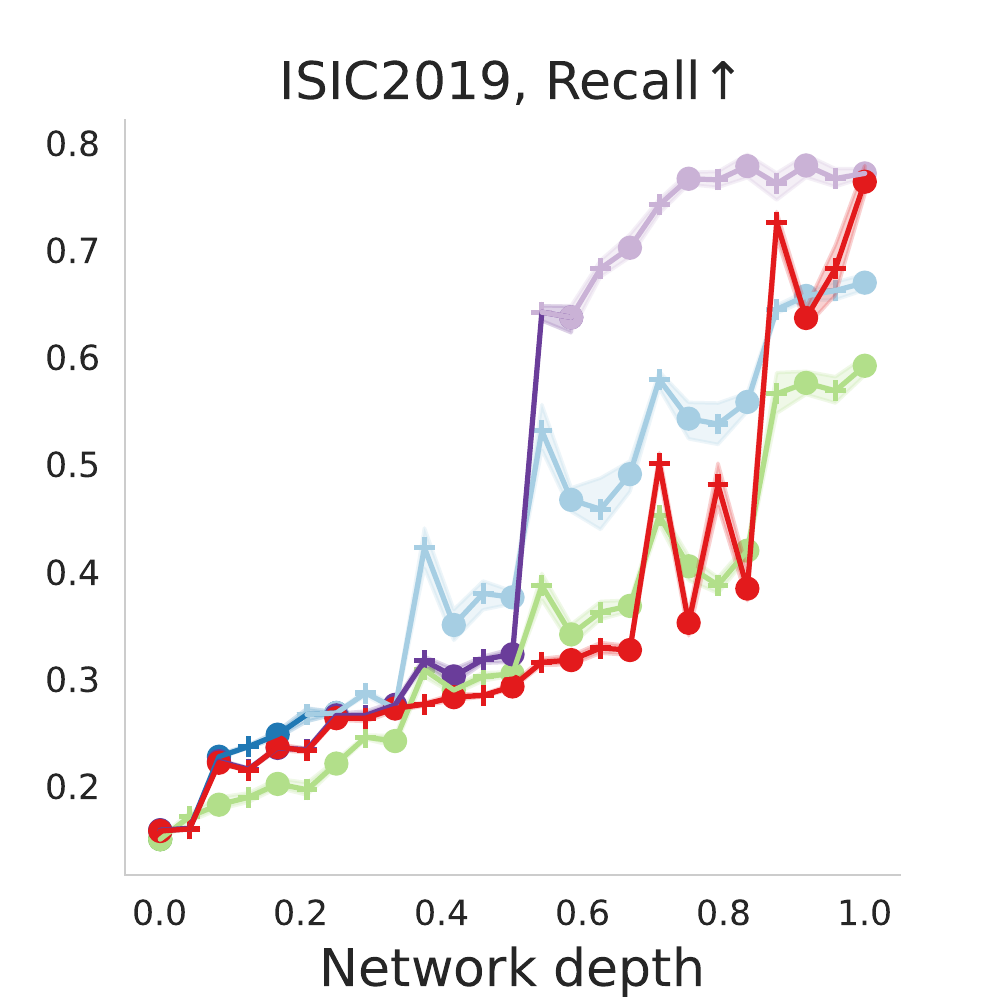} & 
    \includegraphics[width=0.25\columnwidth]{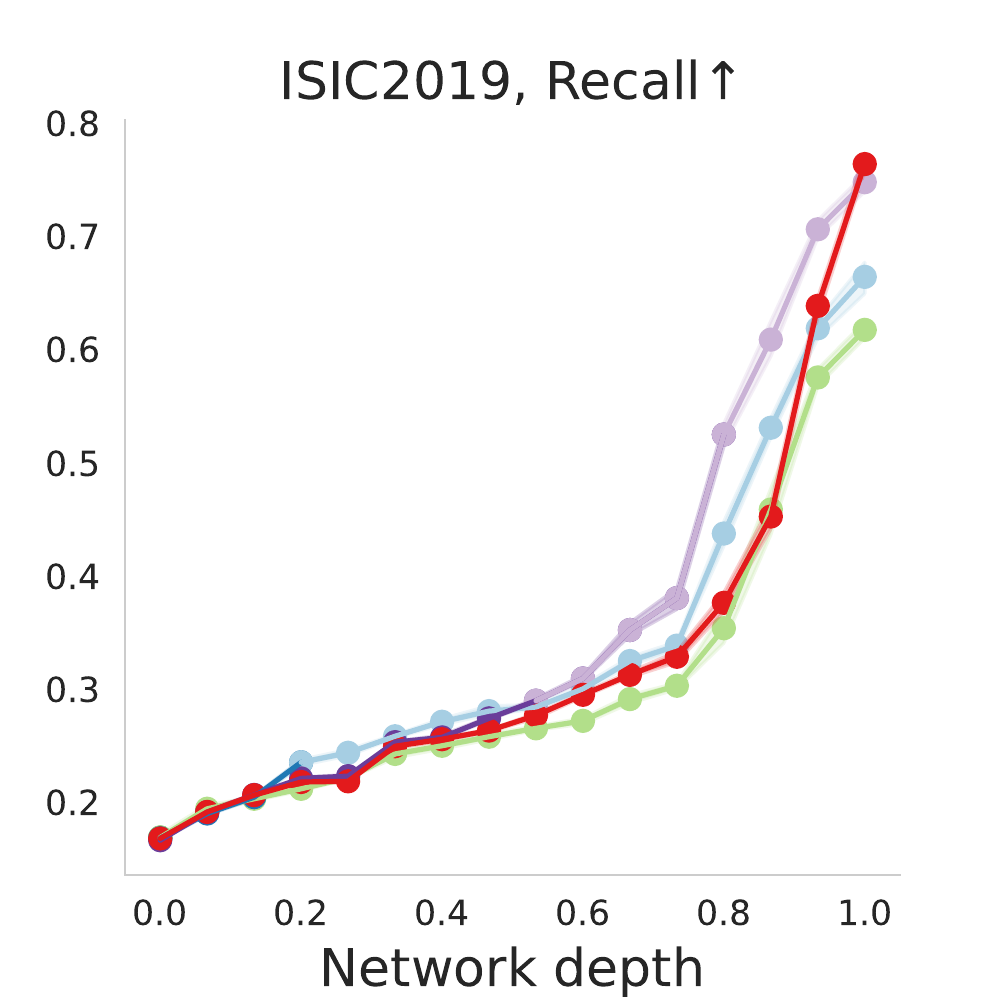} & 
    \includegraphics[width=0.25\columnwidth]{images/knn_wst/knn_wst-ISIC2019-resnet50-all.pdf}\\[-1.5mm]
    \includegraphics[width=0.25\columnwidth]{images/knn_wst/knn_wst-CheXpert-deit_small-all.pdf} & 
    \includegraphics[width=0.25\columnwidth]{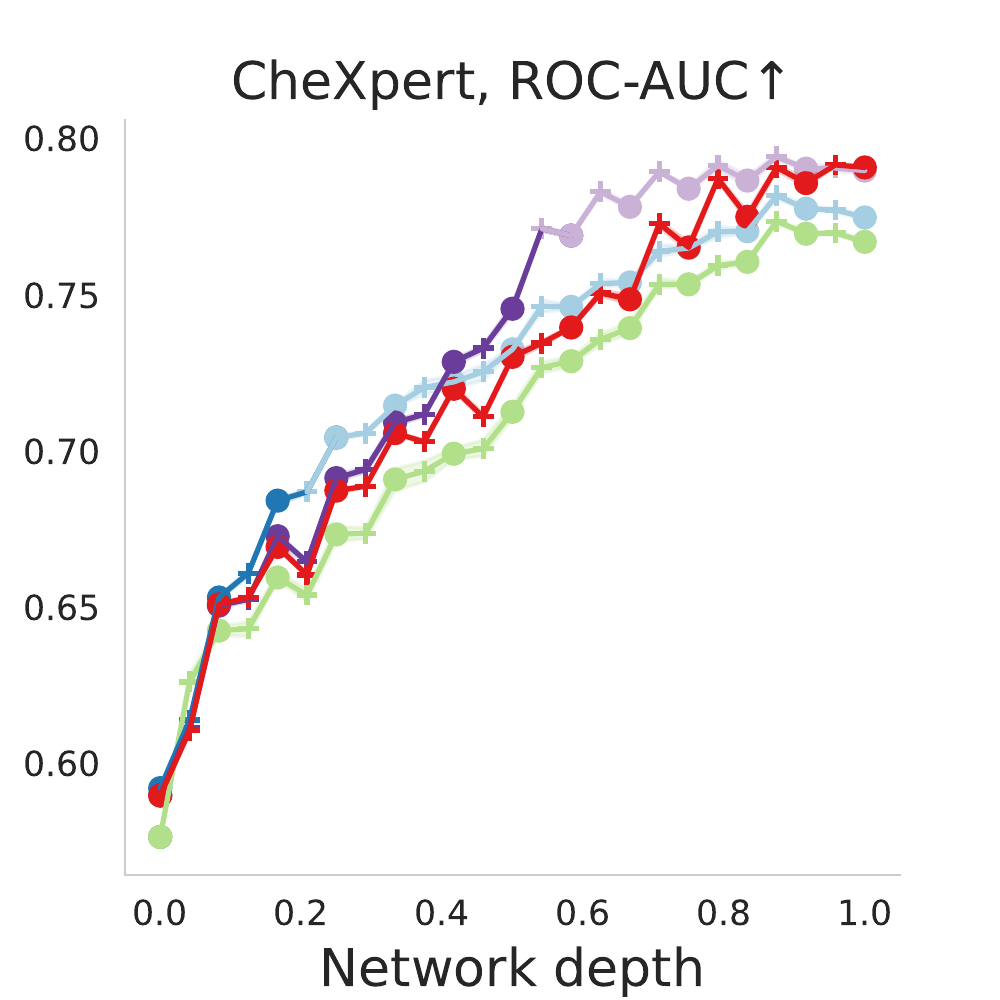} & 
    \includegraphics[width=0.25\columnwidth]{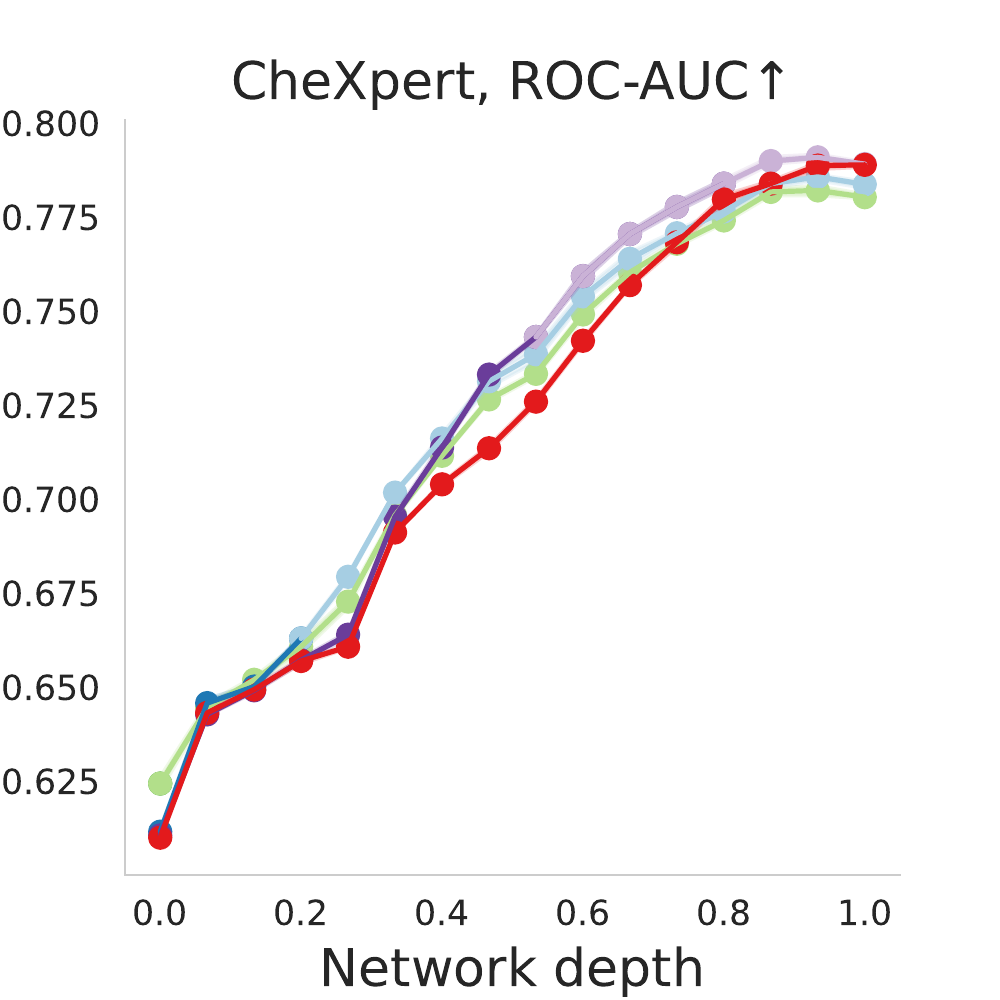} & 
    \includegraphics[width=0.25\columnwidth]{images/knn_wst/knn_wst-CheXpert-resnet50-all.pdf}\\[-1.5mm]    
    \includegraphics[width=0.25\columnwidth]{images/knn_wst/knn_wst-Camelyon-deit_small-all.pdf} & 
    \includegraphics[width=0.25\columnwidth]{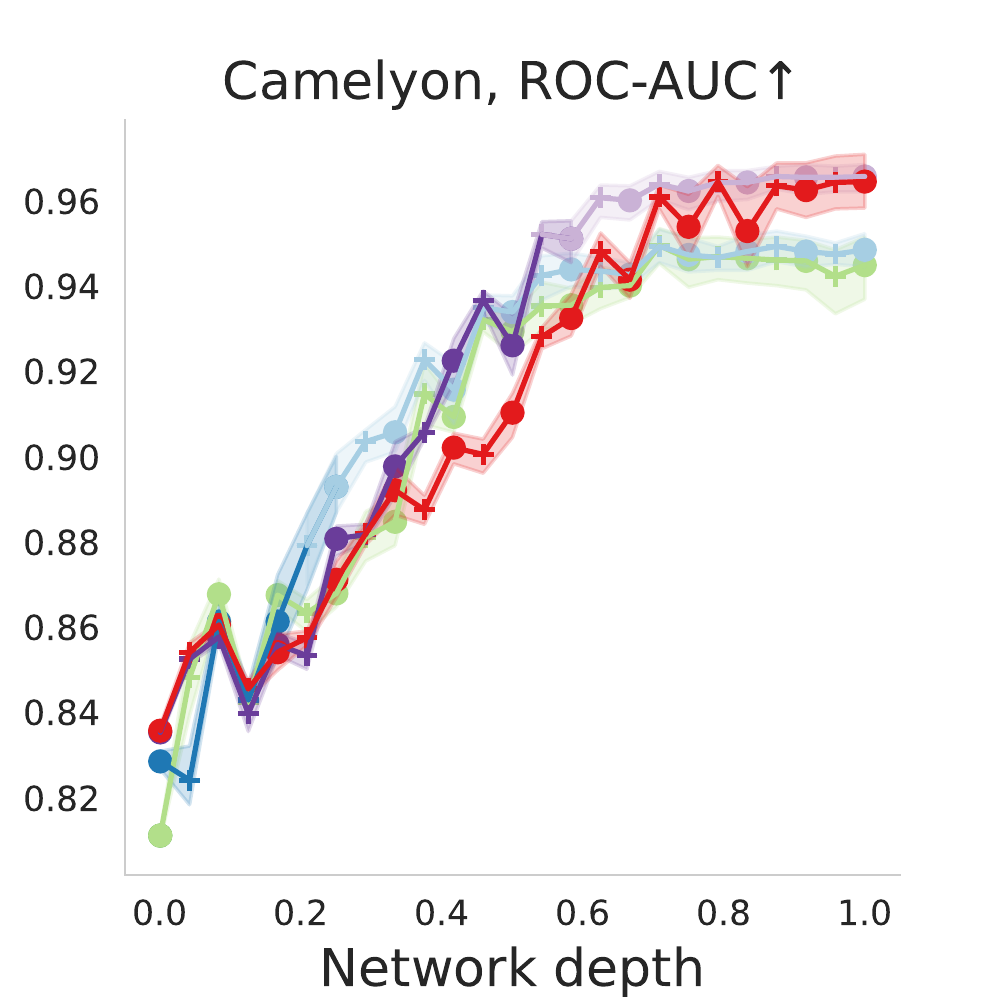} & 
    \includegraphics[width=0.25\columnwidth]{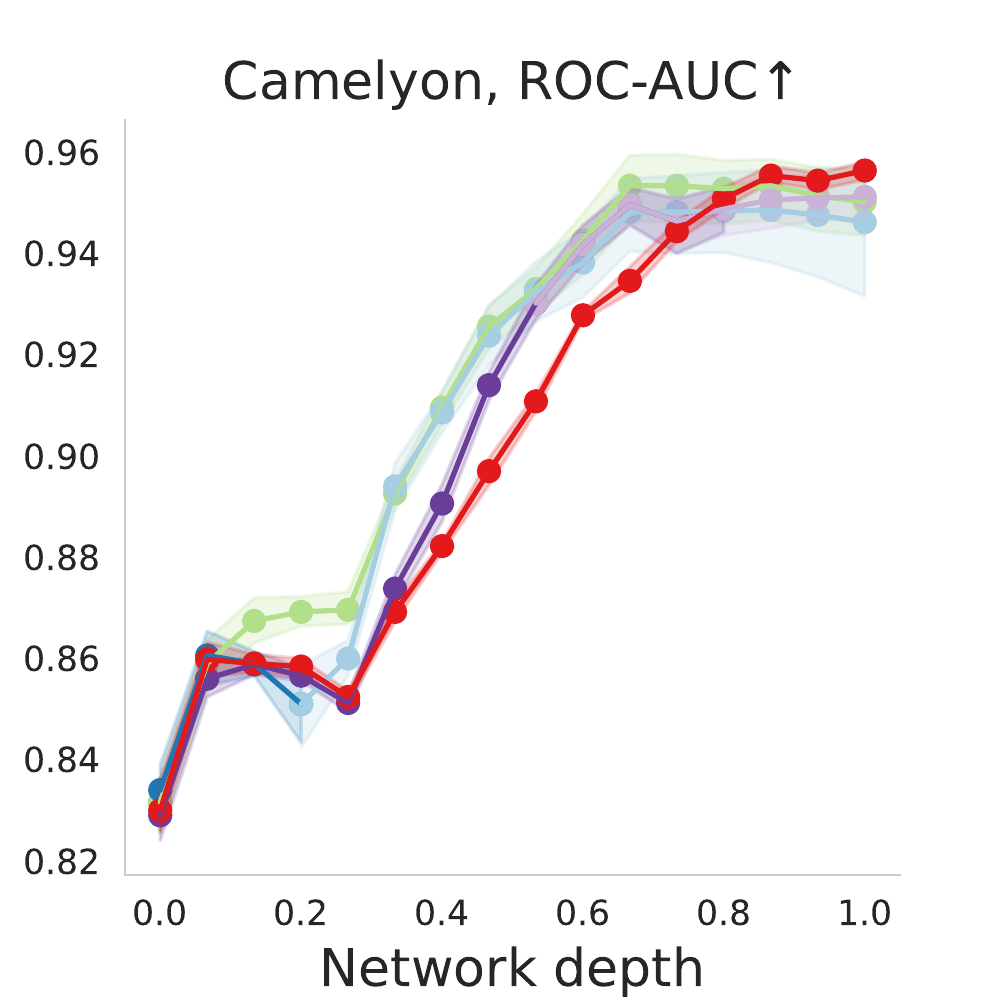} & 
    \includegraphics[width=0.25\columnwidth]{images/knn_wst/knn_wst-Camelyon-resnet50-all.pdf}\\[-1.5mm]        
\end{tabular}
\end{center}
\vspace{-3mm}

\caption{\emph{Predictive performance of features at different depths using \knn evaluation}.
\knn evaluation performance at different depths for models initialized with varying  WT fractions, reported for each dataset (rows) and model type (columns). Overall the \knn performance increases monotonically with depth for all models and datasets. However, relative performance gains from layer to layer exhibit different patterns. CNNs improve progressively, while ViTs increase rapidly in the beginning and then they reach a plateau. This plateau is observed to appear in association with the first ST-initialized layer in the {\wtst} experiments.}

\label{fig:wst_knn_apx}
\vspace{-4mm}
\end{figure}

We use \knn evaluation to investigate the discriminative power at different layers throughout the network.. 
The evaluation is performed by comparing the similarity, in the feature space, of samples from the training and the test set.
In particular, we use cosine similarity as the means to calculate the distance between different data-points.
Then, labels are assigned to the query data-point, from the test set, by considering its $k$ nearest neighbors from the training set.
Throughout this work we use $k=200$ . 
The layers used to extract the embeddings are listed in Table \ref{tab:cnn_layer_details} for CNNs and Table \ref{tab:vit_layer_details} for ViTs.
The results of the {\knn} evaluation experiments can be found in Figure \ref{fig:wst_knn} in the main text and Figures \ref{fig:wst_knn_apx}, \ref{fig:wst_max-knn_apx} and \ref{fig:wst_deit-knn_apx} in the Appendix.

\begin{figure}[t]
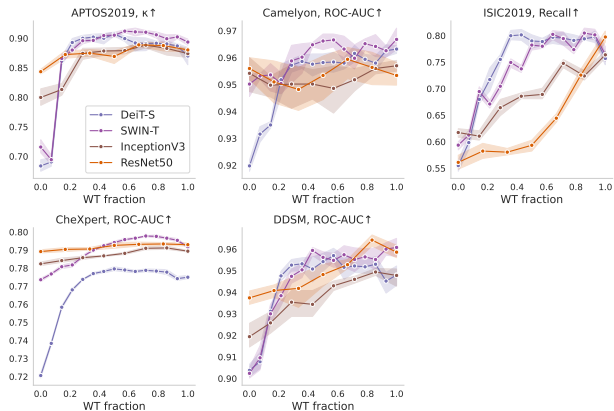

\begin{center}
\begin{tabular}{@{}c@{}c@{}c@{}}
    \includegraphics[width=0.333\columnwidth]{images/knn_wst/max_knn_wst-APTOS2019-all_models-all.pdf} & 
    \includegraphics[width=0.333\columnwidth]{images/knn_wst/max_knn_wst-Camelyon-all_models-all.pdf} &
    \includegraphics[width=0.333\columnwidth]{images/knn_wst/max_knn_wst-ISIC2019-all_models-all.pdf}\\[-1.5mm]
    \includegraphics[width=0.333\columnwidth]{images/knn_wst/max_knn_wst-CheXpert-all_models-all.pdf} &
    \includegraphics[width=0.333\columnwidth]{images/knn_wst/max_knn_wst-DDSM-all_models-all.pdf}\\[-1.5mm]
\end{tabular}
\end{center}
\vspace{-3mm}

\caption{\emph{Maximum $k$-nn predictive performance of intermediate features for different \wtst initialization schemes}.
Similar to Figure \ref{fig:wst_knn_apx} relative gains exhibit different patterns of improvement. Once again ViTs appear to improve quickly early on and then plateau. However, this plateau has a negative slope in some cases, suggesting the presence of strong biases in the high-level features, possibly inherited from the pre-training task.
}
\label{fig:wst_max-knn_apx}
\vspace{-4mm}
\end{figure}

\begin{figure}[t]
\begin{center}
\begin{tabular}{@{}c@{}c@{}c@{}}
    \includegraphics[width=0.333\columnwidth]{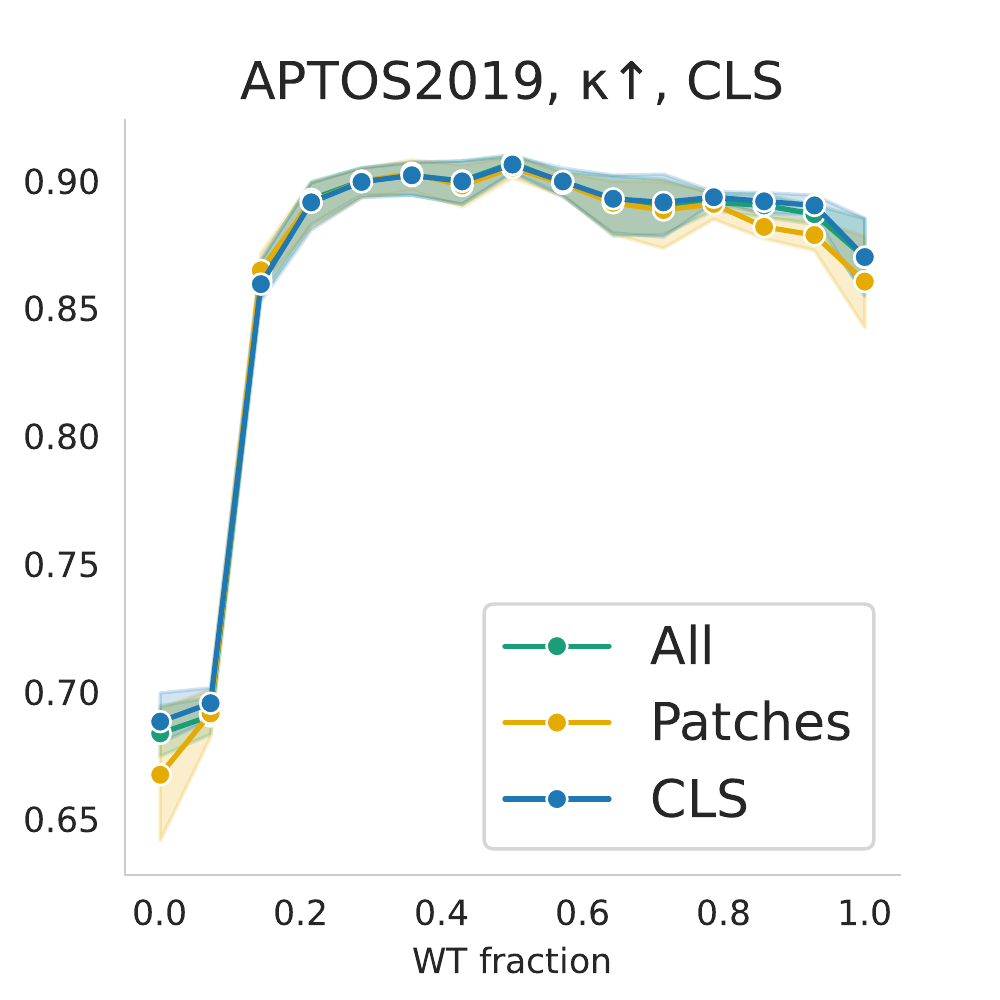} & 
    \includegraphics[width=0.333\columnwidth]{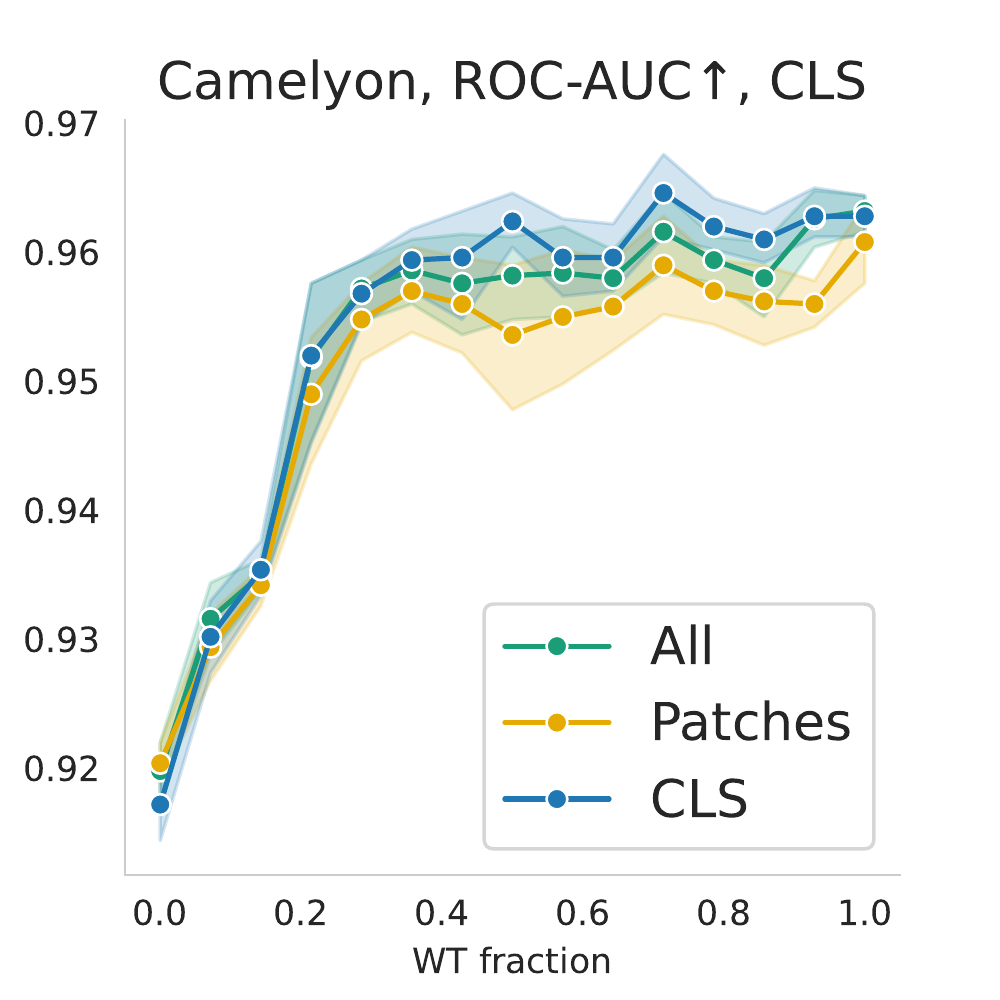} &
    \includegraphics[width=0.333\columnwidth]{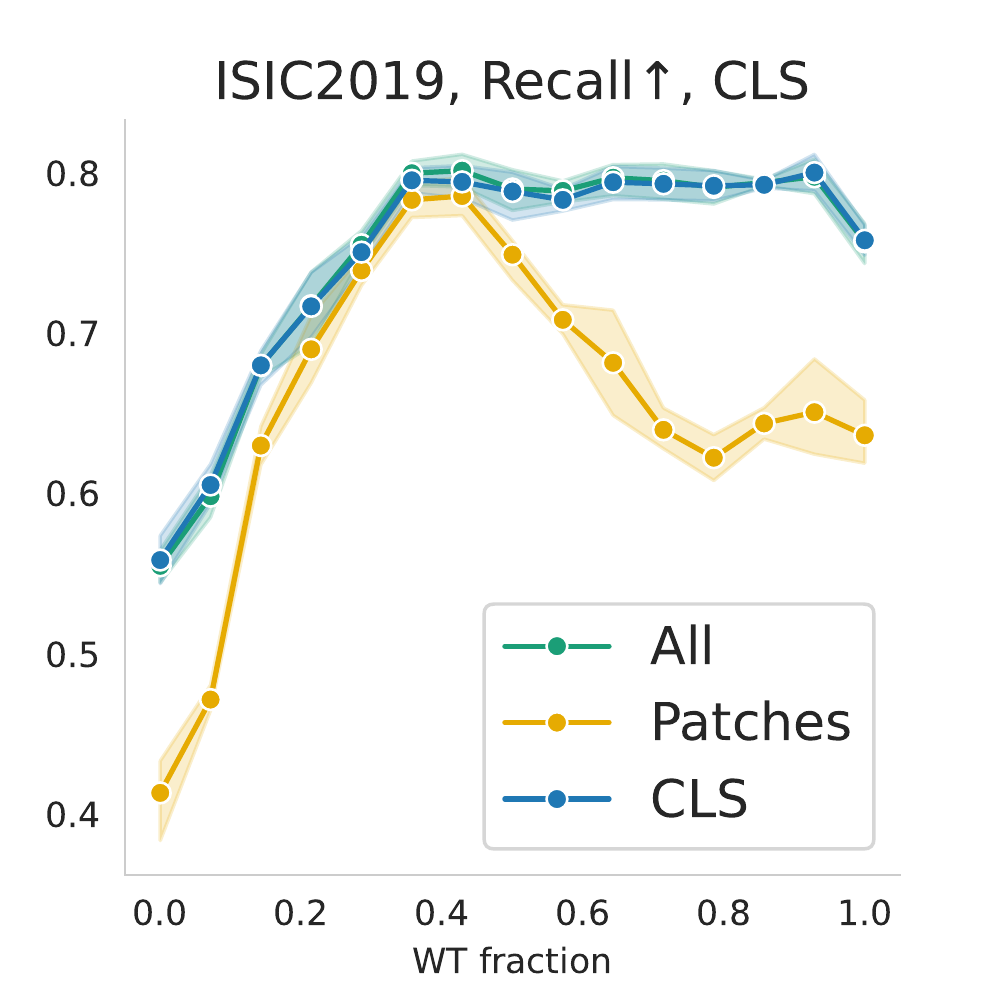}\\[-1.5mm]
    \includegraphics[width=0.333\columnwidth]{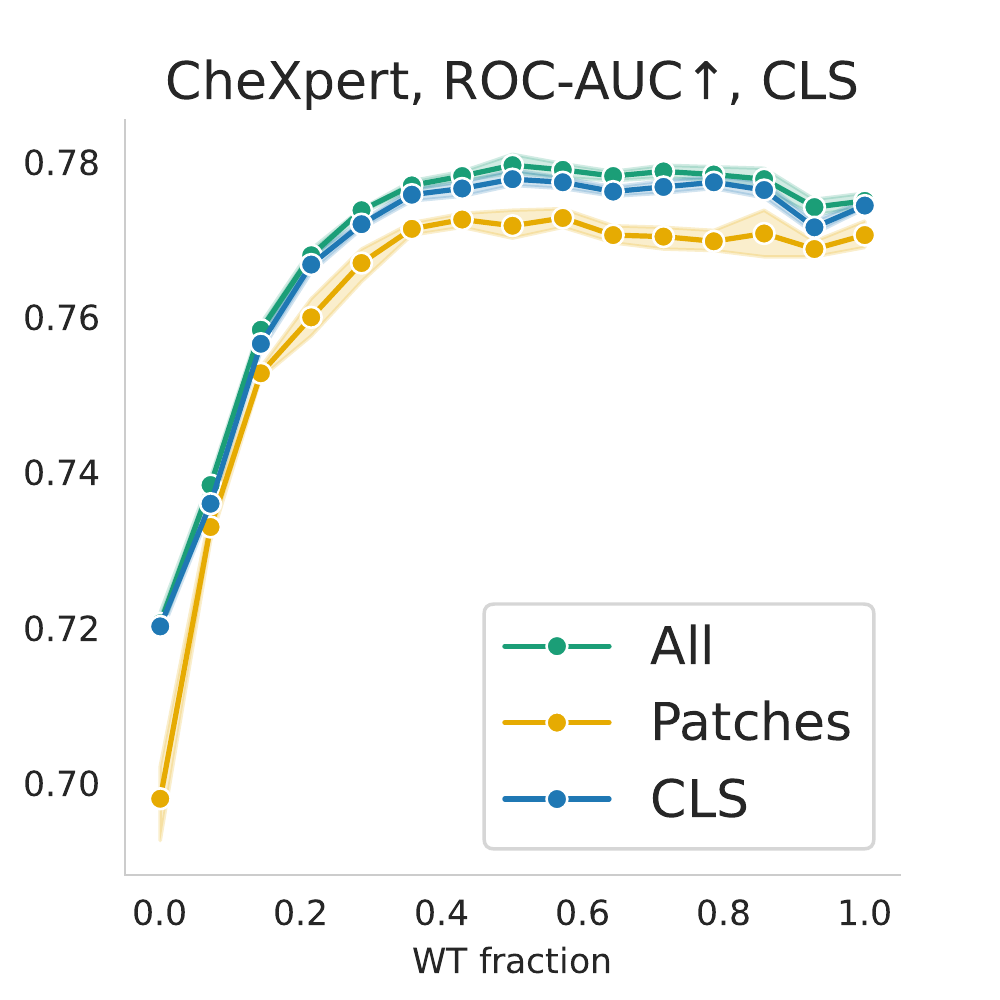} &
    \includegraphics[width=0.333\columnwidth]{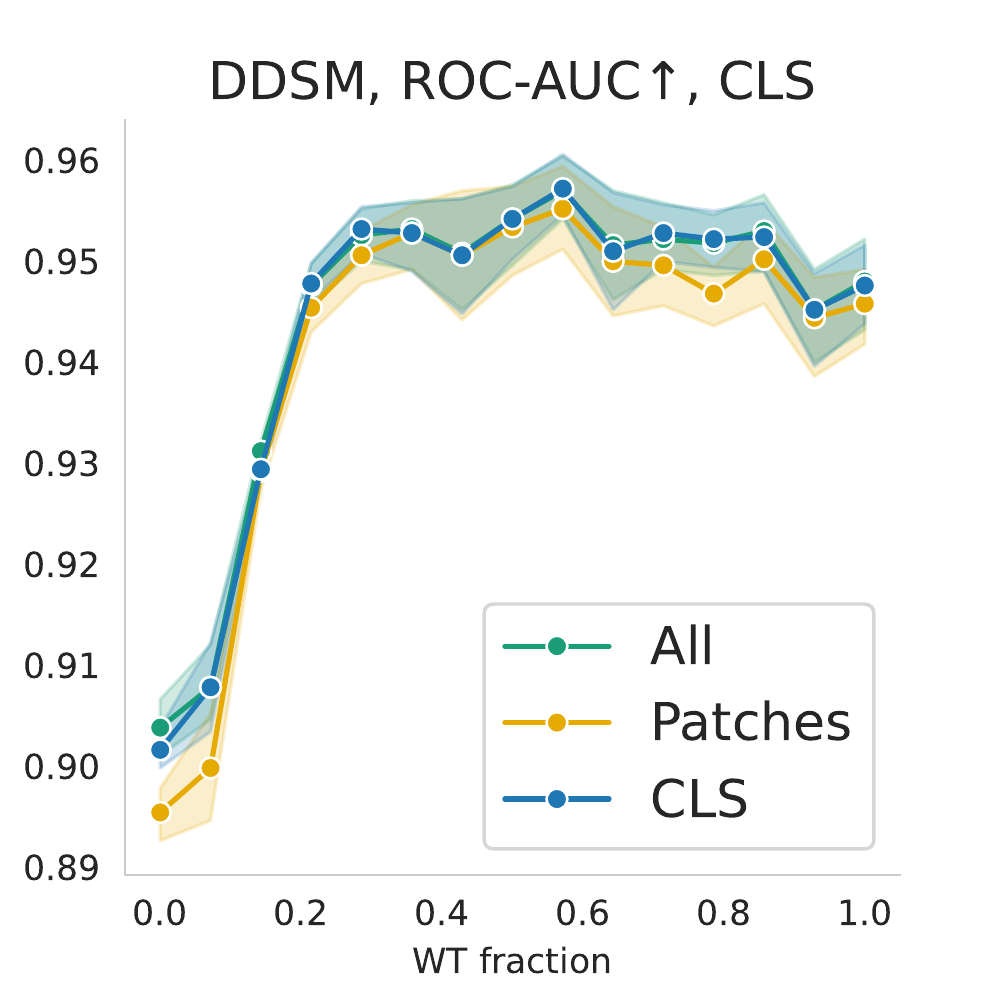}\\[-1.5mm]      
\end{tabular}
\end{center}
\vspace{-3mm}

\caption{\emph{Maximum $k$-nn predictive performance of intermediate features for different \wtst initialization schemes when using different feature types from \deitsmall for evaluation}.
Maximum $k$-nn evaluation score achieved at any depth for corresponding {\wtst} initialization fraction, for \textsc{Deit-S} \textit{(1)} using only the \texttt{cls} token's activations, \textit{(2)} using activations from the spatial tokens, \textit{(3)} concatenating \textit{(1)} and \textit{(2)}.
The different feature embeddings seem to exhibit similar trends 
, but the \texttt{cls} token often seem to outperform the patch embedding.
}
\label{fig:wst_deit-knn_apx}
\vspace{-4mm}
\end{figure}

\section{Re-initialization robustness}
\label{sec:sup-figures-layerwise-importance}

\begin{figure}[t!]
\begin{center}
\begin{tabular}{@{}c@{}c@{}c@{}c@{}}
    \includegraphics[width=0.25\columnwidth]{images/layer_importance/LI_APTOS2019-deit_small.pdf} & 
    \includegraphics[width=0.25\columnwidth]{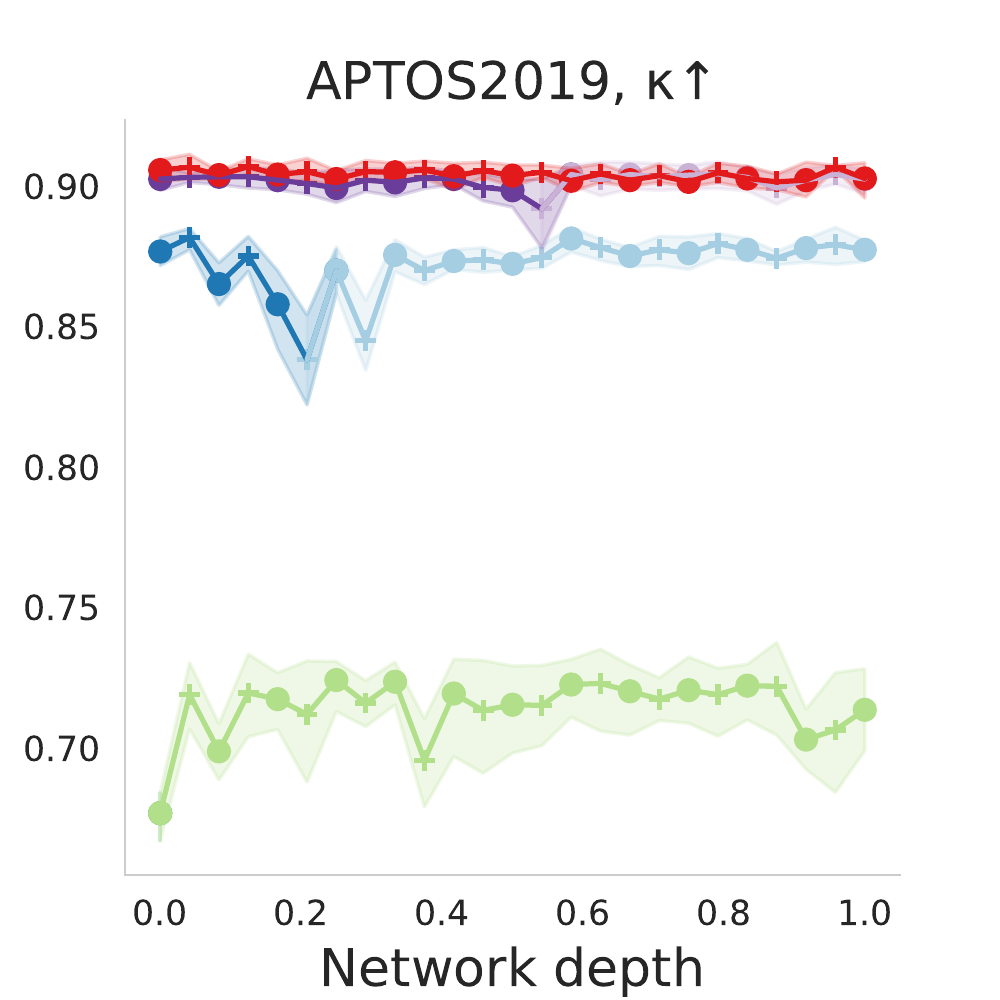} &
    \includegraphics[width=0.25\columnwidth]{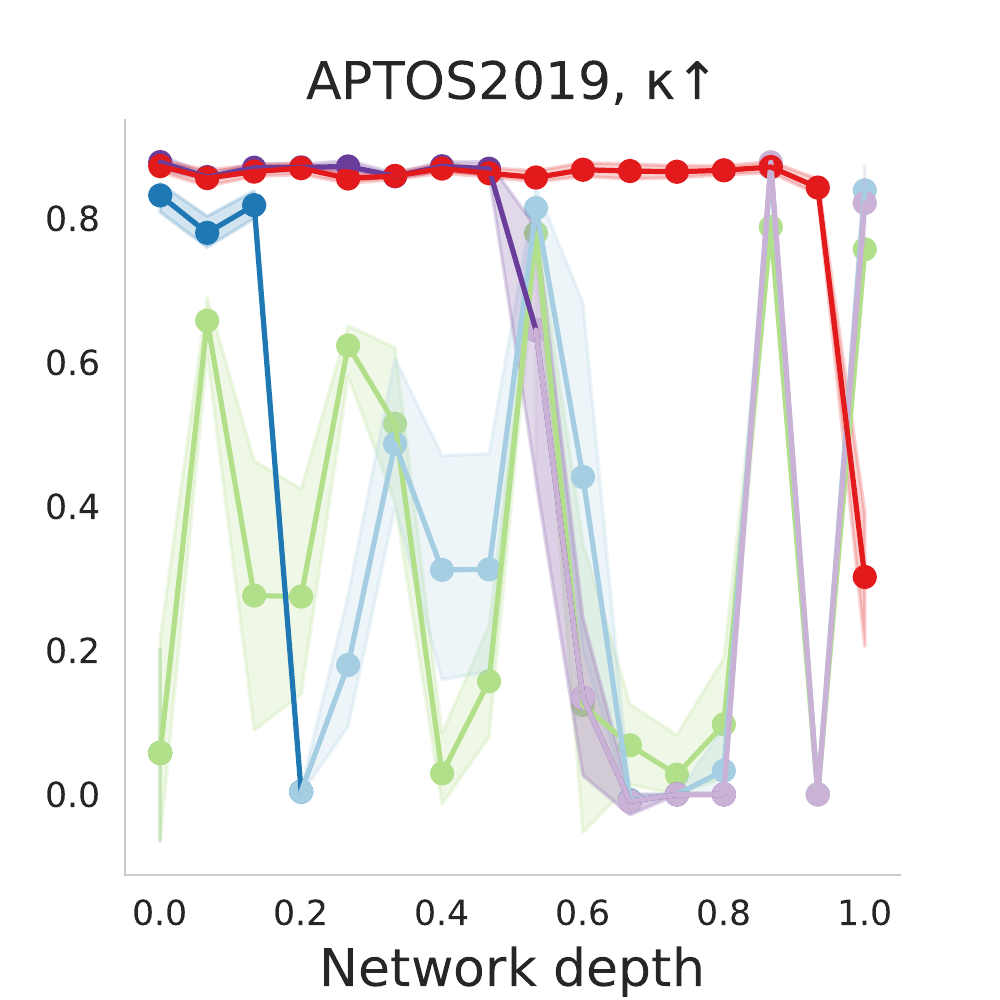} &
    \includegraphics[width=0.25\columnwidth]{images/layer_importance/LI_APTOS2019-resnet50.pdf}\\ [-1.5mm]
    \includegraphics[width=0.25\columnwidth]{images/layer_importance/LI_DDSM-deit_small.pdf} & 
    \includegraphics[width=0.25\columnwidth]{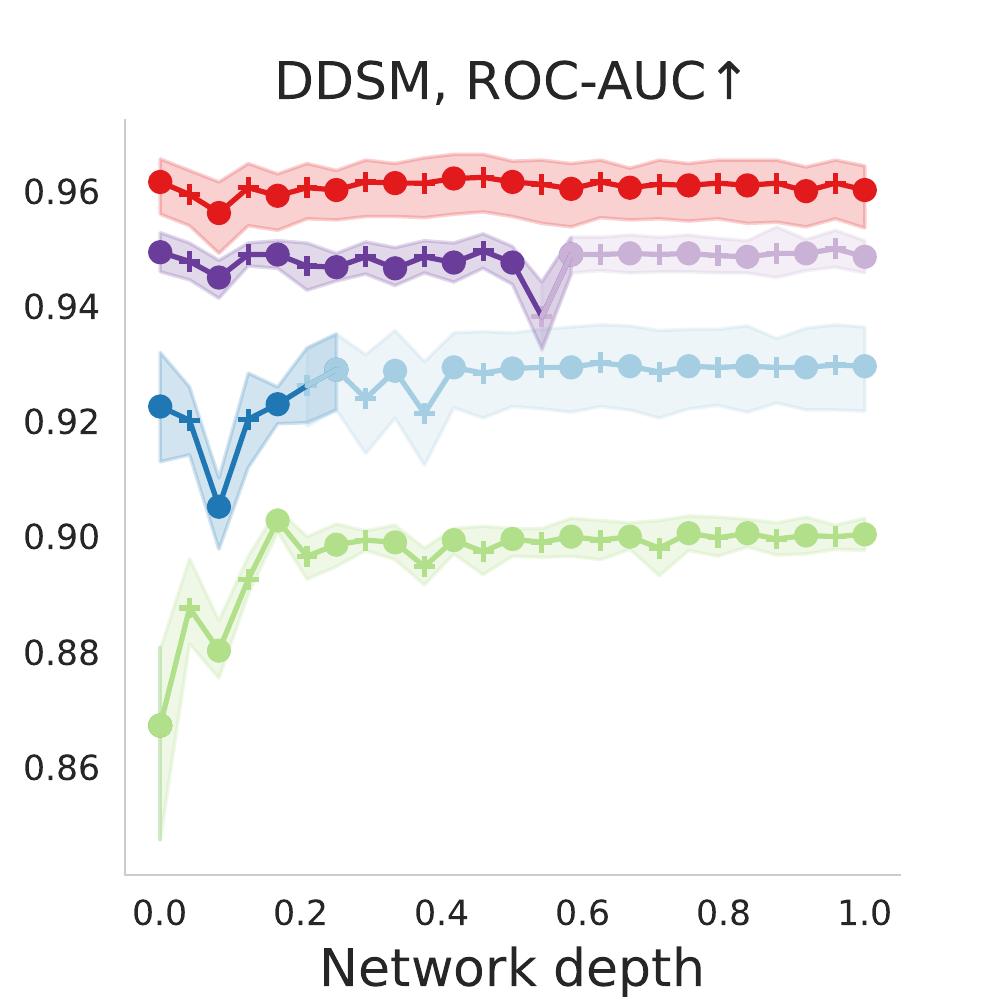} &
    \includegraphics[width=0.25\columnwidth]{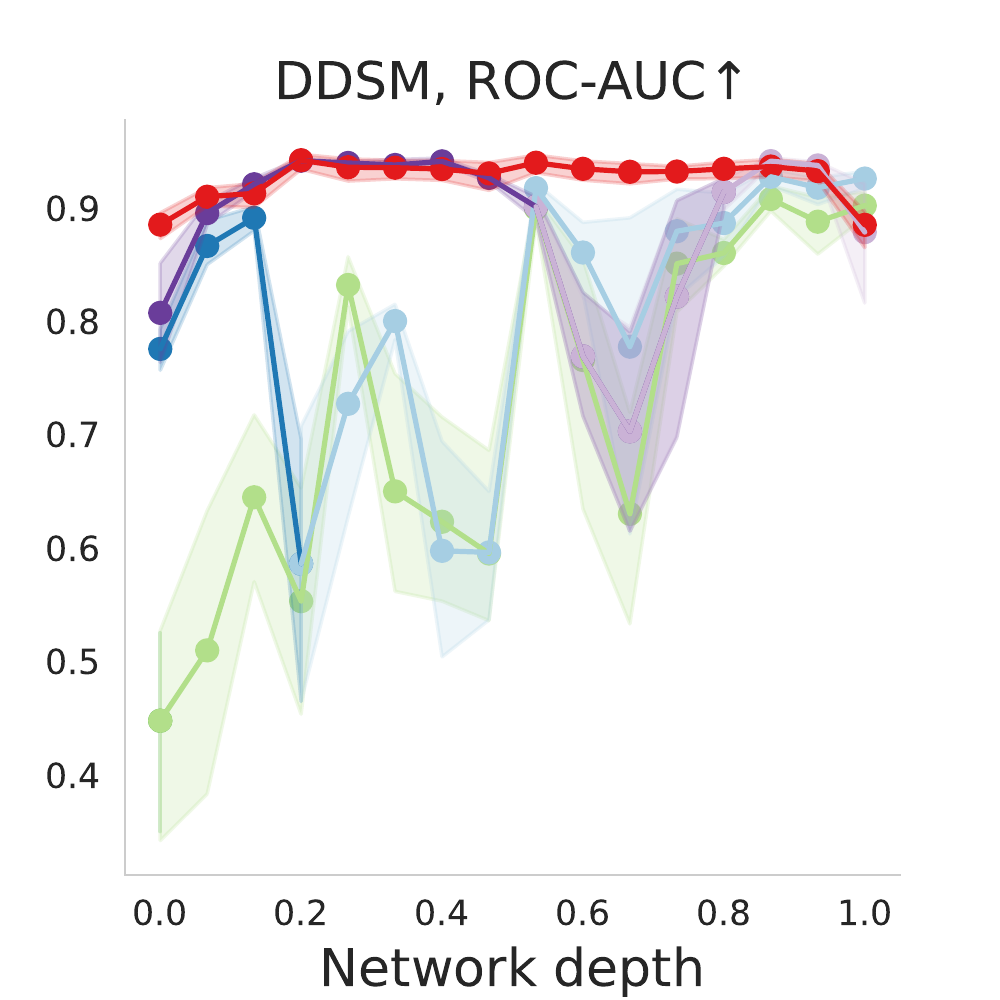} &
    \includegraphics[width=0.25\columnwidth]{images/layer_importance/LI_DDSM-resnet50.pdf} \\[-1.5mm]
    \includegraphics[width=0.25\columnwidth]{images/layer_importance/LI_ISIC2019-deit_small.pdf} & 
    \includegraphics[width=0.25\columnwidth]{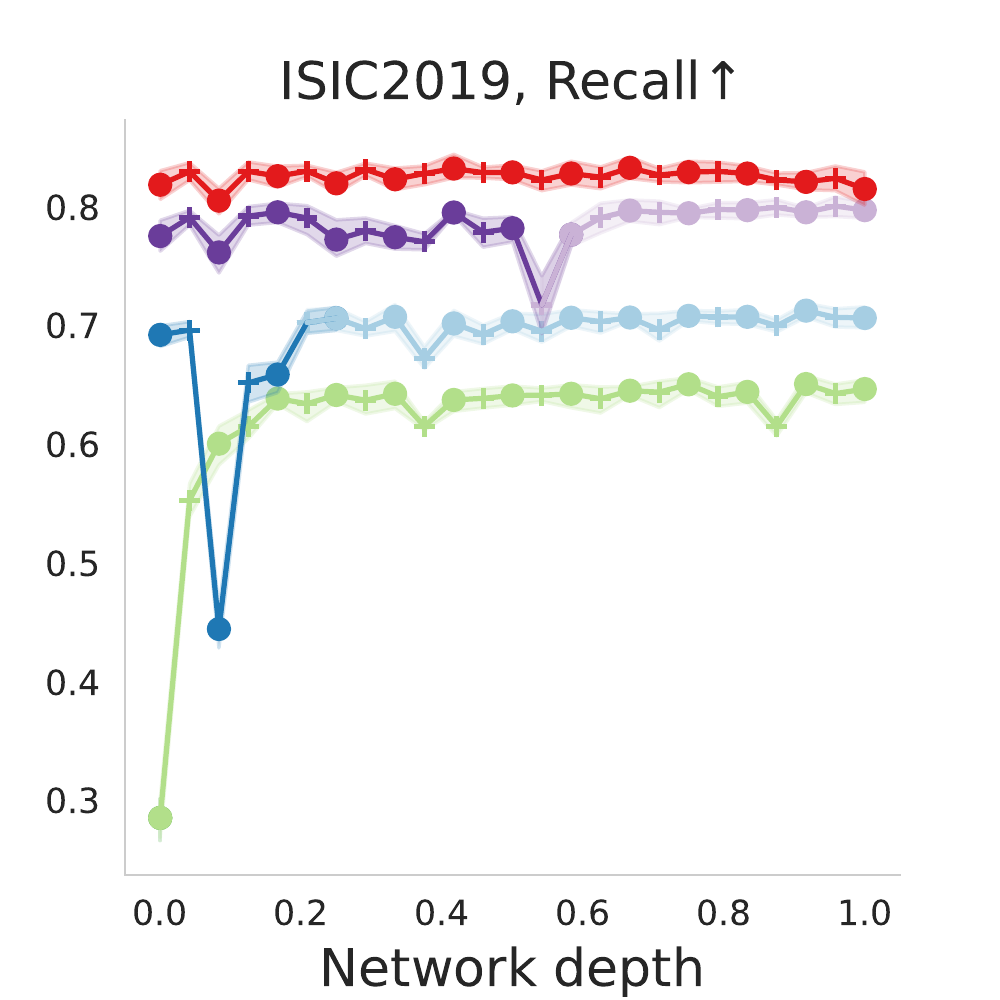} &
    \includegraphics[width=0.25\columnwidth]{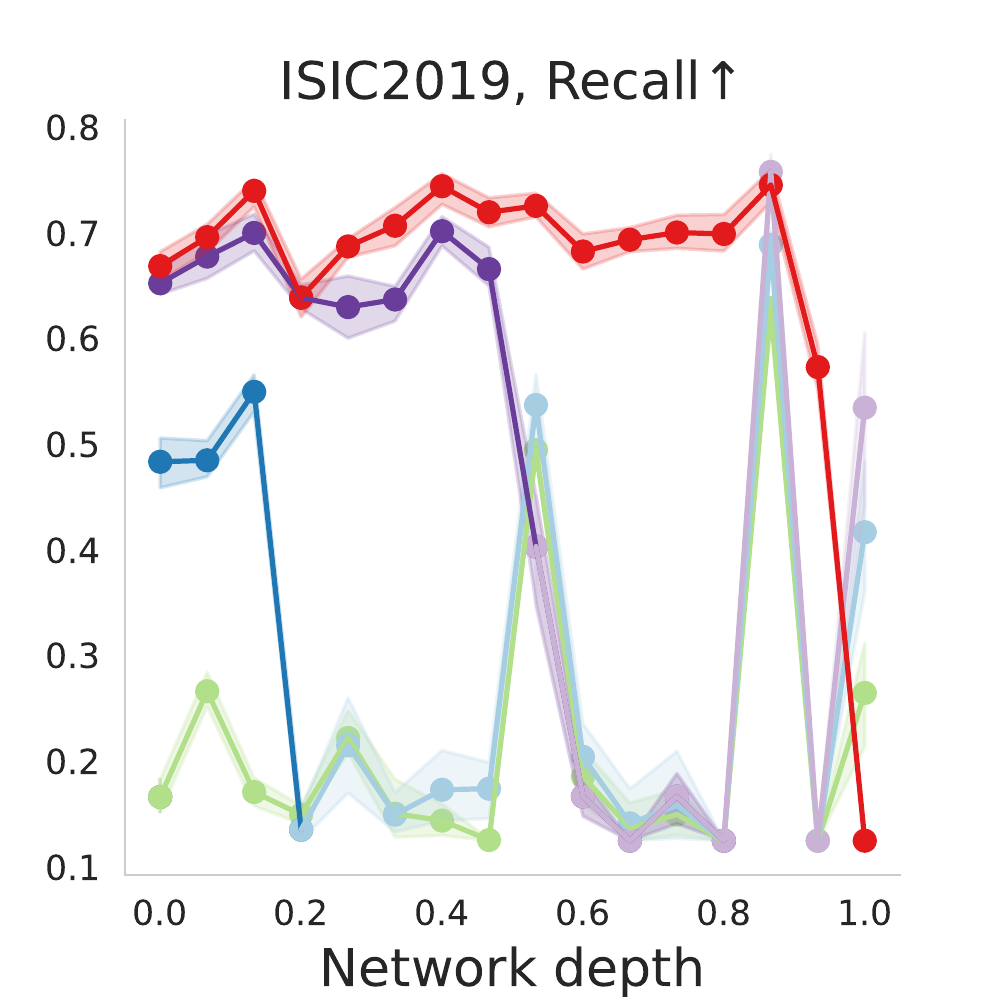} &
    \includegraphics[width=0.25\columnwidth]{images/layer_importance/LI_ISIC2019-resnet50.pdf}\\ [-1.5mm]
    \includegraphics[width=0.25\columnwidth]{images/layer_importance/LI_CheXpert-deit_small.pdf} & 
    \includegraphics[width=0.25\columnwidth]{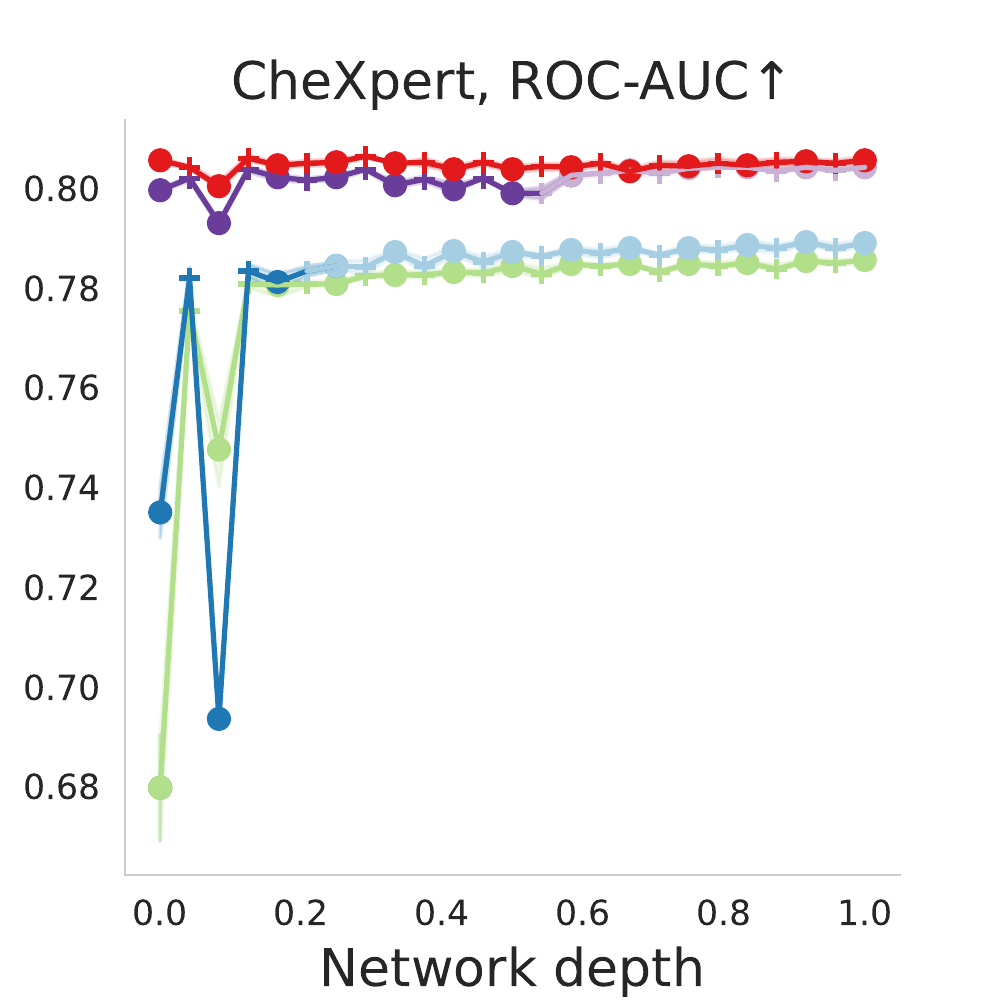} &
    \includegraphics[width=0.25\columnwidth]{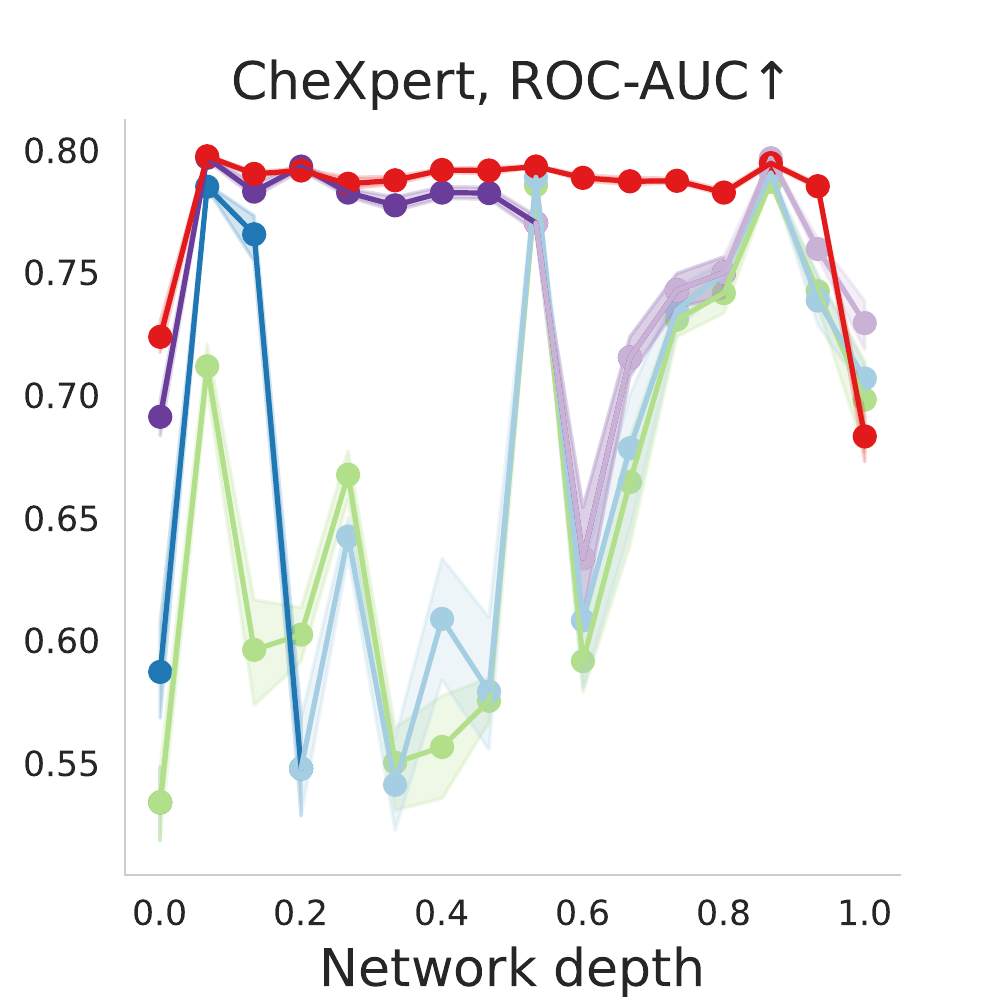} &
    \includegraphics[width=0.25\columnwidth]{images/layer_importance/LI_CheXpert-resnet50.pdf}\\ [-1.5mm]
    \includegraphics[width=0.25\columnwidth]{images/layer_importance/LI_Camelyon-deit_small.pdf} & 
    \includegraphics[width=0.25\columnwidth]{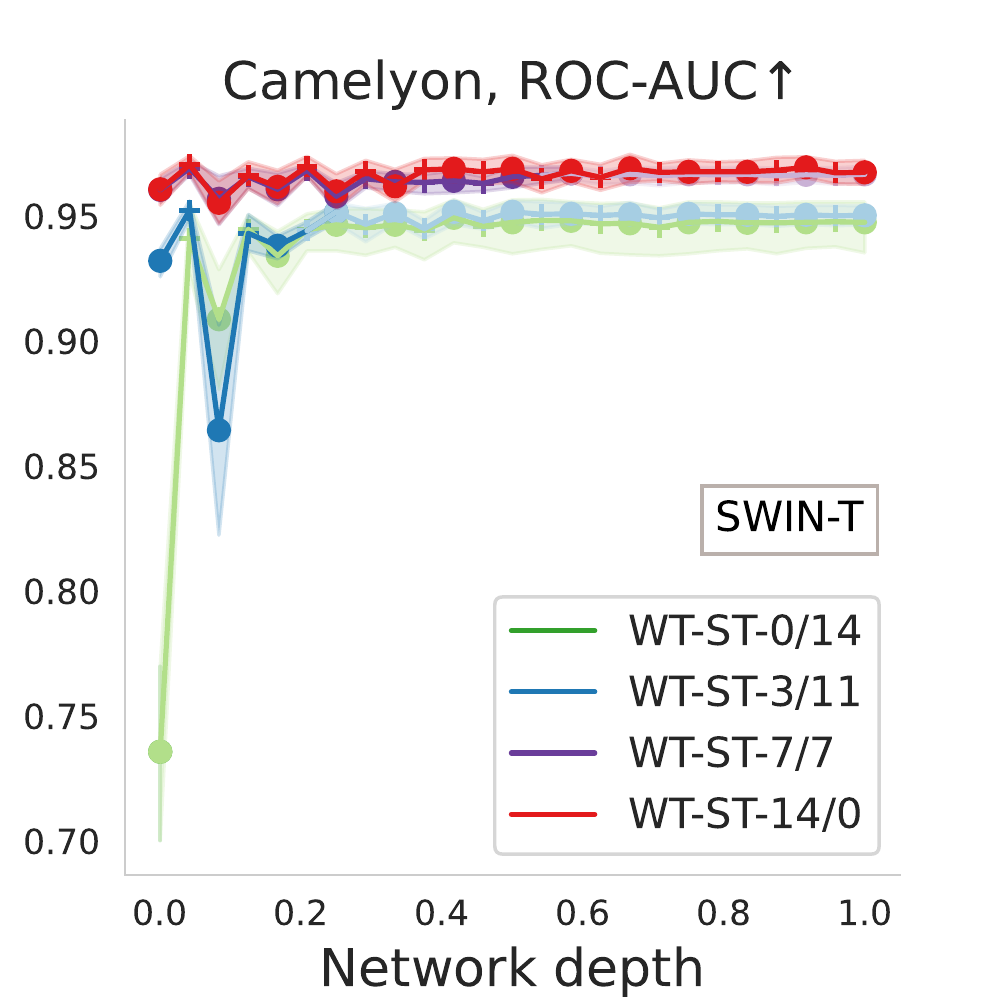} &
    \includegraphics[width=0.25\columnwidth]{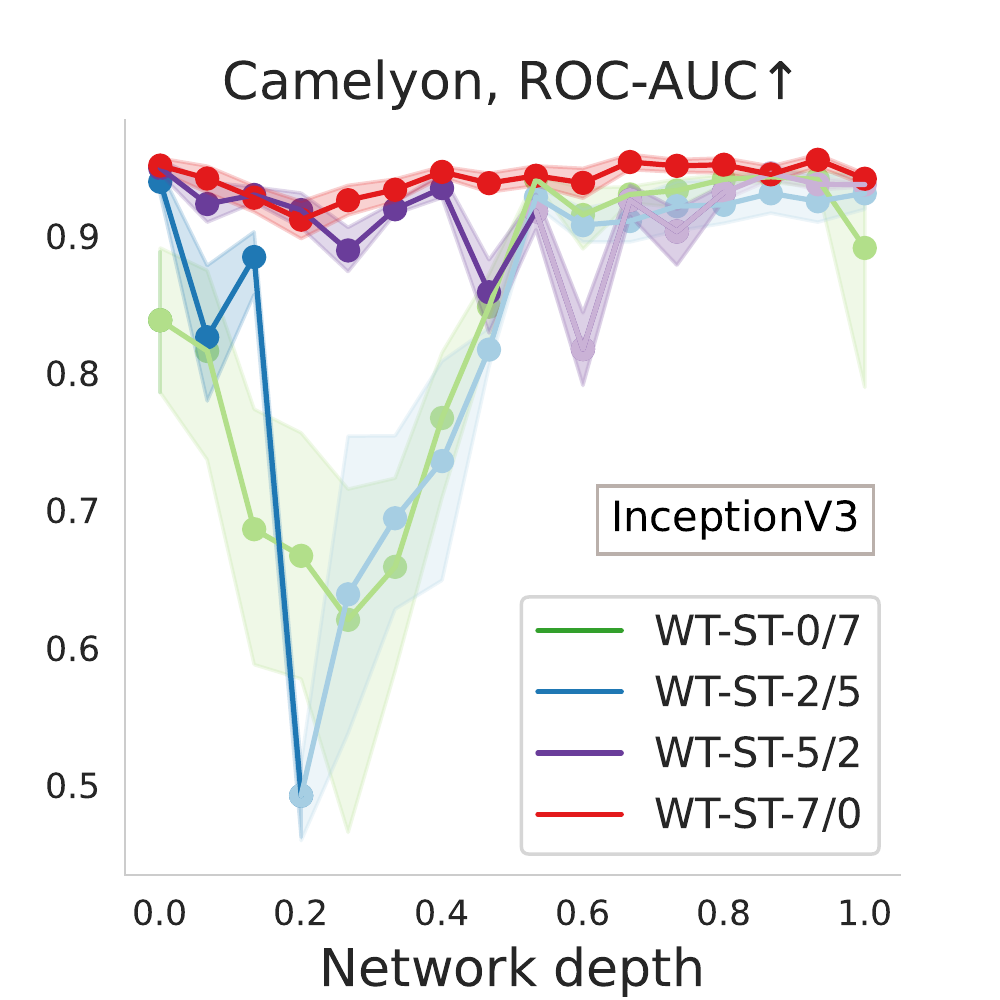} &
    \includegraphics[width=0.25\columnwidth]{images/layer_importance/LI_Camelyon-resnet50.pdf}\\ [-1.5mm]
\end{tabular}
\end{center}
\vspace{-3mm}

\caption{\emph{Resilience of trained layers to change}. We report the performance when reverting a fine-tuned layer back to its original state. This is done using one layer at a time, for each dataset (rows) and model types (columns) for four different {\wtst} initialization strategies.
The results show that layers with low robustness underwent critical changes during fine-tuning. In ViTs, critical layers often appear at the transition between WT and ST. CNNs on the other hand, exhibit poor robustness at the final layers responsible for classification, and also periodically within the network at critical layers. 
}
\label{fig:LI_apx}
\vspace{-4mm}
\end{figure}

In the layer re-initialization experiments, we consider the impact of reverting individual layer of the models to their initial state. 
In detail, we initialize models with different \wtst schemes. Then, after fine-tuning, we reinitialize a single layer at a time to its original state, while keeping all the other layers unchanged.
Finally, we evaluate the model on the test set and measure the drop in predictive performance.

For {\deits} and {\swin}, the intermediate modules include the patchifier, the  self-attention layers of each block separately. For {\resnetfifty}, the modules include the first convolutional layer and the residual blocks of each stage. For {\inception} the modules consist of all the initial convolutional blocks, and all the individual inception modules. A detailed description of the layers that were used can be seen in Table \ref{tab:cnn_layer_details} and Table \ref{tab:vit_layer_details} in Appendix. The results of these experiments can be seen in Figure \ref{fig:layer_importance} in the main text for {\deit} and Figure \ref{fig:LI_apx} for  other model types.

\section{\ltwo distance}
\label{appdx:L2experiment}
In order to understand the extent that model's weights change during training, we calculate the \ltwo norm of the weights before and after fine-tuning.
In practice, for each layer, we calculate the \ltwo distances between the original and fine-tuned weights and then we divide this value by the number of the weights in the layer.
The details of the layers used for each model can be seen in Table \ref{tab:cnn_layer_details} and Table \ref{tab:vit_layer_details} in the Appendix.  Figure \ref{fig:L2_appdx} shows the the results for each model and dataset individually, and Figure \ref{fig:L2_dist} in the main text shows the distances averaged over all the datasets for each model.

\begin{figure}[t]
\begin{center}
\begin{tabular}{@{}c@{}c@{}c@{}c@{}}
    \includegraphics[width=0.25\columnwidth]{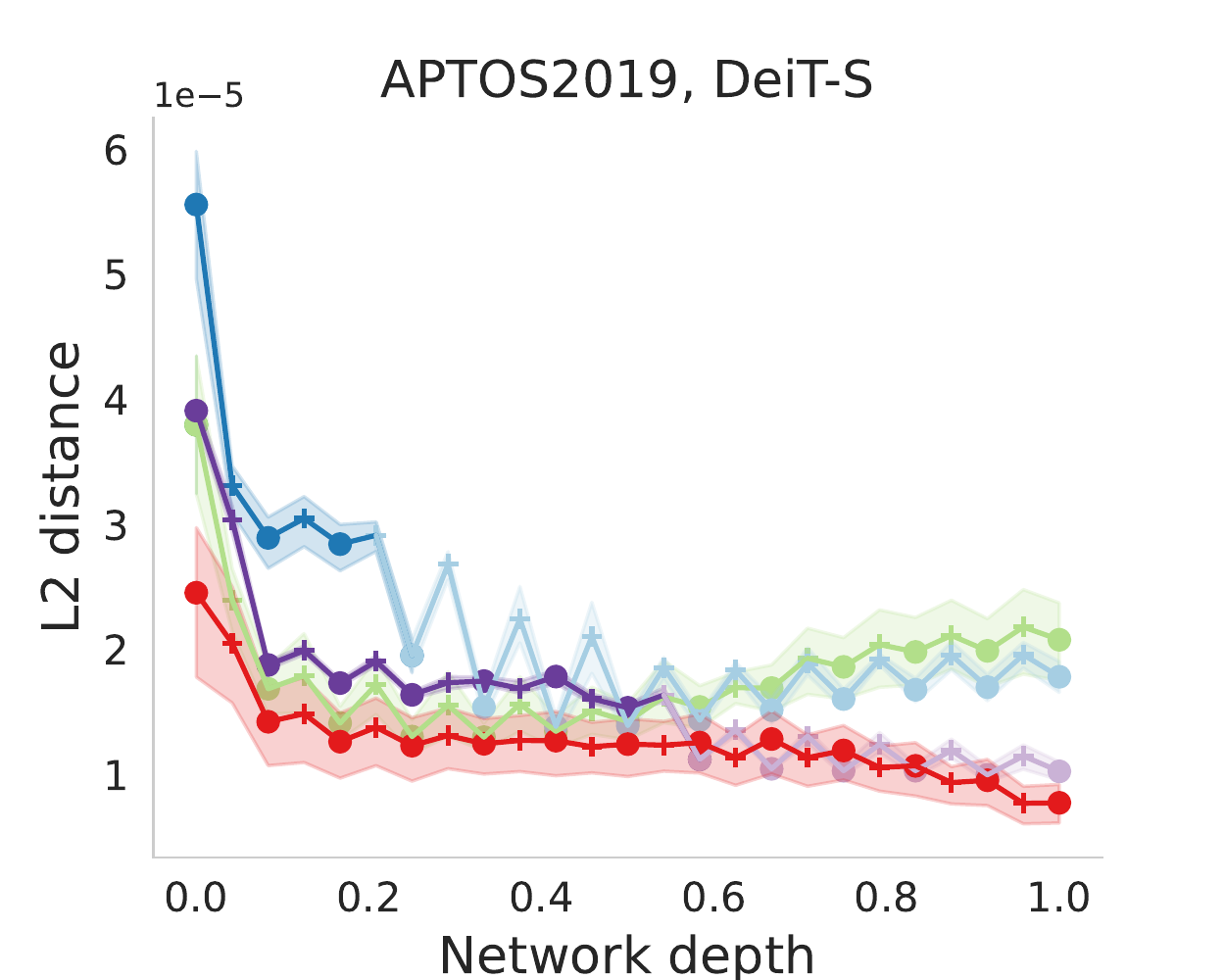} & 
    \includegraphics[width=0.25\columnwidth]{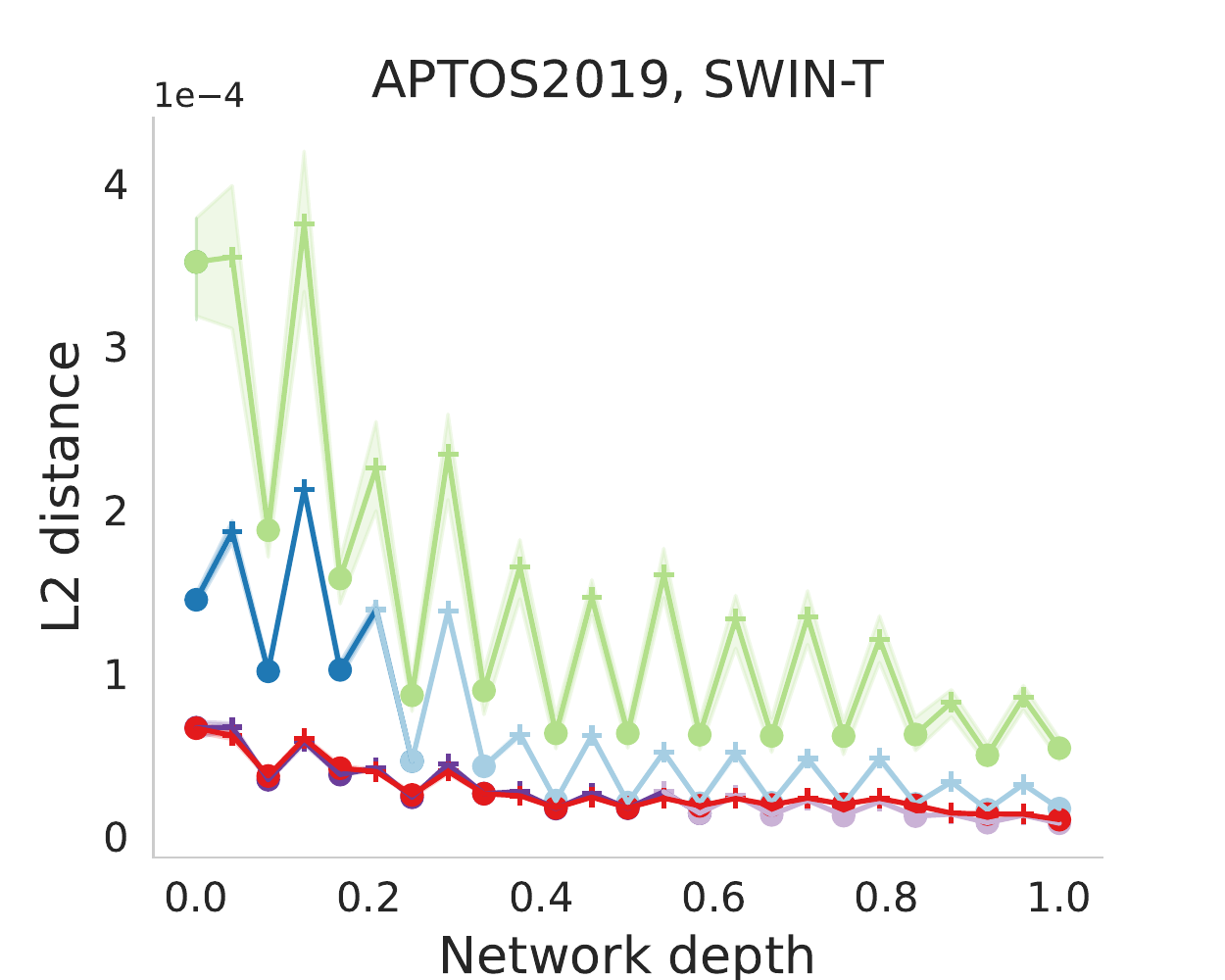} &
    \includegraphics[width=0.25\columnwidth]{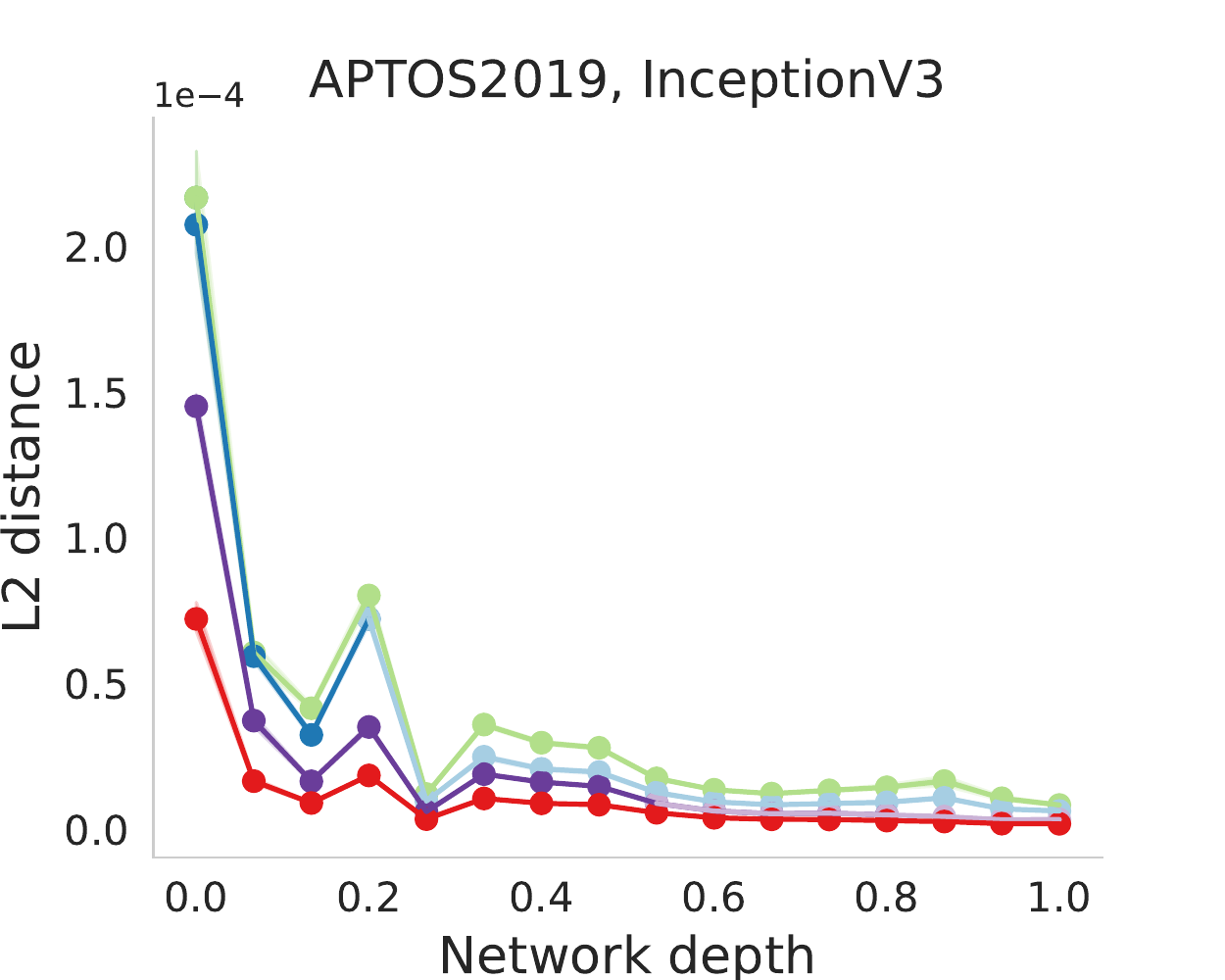} &
    \includegraphics[width=0.25\columnwidth]{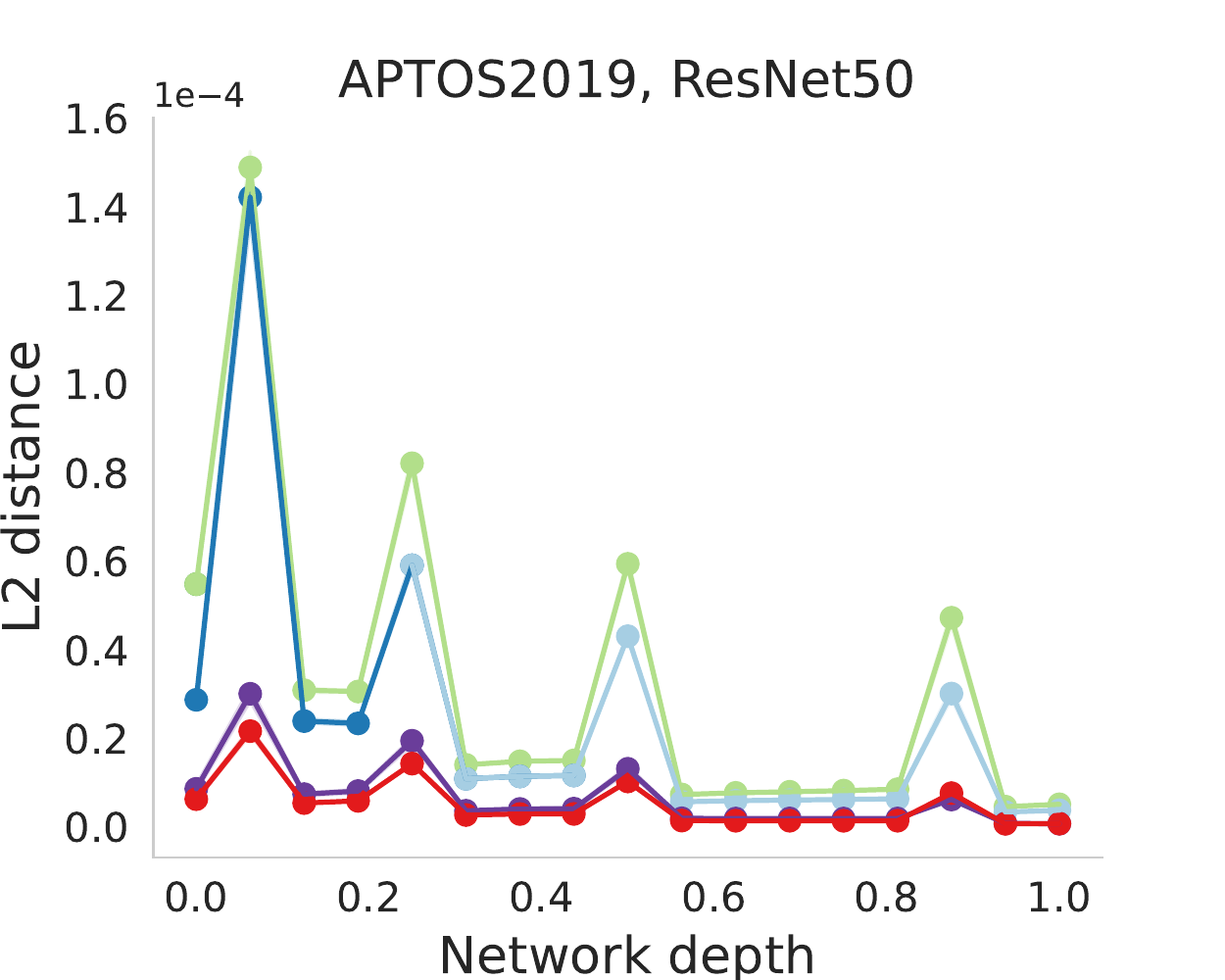}\\ [-1.5mm]
    \includegraphics[width=0.25\columnwidth]{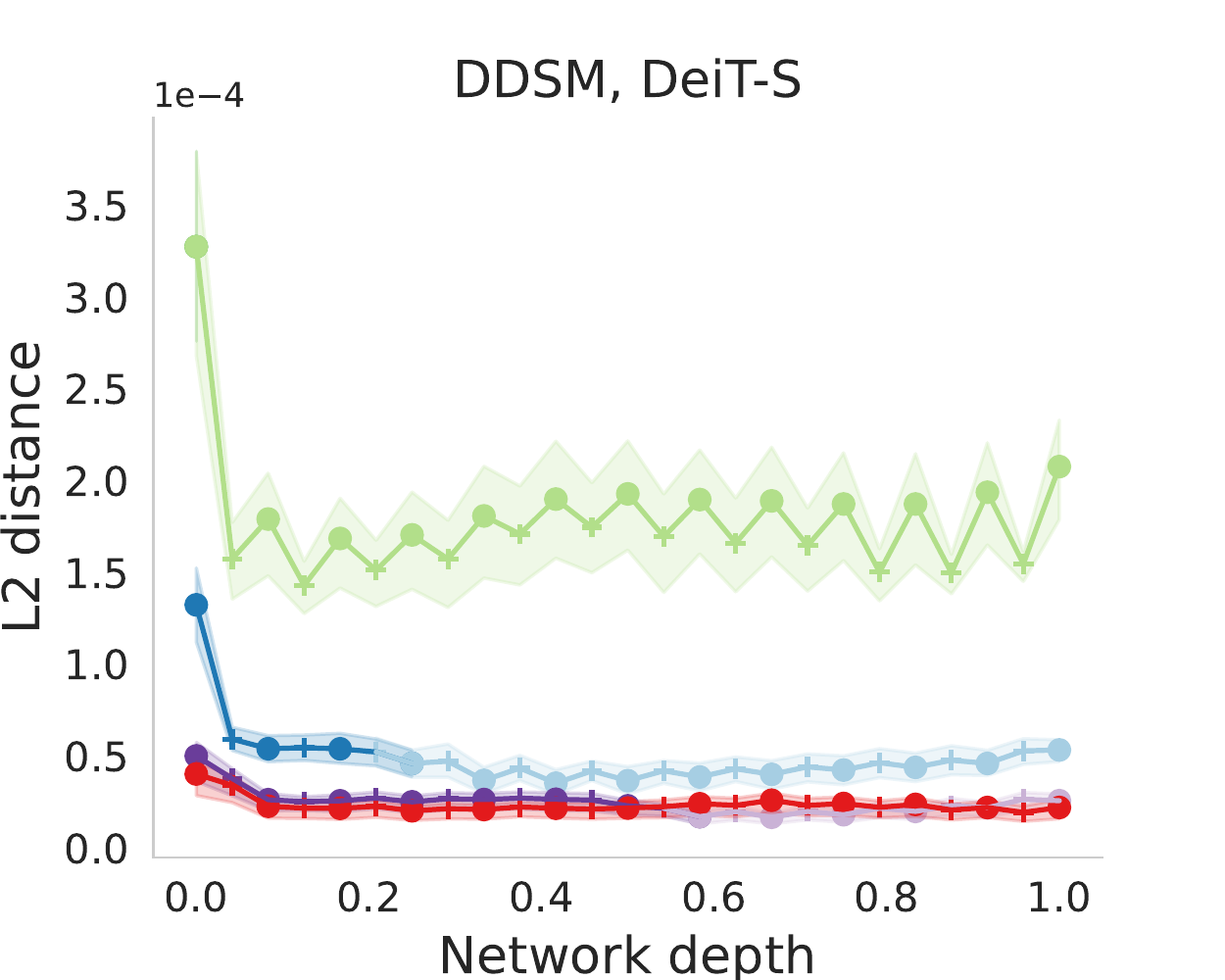} & 
    \includegraphics[width=0.25\columnwidth]{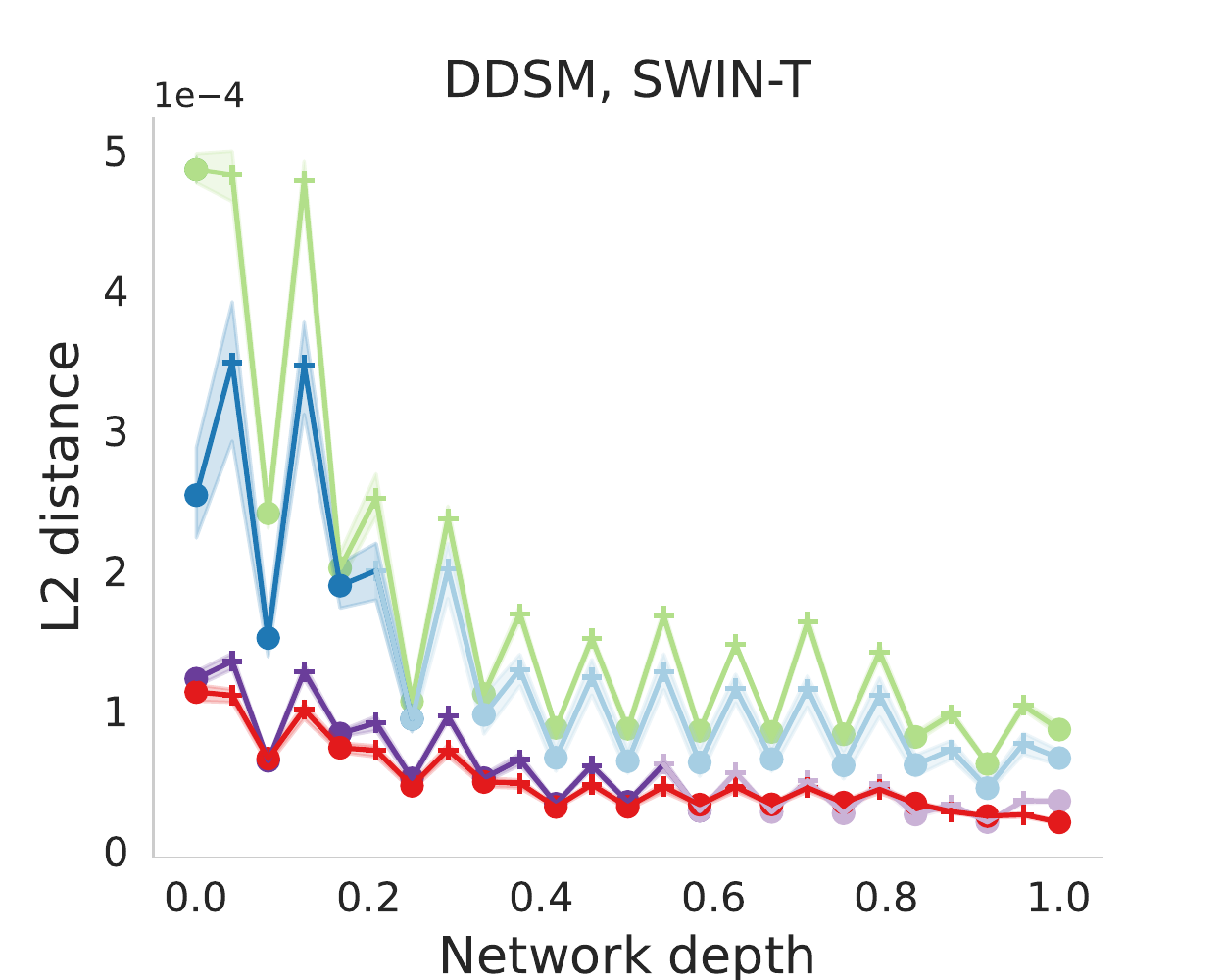} &
    \includegraphics[width=0.25\columnwidth]{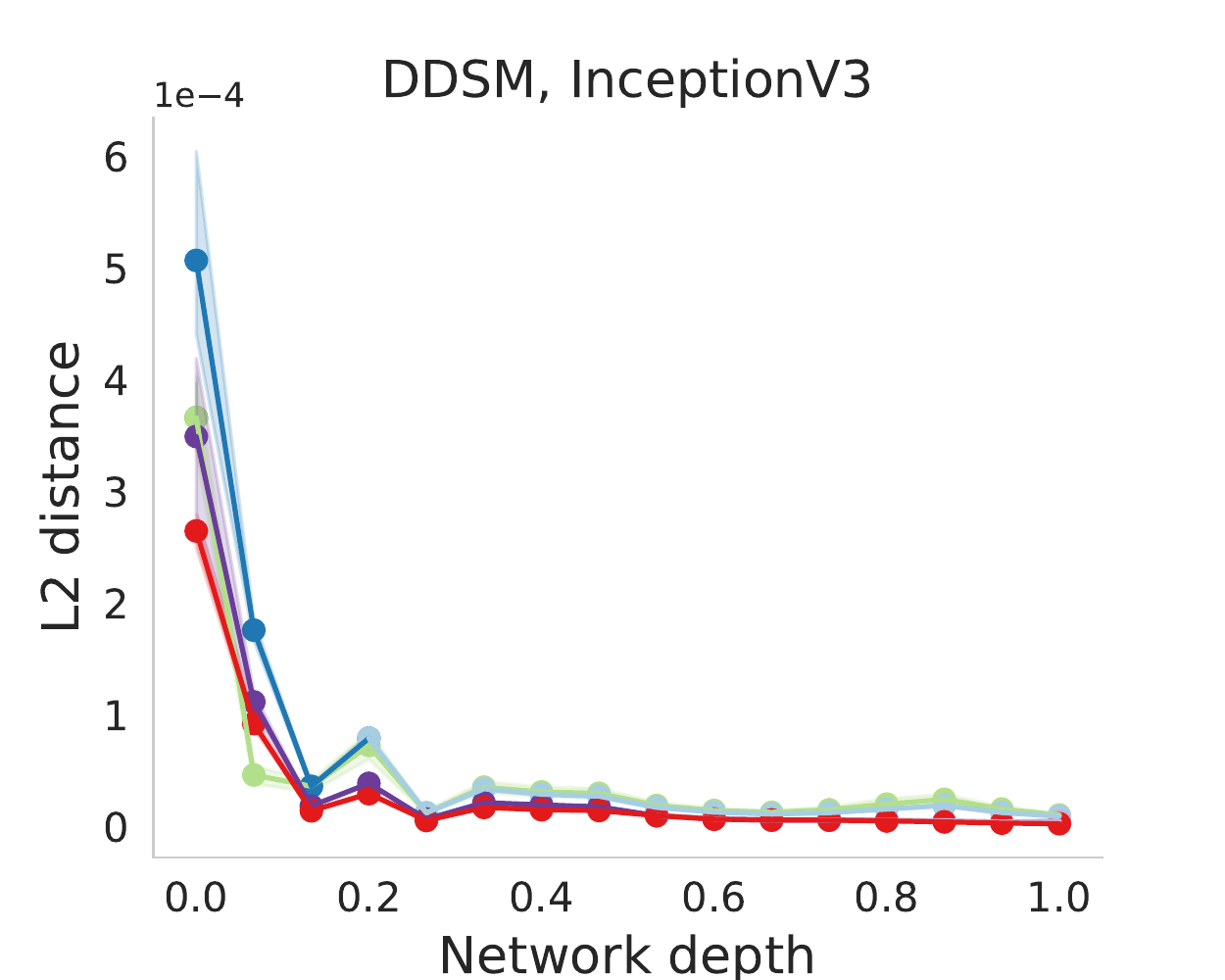} &
    \includegraphics[width=0.25\columnwidth]{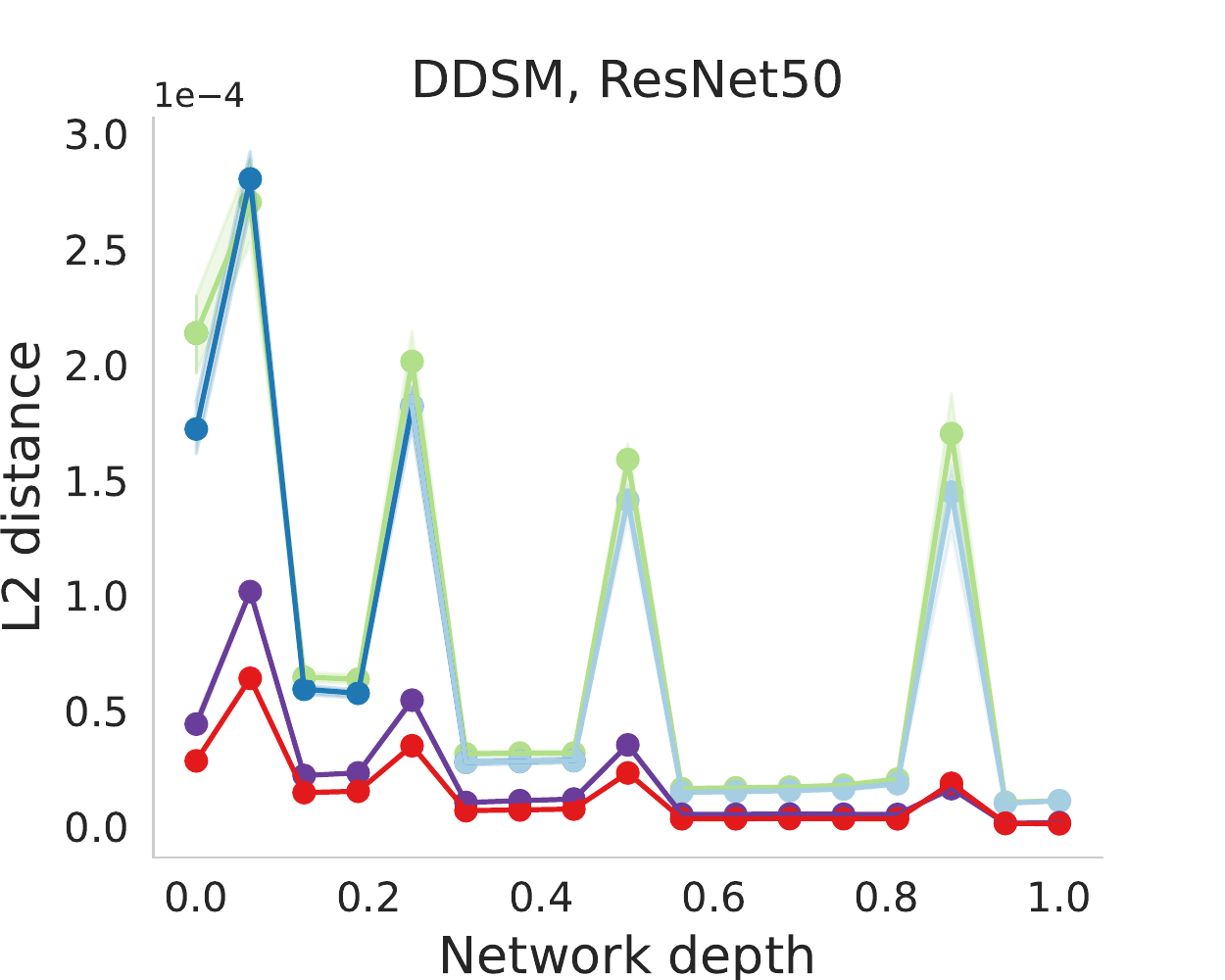} \\[-1.5mm]
    \includegraphics[width=0.25\columnwidth]{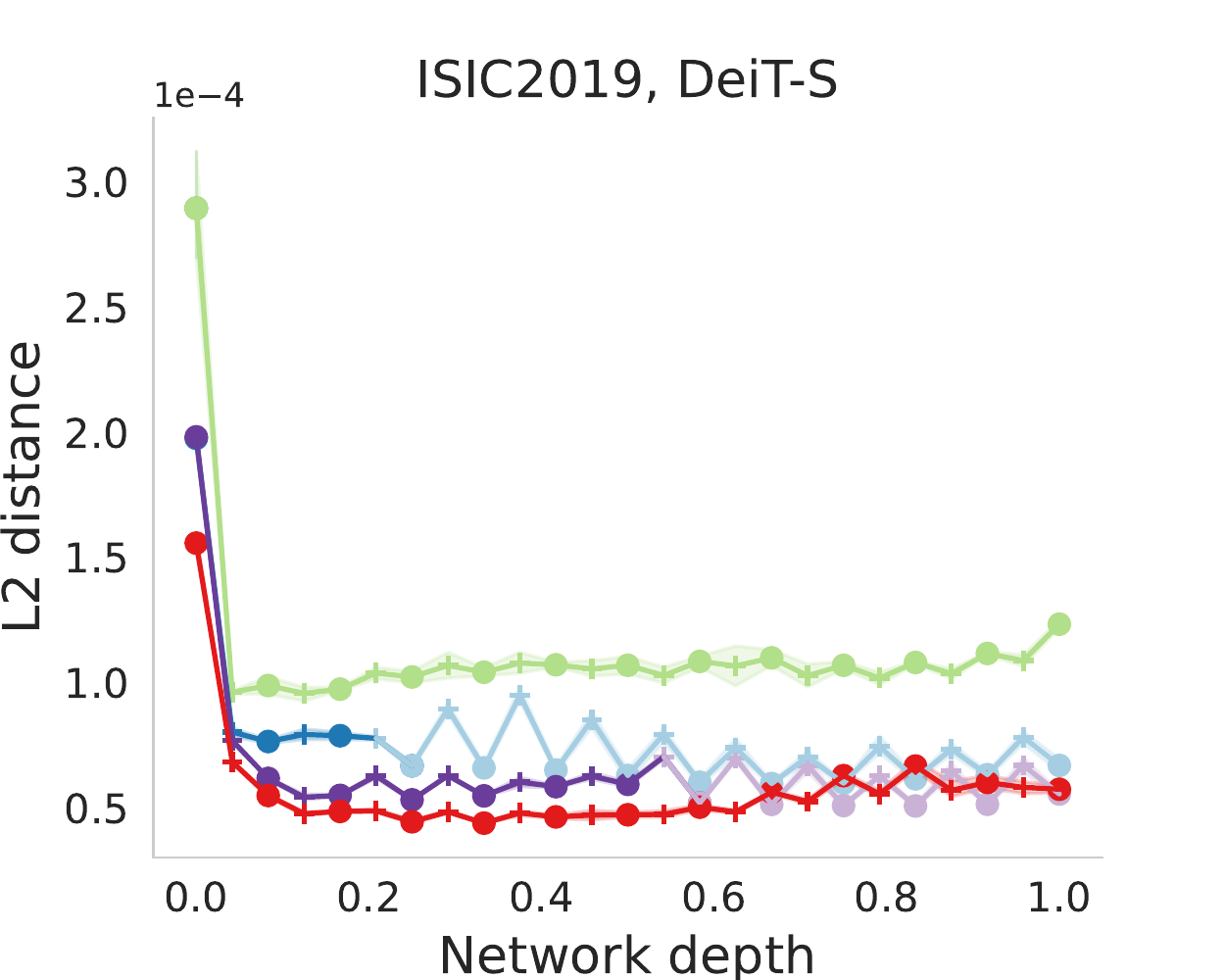} & 
    \includegraphics[width=0.25\columnwidth]{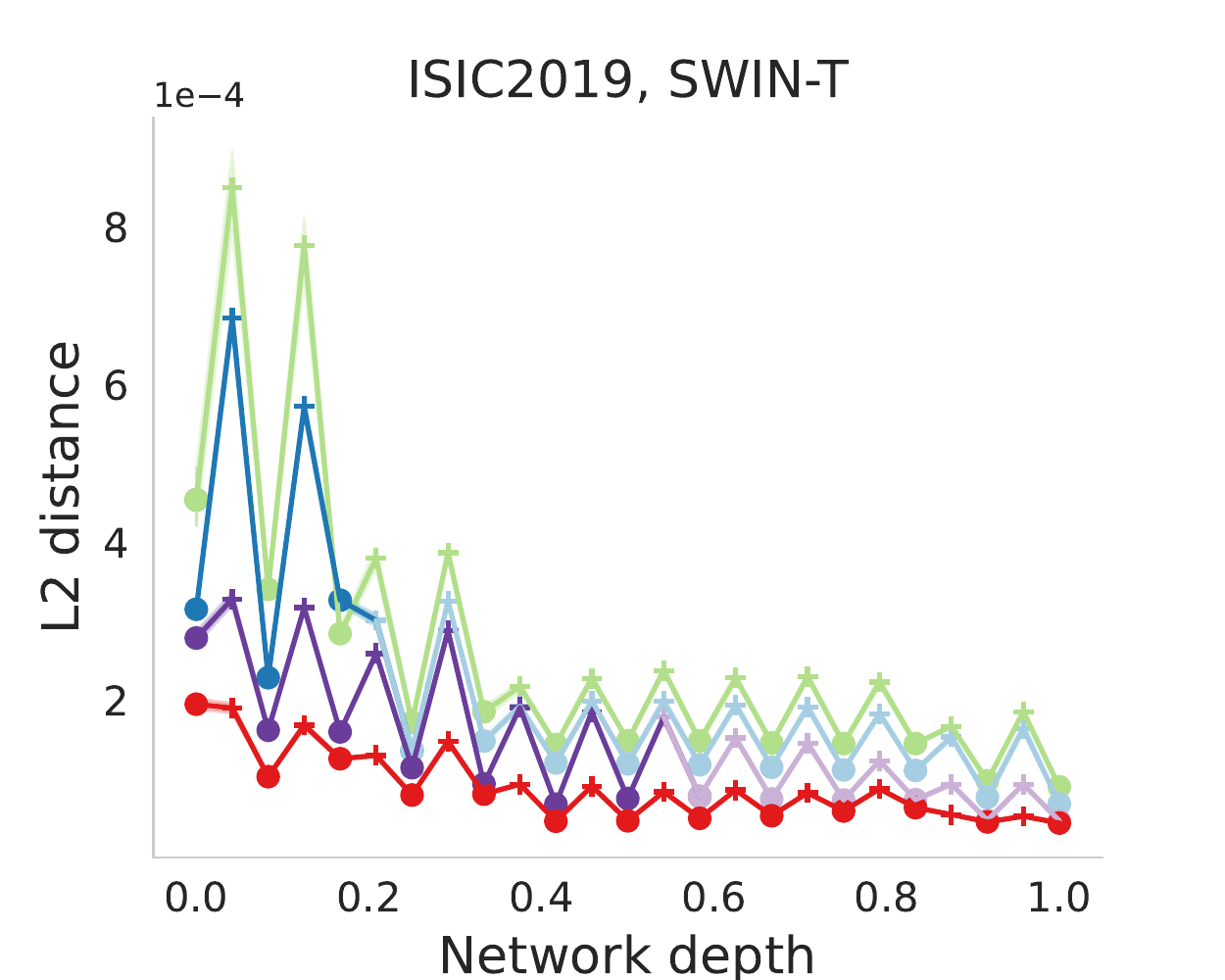} &
    \includegraphics[width=0.25\columnwidth]{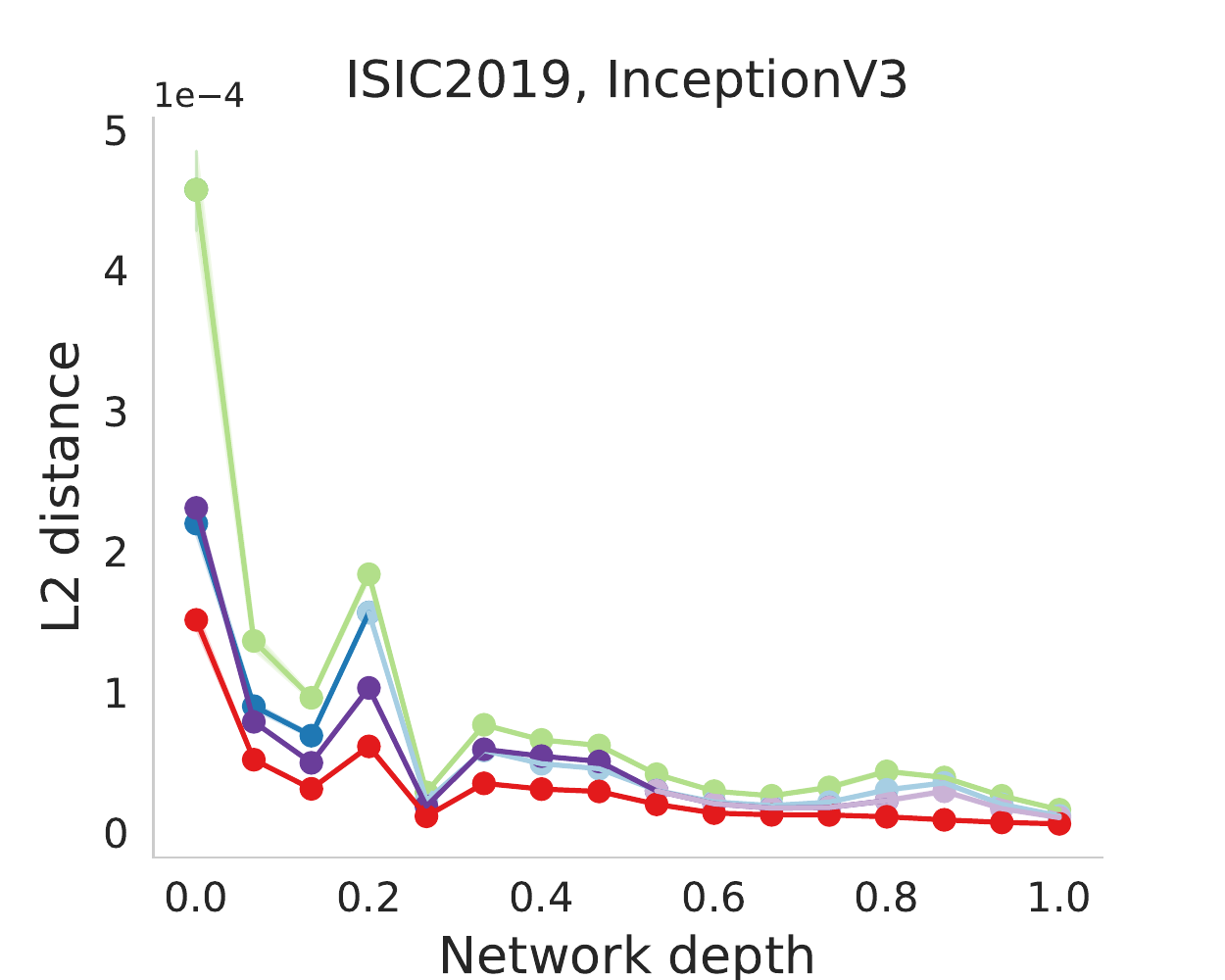} &
    \includegraphics[width=0.25\columnwidth]{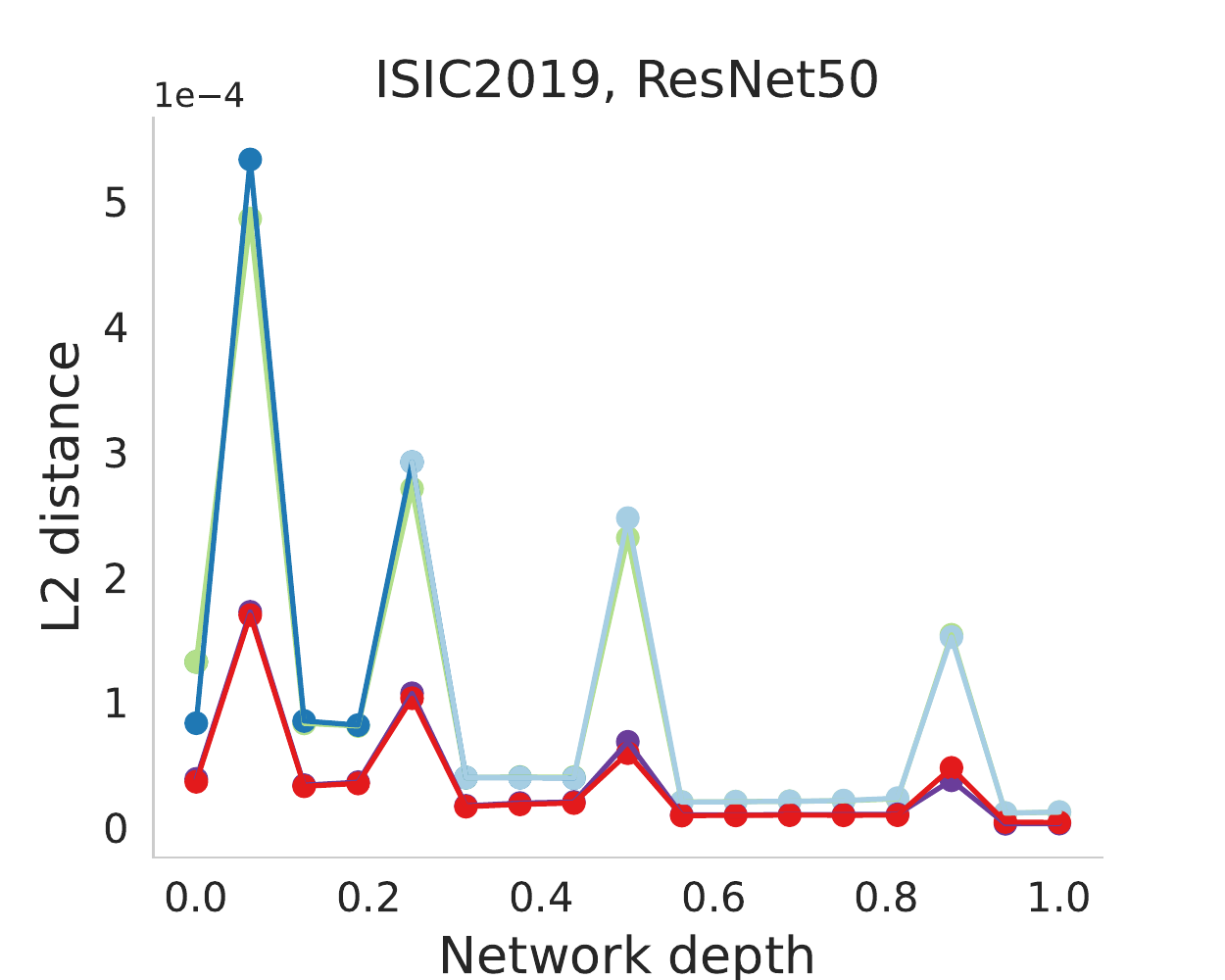}\\ [-1.5mm]
    \includegraphics[width=0.25\columnwidth]{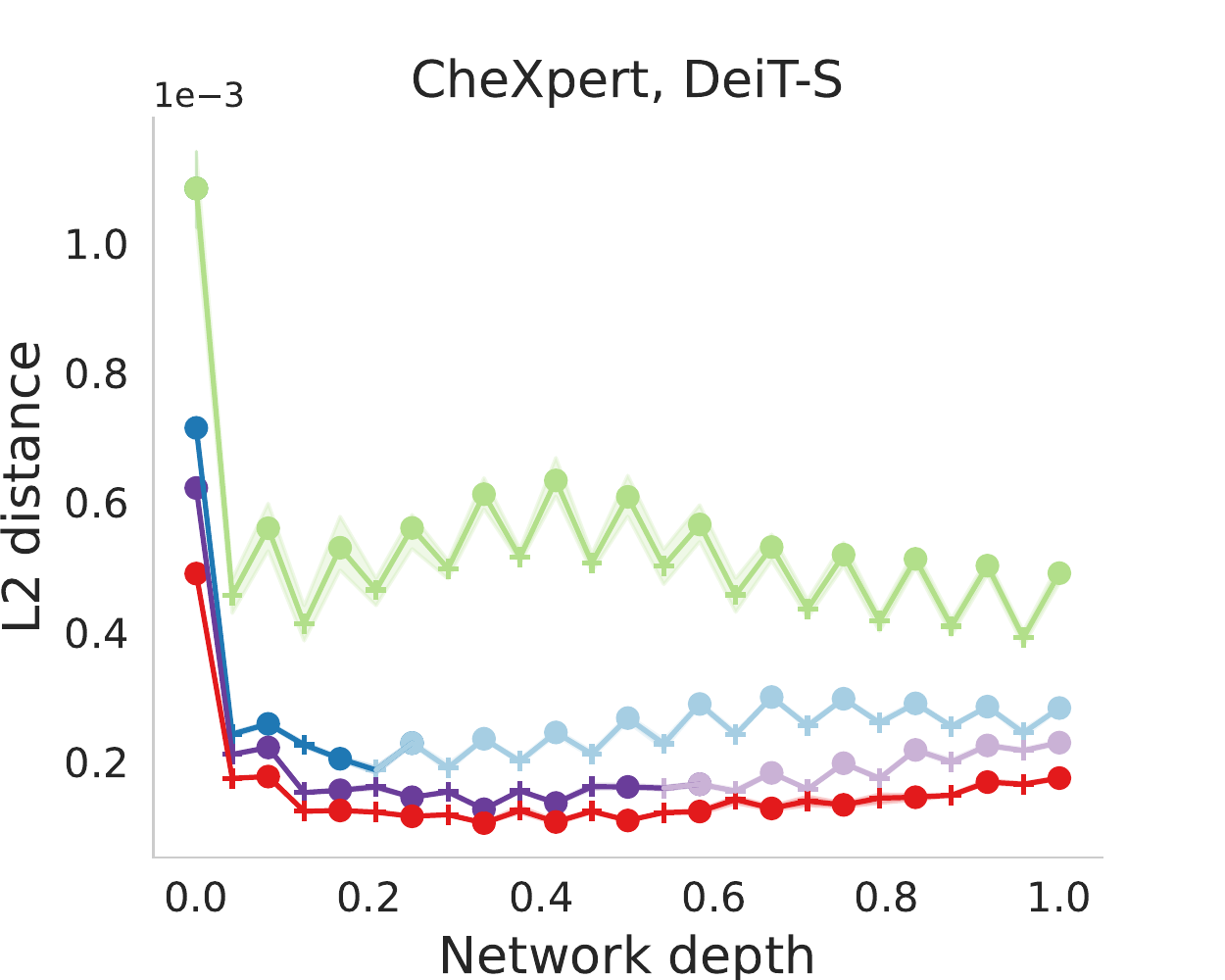} & 
    \includegraphics[width=0.25\columnwidth]{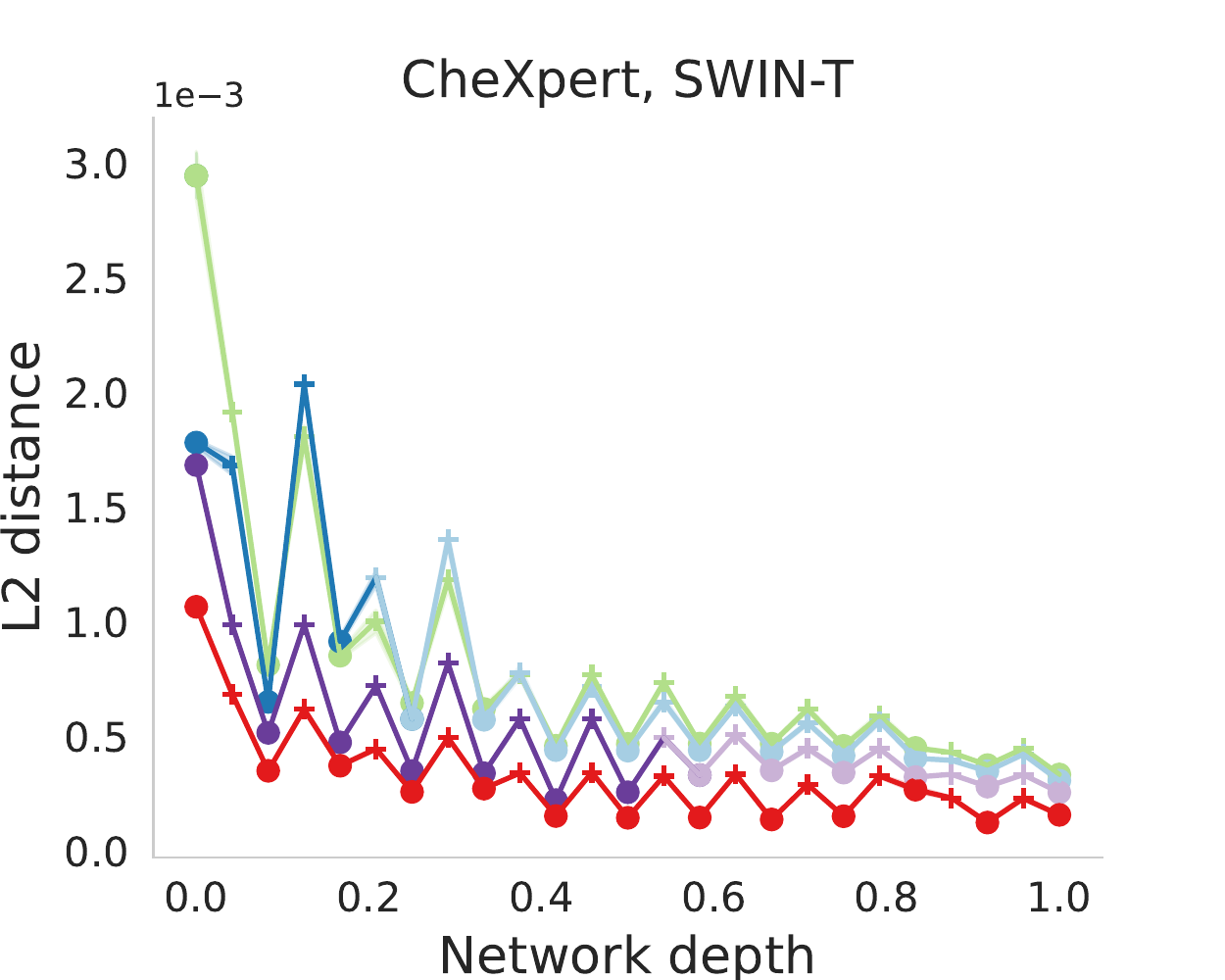} &
    \includegraphics[width=0.25\columnwidth]{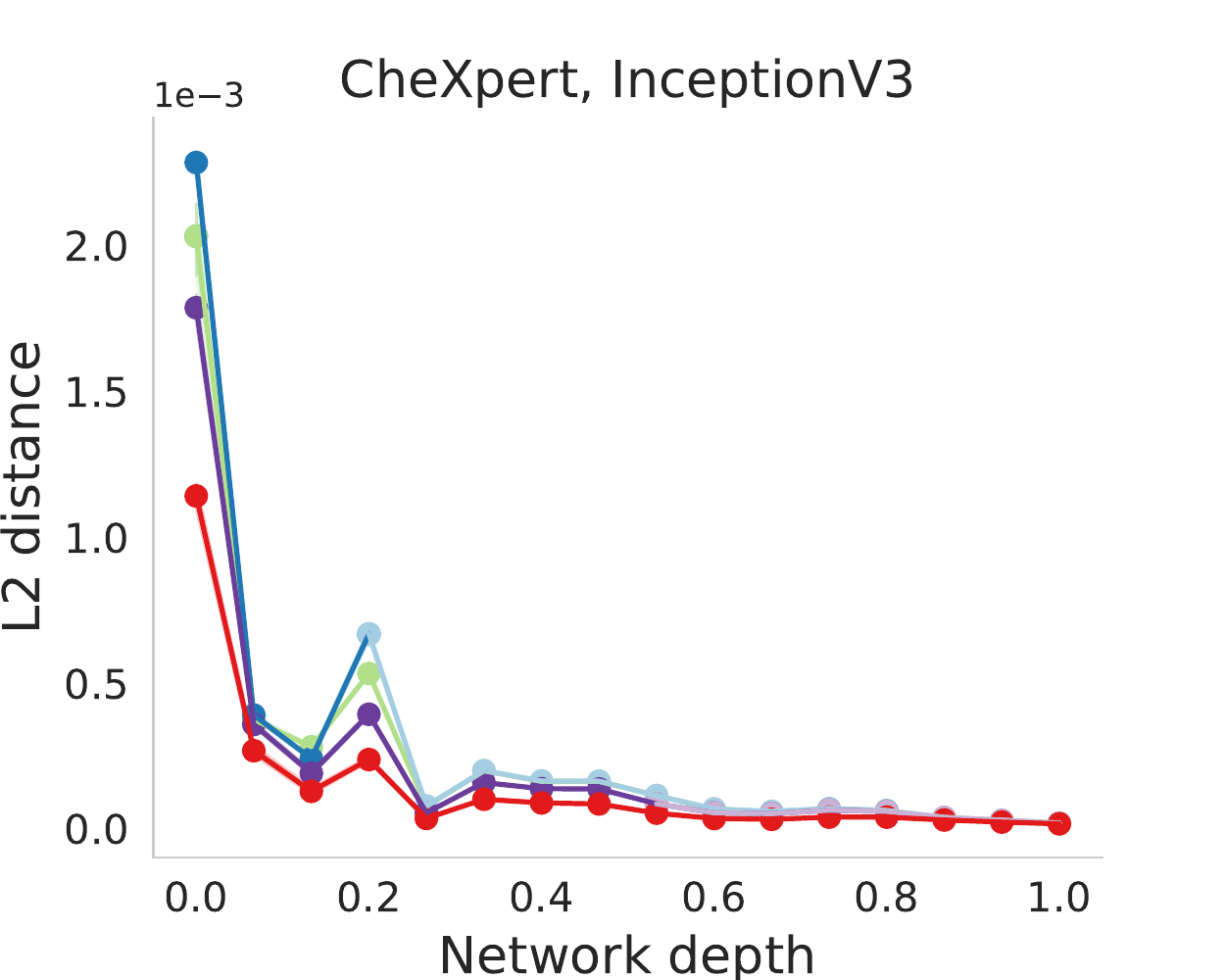} &
    \includegraphics[width=0.25\columnwidth]{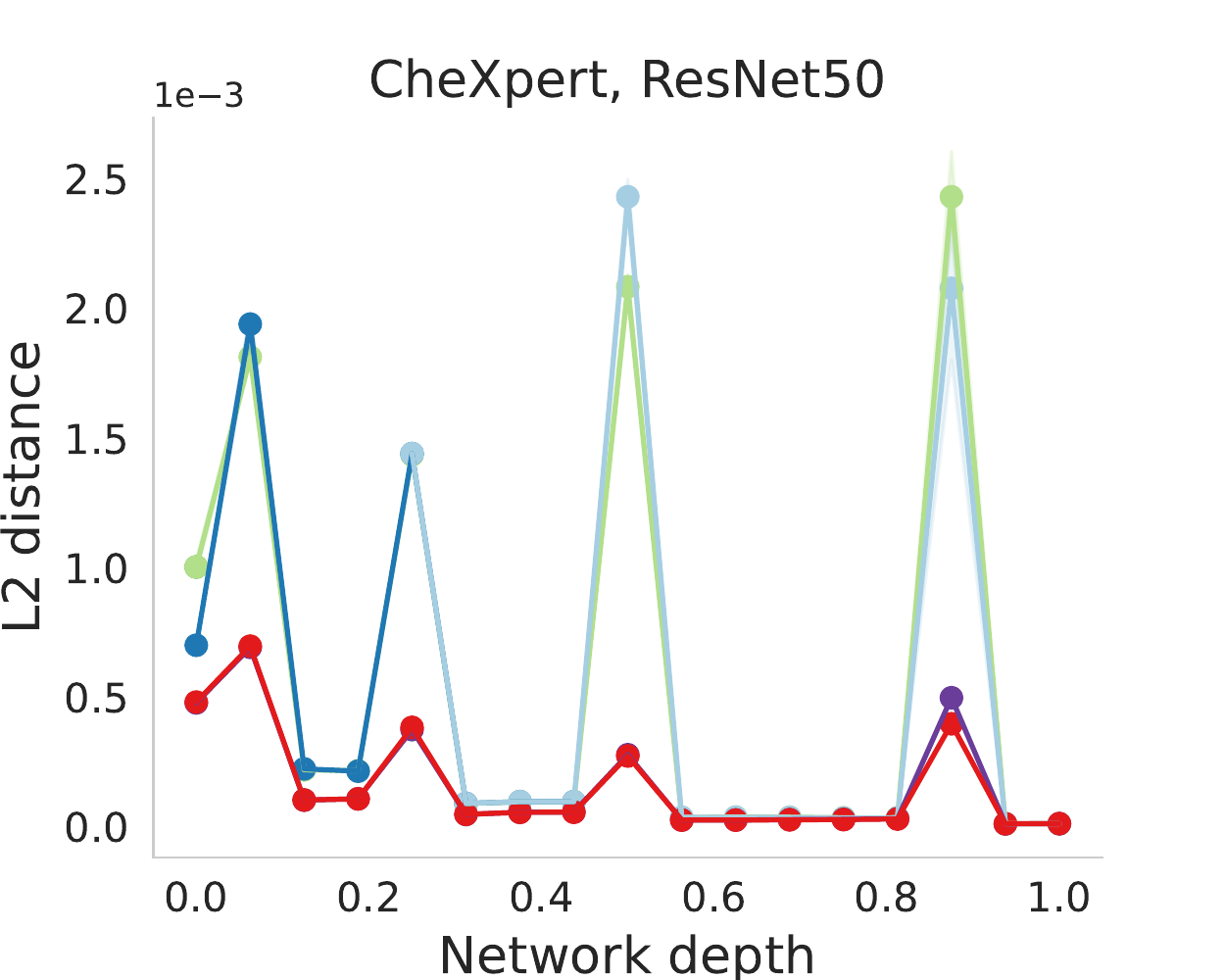}\\ [-1.5mm]
    \includegraphics[width=0.25\columnwidth]{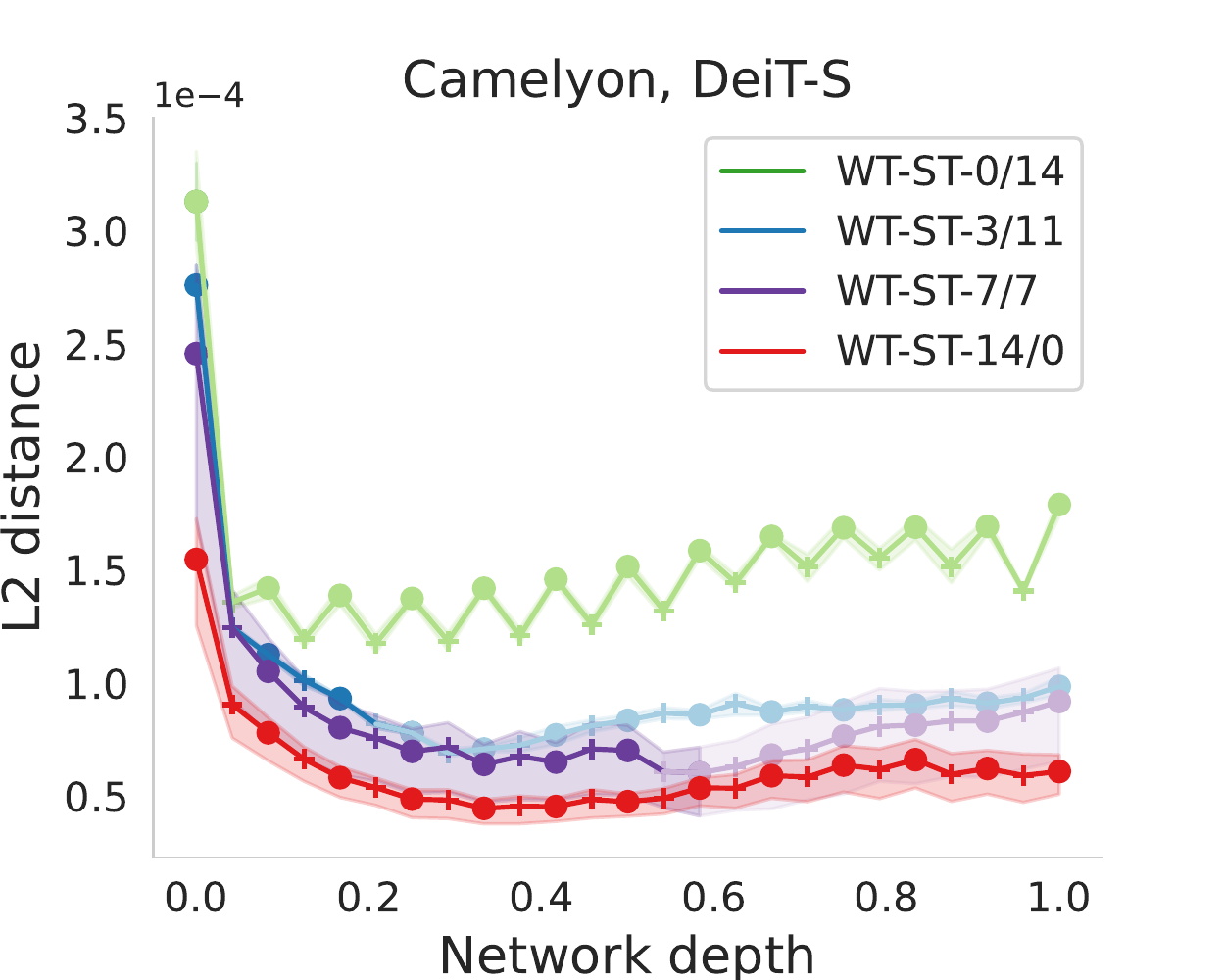} & 
    \includegraphics[width=0.25\columnwidth]{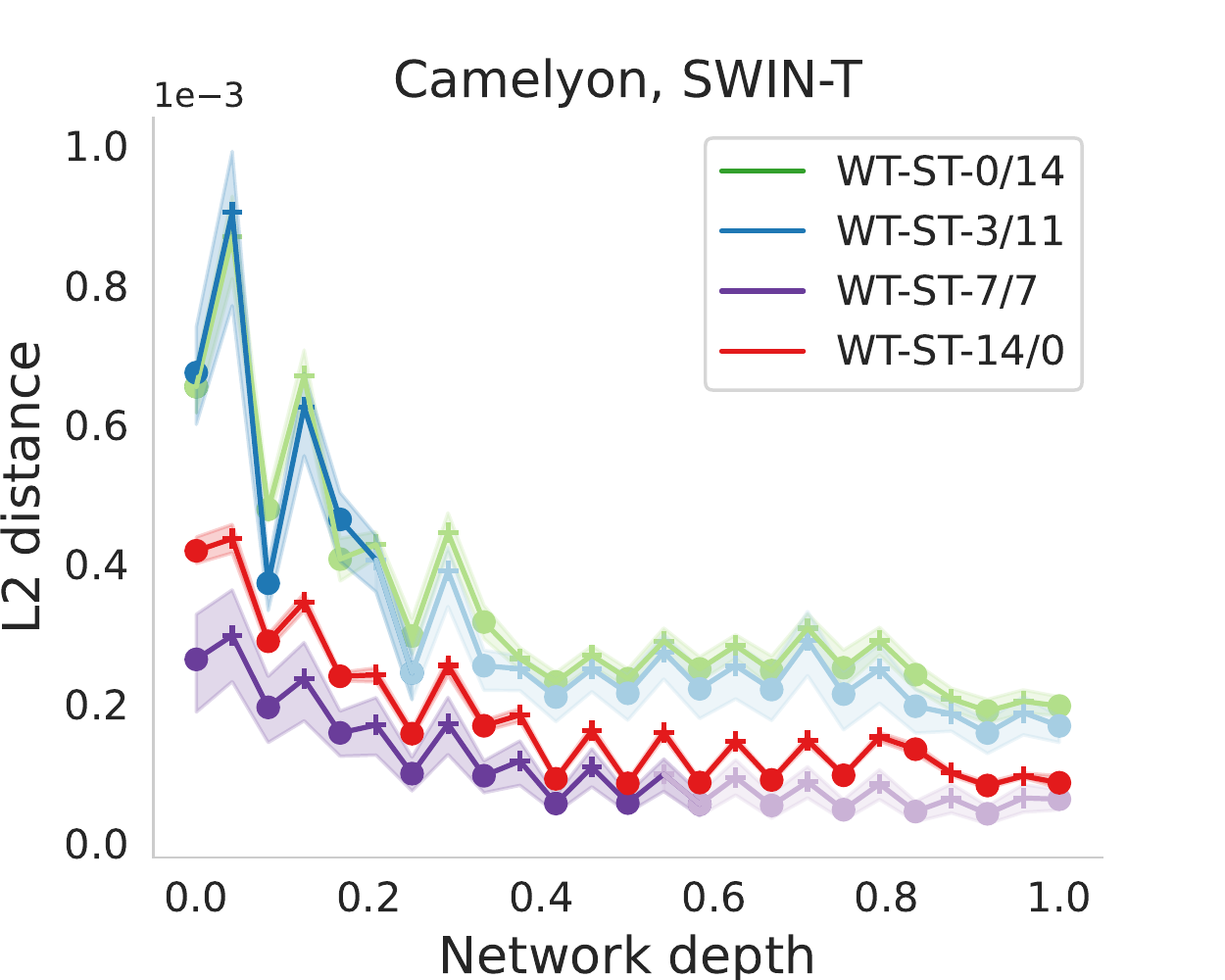} &
    \includegraphics[width=0.25\columnwidth]{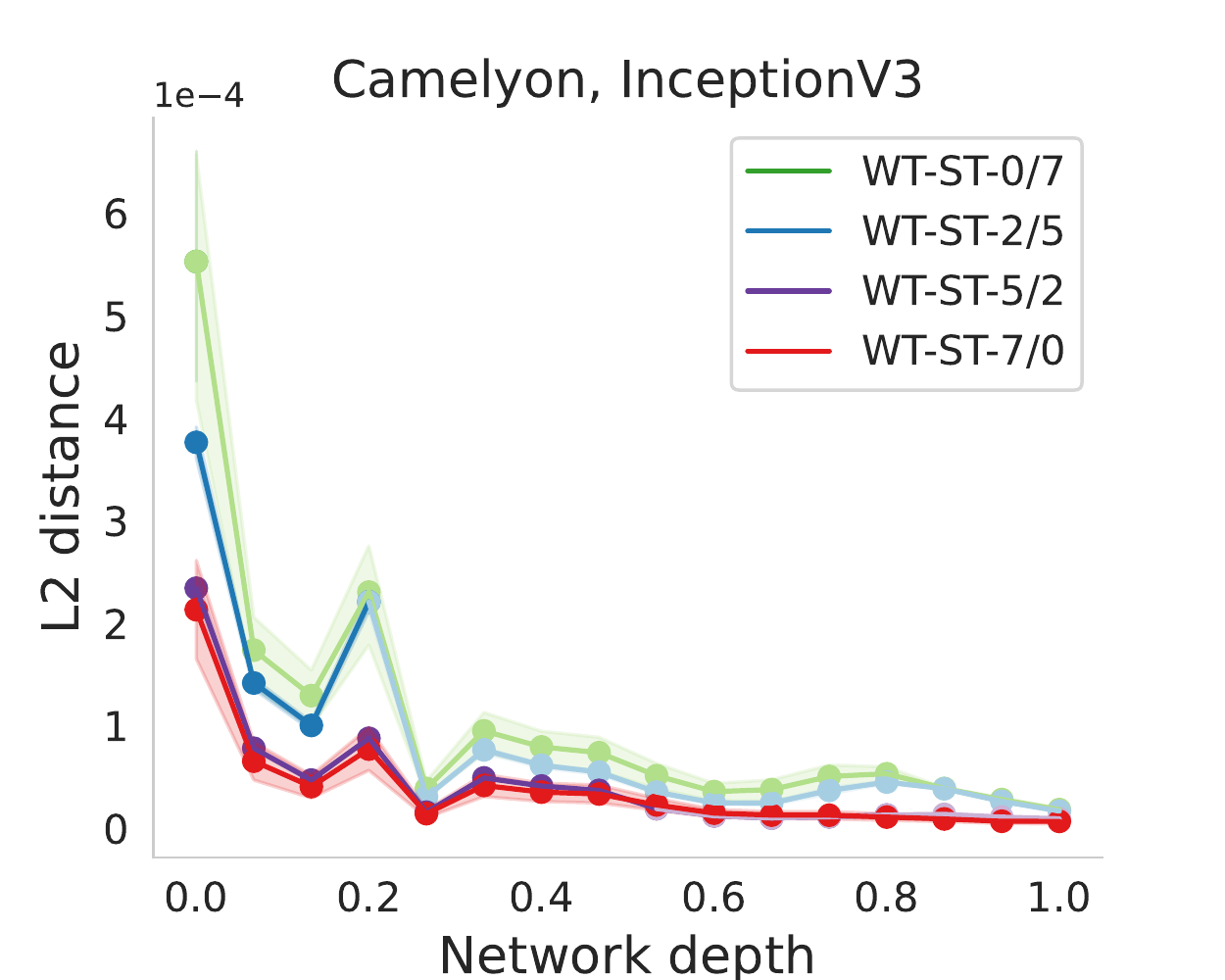} &
    \includegraphics[width=0.25\columnwidth]{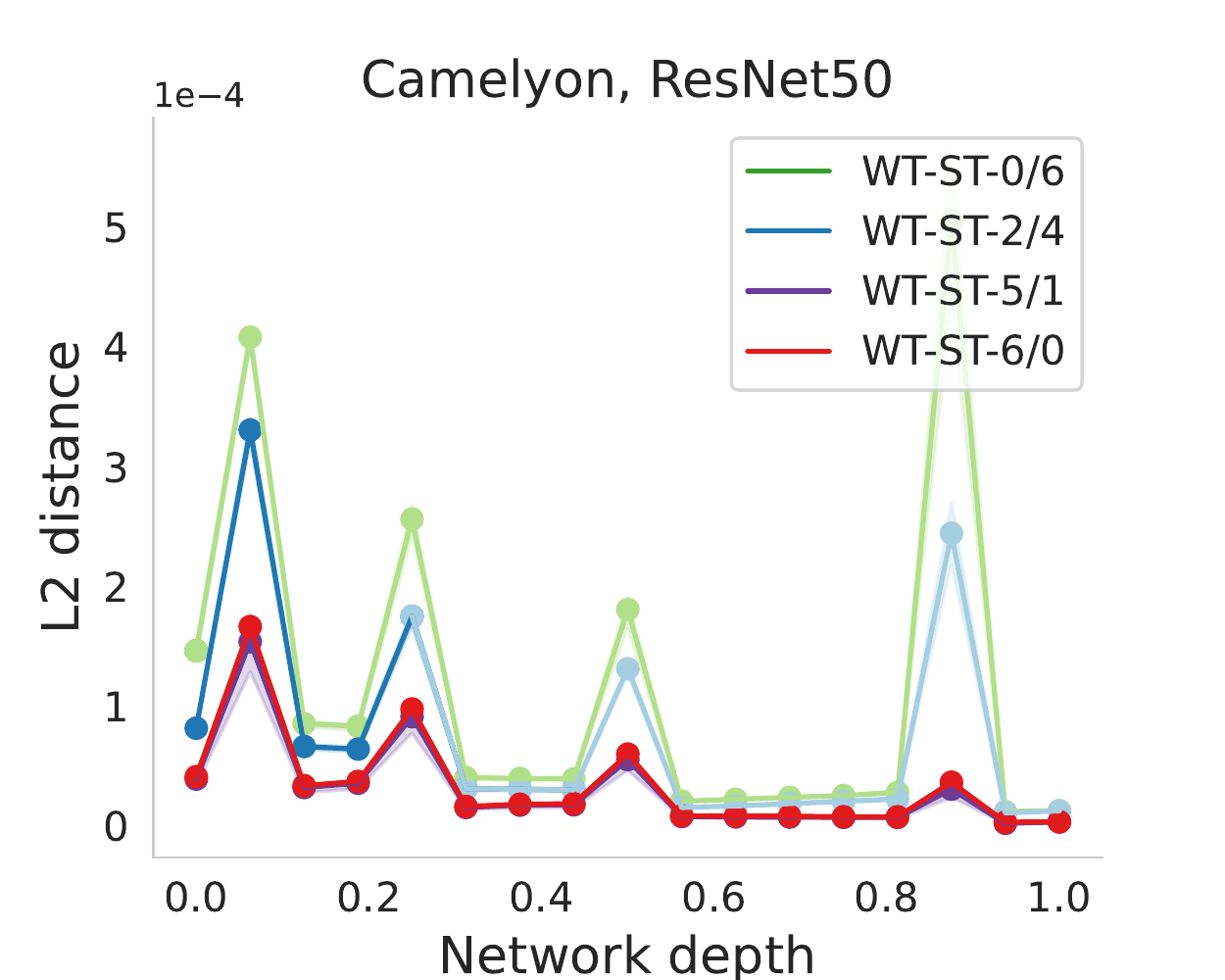}\\ [-1.5mm]
\end{tabular}
\end{center}
\vspace{-3mm}

\caption{\emph{\ltwo distance of the weights.}
We report the the mean \ltwo distances between the initial and trained weights for different models with different initialization schemes. A large distance indicates that the corresponding layer has changed significantly during training.
}
\label{fig:L2_appdx}
\vspace{-4mm}
\end{figure}

\section{Mean attended distance}
\label{appdx:mean_att_distance}
To understand the type of features that emerge at different layer depths as a function of the initialization strategy WT-ST for \deits, we calculate the mean attended distance per layer.
That is, for each of the \deitsmall' attention heads we calculate the mean of the element-wise multiplication of each query token's attention and its distance from the other tokens, similarly to \cite{dosovitskiy2020image}.
Then, we average the calculated distances per layer for all of the WT-ST initialization schemes.
In Figure \ref{fig:distance_appdx} in the Appendix, we report the mean attended distance per layer for all datasets and the average attended distance over all datasets.
The results clearly show that (1) the ST initialization results in global attention throughout the network (2) after the critical layers the attention is mainly global (3) the WT layers introduce a mixture of local and global features.
This suggests that the WT layers are important for a mixture of local and global features that the model is incapable of learning on its own -- due to the small data size.

\begin{figure}[t!]
\begin{center}
\begin{tabular}{@{}c@{}c@{}c@{}c@{}}
    \includegraphics[width=0.31\columnwidth]{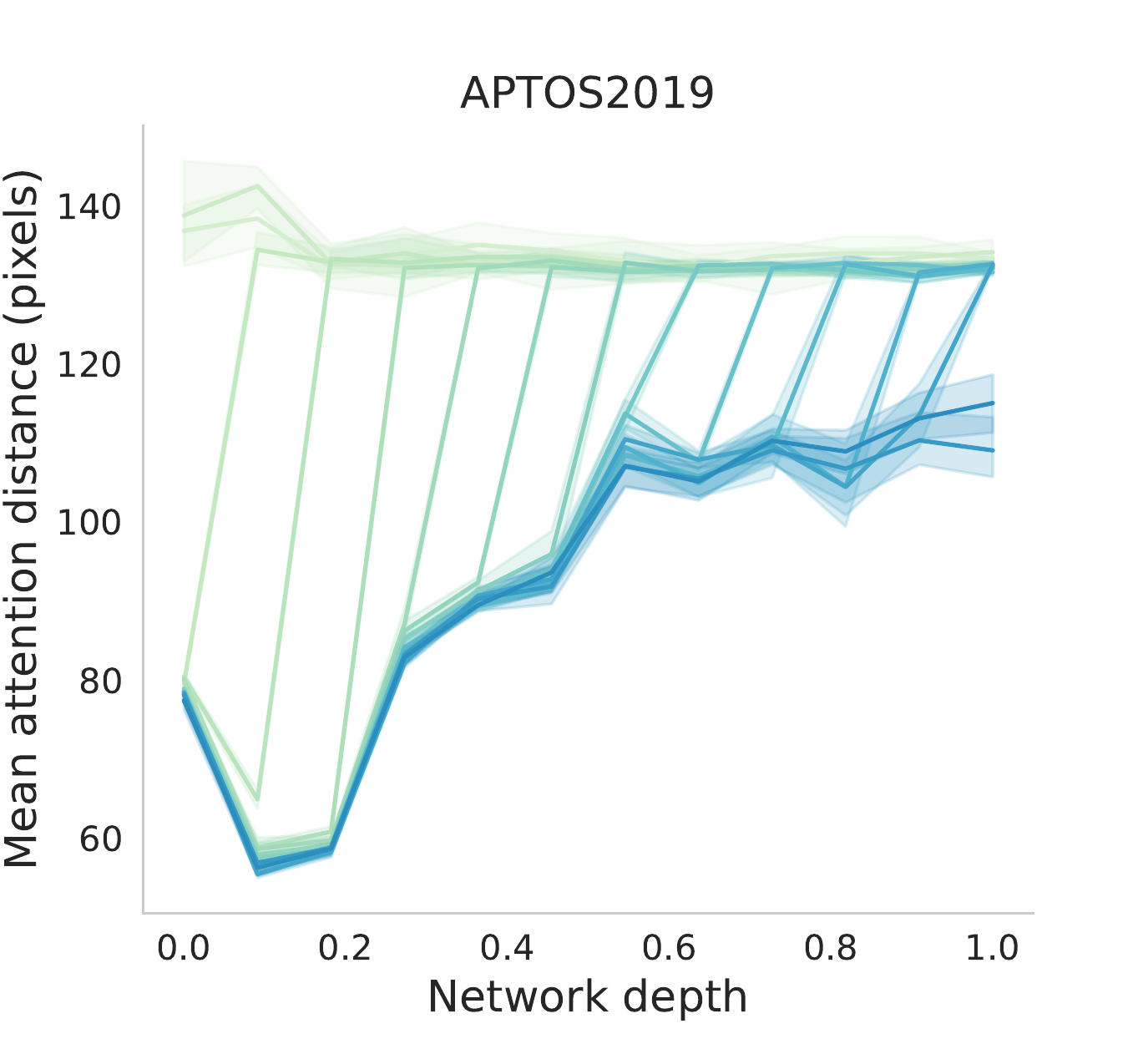} & 
    \includegraphics[width=0.31\columnwidth]{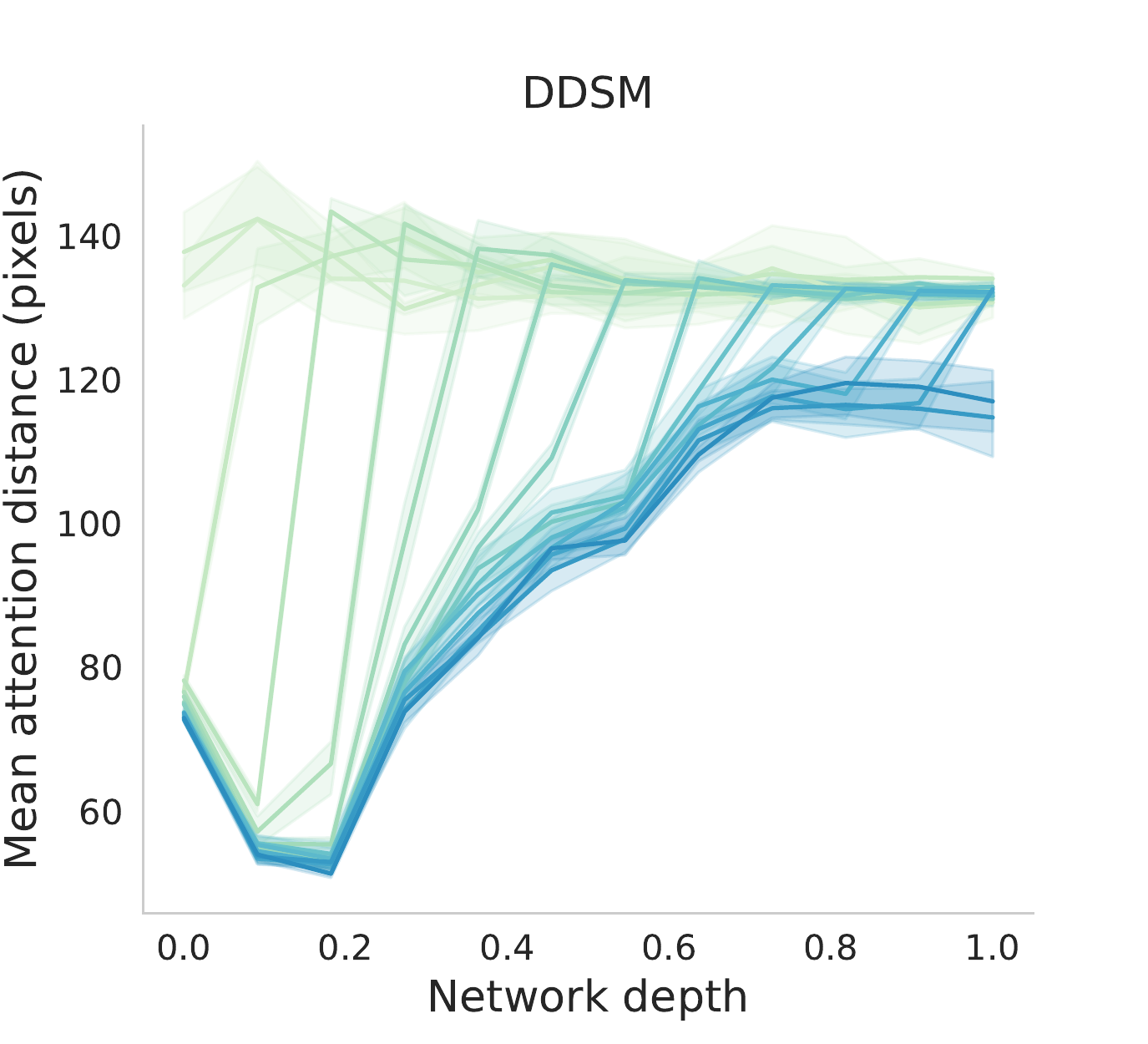} & 
    \includegraphics[width=0.31\columnwidth]{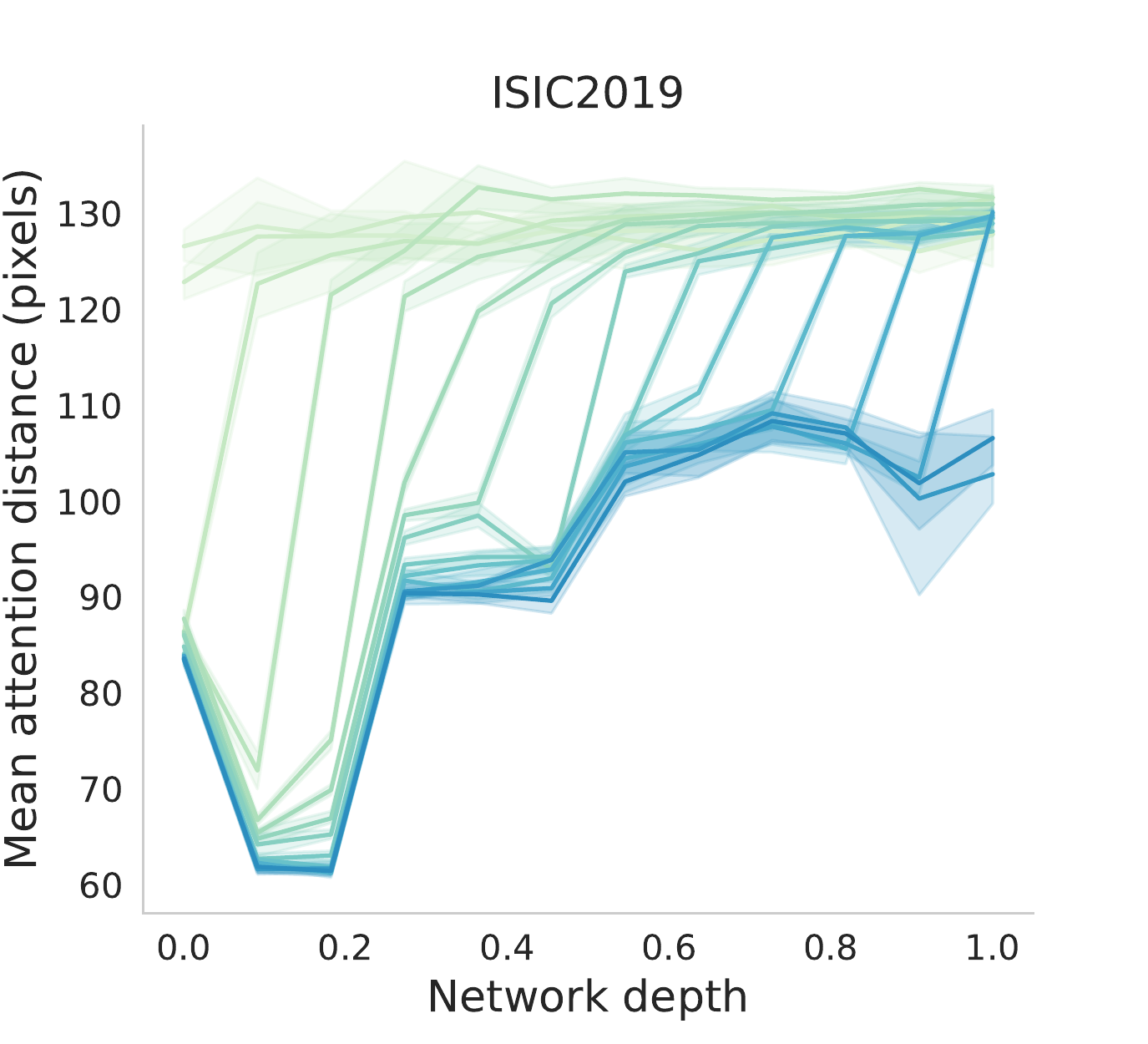} &
    \multirow{2}{*}[1.7cm]{\includegraphics[width=0.075 \columnwidth ]{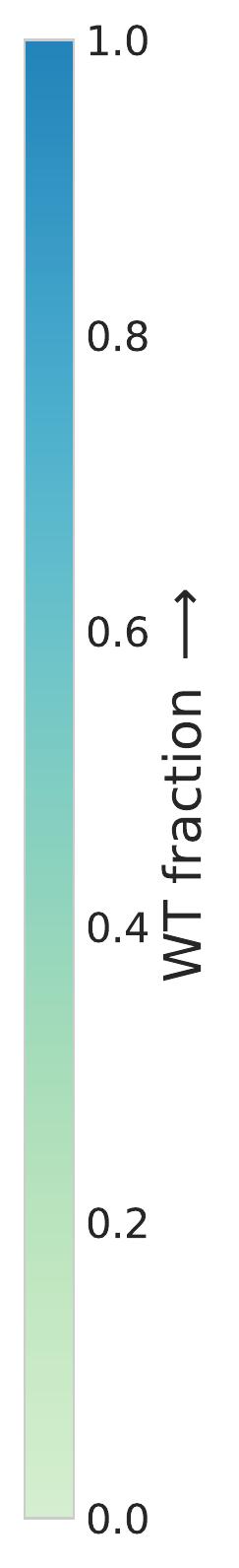}} 
    \\[-1.5mm]
    \includegraphics[width=0.31\columnwidth]{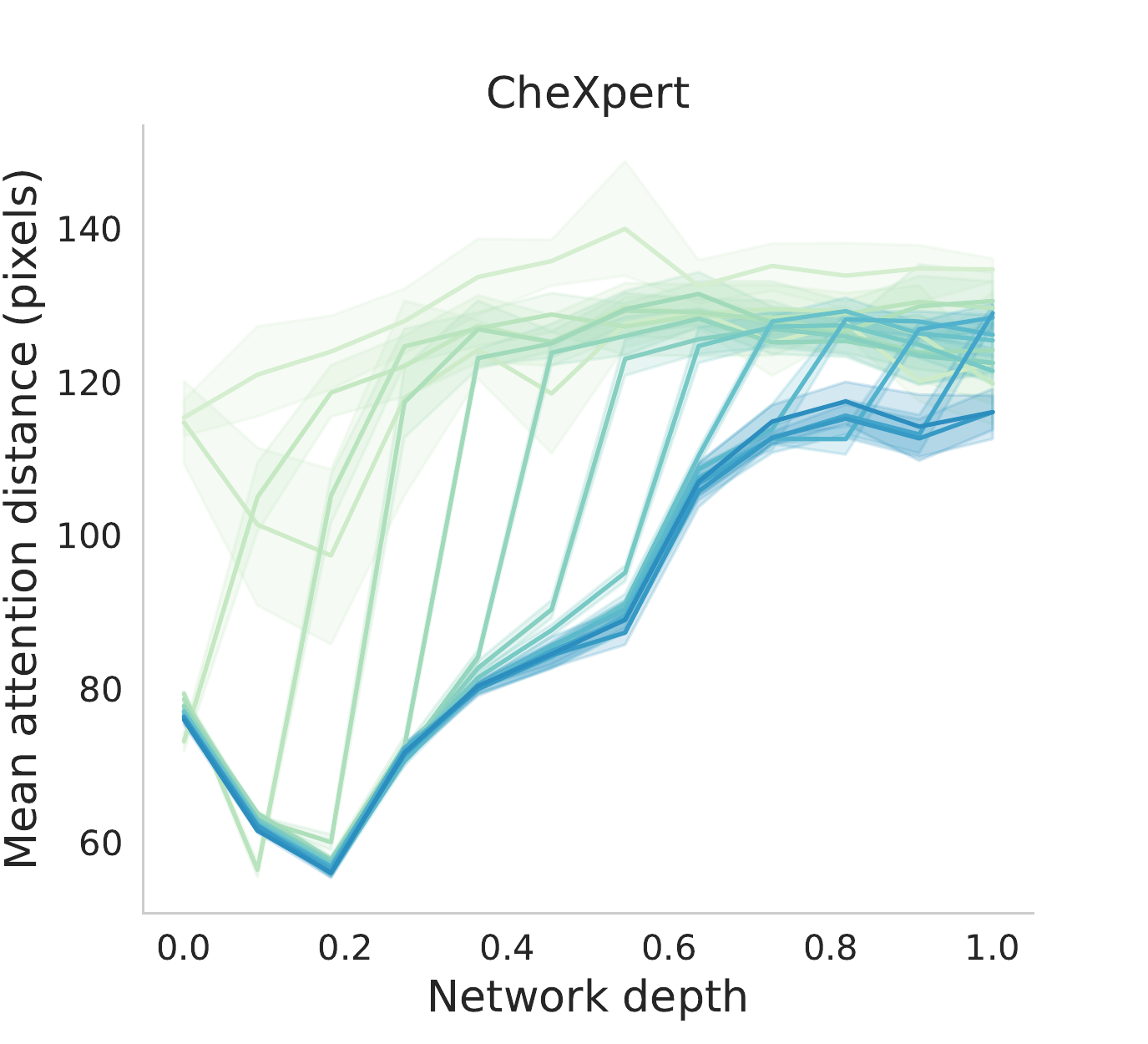} & 
    \includegraphics[width=0.31\columnwidth]{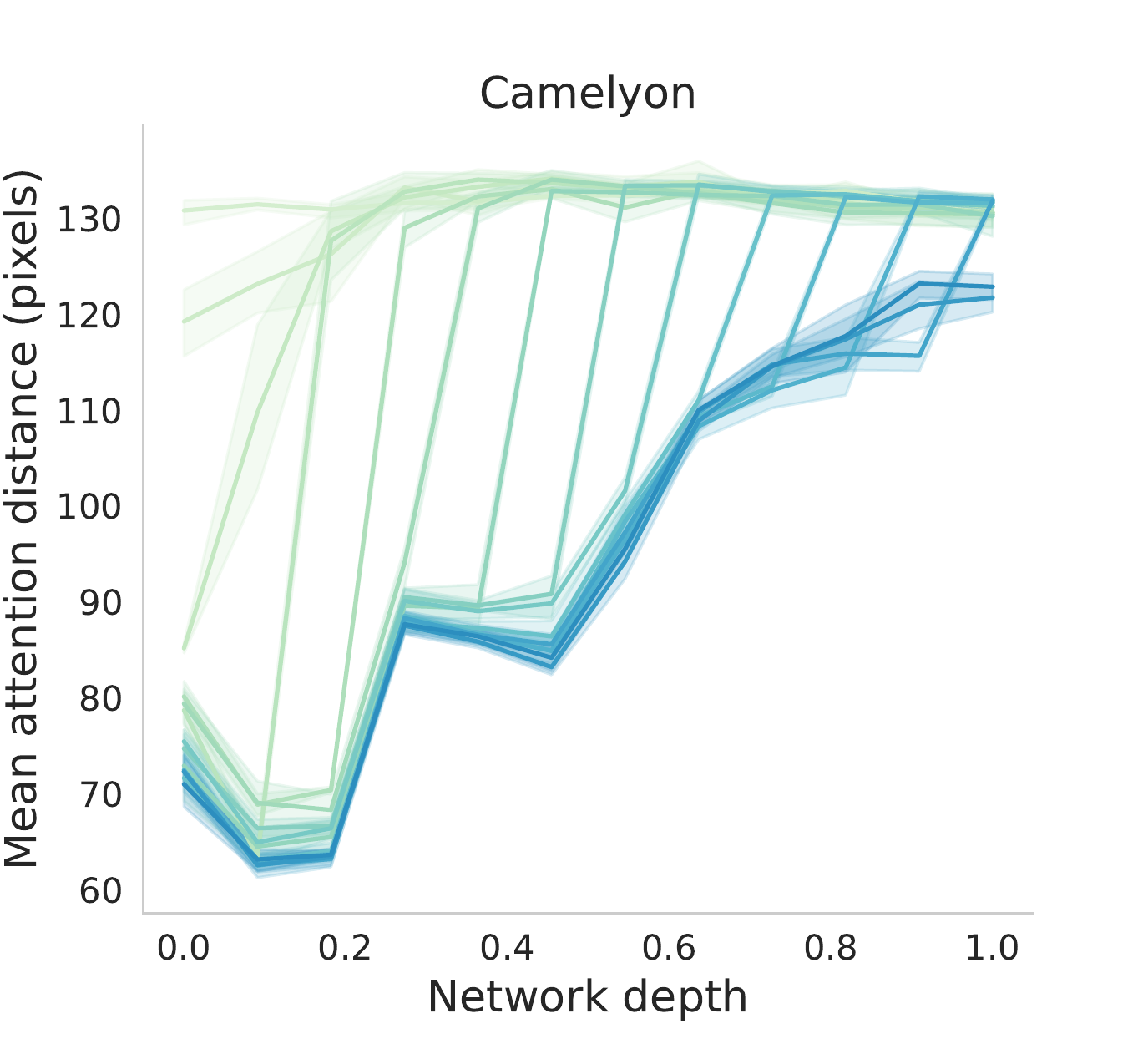} & 
    \includegraphics[width=0.31\columnwidth]{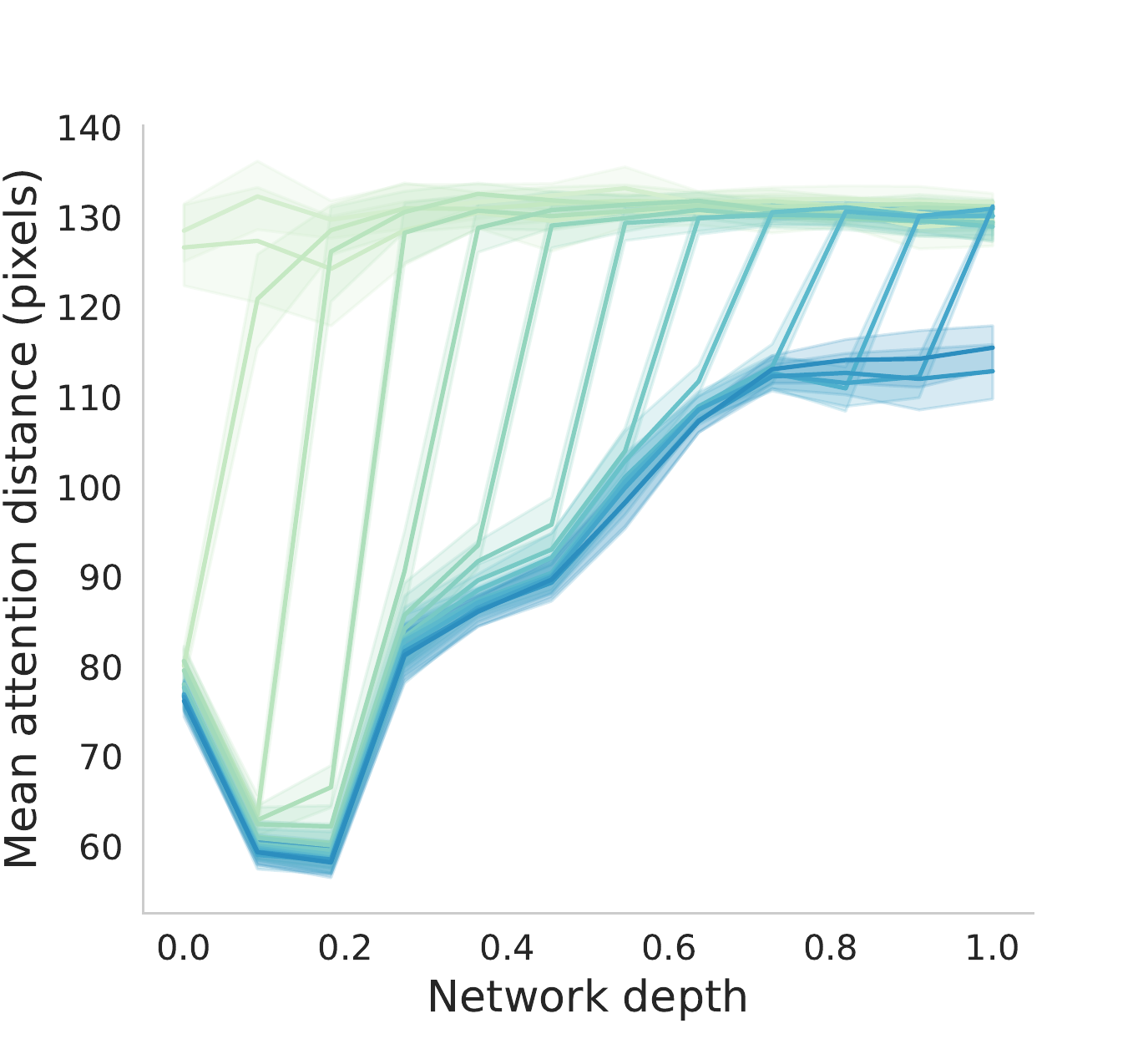} & 
    \\[-1.5mm] 
\end{tabular}
\end{center}
\vspace{-4mm}
\caption{\emph{Mean attended distance for different initializations.}
We report the mean attended distance of the fine-tuned \deitsmall model for all datasets using different WT-ST initializations  (WT fraction from 0 to 1, where 0 = ST and 1 = WT).
The bottom-right figure shows the mean attended distance, averaged over all datasets.
Evidently, in the absence of WT layers the attention is mainly global, whilst the \imagenet pre-trained weights introduce a mixture of local and global attention that the network cannot learn on its own.}
\label{fig:distance_appdx}
\vspace{-2mm}
\end{figure}

\begin{figure}[t]
\begin{center}
\begin{tabular}{@{}c@{}c@{}c@{}}
    \includegraphics[width=0.333\columnwidth]{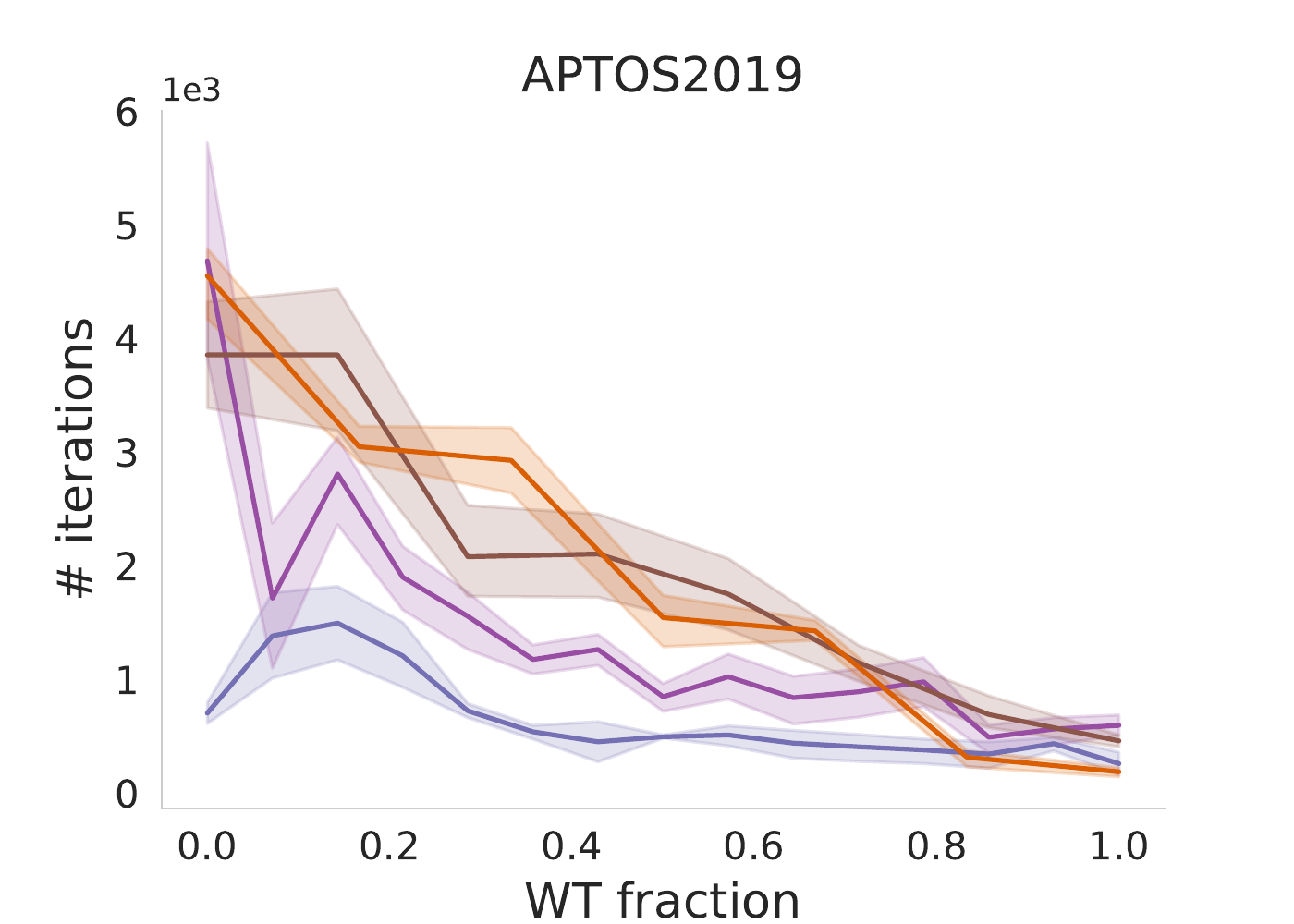} & 
    \includegraphics[width=0.333\columnwidth]{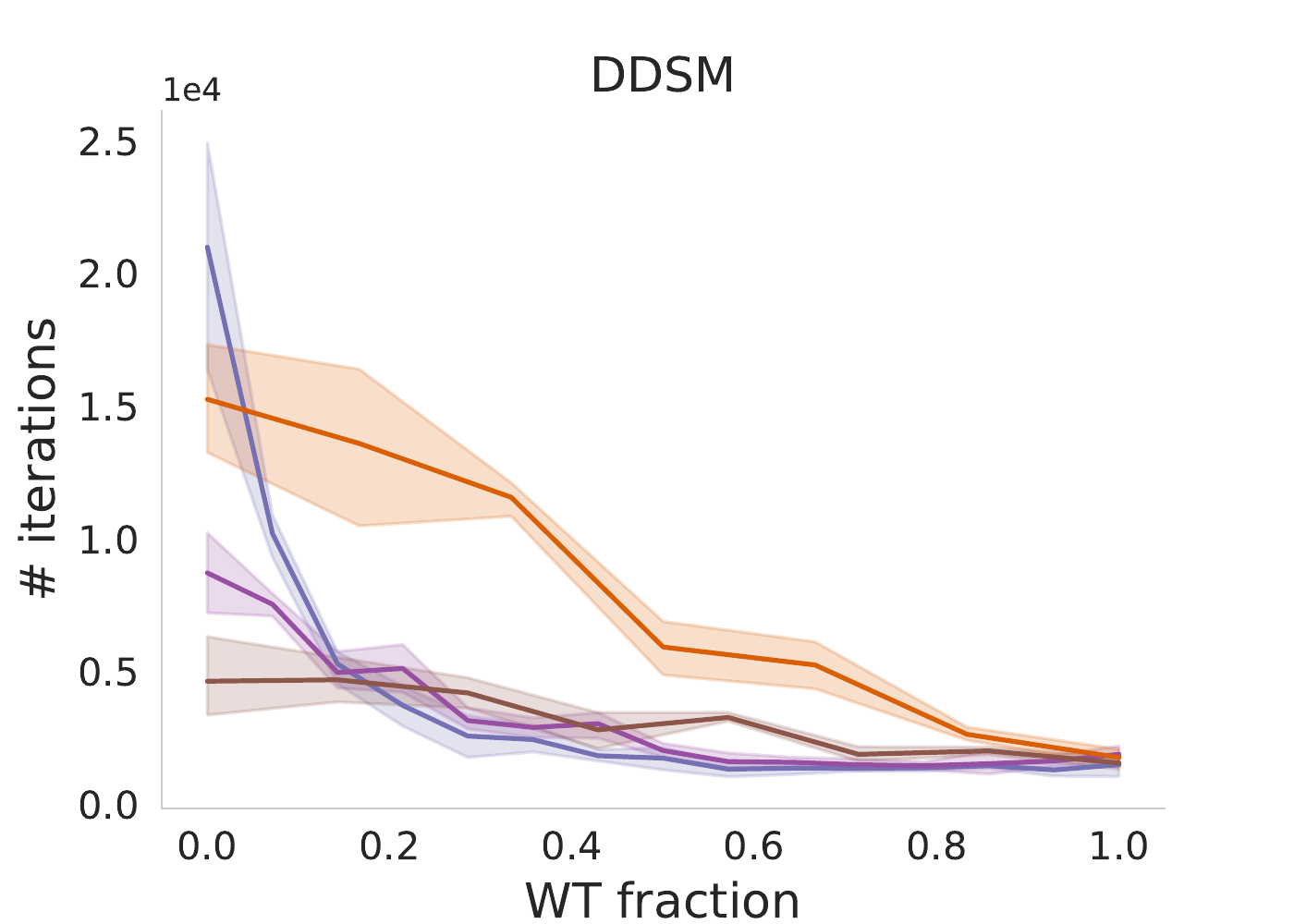} & 
    \includegraphics[width=0.333\columnwidth]{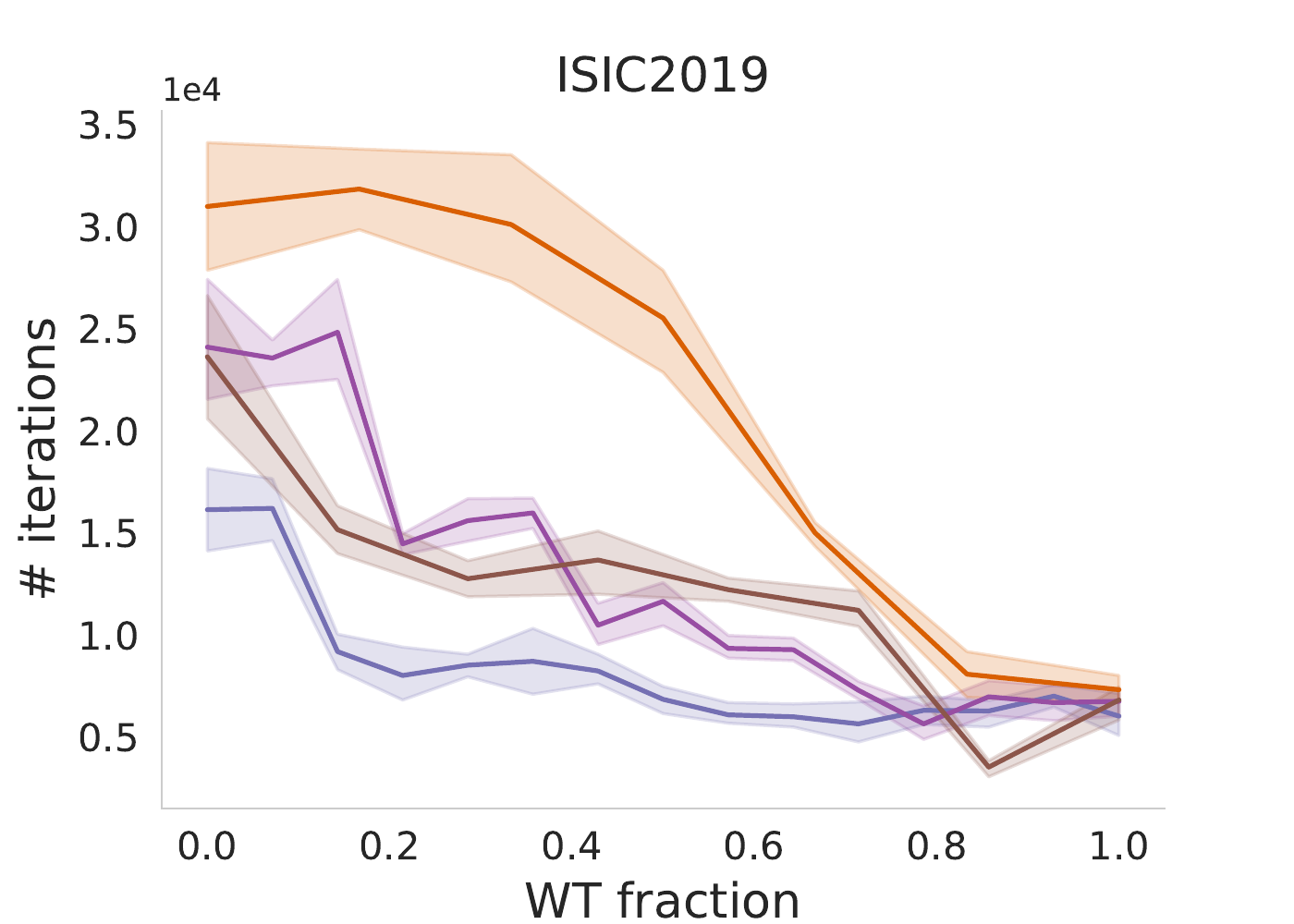} \\[-1.5mm] 
    \includegraphics[width=0.333\columnwidth]{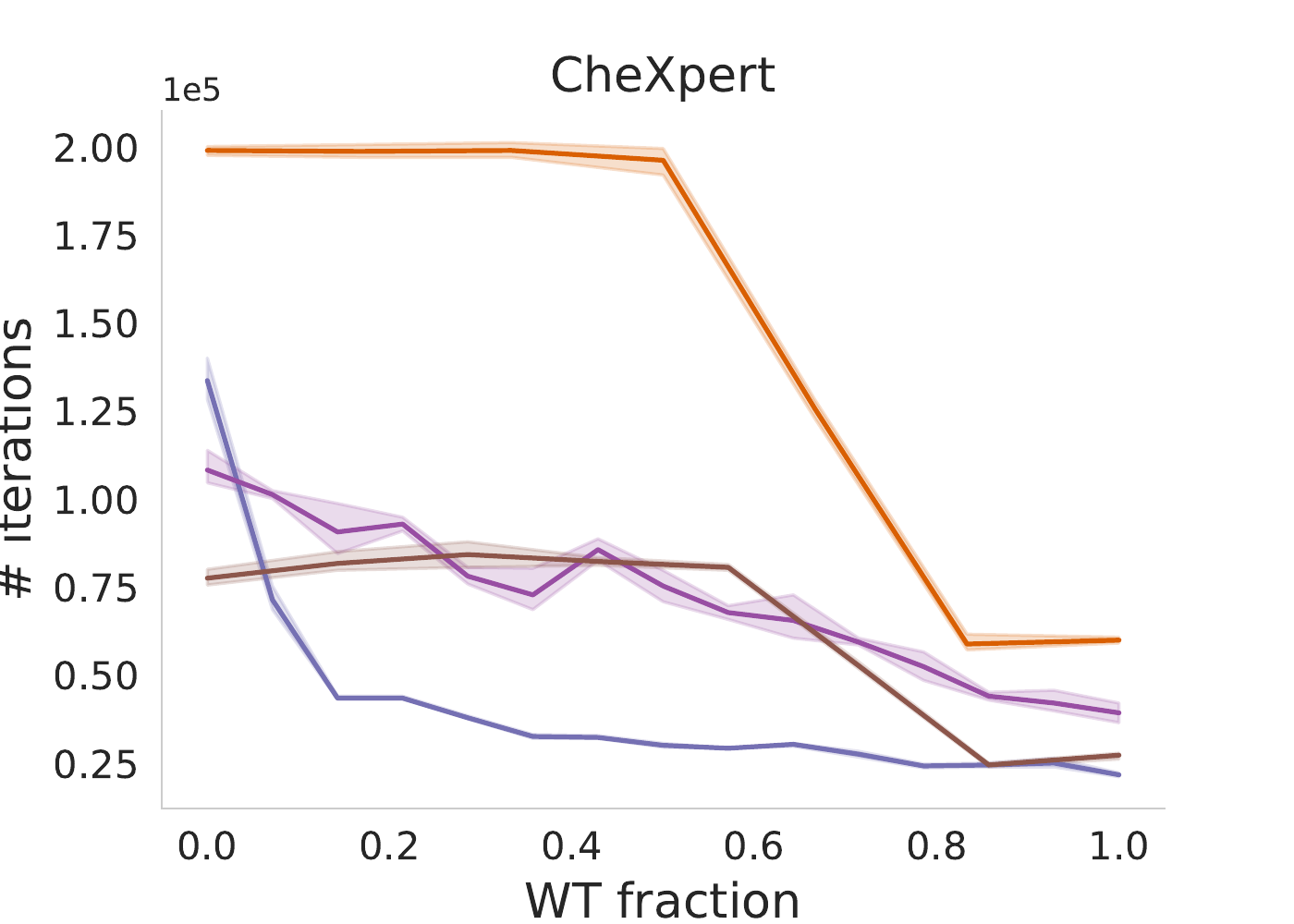} & 
    \includegraphics[width=0.333\columnwidth]{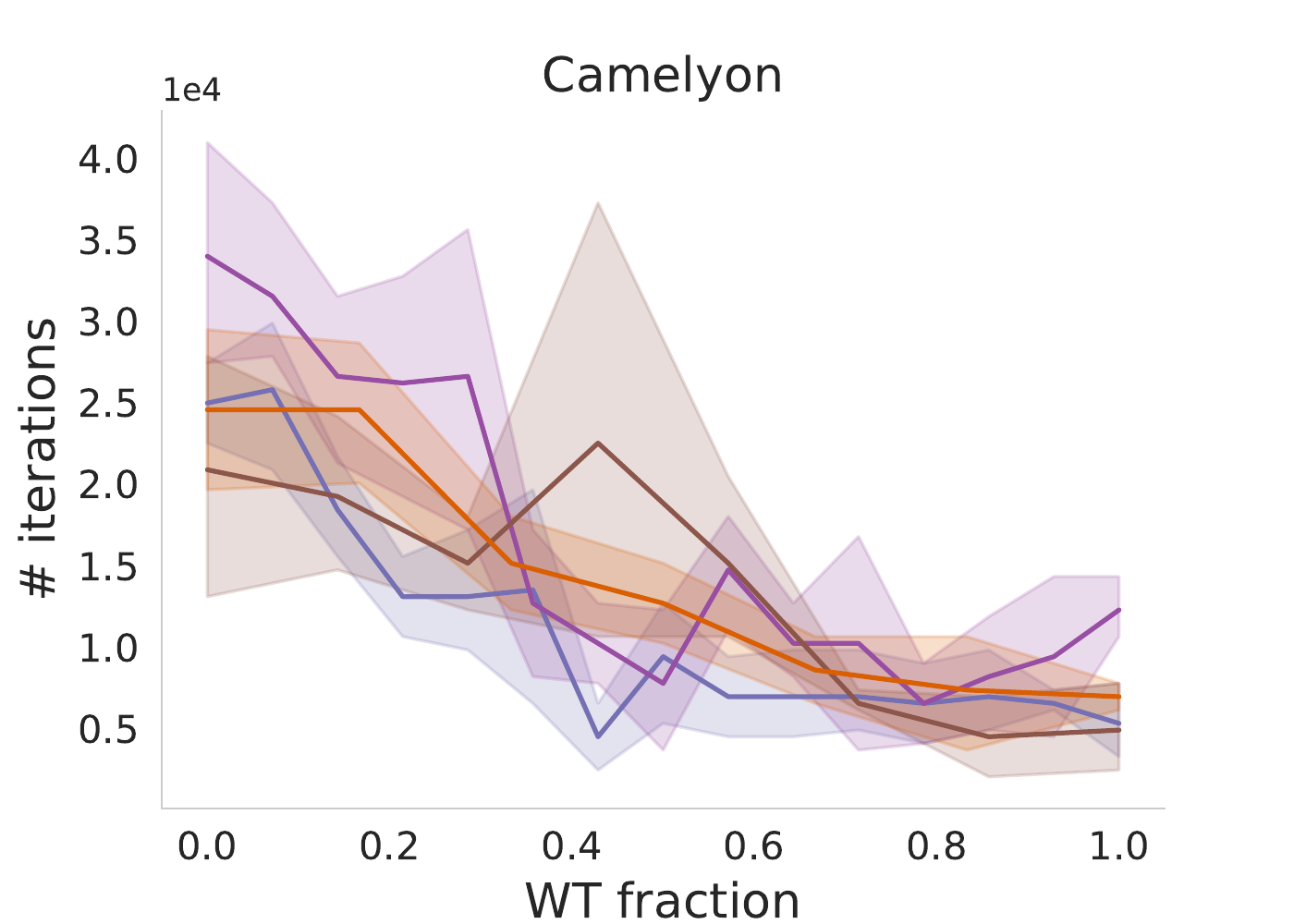} & 
    \includegraphics[width=0.333\columnwidth]{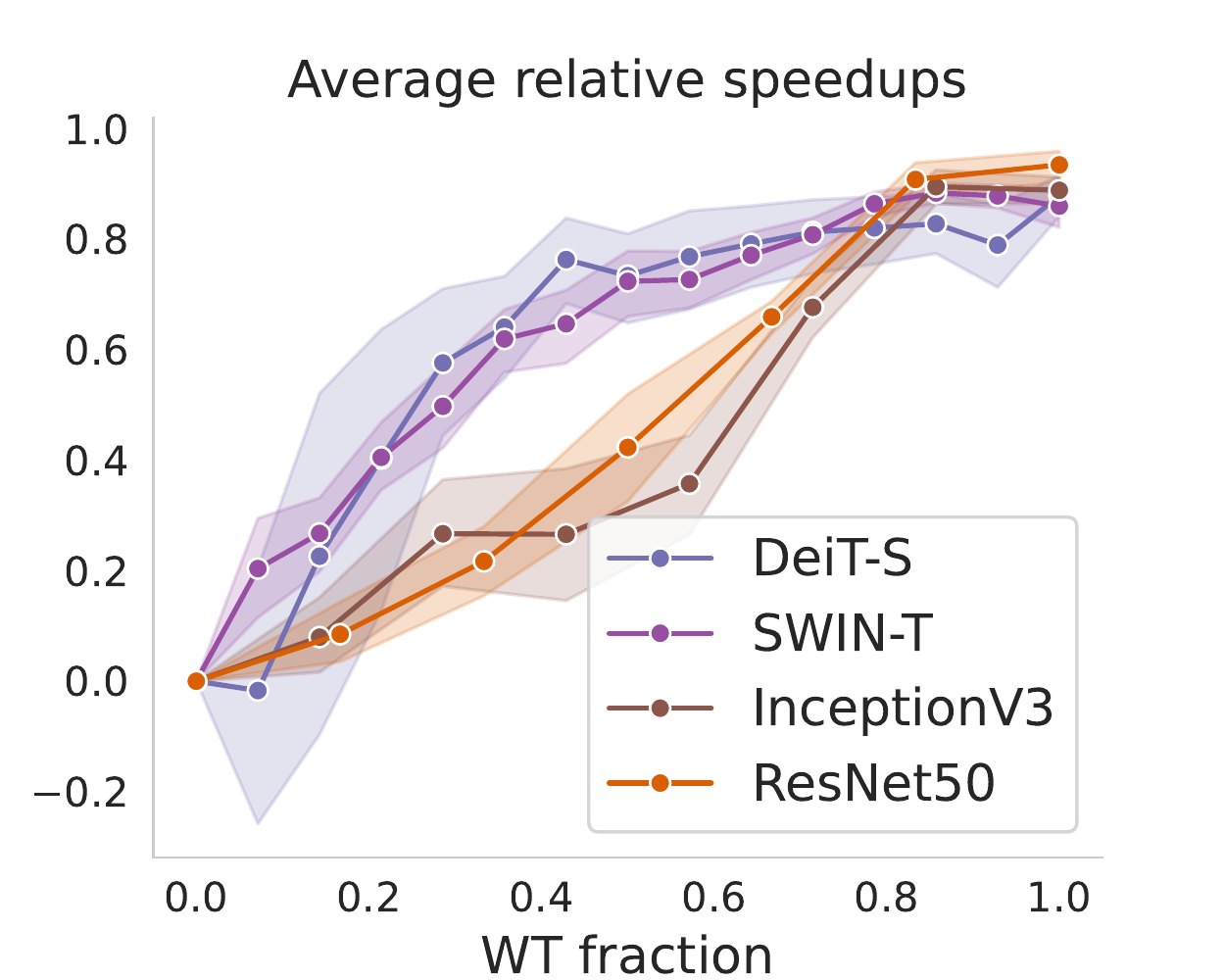}\\[-1.5mm] 
\end{tabular}
\end{center}
\vspace{-3mm}
\caption{\emph{Convergence speed as a function of WT fraction for different models and datasets.} We report the number of iterations it takes for each model to converge for different WT fractions for each individual dataset. The bottom-right figure shows the relative speedups averaged over all datasets. Evidently, the convergence speed monotonically increases with the number of WT layers for all datasets and models.}
\label{fig:convergence_appdx}
\vspace{-4mm}
\end{figure}

\section{Model convergence}
\label{sec:convergence-study}
We investigate how the convergence behavior of the models change as we transfer more layers. Figure \ref{fig:convergence_appdx} in the Appendix and Figure \ref{fig:convergence} in the main text show the number of iterations needed for each model to reach its best validation performance. As a general trend for all models, the higher the WT fraction is, the faster the models converge. Interestingly, for vision transformers, transferring the first few blocks dramatically increases the convergence speed, while transferring further blocks slightly speeds up training. CNNs however, follow a different trend, where transferring more layers improves the convergence speed at a roughly linear rate.

\section{Model capacity}
\label{appdx:capacity-study}
We investigate the impact that the model's capacity has on transfer learning for {\deits} \cite{deit} and {\resnets} \cite{resnet}. To this end, we consider 3 different capacities for each architecture, which are comparable in the number of parameters and compute time. For the {\resnet} family, we considered {\resnetEighteen}, {\resnetfifty}, and {\resnetOneFiftyTwo}. For the {\deit} family, we chose {\deitTiny}, {\deitsmall}, and {\deitBase}. For each model capacity, we carry out the same \wtst experiments. That is, we initialize  each model with different {\wtst} initialization schemes and then we fine-tune them on the target task. The training strategy follows exactly the details mentioned in Section \ref{methods}. Please refer to to Figure \ref{fig:convergence} and Section \ref{experiments} for the results and discussion.

\addtolength{\tabcolsep}{-2pt} 
\begin{table}[t]
\tiny
\begin{tabular}{lll|lll}
\toprule
\multicolumn{2}{l}{\textbf{Initialization for Inception}} &
\multicolumn{1}{l|}{\textbf{Features of Inception}} &
\multicolumn{2}{l}{\textbf{Initialization for ResNets}} &
\multicolumn{1}{l}{\textbf{Features of ResNet50}} 
\\
\midrule
\multicolumn{1}{c}{\textbf{WT-ST-n/m}}  &
\multicolumn{1}{l}{\textbf{Module}}  &
\multicolumn{1}{l|}{\textbf{Intermediate layers}}  &
\multicolumn{1}{c}{\textbf{WT-ST-n/m}}  &
\multicolumn{1}{l}{\textbf{Module}}  &
\multicolumn{1}{l}{\textbf{Intermediate layers}}
\\
\midrule

\multirow{1}{*}{WT-ST-1/6} &
Layer 1 & Conv2d &
\multirow{1}{*}{WT-ST-1/5} &
Layer 1 & Conv2d 
\\[0.5em]
\multirow{2}{*}{WT-ST-2/5} &
\multirow{2}{*}{Layer 2} &
Layer 2, Conv2d  &
\multirow{2}{*}{WT-ST-2/4} &
\multirow{2}{*}{Norm layer (BN)} & \\ & &
Layer 2, Conv2d & & & 
\\[0.5em]

\multirow{1}{*}{WT-ST-3/4} &
Layer 3 &
Layer 3, Conv2d  &
\multirow{3}{*}{WT-ST-3/3} &
\multirow{3}{*}{Stage 1} &
Stage 1, Block 1 \\
\multirow{1}{*}{WT-ST-4/3} &
Layer 4 &
Layer 4, Conv2d & & &
Stage 1, Block 2\\
 & 
 & 
 & & & 
Stage 1, Block 3
\\[0.5em]

\multirow{3}{*}{WT-ST-5/2} & 
\multirow{3}{*}{Stage 1} & 
Stage 1, Block 1 -- Type A &
\multirow{4}{*}{WT-ST-4/2} &
\multirow{4}{*}{Stage 2} &
Stage 2, Block 1 \\ 
& &
Stage 1, Block 3 -- Type A &
& & 
Stage 2, Block 2\\
& & 
Stage 1, Block 3 -- Type A
& & & 
Stage 2, Block 3\\
& & 
& & & 
Stage 2, Block 4
\\[0.5em]

\multirow{5}{*}{WT-ST-6/1} &
\multirow{5}{*}{Stage 2} &
Stage 2, Block 1 -- Type B &
\multirow{6}{*}{WT-ST-5/1} &
\multirow{6}{*}{Stage 3} &
Stage 3, Block 1 \\ & & 
Stage 2, Block 2 -- Type C
& & & 
Stage 3, Block 2\\ & & 
Stage 2, Block 3 -- Type C
& & & 
Stage 3, Block 3\\ & & 
Stage 2, Block 4 -- Type C
& & & 
Stage 3, Block 4 \\ & & 
Stage 2, Block 5 -- Type C
& & & 
Stage 3, Block 5\\ & & 
& & & 
Stage 3, Block 6
\\[0.5em]

\multirow{3}{*}{WT-ST-7/0} &
\multirow{3}{*}{Stage 3} &
Stage 3, Block 1 -- Type D &
\multirow{3}{*}{WT-ST-6/0} &
\multirow{3}{*}{Stage 4} &
Stage 4, Block 1 \\ & &
Stage 3, Block 2 -- Type E
 & & &
Stage 4, Block 2\\ & &
Stage 3, Block 3 -- Type E
 & & &
Stage 4, Block 3
\\[0.5em]

\bottomrule
\end{tabular}

\vspace{-2mm}
\caption{\emph{Implementation details for the CNN models}.
\textbf{(Left)} Modules and layers used for \inception. 
\textbf{(Right)} Modules and layers used for \resnets.
The first column of each side shows the initialization type that we used.
The module that corresponds to each initialization scheme (second column from each side) is the last module that we initialized with WT. The modules after that were initialized with ST.
The third row from each side shows the layers we used for the \knn, \ltwo, re-initialization and representation similarity experiments.
}
\label{tab:cnn_layer_details}
\vspace{-3mm}
\end{table}
\addtolength{\tabcolsep}{3pt}

\section{The WT-ST initialization schemes}
\label{appdx:wtst_details}
Here, we provide additional details regarding the \wtst initialization procedure and the modules we used to investigate where feature reuse occurs within the network.
We transfer weights (WT) up to block $n$ and we initialize the remaining $m$ blocks using ST.
In practice this means that, the first $n$ modules use the exact ImageNet pretrained weights, while the weights of the next $m$ modules are initialised with a Normal distribution $\mathcal{N}(\mu_i,\,{\sigma_i}^{2})$, where $\mu_i$ and ${\sigma_i}^{2}$ are the mean and variance of the $i$th ImageNet pretrained weight.

Due to the architectural differences, we use a different selection of modules for each model. 
For \deits and \swins, we use the input layer (patchifier), each of the transformer blocks and the last normalization layer of the network.
For \inception we use the first four modules which include the layers that operate at the same scale and the inception modules that belong to the same stage.
Finally, for \resnets we include the input layer, the first normalization layer and the resnet blocks from each scale. 
The exact details for each {\wtst} setting for {\resnetfifty},  {\resnetEighteen}, {\resnetOneFiftyTwo} and {\inception} can be found in Table \ref{tab:cnn_layer_details} in the Appendix. Similarly, the details for {\deit}, {\deitTiny}, {\deitBase} and {\swin}-T are found in Table \ref{tab:vit_layer_details} in the Appendix.
The results of these experiments are reported in Figure \ref{fig:wst_all} and \ref{fig:wst_capacity} in the main text.

\addtolength{\tabcolsep}{-2pt} 
\begin{table}[t]
\tiny
\begin{tabular}{lll|lll}
\toprule
\multicolumn{2}{l}{\textbf{Initialization for DeiT}} &
\multicolumn{1}{l|}{\textbf{Features of DeiT-S}} &
\multicolumn{2}{l}{\textbf{Initialization for SWIN-T}} &
\multicolumn{1}{l}{\textbf{Features of SWIN-T}} 
\\
\midrule
\multicolumn{1}{c}{\textbf{WT-ST-n/m}}  &
\multicolumn{1}{l}{\textbf{Module}}  &
\multicolumn{1}{l|}{\textbf{Intermediate layers}}  &
\multicolumn{1}{c}{\textbf{WT-ST-n/m}}  &
\multicolumn{1}{l}{\textbf{Module}}  &
\multicolumn{1}{l}{\textbf{Intermediate layers}}
\\
\midrule

\multirow{1}{*}{WT-ST-1/13} &
Patchifier & Conv2d &
\multirow{1}{*}{WT-ST-1/13} &
Patchifier & Conv2d 
\\[0.5em]

\multirow{2}{*}{WT-ST-2/12} &
\multirow{2}{*}{Block 1} &
Block 1 -- Attention  &
\multirow{2}{*}{WT-ST-2/12} &
\multirow{2}{*}{Stage 1, Block 1} &
Stage 1, Block 1 -- Attention \\ & &
Block 1 -- MLP & & & 
Stage 1, Block 1 -- MLP

\\[0.5em]
\multirow{2}{*}{WT-ST-3/11} &
\multirow{2}{*}{Block 2} &
Block 2 -- Attention &
\multirow{2}{*}{WT-ST-3/11} &
\multirow{2}{*}{Stage 1, Block 2} &
Stage 1, Block 2 -- Attention \\ & &
Block 2 -- MLP & & & 
Stage 1, Block 2 -- MLP
\\[0.5em]
\multirow{2}{*}{WT-ST-4/10} &
\multirow{2}{*}{Block 3} &
Block 3 -- Attention &
\multirow{2}{*}{WT-ST-4/10} &
\multirow{2}{*}{Stage 2, Block 1} &
Stage 2, Block 1 -- Attention \\ & &
Block 3 -- MLP & & & 
Stage 2, Block 1 -- MLP
\\[0.5em]
\multirow{2}{*}{WT-ST-5/9} &
\multirow{2}{*}{Block 4} &
Block 4 -- Attention &
\multirow{2}{*}{WT-ST-5/9} &
\multirow{2}{*}{Stage 2, Block 2} &
Stage 2, Block 2 -- Attention \\ & &
Block 4 -- MLP & & & 
Stage 2, Block 2 -- MLP
\\[0.5em]
\multirow{2}{*}{WT-ST-6/8} &
\multirow{2}{*}{Block 5} &
Block 5 -- Attention &
\multirow{2}{*}{WT-ST-6/8} &
\multirow{2}{*}{Stage 3, Block 1} &
Stage 3, Block 1 -- Attention \\ & &
Block 5 -- MLP & & & 
Stage 3, Block 1 -- MLP
\\[0.5em]
\multirow{2}{*}{WT-ST-7/7} &
\multirow{2}{*}{Block 6} &
Block 6 -- Attention &
\multirow{2}{*}{WT-ST-7/7} &
\multirow{2}{*}{Stage 3, Block 2} &
Stage 3, Block 2 -- Attention \\ & &
Block 6 -- MLP & & & 
Stage 3, Block 2 -- MLP
\\[0.5em]
\multirow{2}{*}{WT-ST-8/6} &
\multirow{2}{*}{Block 7} &
Block 7 -- Attention &
\multirow{2}{*}{WT-ST-8/6} &
\multirow{2}{*}{Stage 3, Block 3} &
Stage 3, Block 3 -- Attention \\ & &
Block 7 -- MLP & & & 
Stage 3, Block 3 -- MLP
\\[0.5em]
\multirow{2}{*}{WT-ST-9/5} &
\multirow{2}{*}{Block 8} &
Block 8 -- Attention &
\multirow{2}{*}{WT-ST-9/5} &
\multirow{2}{*}{Stage 3, Block 4} &
Stage 3, Block 4 -- Attention \\ & &
Block 8 -- MLP & & & 
Stage 3, Block 4 -- MLP
\\[0.5em]
\multirow{2}{*}{WT-ST-10/4} &
\multirow{2}{*}{Block 9} &
Block 9 -- Attention &
\multirow{2}{*}{WT-ST-10/4} &
\multirow{2}{*}{Stage 3, Block 5} &
Stage 3, Block 5 -- Attention \\ & &
Block 9 -- MLP & & & 
Stage 3, Block 5 -- MLP
\\[0.5em]
\multirow{2}{*}{WT-ST-11/3} &
\multirow{2}{*}{Block 10} &
Block 10 -- Attention &
\multirow{2}{*}{WT-ST-11/3} &
\multirow{2}{*}{Stage 3, Block 6} &
Stage 3, Block 6 -- Attention \\ & &
Block 10 -- MLP & & & 
Stage 3, Block 6 -- MLP
\\[0.5em]
\multirow{2}{*}{WT-ST-12/2} &
\multirow{2}{*}{Block 11} &
Block 11 -- Attention &
\multirow{2}{*}{WT-ST-12/2} &
\multirow{2}{*}{Stage 4, Block 1} &
Stage 4, Block 1 -- Attention \\ & &
Block 11 -- MLP & & & 
Stage 4, Block 1 -- MLP
\\[0.5em]
\multirow{2}{*}{WT-ST-13/1} &
\multirow{2}{*}{Block 12} &
Block 12 -- Attention &
\multirow{2}{*}{WT-ST-13/1} &
\multirow{2}{*}{Stage 4, Block 2} &
Stage 4, Block 2 -- Attention \\ & &
Block 12 -- MLP & & & 
Stage 4, Block 2 -- MLP
\\[0.5em]
\multirow{1}{*}{WT-ST-14/0} &
Layer norm & &
\multirow{1}{*}{WT-ST-14/0} &Layer norm &
\\[0.5em]

\bottomrule
\end{tabular}

\vspace{-2mm}
\caption{\emph{Implementation details for the ViT models}.
\textbf{(Left)} Modules and layers used for \deits. 
\textbf{(Right)} Modules and layers used for \swin-T.
The first column of each side shows the initialization type that we used.
The module that corresponds to each initialization scheme (second column from each side) is the last module that we initialized with WT. The modules after that were initialized with ST.
The third row from each side shows the layers we used for the \knn, \ltwo, re-initialization and representation similarity experiments.
}
\label{tab:vit_layer_details}
\vspace{-3mm}
\end{table}
\addtolength{\tabcolsep}{3pt}

\section{The 5-layer {\deitsmall} model}
\label{sec:smaller-deit}
As we showed in Figure \ref{fig:wst_all} and Figure \ref{fig:wst_knn} of the main text, it appears that vision transformers benefit significantly from weight transfer in their initial-to-middle  blocks, while transferring the weights in the later blocks seems to offer little or no benefit.
In fact, transferring weights too deep into the network \emph{may result in worse high-level features}, possibly due to biases learned during the pre-training task.
Furthermore, we noticed from the layer-wise experiments that critical layers often appear at the transition between WT and ST.
This begs  the question: \textit{
Can we use a smaller \deit model that has been initialized with weight transfer without compromising classification performance?} 

To this end, we use a trimmed version of a \deitsmall model that has only five transformer blocks -- effectively reducing the memory and computational requirements by a factor of 2.
We initialize this model, denoted as \deitsmall-5b, with weight transfer from \imagenet and we fine-tune it on the target datasets, using the settings described in Section \ref{methods}.
Surprisingly, our results in Table \ref{tab:trimmed_vs_full} showing no significant changes in classification performance.
This supports the arguments that: \emph{1)} the initial blocks of ViTs contribute the most to the overall performance of the model, \emph{2)} feature reuse in the first layers is so strong for \deits that can compensate for the lack of additional transformer blocks.

This finding might be of further interest for practitioners who work with limited computational and memory budgets.
For example, in medical imaging, there is a need for light-weight models as the large image sizes that are encountered in practice prohibit the utilization of large models.
However, further evaluation is needed to asses the extent to which these benefits are broadly applicable.

\vfill\eject
 
\addtolength{\tabcolsep}{-2.5pt} 
\begin{table}[ht]
\tiny
\begin{tabular}{lccccc}
\toprule
\textbf{Model} & 
\textbf{APTOS2019}, $\kappa \uparrow$  &
\textbf{DDSM}, AUC $\uparrow$ &
\textbf{ISIC2019}, Rec. $\uparrow$ & 
\textbf{CheXpert}, AUC $\uparrow$ &
\textbf{Camelyon}, AUC $\uparrow$ 
\\
& 
$n =$ 3,662  & 
$n =$ 10,239 &
$n =$ 25,333 & 
$n =$ 224,316 &
$n =$ 327,680 \\
\midrule

DeiT-S &
0.894  $\pm$ 0.017 &
0.949  $\pm$ 0.011 &
0.824  $\pm$ 0.008 &
0.792  $\pm$ 0.001 &
0.962  $\pm$ 0.003 
\\
DeiT-S-5b &
0.894  $\pm$ 0.005 &
0.951  $\pm$ 0.001 &
0.812  $\pm$ 0.016 &
0.792  $\pm$ 0.001 &
0.962  $\pm$ 0.004 
\\[0.5em]

\bottomrule
\end{tabular}
\vspace{-2mm}
\caption{
\emph{A trimmed {\deitsmall} with only 5 blocks performs comparably to the full {\deitsmall} model.} We keep only the first 5 (out of 12) blocks of {\deitsmall} and discard the rest. Then, after WT initialization, we fine-tune the model with the same strategy detailed in Section \ref{methods}. It can clearly be seen that the smaller model performs competitively to the full \deitsmall, even when more than half of the blocks are removed.}
\label{tab:trimmed_vs_full}
\vspace{-3mm}
\end{table}
\addtolength{\tabcolsep}{3pt}

\end{appendices}

\end{document}